# ADVANCED MODELLING AND DATA ANALYTICS IN AVIATION

**Aziida Nanyonga**

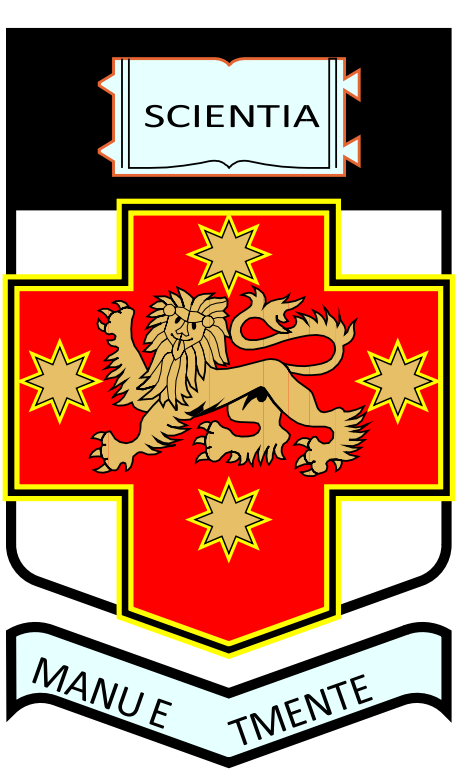

A thesis submitted in partial fulfillment of the requirements for the degree of Doctor of Philosophy at the

School of Engineering & Technology

University of New South Wales at

Australian Defence Force Academy

# Abstract

The aviation industry characterized by its stringent safety standards has seen a growing need for innovative approaches to enhance safety measures. Despite the vast accumulation of aviation safety data over time, its full potential in predicting and preventing incidents has not been fully realized. This research addresses this gap by applying machine learning (ML) and natural language processing (NLP) techniques to analyze aviation safety data from Socrata, the Australian Transport Safety Bureau (ATSB), the National Transportation Safety Board (NTSB), and the Aviation Safety Network (ASN). By leveraging existing ML models, including deep learning and transformer-based architectures alongside NLP methods for mining aviation incident narratives, this study uncovers patterns contributing to safety related incidents such as accidents and near-misses. Additionally, it employs various topic modelling techniques to extract meaningful themes from unstructured safety reports, enhancing the interpretability of incident analysis. Causal inference techniques and interpretable AI frameworks are further explored to improve model transparency and trustworthiness. A key contribution of this work is the deployment of advanced ML methodologies in a structured aviation safety context, assessing their effectiveness and providing insights into their practical implementation. The findings offer valuable insights for aviation stakeholders, including regulators, airlines, and policymakers, by providing data-driven solutions that enhance incident analysis and decision-making. Ultimately, this research supports the industry's ongoing efforts to minimize risks, improve passenger and crew security, and integrate AI-driven methodologies into aviation safety management.

# Acknowledgement

First and foremost, my deepest gratitude goes to my beloved son, whose love, laughter, and presence bring boundless joy to my life. Your innocence and endless energy remind me daily of life's true beauty. Through every challenge, your unwavering love has been my greatest source of strength. This achievement is for you.

To my amazing husband, my unwavering pillar of support—thank you for believing in me even when I doubted myself. Your patience, understanding, and steadfast encouragement have carried me through every hurdle. Your love and reassurance have been a guiding light, and I am beyond grateful to share this journey with you.

I extend my heartfelt appreciation to my primary supervisor, Dr. Graham Wild. Your dedication and unwavering support, especially during the most challenging moments, have been instrumental in the completion of this research. Your profound knowledge, pragmatic approach, and invaluable advice have not only refined my academic work but have also inspired me in ways I will always cherish. If I had to do my PhD all over again, I would choose you without hesitation, countless times. You have been everything I could have ever hoped for in a supervisor. Words cannot fully express my gratitude for your mentorship, guidance, and encouragement thank you. Your expertise, constructive feedback, and unwavering support have shaped the direction of my research and helped me grow both as a scholar and as a person. I am incredibly fortunate to have had the opportunity to work under your guidance, and your commitment to excellence has inspired me every step of the way. I hope that while reading this section, you feel just how much I appreciate you.

I am also deeply thankful to my secondary supervisor, Dr. Ugur Turhan, who has been with me since the beginning of this journey. Your continuous support, insightful feedback, and dedication to my development have been invaluable. Your guidance has played a crucial role in shaping this research, and I am truly grateful for your unwavering encouragement.

To Dr. Keith Joiner, even though you joined my supervisory team later in the process, I sincerely appreciate your expertise and the constructive feedback that significantly enhanced my research. Your input has been invaluable, and I am truly grateful for the time and effort you dedicated to refining this work.

To my dear friend Hassan, thank you for always being there when I needed support. Whether it was providing access to your server, which allowed me to run my work smoothly, reading through my papers before I submitted them to my supervisors, or even coming to the university on weekends so I wouldn't have to be there alone while running my experiments, your generosity, patience, and kindness have made a world of difference. I deeply appreciate everything you have done for me.

To my wonderful mother, your unwavering love and willingness to listen, even in the late hours when I had no one else to talk to, mean more to me than words can express. Your presence has been a source of comfort and strength, and I am endlessly grateful for you.

Each of you has played a unique and irreplaceable role in this journey. Your support, belief, and kindness have made this achievement possible, and I carry my gratitude for you in my heart always.

This research was supported by the UNSW Tuition Fee Scholarship (TFS).

# Certificate of Originality

I hereby declare that this submission is my work and that, to the best of my knowledge and belief, it contains no material previously published or written by another person, nor material which to a substantial extent has been accepted for the award of any other degree or diploma at UNSW or any other educational institution, except where due acknowledgement is made in the thesis. Any contribution made to the research by colleagues, with whom I have worked at UNSW or elsewhere, during my candidature, is fully acknowledged.

I also declare that the intellectual content of this thesis is the product of my work, except to the extent that assistance from others in the project's design and conception or in style, presentation and linguistic expression is acknowledged.

Aziida Nanyonga

# Table of Contents

Contents

# List of Figures

# List of Tables

# List of Publications

1. Nanyonga, A., Turhan, U., Joiner, K., & Wild, G. A Systematic Review of Machine Learning Analytic Methods for Aviation Accident Research, Submitted to TRIP.
2. Nanyonga, A., Wasswa, H., Turhan, U., Joiner, K., & Wild, G. A Multi-head Attention-Based Transformer Model for Predicting Causes in Aviation Incidents. Submitted to Modelling.
3. Nanyonga, A., Wasswa, H., Turhan, U., Joiner, K., & Wild, G. Predicting the Indicative Likely Causal Factors of an Aviation Incident Using NLP and Variational Auto-encoder. Submitted to IEEE Access
4. Nanyonga, A., Wasswa, H., Turhan, U., Joiner, K., & Wild, G. Explainable Supervised Learning Models for Aviation Predictions in Australia. Submitted to Journal of Aerospace
5. Nanyonga, A., Joiner, K., Turhan, U., & Wild, G. (2025). Applications of natural language processing in aviation safety: A review and qualitative analysis. In AIAA SCITECH 2025 Forum (p. 2153).
6. Nanyonga, A., Wasswa, H., Molloy, O., Turhan, U., & Wild, G. (2023, September). Natural language processing and deep learning models to classify phase of flight in aviation safety occurrences. In 2023 IEEE Region 10 Symposium (TENSYMP) (pp. 1-6). IEEE.
7. Nanyonga, A., Wasswa, H., Turhan, U., Molloy, O., & Wild, G. (2023). Sequential classification of aviation safety occurrences with natural language processing. In AIAA AVIATION 2023 Forum (p. 4325).
8. Nanyonga, A., Wasswa, H., & Wild, G. (2024, March). Comparative Study of Deep Learning Architectures for Textual Damage Level Classification. In 2024 11th International Conference on Signal Processing and Integrated Networks (SPIN) (pp. 421-426). IEEE.
9. Nanyonga, A., & Wild, G. (2023, December). Impact of Dataset Size & Data Source on Aviation Safety Incident Prediction Models with Natural Language Processing. In 2023 Global Conference on Information Technologies and Communications (GCITC) (pp. 1-7). IEEE
10. Nanyonga, A., Wasswa, H., & Wild, G. (2023, December). Phase of flight classification in aviation safety using lstm, gru, and bilstm: a case study with asn dataset. In 2023 International Conference on High Performance Big Data and Intelligent Systems (HDIS) (pp. 24-28). IEEE.

# List of Common Acronyms

- AI – Artificial Intelligence
- ASN – Aviation Safety Network
- ASRS – Aviation Safety Reporting System
- ATSB – Australian Transport Safety Bureau
- ATM – Air Traffic Management
- BERT – Bidirectional Encoder Representations from Transformers
- BLEU – Bilingual Evaluation Understudy
- BLSTM – Bidirectional Long Short-Term Memory
- CNN – Convolutional Neural Network
- CVR – Cockpit Voice Recorder
- EASA – European Union Aviation Safety Agency
- FAA – Federal Aviation Administration
- FDR – Flight Data Recorder
- GA – General Aviation
- GloVe – Global Word Vectors
- GRU – Gated Recurrent Unit
- IATA – International Air Transport Association
- ICAO – International Civil Aviation Organization
- KNN – K-Nearest Neighbors

- LDA – Latent Dirichlet Allocation
- LIME – Local Interpretable Model-agnostic Explanation
- LR – Logistic Regression
- LSA – Latent Semantic Analysis
- LSTM – Long Short-Term Memory
- ML – Machine Learning
- NB – Naive Bayes
- NLP – Natural Language Processing
- NMF – Non-negative Matrix Factorization
- NN – Neural Networks
- NTSB – National Transportation Safety Board
- PCA – Principal Component Analysis
- RF – Random Forest
- RNN – Recurrent Neural Networks
- ROUGE – Recall-Oriented Understudy for Gisting Evaluation
- Seq2Seq – Sequence to Sequence
- SHAP – SHapley Additive exPlanations
- SLR – Systematic Literature Review
- SRNN – Simple Recurrent Neural Network
- SVM – Support Vector Machine
- VAEs – Variational Autoencoders
- XAI – Explainable AI

# Section I: General Introduction and Literature Review

## Chapter 1. Introduction

This chapter provides an overview of the research, outlining its overarching aim and the relevant background. The significance and motivation behind the study are discussed, highlighting the need for advancements in aviation safety. Additionally, the research questions and objectives are presented, establishing the foundation for the study. The chapter concludes with a discussion of the structure of the thesis and the published papers that contribute to its findings.

### 1.1 Background

Aviation safety remains a critical concern in the air transport industry, where accidents and incidents can result in devastating consequences, including loss of life, financial implications, and reputational damage [1-4]. Traditional safety approaches, which rely heavily on historical data analysis, expert assessments, and regulatory compliance, have limitations such as human biases, delays in identifying risks, and challenges associated with large-scale data management [5, 6]. The integration of advanced artificial intelligence (AI) and machine learning (ML) techniques presents new opportunities for improving aviation safety by enabling automated risk assessments, predictive analytics, and real-time safety monitoring [610].

Natural language processing (NLP)[11] has emerged as a pivotal tool in extracting meaningful insights from unstructured textual data, particularly in aviation safety reports. AI-driven methodologies, such as transformers, variational autoencoders (VAEs) [12], and explainable AI (XAI) [13] frameworks, offer a promising approach for gaining deeper insights into incident causes, improving predictive modelling, and enhancing transparency in AI-driven safety assessments. Despite these advancements, challenges persist in integrating AI into safety-critical applications, particularly concerning model interpretability, dataset imbalances, and trust in AI-generated predictions [14].

In particular, topic modelling and deep learning techniques have gained traction in aviation safety research due to their ability to uncover hidden patterns in large volumes of incident reports. Topic modelling techniques, such as Probabilistic Latent Semantic Analysis (pLSA) and BERTopic, facilitate the automatic discovery of key themes within aviation safety narratives, providing valuable insights into recurrent incident causes and emerging safety concerns. Meanwhile, deep learning models, including recurrent neural networks (RNNs), transformers, and convolutional neural networks (CNNs), have demonstrated significant potential in classifying aviation safety occurrences, predicting risk factors, and automating incident analysis [15].

The investigation of aviation incidents remains a time-intensive and resource-demanding task, requiring expert analysis of structured reports and textual narratives. Although ML and AI models have shown promise in automating and enhancing safety assessments, their adoption in aviation safety remains limited due to concerns about model interpretability, predictive accuracy, and data imbalance. Current ML approaches predominantly focus on structured datasets, leaving a gap in effectively processing and deriving insights from unstructured textual narratives. Moreover, the lack of transparency in existing models further inhibits their practical implementation in real-world decision-making [16-19].

Given these challenges, this research will apply predictive models to identify incident causes, optimize AI transparency, and validate model effectiveness on real-world aviation safety datasets. The findings will contribute to the development of interpretable, data-driven safety assessment tools that support proactive risk mitigation and enhance aviation safety management.

## 1.2 Significance

Aviation safety is a critical concern that requires continuous improvement to mitigate risks associated with aircraft operations. Despite the extensive data collected from aviation incidents, leveraging this information effectively remains a challenge. This research contributes significantly to aviation safety by integrating machine learning (ML) and natural language processing (NLP) techniques to enhance aviation safety models' predictive capabilities and interpretability.

The significance of this study is threefold. First, it addresses the limitations of traditional aviation safety analysis methods, which rely heavily on manual assessments and structured data processing. By incorporating transformer-based models, variational autoencoders (VAEs), and explainable AI (XAI) techniques, this research advances automated incident cause identification and predictive modelling, improving aviation risk assessment accuracy [20].

Second, this study bridges the gap in analyzing unstructured textual narratives, a largely underutilized data source in aviation safety research. Most existing ML applications focus on structured datasets, limiting insights derived from pilot reports, air traffic control communications, and accident investigation summaries. This research employs NLP-driven techniques to extract meaningful information from unstructured aviation safety reports, thereby providing deeper insights into the root causes of incidents.

Third, the findings of this study offer practical implications for aviation stakeholders, including airlines, regulatory bodies, and accident investigators. The development of interpretable and predictive ML models enhances transparency and trust in AI-driven safety assessments. These models can be integrated into safety management systems (SMS) to support real-time monitoring, proactive risk mitigation, and datadriven decision-making. Ultimately, this study contributes to the ongoing efforts to improve aviation safety, reduce accident occurrences, and ensure the well-being of passengers and crew members.

## 1.3 Motivations

The motivation for this research stems from the growing need to enhance aviation safety using advanced data-driven approaches. The aviation industry generates vast amounts of data daily, ranging from sensor readings and flight logs to pilot and crew reports. However, despite the availability of extensive safety data, its potential remains largely untapped due to the complexity of processing heterogeneous and unstructured information. This research is driven by the need to leverage AI and ML technologies to unlock valuable insights from aviation safety data.

Several key factors motivated this study. First, the increasing global air traffic volume necessitates the adoption of more efficient and accurate safety monitoring techniques. With air travel expected to continue growing, traditional safety analysis methods may become inadequate in handling the escalating volume of incident reports. ML and NLP offer scalable solutions to automate safety assessments and identify patterns in aviation incident data more effectively.

Second, recent advancements in AI have demonstrated the potential of ML models in safety-critical applications, such as autonomous vehicles, medical diagnostics, and industrial automation. The successful implementation of AI-driven predictive analytics in these fields suggests that similar approaches can be applied to aviation safety. By integrating state-of-the-art AI techniques, this research aims to enhance aviation risk assessment methodologies and improve the reliability of safety predictions.

Third, the lack of interpretability in existing ML models has hindered their adoption in high-stakes industries like aviation. Black-box models raise concerns regarding trust, regulatory compliance, and decision-making accountability. This research is motivated by the need to develop transparent, explainable AI models that aviation stakeholders can trust. By incorporating SHapley Additive exPlanations (SHAP) and other explainability techniques, this study enhances the interpretability of ML-driven aviation safety assessments, ensuring their practical applicability in real-world decision-making.

Finally, this research is inspired by the broader goal of reducing aviation accidents and saving lives. By developing more accurate, interpretable, and data-driven safety models, this study contributes to making air travel safer for passengers and crew members worldwide. The findings of this research have the potential to inform regulatory policies, enhance incident investigation processes, and support the aviation industry's commitment to continuous safety improvement

## 1.4 Research Aim and Objectives

This thesis aims to apply advanced ML techniques specifically, transformers, VAEs, and explainable AI frameworks to improve aviation safety analysis by addressing critical challenges such as predictive accuracy, model interpretability, and the utilization of under-explored models in the aviation sector. This research seeks to:

1. Conduct a systematic literature review to evaluate the current state of ML applications in aviation safety and identify gaps in existing research.
2. Apply transformer-based models to predict probable causes of aviation incidents from textual narratives.
3. Utilize VAEs to improve the prediction of causal factors from aviation incident reports.
4. Incorporate explainable AI methods, such as SHapley Additive exPlanations (SHAP), to improve model transparency and enhance trust in the results.
5. Assess the effectiveness of these models using real-world aviation safety datasets, such as those from the NTSB and ATSB

## 1.5 Research Questions

The overarching research question of this thesis is:

How can machine learning (ML) and natural language processing (NLP) techniques be leveraged to enhance aviation safety by predicting incident causes and improving interpretability in aviation safety models?

To address this question, the following secondary research questions are formulated:

1. What are the current trends, limitations, and future directions of ML applications in aviation safety research? (Addressed in Chapter 2)
2. How can transformer-based models be leveraged to improve the predictive accuracy of aviation incident cause analysis from textual narratives? (Addressed in Chapter 4).
3. How can variational autoencoders (VAEs) and NLP techniques be used to infer causal factors from aviation incident reports? (Addressed in Chapter 5)
4. How can explainable AI methods enhance the interpretability and trustworthiness of ML models used in aviation safety predictions? (Addressed in Chapter 6)

## 1.6 Thesis Structure

This thesis comprises twelve conference papers and four journal articles, systematically organized into four sections, each contributing to different aspects of AI-driven aviation safety analysis.

Section I: Literature Review (Chapters 2–3). This section provides a comprehensive review of ML applications in aviation safety, highlighting key methodologies, research gaps, and challenges.

Chapter 2: A review paper, currently under consideration for publication in Transportation Research Interdisciplinary Perspectives, that examines the state of ML applications in aviation safety. This chapter identifies research gaps and limitations within existing studies, offering a foundation for the subsequent contributions of this thesis. Chapter 3: A conference paper published in the AIAA SciTech 2025 Forum (p.

2153), which explores the use of NLP in aviation safety. The study conducts a systematic analysis of 34 Scopus-indexed publications on NLP applications in aviation safety as of May 2024. Through qualitative and quantitative evaluations, it identifies key trends, research motivations, and challenges, such as the need for large, annotated datasets and the interpretability of complex models. The chapter further proposes solutions, including active learning for data annotation and explainable AI for model transparency, and presents case studies demonstrating the successful application of NLP in enhancing aviation safety.

Section II: Core Contributions – Machine Learning and NLP for Aviation Safety (Chapters 4–6). This section presents the primary research contributions, focusing on the application of machine learning and NLP techniques for aviation safety analysis. Chapter 4: A study on the application of variational autoencoders (VAEs) for causal factor inference. This research addresses the challenge of determining likely causes of aviation incidents by deploying NLP techniques to extract numerical encodings of textual narratives. Chapter 5: A study on the integration of explainable AI (XAI) techniques to improve model transparency and trustworthiness. Despite AI's success across various industries, its adoption in aviation safety remains limited due to the opacity of black-box models. This paper has been accepted for publication in the Aerospace Journal and is currently undergoing revisions. Chapter 6: A study on the application of transformer-based models for incident cause prediction. This research investigates the effectiveness of transformer architectures in analyzing aviation safety narratives and predicting risk factors with improved accuracy and contextual understanding. This section includes three journal papers, all currently in the publication process.

Section III: Deep Learning for Aviation Safety (Chapters 7–13). This section focuses on deep learning applications across various aviation safety datasets, presenting six conference papers published between 2023 and 2025. Chapter 7: A conference paper presented at IEEE TENSYMP 2023, where the research was awarded Best Student Paper. This study explores the application of deep learning techniques in aviation safety analysis, demonstrating their effectiveness in processing and classifying safety incidents. Chapters 8–13: A collection of additional conference papers that examine various deep learning methodologies for aviation safety data analysis.

Section IV: Topic Modeling for Aviation Safety (Chapters 14–18). This section presents topic modelling techniques applied to aviation safety datasets, utilizing methods such as probabilistic latent semantic analysis (pLSA) and BERTopic to extract hidden patterns from aviation safety reports. These chapters contribute novel insights into automated safety analysis by identifying recurring themes and risk factors in unstructured textual narratives.

Section V: Synthesis, Implications, and Future Research (Chapter 19). The final section synthesizes key findings from previous chapters, discusses their implications for aviation safety, and provides recommendations for future research. It offers a critical reflection on the limitations of current

AI-driven approaches and suggests avenues for further exploration in the field of aviation safety analytics.

Each chapter, except for Chapter 1 (Introduction) and Chapter 19 (Conclusion), is structured as an individual publication, ensuring that each research contribution stands independently while maintaining coherence with the overarching theme of AI-driven aviation safety analysis. All references are compiled in the bibliography section at the end of the thesis.

# Chapter 2. A Systematic Review of Machine Learning Analytic Methods for Aviation Accident Research

The contents of this chapter, with only minor modifications, have been submitted and are currently under review in the Journal of Transportation Research: Interdisciplinary Perspectives. The recommended citation for the original publication is:

Nanyonga, A., Turhan, U., Joiner, K., & Wild, G. (2024). A Systematic Review of Machine Learning Analytic Methods for Aviation Accident Research. Journal of Transportation Research: Interdisciplinary Perspectives.

## 2.1 Abstract

The aviation industry prioritizes safety and has embraced innovative approaches for both reactive and proactive safety measures. Machine learning (ML) has emerged as a useful tool for aviation safety. This systematic literature review explores ML applications for safety within the aviation industry over the past 25 years. Through a comprehensive search on Scopus and backward reference searches via Google Scholar, 87 of the most relevant papers were identified. The investigation focused on the application context, ML techniques employed, data sources, and the implications of contextual nuances for safety analysis outcomes. ML techniques have been effective for post-accident analysis, predictive modelling, and real-time incident detection across diverse aviation scenarios. Supervised, unsupervised, and semisupervised learning methods, including neural networks, decision trees, support vector machines, and deep learning models, have all been applied for analyzing accidents, identifying patterns, and forecasting potential incidents. Notably, data sources such as the Aviation Safety Reporting System (ASRS) and the National Transportation Safety Board (NTSB) datasets were the most used. Transparency, fairness, and bias mitigation emerge as critical factors that shape the credibility and acceptance of ML-based safety research in aviation. The review revealed seven future research directions recommended: 1) interpretable AI, 2) real-time prediction, 3) hybrid models, 4) handling of unbalanced datasets, 5) privacy and data security, 6) human-machine interface for safety professionals, and 7) regulatory implications. These directions provide a blueprint for further ML-based aviation safety research. This review underscores the role of ML applications in shaping aviation safety practices, thereby enhancing safety for all stakeholders. It serves as a constructive and cautionary guide for researchers, practitioners, and decision-makers, emphasizing the value of ML when used appropriately to transform aviation safety to be more data-driven and proactive.

## 2.2 Introduction

Air transport stands at the Centre of modern global connectivity, revolutionizing the way people traverse the World and facilitating rapid travel [21]. However, within this critical industry, the paramount concern is ensuring safety. The global aviation industry's complex procedures and high-stakes operations leave little margin for operational errors. Even the slightest mistake or system misconfiguration can lead to catastrophic accidents, with significant loss of life, potential property damage, substantial financial impacts, and eroded customer trust [22]. The aviation industry has consistently demonstrated a comparatively remarkable commitment to safety, continually evolving and embracing technological advancements and comprehensive protocols. This dedication is particularly evident in the aftermath of accidents, where in-depth post-accident analyses play a pivotal role in understanding the root causes, mitigating future risks, and fortifying safety measures [23]. Government bodies like the International Civil Aviation Organization (ICAO), the European Union Aviation Safety Agency (EASA), and the U.S. Federal Aviation Administration (FAA) establish and promote aviation safety standards. Their guidelines, protocols, and procedures safeguard passengers, crew, and aircraft operations, forming the foundation of aviation safety.

The journey towards enhanced safety within the aviation sector is punctuated by significant milestones, each shaped by lessons derived from post-accident analyses of major (and sometimes minor) aviation accidents. These analyses entail multifaceted collaborations between regulatory bodies, aircraft manufacturers, airlines, aviation experts, and investigators [24]. While immediate safety occurrence causes are quickly examined, the analyses routinely extend beyond, aiming to uncover contributing factors and systemic issues. This approach has fostered advancements spanning aircraft design, maintenance engineering, air traffic management, crew training, and human factors analysis [25]. Addressing these challenges necessitates collaborative efforts. For example, NASA's Aviation Safety Reporting System (ASRS) collects safety occurrence reports voluntarily submitted by aviation professionals, offering insights into potential safety concerns and proactive accident prevention (ASRS program briefing, "https://asrs.arc.nasa.gov/overview/summary.html") [26]. Additionally, the FAA and NTSB play integral roles in investigating and enhancing aviation safety standards. While the FAA enforces regulations to maintain airspace safety, the NTSB investigates accidents, supplying invaluable data for safety improvement [27]. Collectively, these efforts have substantially reduced the rate of accidents [28]. The industry's commitment to continuous improvement reflects its determination to uphold a high safety record despite inherent risks.

The aviation sector's recent surge in growth fueled by escalating demand for air travel and infrastructure constraints, has ushered in both opportunities and challenges [29]. The International Air Transport Association (IATA) forecasts global air travel demand doubling within the next two decades [30]. While this growth promises economic potential, it also increases pressure on existing infrastructure, underscoring safety as a paramount concern. IATA, a significant industry body, plays a pivotal role in advocating industry initiatives and standards. However, concerns have arisen regarding its representation of smaller airlines and potential conflicts of interest among its members. Increasing air traffic congestion and the strain on Air Traffic Management (ATM) systems further escalates safety risks where infrastructure developments often lag demand [31]. Considering aviation's recent growth surge, various approaches are being researched to bolster safety. These range from investigative methods that analyze data from accidents, incidents, and near misses to predictive strategies that proactively identify potential risk factors before accidents occur. Regulatory bodies and airlines maintain extensive repositories of data, and researchers diligently analyze these datasets to facilitate proactive measures against potential risks [32].

According to Annex 13 to the Convention on International Civil Aviation: Aircraft Accident and Incident Investigation (10th ed.) by the International Civil Aviation Organization (ICAO), an "accident" is defined as an occurrence associated with the operation of an aircraft that takes place between the time any person boards the aircraft with the intention of flight until all such persons have disembarked, where a person is fatally or seriously injured, the aircraft sustains damage or structural failure, or the aircraft is missing or inaccessible. On the other hand, an "incident" is defined as an occurrence other than an accident, associated with the operation of an aircraft, which affects or could affect the safety of operations. An "occurrence" is an accident, incident or any other operational issue that could impact aviation safety. These definitions provided by ICAO serve as fundamental frameworks for the investigation and analysis of aviation-related events, guiding safety protocols and procedures within the aviation sector [33].

Machine Learning (ML) has emerged as a transformative force in data analysis. A subset of Artificial Intelligence (AI), ML aims to equip systems to learn from data, make informed decisions, and detect patterns without explicit programming. With the patterns and process of vast datasets and employing advanced algorithms, ML offers substantial new potential to predict anomalies, extract patterns, and identify factors contributing to events [34]. As aviation continues to evolve, integrating ML into aviation safety occurrence investigation and analysis should help sustain or improve safety practices and operational procedures.

This systematic literature review explores ML's role in aviation safety, with a focus on post-accident analysis. Employing a systematic approach, the paper categorizes and synthesizes primary studies to address well-defined research questions. By analyzing 87 carefully selected studies from a pool of 3832 papers, this review aims to explore the state-of-the-art in the domain of ML for aviation accident analysis. Through the synthesis of these studies, this paper serves as a foundational resource for researchers, aviation stakeholders, and regulatory bodies seeking to harness the potential of machine learning to bolster safety practices. Formally, the primary objective of this systematic literature review (SLR) was to gather, evaluate, and synthesize existing research studies that use ML applications in post-accident analysis within the aviation sector. The associated research questions to be answered were:

1. How do variations in the demographic factors such as aviation application, electronic database, publication type, study year, trends, data sources, and authors influence the comprehensiveness, reliability, and applicability of findings in aviation accident research?

2. What types of machine learning approaches have been applied in post-accident analysis within the aviation industry?

    a. What specific machine learning types have been employed in this context?
    b. Which machine learning tasks have been targeted to enhance post-accident analysis?
    c. What are the predominant machine learning algorithms used in the aviation industry for postaccident analysis?

3. How have these machine learning techniques contributed to enhancing safety measures and providing insights into aviation accidents?

In the subsequent sections, the background and related work (Section 2) are covered, the methodology used in this systematic review (Section 3) is described, the results are presented and discussed (Sections 4 and 5) and conclusions based on the findings are presented (Section 6). This review aims to be a benchmark overview of machine learning applications in post-accident analysis within the aviation sector and identify associated opportunities for future research.

## 2.3 Background

In this section, the fundamental concepts that underpin the exploration of machine-learning techniques in the aviation sector are examined. It is essential to understand the difference between machine learning (ML) and artificial intelligence (AI) and to have an awareness of the various types of ML and the tasks they encompass.

### *2.3.1 Artificial intelligence*

Artificial Intelligence (AI) is the field dedicated to the study of intelligent agents, encompassing machines that mimic human-like behaviours in learning and problem-solving processes [35]. Utilizing computerprocessing techniques, AI enables machines to learn, perceive, and process natural language, while also making decisions in ways reminiscent of human cognitive processes [36]. This transformative technology excels at processing substantial volumes of complex data and performing real-time decision-making, thus finding applications in a multitude of management domains.

Within the realm of AI, one prominent subset is machine learning (ML) because it plays a pivotal role in various industries, including aviation. The different types of ML and the diverse tasks they can accomplish will be explored in the following sections. With a clear understanding of these concepts, an exploration into the integration of ML techniques to improve aviation safety and operational efficiency becomes feasible.

#### *2.3.2 Machine Learning*

Machine learning (ML) empowers computer programs to execute intricate tasks like prediction, diagnosis, planning, and recognition by assimilating insights from historical data. The synergy between data and algorithms is paramount for the proficiency of ML models. Enhanced accuracy often results from highquality data and large datasets, so long as they are appropriately balanced to the key factors and the learning cases are appropriately sequenced [37, 38]. Equally crucial is the selection of suitable algorithms tailored to the specific problems at hand, accommodating diverse types of datasets [35, 39].

#### *2.3.3 Machine Learning Types*

ML encompasses several types, each with its unique characteristics and applications. These types enable the creation of models that can adapt and learn from data, making them versatile tools for addressing various challenges. These types are depicted in Figure 1. The following ML types are explored:

1. Supervised Learning: In supervised learning, models are trained on labelled datasets, where input data is associated with corresponding desired outputs. The model learns the relationship between inputs and outputs, enabling it to make predictions on new, unseen data. Supervised learning is widely employed in tasks such as classification, where the goal is to assign input data to predefined categories, and regression, where continuous values are predicted based on inputs [40].

2. Unsupervised Learning: Unsupervised learning involves training models on unlabelled data, to discover patterns and structures within the data. Clustering is a common task in unsupervised learning, where similar data points are grouped. Dimensionality reduction is another application, simplifying the data while preserving its key characteristics [41, 42].

3. Semi-Supervised Learning: This type of learning combines labelled and unlabelled data to enhance model performance. Often, labelled data is scarce, but unlabelled data is more abundant. By leveraging both types of data, models can generalize better and make more accurate predictions [43].

4. Reinforcement Learning: Reinforcement learning is rooted in the concept of learning through interactions, often simulation of varying fidelity and often in environments with significant variability and uncertainty [44]. An algorithm interacts with a system and optimizes variables to maximize cumulative rewards based on an objective function and through trial and error, the system learns optimal strategies to achieve its goals [44].

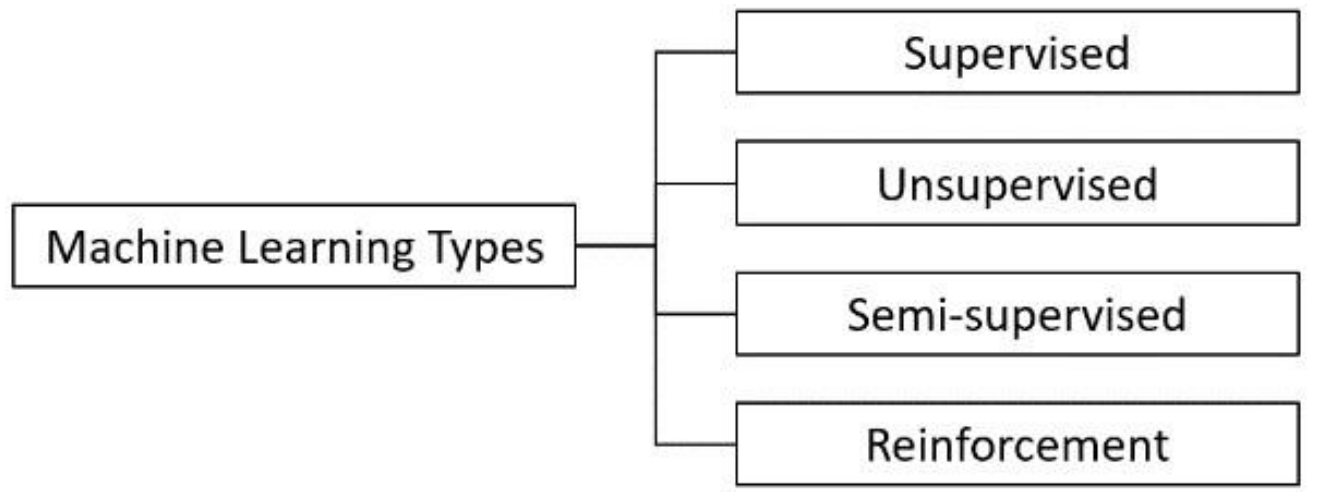

Figure 1. The four most common board machine-learning types

*2.3.4 Machine Learning Tasks*

ML encompasses a diverse array of tasks, each serving a specific purpose and leveraging data-driven techniques to extract valuable insights. These tasks enable the development of models that can autonomously make decisions, predictions, and classifications as shown in Figure 2. The following six common ML tasks are explicated:

1. Classification: Classification involves assigning input data to predefined categories or classes. The model learns patterns from labelled training data, enabling it to classify new, unseen data accurately. Applications range from email spam detection to medical diagnosis [40].

2. Regression: Regression is concerned with predicting continuous values based on input features. The model learns the relationships between variables from training data and can then make predictions on new data points. Examples include predicting housing prices or stock market trends [45].

3. Clustering: Clustering involves grouping similar data points based on inherent patterns in the data. The model identifies clusters or segments within the data without requiring predefined categories. This task finds applications in customer segmentation and anomaly detection [46].

4. Anomaly Detection: Anomaly detection focuses on identifying rare or unusual instances within a dataset. The model learns the normal patterns and detects deviations from these patterns, making it valuable for fraud detection and network security [27, 47].

5. Data Reduction: Data reduction involves techniques to reduce the complexity of large datasets while retaining crucial information. These techniques help minimize computational overhead, improve model efficiency, and avoid overfitting. Methods like Principal Component Analysis (PCA) and feature selection are commonly used for data reduction [42].

6. Natural Language Processing (NLP): NLP is a specialized task involving the interaction between computers and human language. It encompasses various sub-tasks such as sentiment analysis, text generation, and language translation. NLP enables machines to understand, interpret, and generate human language, with applications spanning from chatbots to language translation services [48, 49]. Because of its role in communicating, NLP is often a necessary sequential addition to enable the other listed ML tasks.

Due to the diverse ontology at play in each domain, it is often necessary to systematically tailor NLP to new domains like extending a dictionary or lexicon [50, 51].

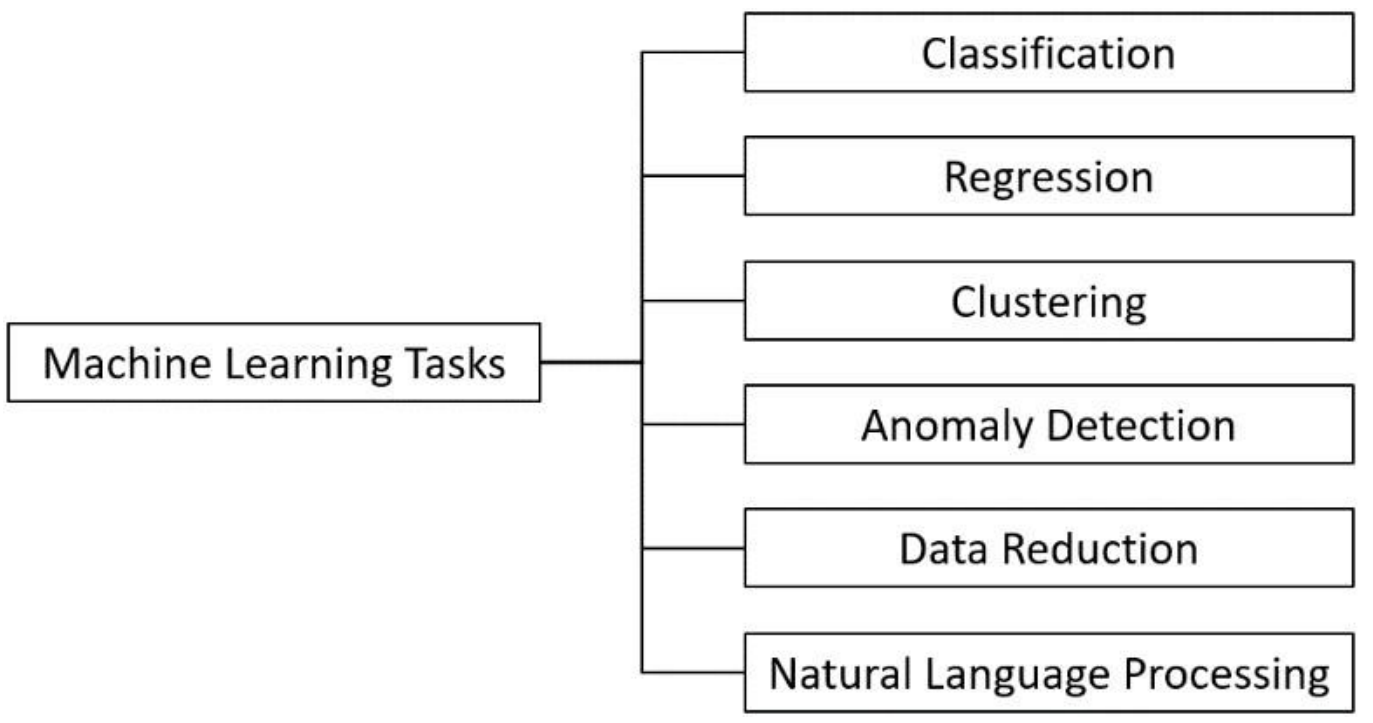

Figure 2. The six most common machine learning tasks

*2.3.5 Machine Learning Techniques in Aviation Safety*

ML techniques enable the analysis of vast aviation datasets to uncover patterns, make predictions, and extract valuable information. Below are some explanations of key techniques along with notable studies that have employed specific ML techniques in aviation safety research:

1. Support Vector Machine (SVM). SVM is a powerful ML technique commonly used for classification and regression tasks. It works by finding the optimal hyperplane that maximally separates data points belonging to different classes. This hyperplane is selected in such a way that the margin between the classes is maximized, allowing for the effective classification of new, unseen clusters. In the context of aviation safety analysis, SVM has proven to be a valuable tool for categorizing aviation accidents based on their characteristics, such as severity, causal factors, and contributing variables [52]. In that research, SVM was used alongside other techniques such as Random Forest and Naive Bayes to classify aviation accidents over 20 years using the Aviation Safety Reporting System (ASRS) dataset.

2. Decision Tree: This is a powerful tool used for both classification and regression tasks in ML. It takes the form of a tree-like structure where each internal node represents a decision based on a feature, each branch represents an outcome of that decision, and each leaf node represents a classification label or a predicted value. Decision Trees have found applications in aviation safety and analysis due to their ability to handle complex decision-making processes and their transparency. For instance, [53, 54] utilized Decision Trees, along with other ML algorithms, to classify and reduce data from airline databases spanning 42 years. This research enabled the classification of aircraft accident data into different categories, contributing to insights into warning levels in accidents. Also, another study extended the use of Decision Tree in the aviation safety domain by applying them in conjunction with Natural Language Processing (NLP) techniques. Their study involved extracting word-level meaning from safety report narratives and using these meanings to analyze a large dataset of 186,000 reports over 39 years. The combination of NLP and Decision Trees allowed for the extraction of valuable insights from textual data, contributing to improved safety analysis and risk assessment in the aviation industry [55].

3. K-Nearest Neighbors (KNN): KNN algorithm is a fundamental ML technique used for classification and regression tasks. It is a non-parametric, instance-based learning method that makes predictions based on the similarity between input data points. KNN is intuitive and easy to understand, making it a popular choice for beginners in ML, noting it leverages the ML to distinguish nearest Neighbor in often very multifactor dimensional space (10 or more). The algorithm operates on the principle that similar data points tend to share common characteristics and attributes. The KNN algorithm's performance depends on the choice of the parameter "k" and the distance metric used to measure similarity. A smaller "k" value can lead to a noisy prediction, while a larger "k" value can result in a smoother but potentially biased prediction. Additionally, selecting an appropriate distance metric is crucial, as it affects how the algorithm measures similarity between data points. Koteeswaran et al., [56] employed the KNN technique as part of their study on predicting the topmost causes of aviation accidents using data mining algorithms. In this research, along with other ML algorithms like Naive Bayes (NB) and Decision Trees (DT), KNN was used to classify aviation accidents spanning 95 years using data from the Federal Aviation Administration (FAA). By applying KNN to the FAA dataset, the study identified commonalities and patterns in accidents that could contribute to better accident prevention strategies and safety measures.

4. Neural Networks: This is a class of ML algorithms inspired by the structure and functioning of the human brain's NN (Neural Networks). They are designed to process and recognize patterns in data, making them highly suitable for tasks involving complex relationships and non-linear interactions. The determinations from neural networks can be difficult to trace, even when they satisfy objective functions and outperform baseline performance of humans or other more explainable AI [57]. NNs consist of interconnected nodes, or artificial neurons, organized in layers, each responsible for specific computations. In the context of aviation safety research, [58] harnessed the power of NNs to address a critical task in aviation safety management systems. Their study involved classifying risk factors within the aviation domain using data obtained from the Aviation Safety Reporting System (ASRS) over 24 years. Also, the study conducted by [59]involved the application of NN to predict Human Factors Analysis and Classification System (HFACS) unsafe acts based on the pre-conditions of those unsafe acts.

5. Naive Bayes (NB): This technique is a probabilistic ML algorithm rooted in Bayes' theorem. It is particularly well-suited for classification tasks, where the goal is to assign predefined categories or labels to input data based on observed features. Naive Bayes assumes feature independence, which simplifies calculations and makes it a relatively efficient algorithm for text classification and other categorical data [60]. In their study, Koteeswaran et al., [56] used NB to predict the topmost cause of accidents in aviation as a data mining technique. This technique was compared to other techniques such as SVM and KNN explained above and in this instance was the most effective when for their classification tasks.

6. Random Forest (RF): RF “combines several randomized decision trees and aggregates their predictions by averaging” [61]. It is particularly effective for classification and regression tasks. RF is based on the concept of creating multiple decision trees during the training phase and combining their predictions to improve accuracy. Each decision tree is constructed using a random subset of the training data and a random subset of features, which helps to introduce diversity and robustness to the model [62]. Zhang & Mahadevan, [30] utilized the Random Forest technique along with Deep Neural Networks (DNN) and SVM for Natural Language Processing (NLP) and classification of aviation incident reports from the Aviation Safety Reporting System (ASRS) dataset. The study focused on analyzing and classifying aviation incident reports spanning an 11-year period. By incorporating RF in their analysis, the researchers leveraged its ensemble capabilities to improve the accuracy and reliability of their classification model, ultimately enhancing the understanding of safety incidents in aviation.

7. Latent Dirichlet Allocation (LDA): LDA is a widely used ML technique for topic modeling and document clustering. It's particularly applicable to textual data analysis, such as the analysis of aviation accident reports [63, 64]. LDA assumes that each document is a mixture of a small number of topics or themes, and each topic is characterized by a distribution of words. The goal of LDA is to uncover these hidden topics and their associated word distributions from a collection of documents. In [65] study, "Text Mining Classification and Prediction of Aviation Accidents Based on TF-IDF-SVR Method," various machine learning techniques, including LDA, SVM, NB, RF, and LR, are employed to analyze 20,000 aviation accidents spanning 59 years from NTSB data. LDA, a natural language processing technique, aids in uncovering underlying topics within accident reports, contributing to NLP-driven classification efforts for improved accident prediction and prevention.

8. Long Short-Term Memory (LSTM): LSTM is a type of Recurrent Neural Network (RNN) architecture that is well-suited for processing sequences of data, such as time series or sequences of text. It is particularly effective in capturing long-range dependencies and patterns within sequential data due to its ability to maintain and update information over extended sequences. This makes LSTM suitable for tasks that involve sequential data, where past information can significantly impact future predictions and potentially where variability or uncertainty tends to obfuscate or confuse classical algorithms [66]. The research conducted by [67] focuses on the application of Long Short-Term Memory (LSTM) techniques for aviation safety prediction. Specifically, the study employs LSTM with variable selection methods, including LASSO (Least Absolute Shrinkage and Selection Operator), to predict aviation safety-related outcomes. The dataset used in this research is sourced from the Aviation Safety Reporting System (ASRS), which collects and analyzes incident and accident reports from aviation professionals. In the context of aviation safety prediction by Zeng et al., LSTM is used to model and analyze the temporal patterns of safety-related incidents spanning 10 years. The LSTM network is designed to learn from historical data and capture complex relationships between variables, allowing it to make predictions about potential safety outcomes based on past incident reports.

9. Principal Component Analysis (PCA): PCA is a widely used dimensionality reduction technique in ML and data analysis. It aims to transform high-dimensional data into a lower-dimensional space while preserving as much of the original data's variance as possible. This reduction in dimensionality helps in simplifying the dataset and removing redundant or less informative features, making it easier to work with and potentially improving the performance of machine learning algorithms [68]. In the study conducted by İnan & Gökmen İnan, [69], PCA was applied in conjunction with NNs and DTs (Decision Trees) to classify survivor and non-survivor passengers in fatal aviation accidents based on data reduction. The researchers aimed to identify significant patterns and features that could distinguish between passengers who survived accidents and those who did not.

## 2.4 Methodology

This research employed a rigorous Systematic Literature Review (SLR) approach, adhering to the comprehensive guidelines established by Kitchenham et al., [70]which are still widely used today [71].

This section provides details of the methodology, encompassing the formulation of research questions, the development of the search strategy, the establishment of study selection criteria, the application of a quality assessment methodology, the data collection procedures, and effective data synthesis techniques. Throughout the research process, the SLR guidelines and protocols outlined by Kitchenham et al., [70] were adhered to, ensuring a systematic and methodical process. Based on the findings by [71], the SLR analysis and reporting was done manually despite the possibility of using AI ML.

Upon defining the research questions, a robust search strategy was designed to retrieve pertinent articles from the Scopus database. Subsequently, specific study selection criteria were applied to the retrieved articles, refining the initial pool into a focused subset for more rigorous evaluation during the quality assessment phase. Details of the search, selection, and analysis (as illustrated in Figure 3) are presented in the following sections. The flow chart for the process is depicted in Figures 4.

### *2.4.1. Search Strategy*

The data source selection and search strategy were designed to ensure the thorough inclusion of pertinent studies. The search process involved several key steps, including database searches, backward search from references, and a structured data extraction process. The initial search commenced with 3832 articles identified. After the removal of duplicates (33 articles) and irrelevant records (3731 articles), a refined dataset was obtained for further analysis. The methodology employed for this systematic literature review (SLR) was designed to ensure a thorough exploration of ML applications in post-accident analysis within the aviation industry. The review's purpose was to gather a diverse range of relevant studies that could enhance the understanding of how ML contributes to aviation safety and insights. The search process encompassed a series of well-defined and structured steps, which are detailed as follows.

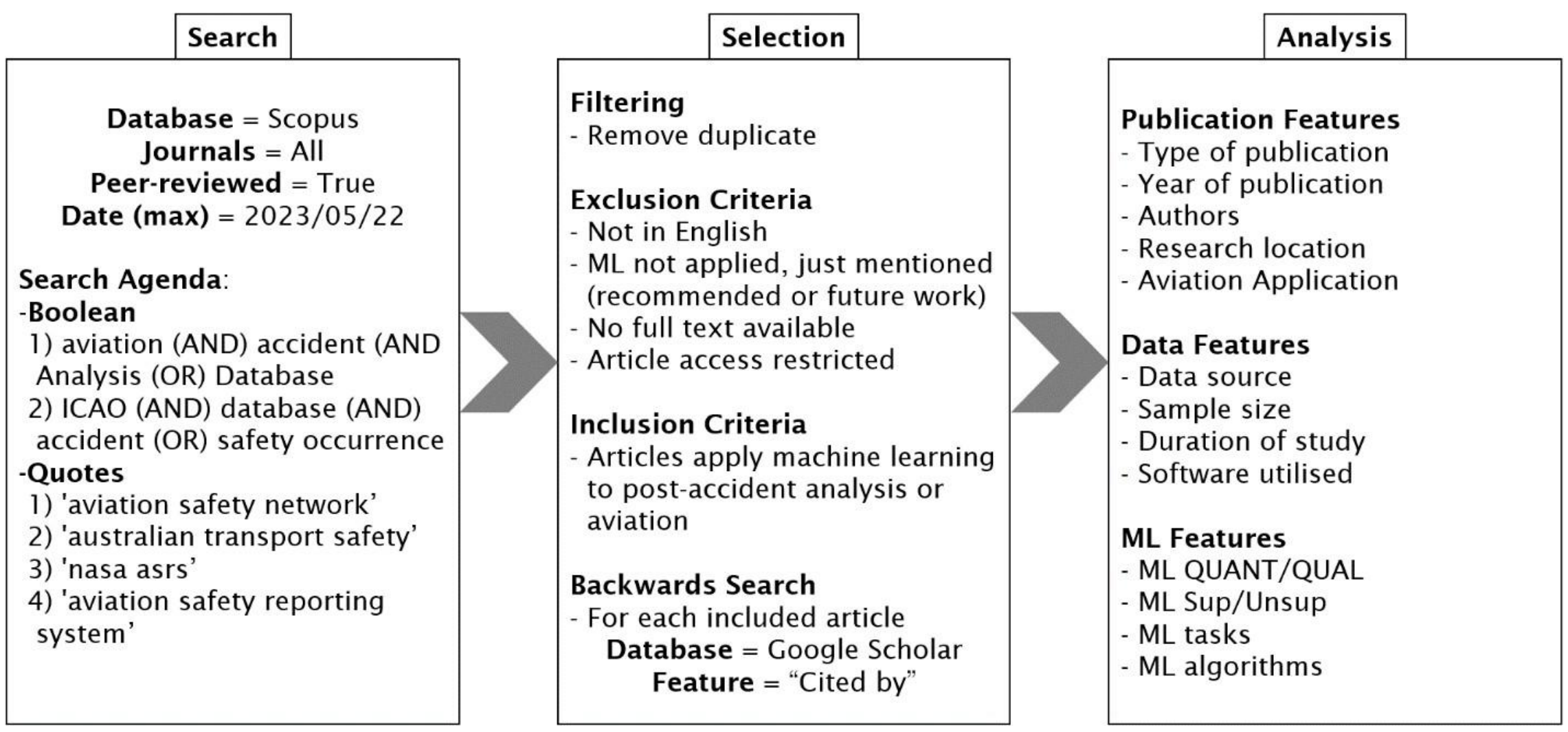

Figure 3. General framework of the process, from search, through selection, to analysis

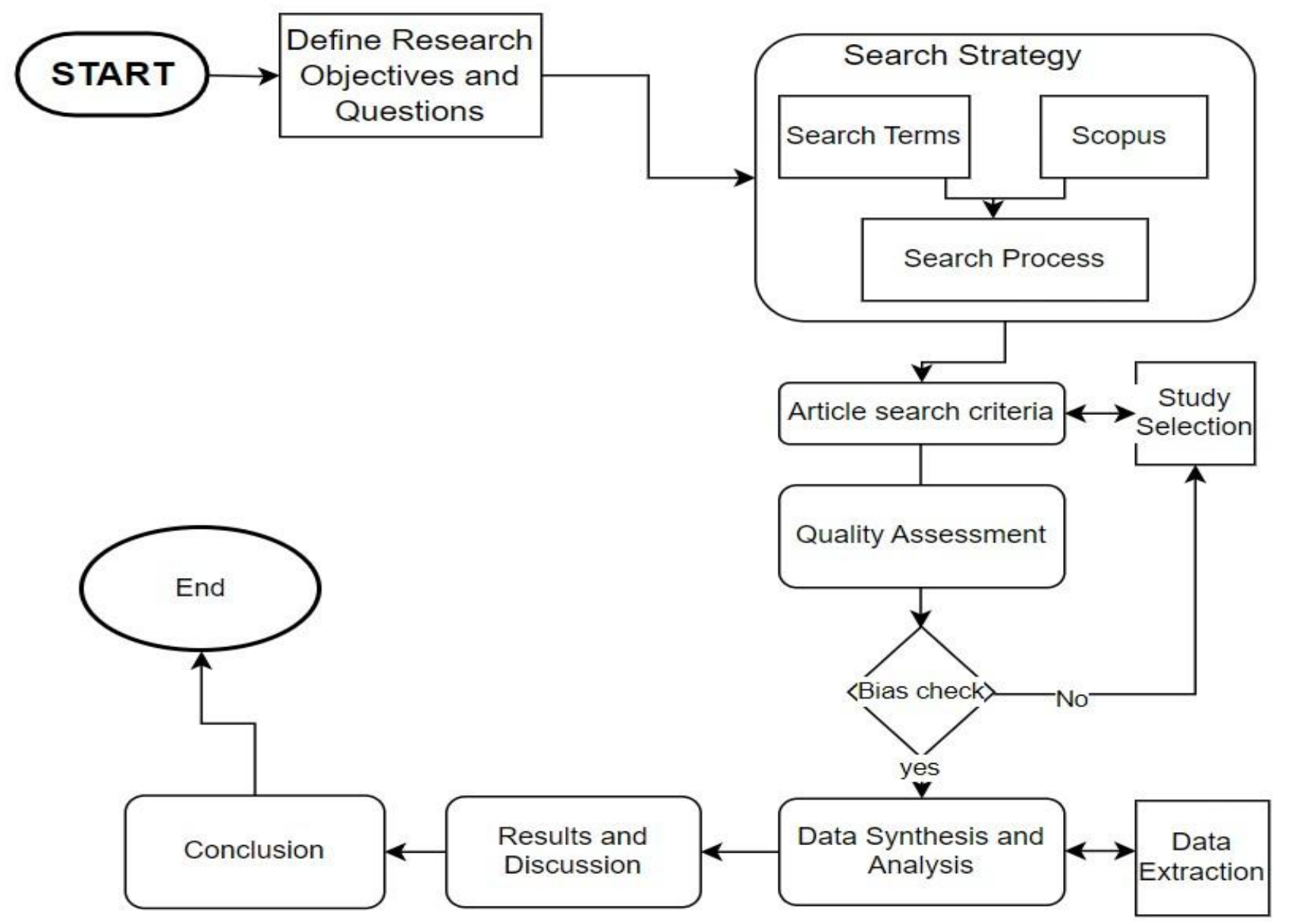

Figure 4. Flowchart showing the entire literature review process followed in this study.

*2.4.2. Database Searches*

The search was primarily conducted in the Scopus database due to its coverage of academic literature across multiple disciplines, making it a valuable resource for conducting systematic literature reviews (SLRs) [72, 73]. This choice is aimed at ensuring the comprehensiveness of the review. The initial search using carefully chosen keywords yielded many articles, which were then refined through a stepwise process. After the removal of irrelevant articles (3731) and duplicates (33), the number was reduced to 68 articles that met the preliminary criteria for further evaluation.

The database searched included key sources such as "NASA (National Aeronautics and Space Administration) Aviation Safety Reporting System," "ICAO Accident Database," "Accident Safety Network," "NTSB AND Aviation AND Accident AND Database," "ICAO AND Safety AND Occurrence AND Database," and "Australian AND Transport AND Safety". The chosen keywords were designed to capture articles that may reference or discuss information from these sources within broader academic literature. The search criteria were expertly devised using Boolean operators (AND) and quotation marks ("") to precisely refine the search parameters, ensuring a high level of accuracy, as illustrated in Table 1.

*2.4.3. Backward Search*

Beyond the database searches, an additional approach was used to enhance the study's breadth. A backward search strategy was executed by reviewing the references cited within the identified studies. This approach, recommended for SLRs, helps identify relevant articles that may not have been captured by the initial search [70, 74, 75]. By including this strategy, an additional 26 articles were retrieved and added to the review, thereby broadening the scope and depth of analysis, and ensuring a comprehensive coverage of the research landscape as depicted in Table 1.

Table 1. Number of papers initially identified by each search

| Search Criteria | Initial no. | Duplicates | Irrelevant | Final |
|---|---|---|---|---|
| Nasa asrs | 9 | - | 6 | 3 |
| ICAO accident database | 15 | - | 13 | 2 |
| Accident safety network | 8 | - | 7 | 1 |
| NTSB Aviation accident database | 86 | - | 78 | 8 |
| ICAO safety occurrence database | 4 | - | 3 | 1 |
| Australian transport safety | 29 | - | 29 | 0 |
| Aviation Accident Analysis AI | 12 | - | 12 | 0 |
| Aviation safety reporting system | 159 | 2 | 132 | 25 |
| Aviation Accident Analysis ML | 61 | 5 | 44 | 10 |
| Aviation accident database | 457 | 10 | 440 | 7 |
| Aviation accident analysis | 2992 | 16 | 2972 | 4 |
| **Scopus article Total** | **3832** | **33** | **3736** | **61** |
| Backward search (Google Scholar) | - | - | - | 26 |
| **Total** | | | | **87** |

*2.4.4 Selection criteria*

The criteria for inclusion and exclusion were established to ensure the relevance and quality of the selected articles [70]. The selection process involved multiple steps, including the initial screening of abstracts and titles, followed by a more detailed assessment of the full-text articles. The criteria outlined below guided the selection process. • Inclusion Criteria

- o Relevance to Aviation Accidents: Articles must directly address the application of machine learning techniques within the aviation industry.
- o Publication Language: Only articles written in the English language were considered to ensure accessibility for the research team.

- Study Focus: The primary focus of the article should be on accident analysis within the aviation/air transport sector.
- Study Type: Peer-reviewed journal articles, book chapters, theses, and conference proceedings were included to ensure academic rigor.

• Exclusion Criteria

- Non-English Language: Articles published in languages other than English were excluded due to language limitations.
- Irrelevant Focus: Studies unrelated to aviation accident analysis were excluded.
- Non-Aviation Applications: Articles discussing machine learning applications in contexts other than aviation accident analysis were excluded.
- Publication Type: Reports and other non-peer-reviewed sources were excluded to ensure scholarly rigor.
- Duplication: Duplicated records were removed to maintain the uniqueness of the dataset. o Restricted Access: Articles for which full-text content was not accessible due to restrictions were excluded.

#### *2.4.5. Data Extraction, Coding, and Quality*

The data extraction process was conducted employing a structured tool developed through Microsoft Forms. This form encompassed a comprehensive set of 22 questions, addressing the selection criteria, statements assessing the quality of the article, and other important dimensions. These aspects included study methodologies, utilized datasets, primary research goals, participant details, publication year, author information, techniques, and relevant keywords. To ensure consistency and coherence, the extracted data were systematically organized using Microsoft Excel [71]. This approach yielded a well-structured overview of the research landscape, facilitating efficient analysis and synthesis of the acquired information.

To evaluate the quality of the selected papers, a quality assessment process was included in the data extraction tool. This process aimed to establish the robustness and credibility of the studies, thereby bolstering the overall reliability of the review's findings. The quality assessment was systematically conducted by closely scrutinizing each selected paper, guided by a set of predefined quality assessment questions as presented in Table 2. These questions were designed to assess diverse dimensions of study quality, encompassing research methodology, data collection procedures, analysis techniques, and presentation of results.

The quality assessment questions were formulated by drawing from well-established criteria in the realm of systematic reviews and were tailored to the specific context of machine learning applications in aviation post-accident analysis. Each paper underwent a meticulous assessment, with the evaluations' outcomes recorded to ensure consistency and accuracy throughout the evaluation process. It is important to note that the quality assessment was not solely reliant on quantitative metrics; rather, it encompassed a holistic evaluation of factors such as the alignment of the study with research objectives, the clarity of research questions, the validity of methodologies, and the appropriateness of interpretations.

The quality assessment procedure led to the exclusion of an additional 7 papers. This stringent process ultimately resulted in the final selection of 87 papers that met the rigorous quality standards established by the review[70].

Table 2. Quality assessment questions and the corresponding scope of papers they were applied to.

| Quality Control Questions | Scope |
|---|---|
| 1. Is the research objective clearly defined? | All |
| 2. Is the context of the research clearly defined? | All |
| 3. Does the study bring value to academia or industry? | All |
| 4. Are the findings clearly stated and supported by the results? | All |
| 5. Are limitations explicitly mentioned and analyzed? | All |
| 6. Is the methodology clearly defined and justified? | All |
| 7. Is the experiment clearly defined and justified? | All |
| 8. Has the utilization of ML techniques been comprehensively described and justified? | ML |
| 9. Are the chosen ML types and tasks appropriate for addressing the research questions? | ML |
| 10. Is there clarity on the ML algorithms employed, including their rationale and suitability? | ML |
| 11. Have biases and ethical considerations in the application of ML techniques been addressed? | ML |
| 12. Are the implications of utilizing ML in post-accident analysis discussed? | ML |
| 13. Is the integration of ML insights with safety measures thoroughly explored and elucidated? | ML |

*2.4.6. Data Synthesis*

The process of data synthesis involved a systematic aggregation and analysis of the extracted data from the curated subset of 87 papers. This step aimed to distill and concisely summarize the key findings, trends, and insights across the studies to provide an overview of ML applications in aviation analysis. The extracted data were systematically organized and categorized based on common themes and research aspects. Patterns and recurring trends were identified by comparing the methodologies, approaches, algorithms, and outcomes of the studies. This facilitated the development of a coherent narrative illuminating diverse ways ML techniques enhance aviation safety through the post-accident analysis and identified gaps, challenges, and future research directions. Meaningful conclusions were derived, offering a holistic perspective and contributing to an understanding of ML implications in aviation post-accident analysis. The next section presents outcomes of the systematic literature review, offering insights into a wide array of learning approaches, techniques, and their implications for aviation safety in post-accident analysis.

## 2.5 Results

The SLR on machine learning (ML) applications in post-accident analysis within the aviation industry has uncovered 87 relevant articles. These studies explain various ML techniques, their potential safety enhancements, and the profound insights they offer into aviation accidents. This section addresses Research Question 1 and subsequent inquiries, with the remaining questions answered in the following section.

### *2.5.1 Aviation Application*

The systematic review categorized the post-accident analysis studies based on their specific aviation applications. The included studies were thoroughly examined to identify and analyze the aviation applications employed. Most of the studies focused on ALL aviation applications (44%), Regular Passenger Transport (S/RPT) (31%), and general aviation (GA) (7%). Other studies addressed Military aviation (3%), and Cargo (1%). This distribution is visually represented in Figure 5 as a Pareto plot, which illustrates that ALL received a more comprehensive focus in the studies compared to other aviation applications, at around 55% of all the studies included.

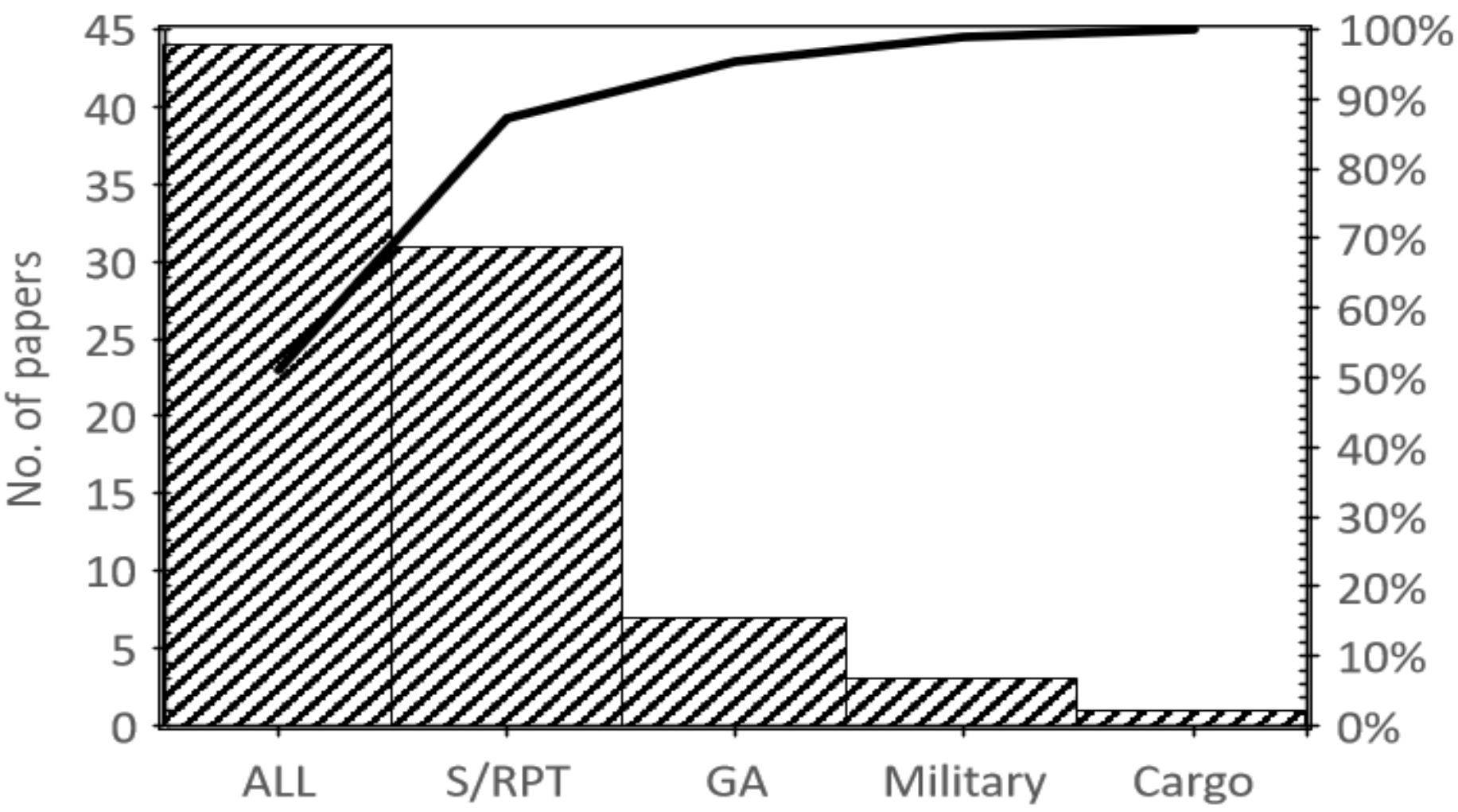

Figure 5. Pareto plot of the publication counts for different aviation types identified in this study where GA = General Aviation, and S/RPT = Regular Passenger Transport.

### *2.5.2. Sources and Types of Publications*

Although the Scopus database was used exclusively to search for relevant studies, the inclusion of the backward search offered a slightly wider pool of readily identifiable studies, using a reproducible approach. As such, Scopus was the primary source for included studies of suitable quality, with 69% of the articles found directly though Scopus. However, for those 61 articles from Scopus, the backward search in Google Scholar, using the “cited by” feature, a further 26 unique articles that directly referenced one or more of those 61 Scopus articles were also relevant and of suitable quality. Figure 6 below shows the breakdown of both the source and type of publication. As expected, a larger proportion of Google Scholar results are conferences, due to the limited number of such events and proceedings which are indexed by Scopus. The same is true of research theses.

The categorization of AI/ML post-accident analysis studies based on their types of publication revealed distinct patterns. The included studies were examined to identify and analyze their distribution across various publication types. Most of the studies were centered on journal articles, constituting 52% of the total. Conference papers accounted for 44% of the publications, signifying their significant presence in the field. A smaller portion of the studies were Theses (4%). Figure 6 visually represents the distribution of these publication types, clearly indicating that journal articles were the primary publication type.

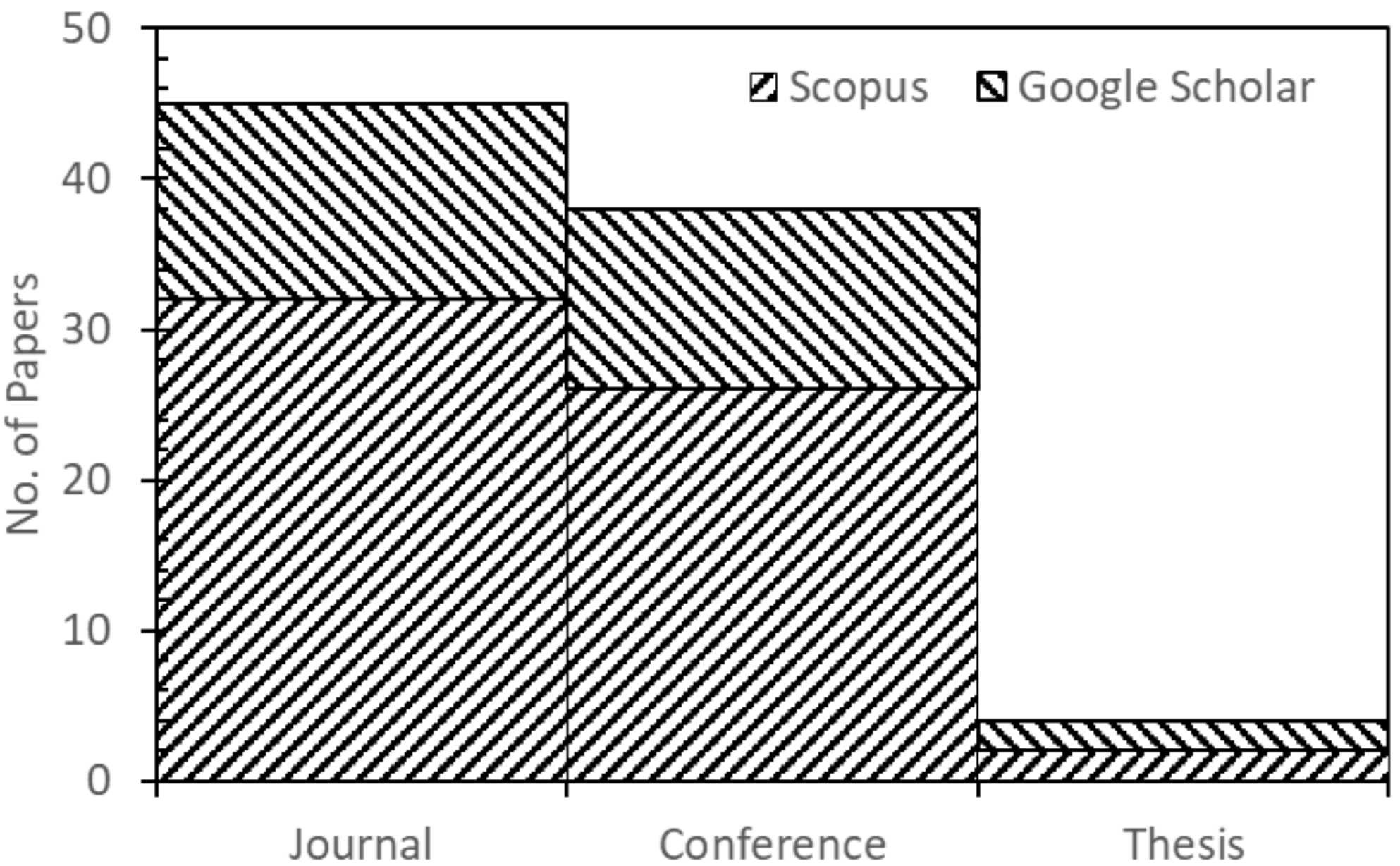

Figure 6. A stacked bar chart showing the distribution of source and type of publication.

*2.5.3 Duration of Studies*

The dataset periods ranged widely across the studies, encapsulating various periods. Some studies utilized extensive datasets that spanned many years or even multiple decades, such as those covering 62 years [53] and 101 years [76]. Other studies focused on more recent periods, with dataset durations of 1 year [77] and 2 years [78, 79]. Also of note is the most extreme example being that by Koteeswaran et al spanning 96 years [56]. Figure 7 visually represents the distribution of the number of publications/studies by the number of years of accidents or incidents studied, i.e. the duration of the study in years. The distribution is skewed to shorter durations, with a peak in the 10-to-19-year category, when broken down into decades. The mean of the 67 studies that included clear information about their duration was 24.7 years, with a median of 15 years. Duration is estimated as the top year minus the start year plus 1, so a study is at least one year long, although this makes the error for each year at ± 0.5 years.

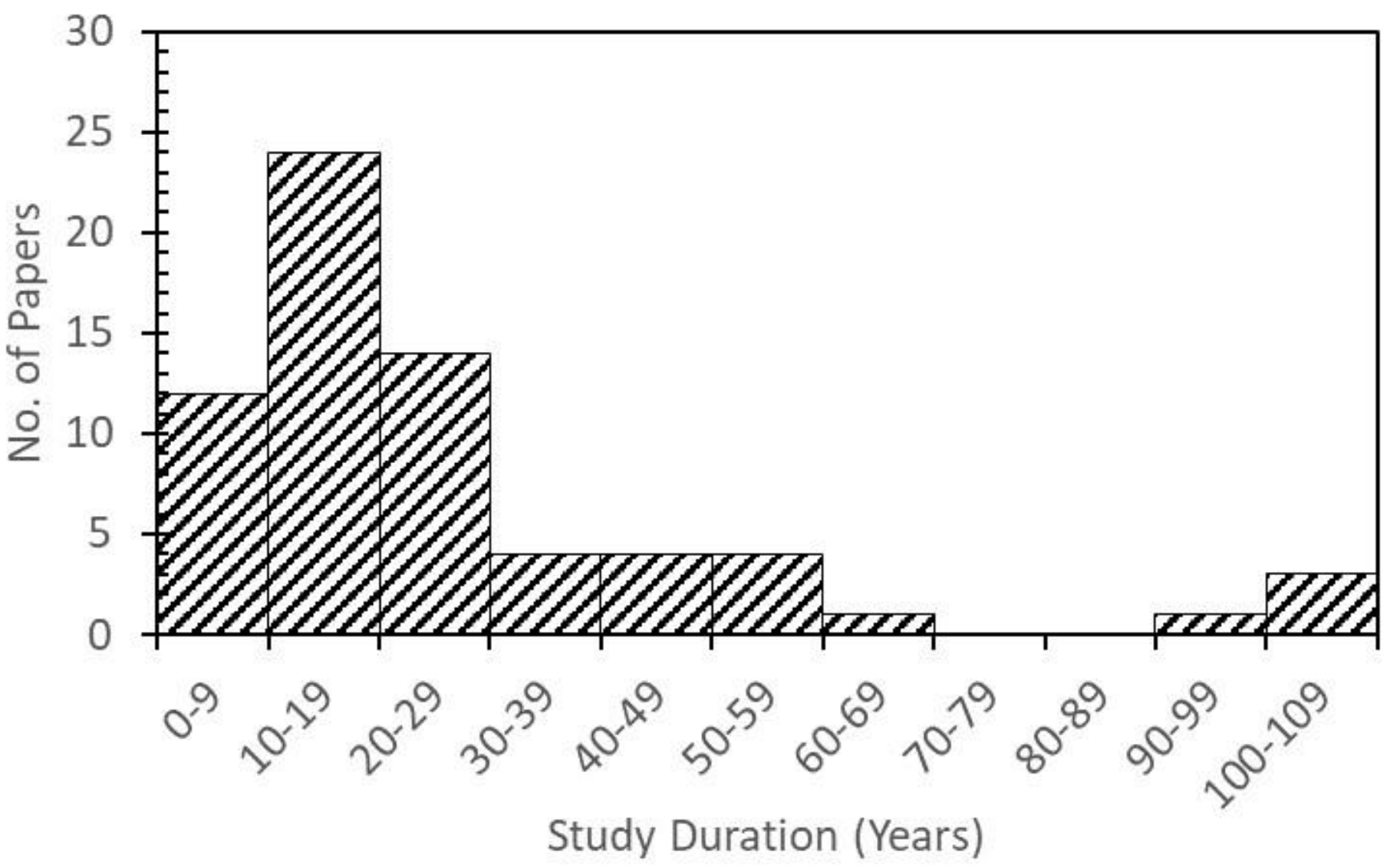

Figure 7. The bar chart shows the distribution of the publication count relative to the study duration in years.

*2.5.4 Publication Trends*

The analysis also revealed diverse variations in the years of publication. Studies encompassing earlier years, such as [80], provided insights into the field's historical progression. More recent studies, like those [55, 81], added contemporary perspectives to the evolving landscape of ML in aviation safety analysis.

This study focused on data from 1998 to 2023 to analyze the trends in ML applications in post-accident analysis within the aviation industry. The analysis revealed interesting patterns in the publication of relevant studies over the years. Notably, the year 2020 exhibited a peak in publication activity, followed by 2021 and 2019, as illustrated in Figure 8. The graph depicting the publication trends highlights a distinct evolution. From 1998 to 2008, there was a limited number of studies applying ML techniques to postaccident analysis in aviation, with only one paper being published per year. Subsequently, a gradual increase was observed, with the number of published papers reaching six by 2018. The most significant surge in publications occurred from 2019 onwards, when the rate reached ten papers per year. It is important to note that the range from 1998 to the present is due to the availability of qualifying papers within the Scopus database, signifying the emergence and subsequent growth of this field after 1998.

The stratification of data in Figure 8 suggests further analysis is warranted. That is, three distinct regions in terms of publication counts can be identified looking at the plot; these are three generations of research, each marked by notable shifts in focus and intensity. The first generation, spanning from 1998 to 2008, reflects a foundational period characterized by an average of 0.4 papers per year. This era witnessed the emergence of key methodologies starting with decision trees in 1998 and logistic regression in 2004. In the second generation, extending from 2009 to 2018, research activity increased, with an average of 3.4 papers per year. During this period, methodologies such as support vector machines in 2010 and deep learning in 2017 gained prominence. Finally, the current third generation, covering 2019 to 2023 (2022), exhibits a significant surge in research output, averaging 12.5 papers per year. This phase signifies a rapid increase in research efforts, underscoring the dynamic evolution of AI/ML methodologies applied to postaccident analysis in aviation.

This trajectory indicates the growing recognition and adoption of ML in post-accident analysis within the aviation industry. The steady increase in publications from 2009 to 2018 could be attributed to the gradual integration of data-driven approaches in accident investigation. The substantial rise in publications from 2019 to 2020 signifies a significant shift towards more sophisticated ML techniques and their applications, reflecting the industry's acknowledgment of the potential of these methods in enhancing safety measures and accident insights. The trend is also likely driven by the fidelity of the NLP techniques to infer the ontology of aviation safety.

In general, the evolution of research generations, with increasing efforts, indicates the growing recognition and adoption of ML in post-accident analysis within the aviation industry. The steady increase in publications from 2009 to 2018 could be attributed to the gradual integration of data-driven approaches in accident investigation. The substantial rise in publications from 2019 to 2020 signifies a significant shift towards more sophisticated ML techniques and their applications, reflecting the industry's acknowledgment of the potential of these methods in enhancing safety measures and accident insights. The trend is also likely driven by the fidelity of the NLP techniques to infer the ontology of aviation safety as shown.

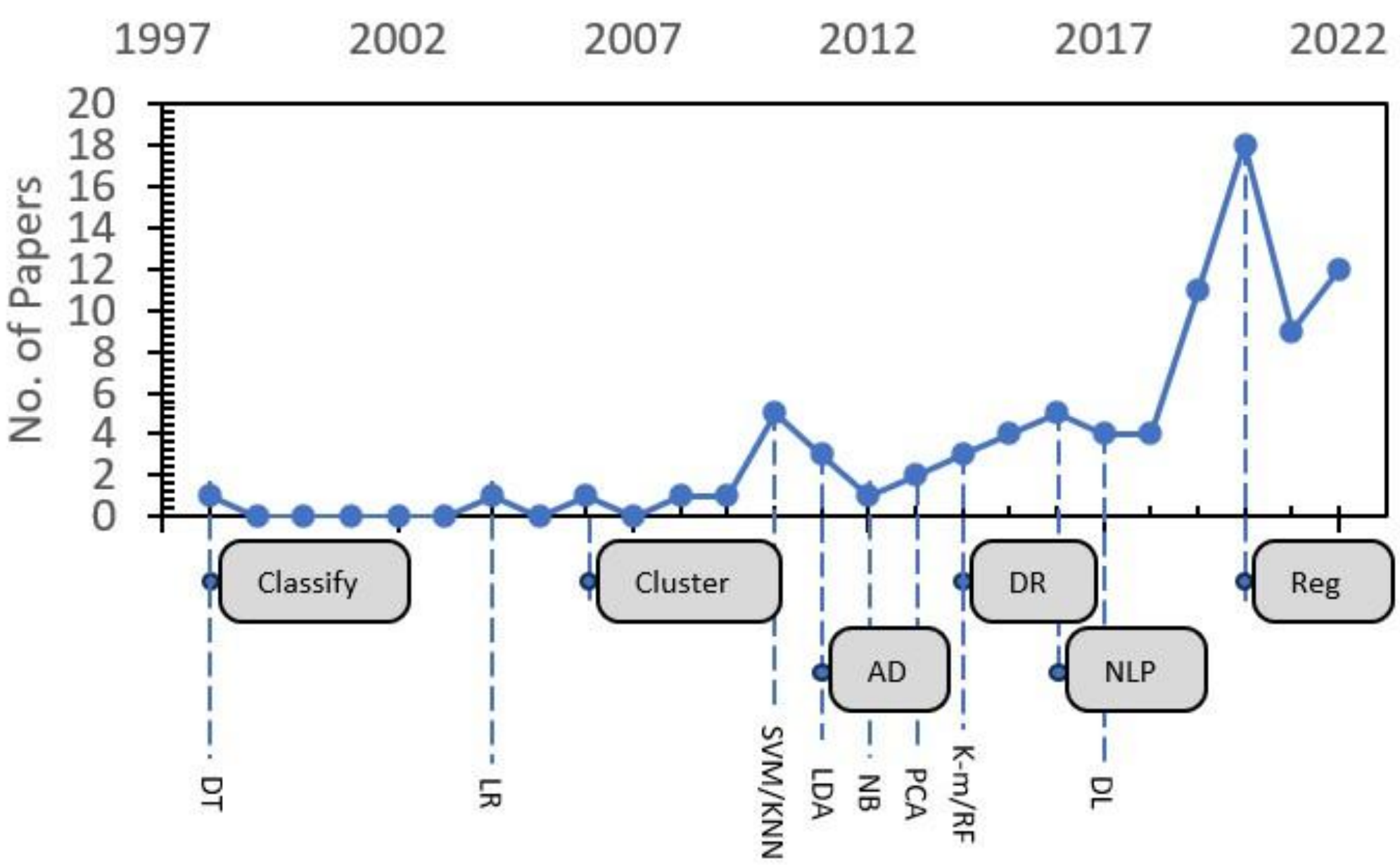

Figure 8. Line chart illustrating the publication trend from 1998 to 2022. Also shown are the first implementation of each ML task (boxes) and the first implementation of each specific ML algorithm (vertical text).

*2.5.5 Software Packages Utilized for Analysis*

The analysis of software packages employed in the reviewed studies highlights the diverse tools utilized for conducting machine learning research within the context of aviation post-accident analysis. Among the various options available, Python programming language emerged as the most prominently favoured choice, accounting for 40% of the studies. Python's popularity since it was released in 1995 can be attributed to its versatility, extensive ML libraries, batch scripting capabilities, and the broader data science community it supports [82].

In addition to Python, several other software packages were prevalent in the studies. The Waikato Environment for Knowledge Analysis (WEKA) software package was identified as a key tool, utilized in 20% of the studies [83]. WEKA's user-friendly interface and comprehensive suite of ML algorithms made it a preferred option for researchers. MATLAB, known for its robust numerical computing capabilities, was another significant choice, adopted by 16% of the studies.

While most studies leaned towards Python, a notable subset opted for R programming packages, constituting 11% of the total. R's statistical capabilities and data visualization functionalities were the reasons behind its selection.

Beyond these primary software packages, a smaller proportion of studies incorporated other tools. Structured Query Language (SQL), often used for managing and querying databases, was employed in 4.5% of the studies. Similarly, the Statistical Package for the Social Sciences (SPSS) and the Statistical Analysis System (SAS), renowned statistical software packages, each contributed to 4.5% and 2% of the studies, respectively. RM Studio, a platform for statistical analysis and graphical representation, was utilized in 2% of the studies.

Visualized in Figure 9, the distribution of software packages underscores Python's commonality, followed by WEKA and MATLAB. The fact that 80% of the studies correspond to almost 40% of the different software packages indicates that there is currently no consensus on the software that is best to utilize for aviation safety analysis; likely due to a variety of selection criteria related to goals and the specific requirements. However, Python, as a free open-source package, appears to be becoming dominant.

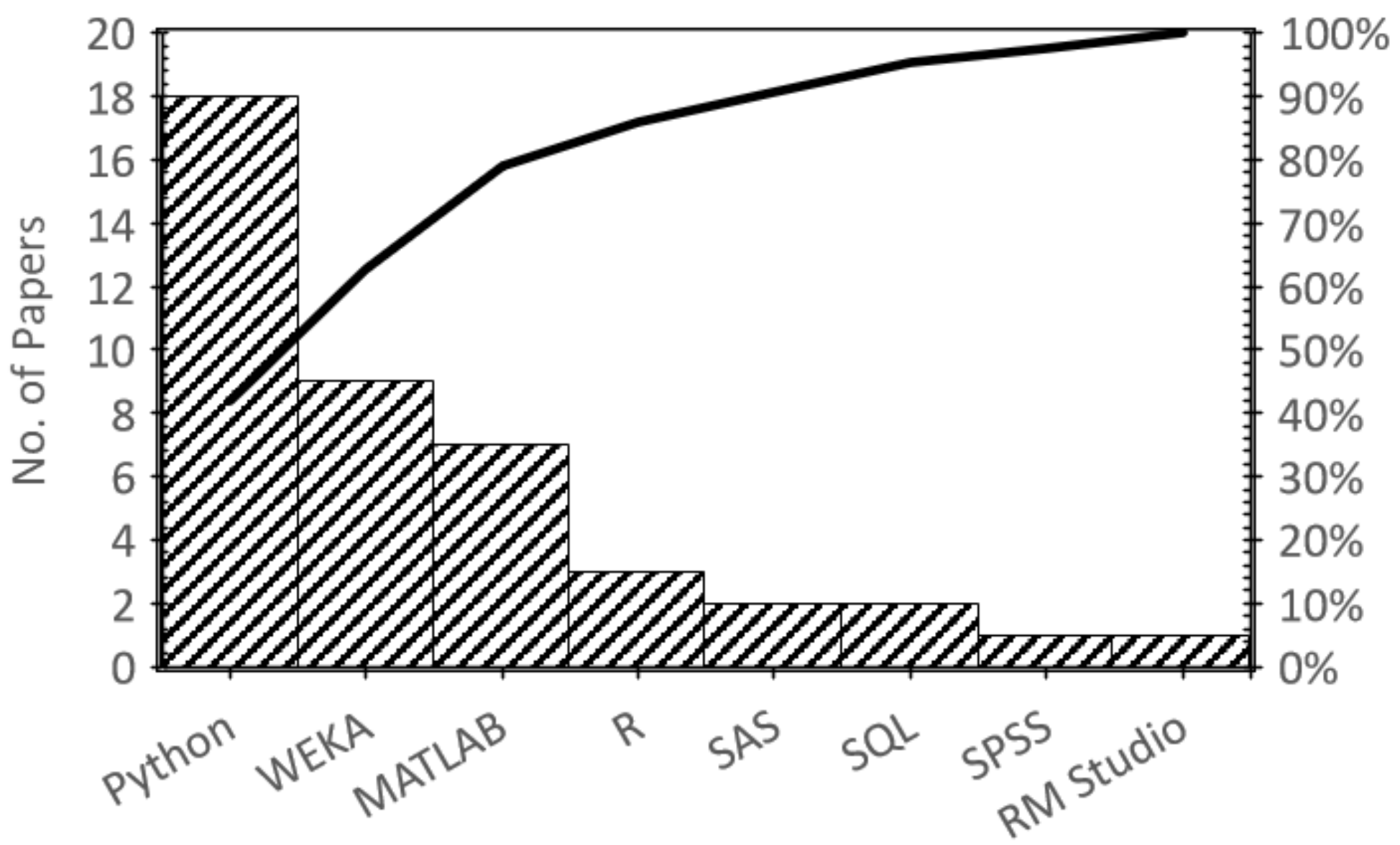

Figure 9. Pareto plot of publication counts as categorized by the software package utilized in the study, to implement the relevant ML tools and techniques.

*2.5.6 Data Source Utilized in the Studies*

The Aviation Safety Reporting System (ASRS) dataset emerged as the most prominently utilized data source, featuring in 41% of the studies. This was followed by the National Transportation Safety Board (NTSB) dataset, which constituted another significant data source, contributing to 25% of the studies. Other data sources collectively constituted 18% of the total. This category encompassed a diverse array of sources that researchers drew upon to address their specific research questions.

A smaller fraction of studies incorporated additional data sources to enrich their analyses. The Federal Aviation Administration (FAA) dataset, comprising aviation-related data, was employed in 8% of the studies. Some researchers conducted surveys to gather primary data, representing 3% of the total. Accessing data from the Aviation Safety Network (ASN) and the Australian Transport Safety Bureau (ATSB) contributed to 4% and 1% of the studies, respectively. Figure 10 visually encapsulates the distribution of data sources, clearly illustrating the prevalence of the ASRS and NTSB datasets across the studies and reflecting the 41 and 25 per cent of overall air miles flown in these air domains.

*2.5.7. Country of Publication Analysis*

The geographic distribution of research output offers insights into the global engagement with ML applications in aviation post-accident analysis, noting of course air travel is still dominant in developed countries or regions like North America and Europe. Notably, the United States (USA) emerges as a dominant contributor with 45 publications, reflecting active research in advancing ML in aviation safety. India follows with 12 publications, highlighting robust engagement. China, with 11 publications, underscores investment in ML and aviation safety analysis. Canada, Türkiye, Italy, and Saudi Arabia contribute 3 to 4 publications each, highlighting diverse regional interests. Additionally, several countries, including Australia, UK, and Germany, each contribute one publication, enriching the global discourse on ML in aviation safety. Figure 11 visually represents the distribution of research output across these prominent countries. The diverse representation of nations underscores the collaborative and global nature of research efforts aimed at enhancing aviation safety through ML applications.

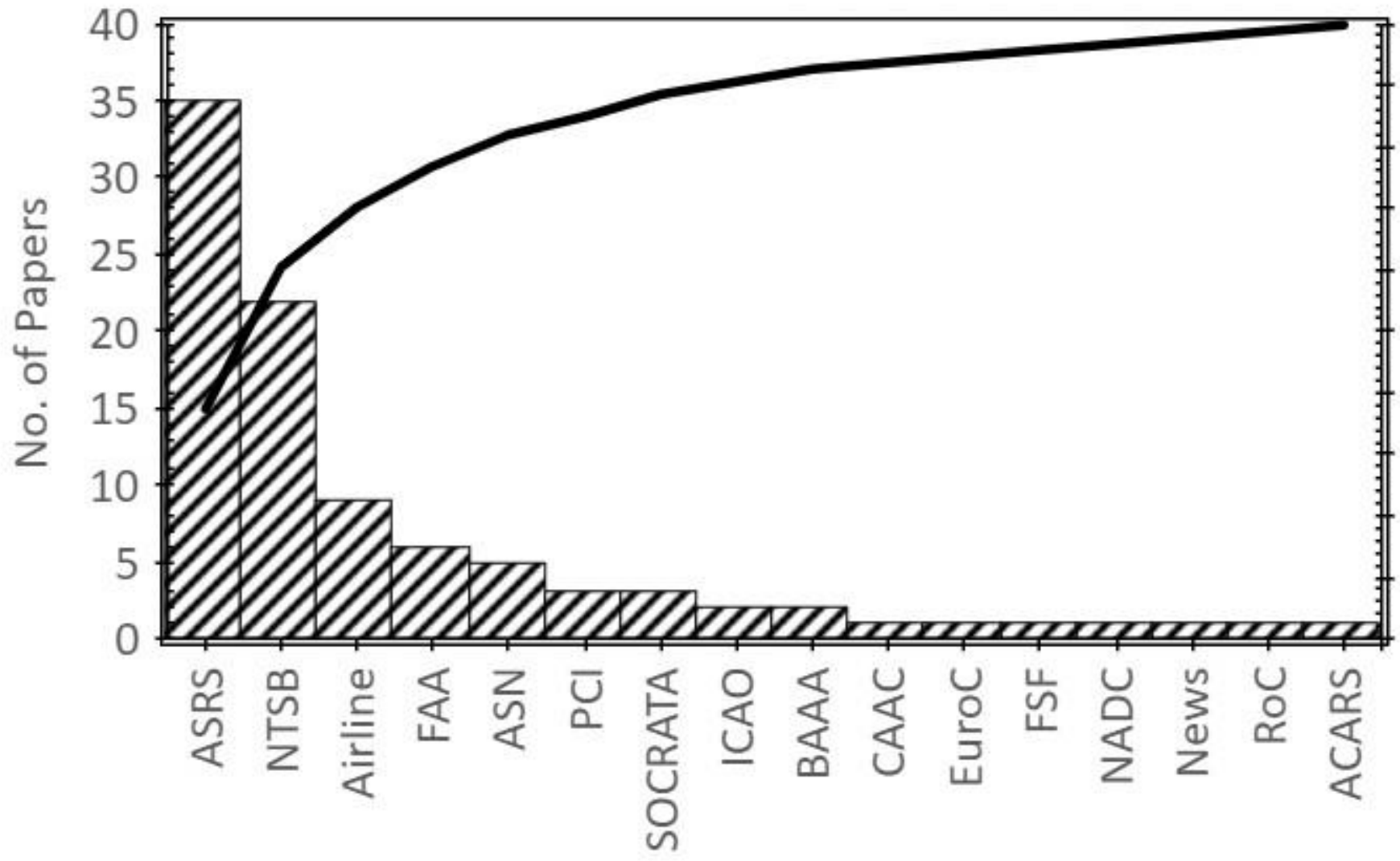

Figure 10. Pareto plot showing the publication count as categorized by the data sources utilized.

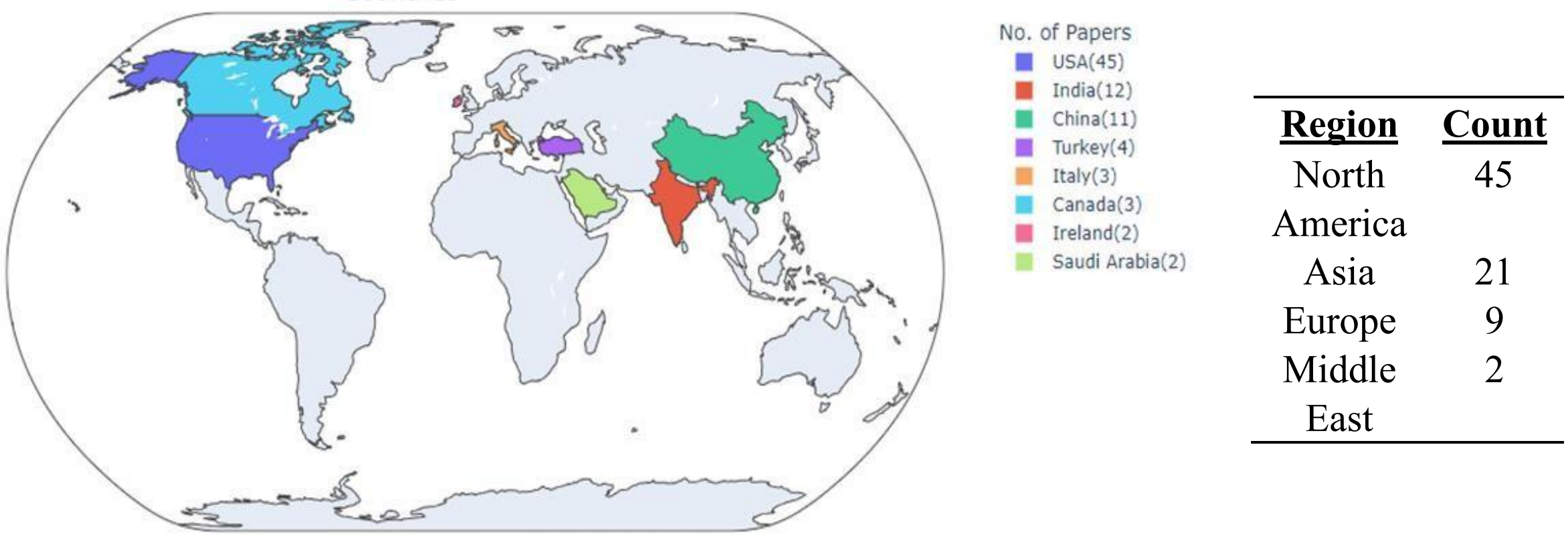

| Region | Count |
| --- | --- |
| North America | 45 |
| Asia | 21 |
| Europe | 9 |
| Middle East | 2 |

Figure 11. Geographic distribution of publications by country (left), and count by ICAO Region/Continent (right)

*2.5.8. Authors*

In this study, authors who have made noteworthy contributions to the realm of ML applications in aviation post-accident analysis have been identified. Table 3 presents an overview of authors who have exceeded two publications in this domain, accompanied by their respective affiliations. The recurring presence of certain authors across multiple papers accentuates their active involvement and proficiency in driving the advancement of research within this field. Notably, Khan, L and Mavris, DN stand out as prolific contributors, each boasting five papers in the study period of 1998-2023. Likewise, Christopher, AA, and Puranik, TG have each authored four papers, while Alias Balamurugan, SA, Mahadevan, S, Rao, AH, Robinson, SD, and Zhang, X have made significant contributions with three papers each.

Table 3. Authors with notable contributions, showing their relevant number of publications, along with the years of the first and last (most recent) publication.

| Author | Count | First | Last | Affiliation |
| --- | --- | --- | --- | --- |

| Khan, L | 5 | 2008 | 2020 | Department of Computer Science, The University of Texas at Dallas |
|---|---|---|---|---|
| Mavris, DN | 5 | 2020 | 2022 | Georgia Institute of Technology, Atlanta, GA 30332, United States |
| Christopher, AA | 4 | 2013 | 2022 | Research scholar Anna University Tamil Nadu, India |
| Puranik, TG | 4 | 2020 | 2022 | Universities Space Research Association, NASA Ames Research Center, Moffett Field, CA, USA |
| alias Balamurugan, SA | 3 | 2013 | 2022 | Research Scholar Anna University, Tamilnadu, India |
| Mahadevan, S | 3 | 2015 | 2021 | Department of Civil and Environmental Engineering, Vanderbilt University, Nashville, TN, USA |
| Rao, AH | 3 | 2018 | 2020 | Collins Aerospace, 400 Collins RD, MS 124-319 Cedar Rapids, USA |
| Robinson, SD | 3 | 2015 | 2019 | Parks College of Engineering, Aviation and Technology, Saint Louis University, Saint Louis, MO 63103, USA |
| Zhang, X | 3 | 2015 | 2018 | Department of Civil and Environmental Engineering, School of Engineering, Vanderbilt University, Nashville, TN, 37235, USA |

#### *2.5.9. Machine Learning Dimensions*

This section addresses Research Questions 2a, 2b, and 2c by providing insights into various dimensions of machine learning, specifically focused on data analysis in accident research within the aviation industry. Through systematic analysis of collected data and literature, each dimension is explored to understand the intricacies of machine learning applications in post-accident analysis. This includes investigating machine learning types, tasks, and algorithms utilized in the field.

##### 2.5.9.1 Machine Learning Approaches

In response to Research Question 2, this section shows the research approaches adopted in the selected studies. It was found that most of these studies (62%) employed a quantitative approach, emphasizing the collection and analysis of numerical data to draw conclusions and make inferences. Also, a considerable proportion of studies (30%) utilized a mixed-methods approach, integrating both quantitative and qualitative methods while a smaller portion of studies (8%) employed a qualitative-only approach, focusing on understanding underlying meanings, motivations, and nuances of the phenomena under investigation.

The prevalence of quantitative and mixed methods approaches signifies a commitment to robust empirical data collection and analysis, enhancing the credibility and validity of the research findings.

##### 2.5.9.2 Machine Learning Types

In response to Research Question 2a, the researchers explored various ML types used across the selected studies. The analysis revealed that a large portion of the studies (65%) applied supervised learning techniques. Supervised learning involves training models using labelled data, allowing them to make predictions or classifications based on patterns identified in the training set [40].

Furthermore, 30% of the studies embraced unsupervised learning, a technique aimed at discovering patterns or structures in data without the need for predefined labels. Unsupervised learning algorithms enable researchers to uncover hidden insights and relationships within datasets [41, 42]. While a smaller portion of studies (5%) incorporated semi-supervised learning, which combines elements of both supervised and unsupervised learning. Semi-supervised learning leverages a limited amount of labelled data with a larger pool of unlabelled data to enhance the efficiency and performance of the model [43].

To provide a visual representation of these findings, Figure 12 depicts the distribution of ML types employed in the context of aviation post-accident analysis.

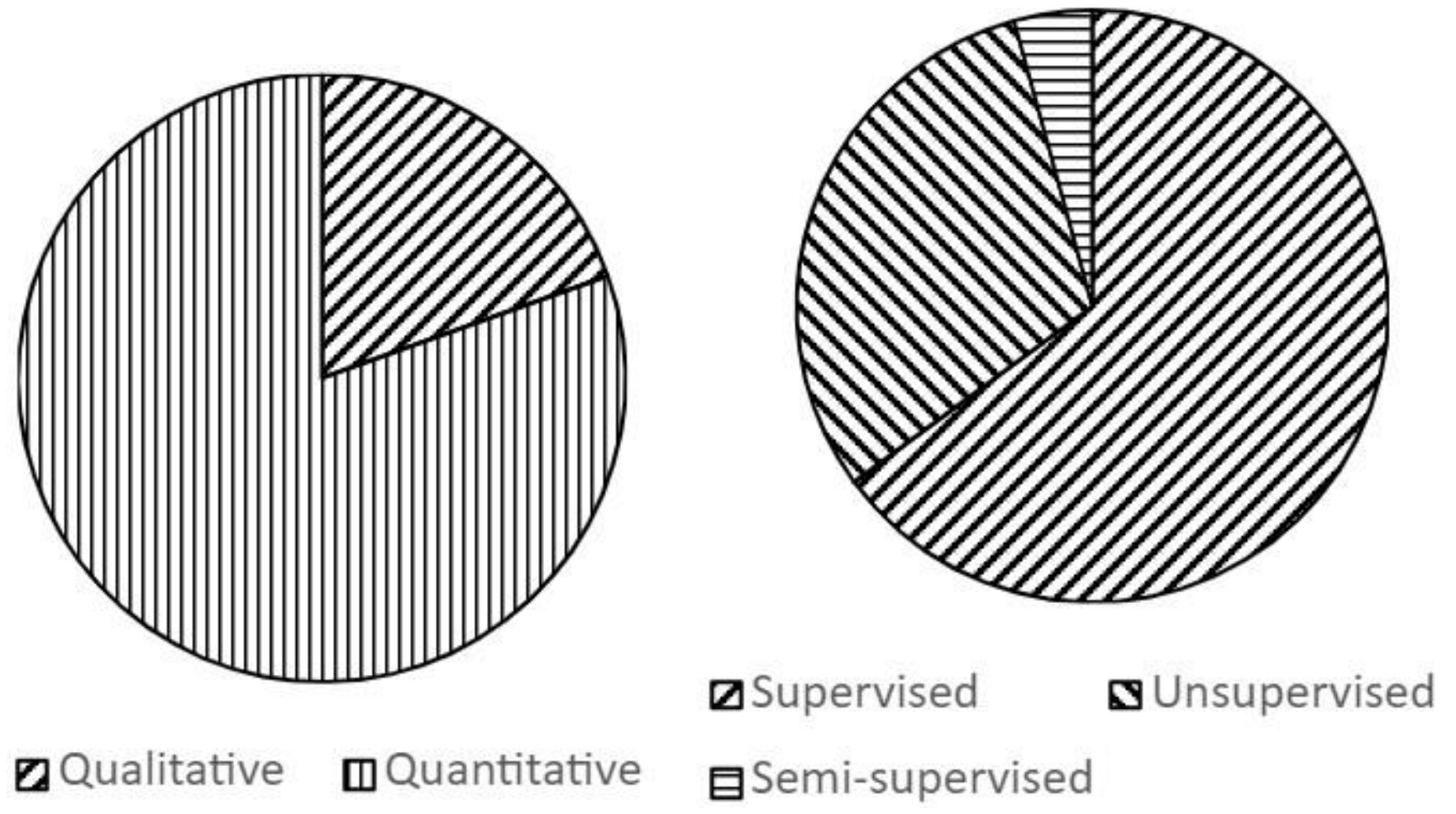

Figure 12. Left, is the breakdown of data types used in the ML studies, and Right, is the broad type of ML used.

2.5.9.3 Machine Learning Tasks

Addressing Research Question 2b, researchers investigated machine learning tasks employed to enhance post-accident analysis within the aviation industry. The investigation revealed a range of tasks that researchers engaged with to extract valuable insights from accident data.

The most prominent ML task undertaken was classification, accounting for 59% of the studies. Classification involves categorizing data into predefined classes or categories, aiding in the identification of patterns that differentiate various accident scenarios [84, 85].

Furthermore, 18% of the studies focused on natural language processing (NLP), a task that involves the extraction of meaningful information from textual data. NLP techniques are instrumental in unravelling insights embedded in textual descriptions of accidents and incidents [86]. Anomaly detection, with 8% representation, emerged as another significant task. This involves identifying unusual or exceptional patterns in the data that could indicate potential safety hazards or exceptional circumstances [77]. Clustering, contributing to 9% of the studies, entails grouping similar data points together, helping to unveil inherent structures within accident datasets [87].

A smaller fraction of studies (5%) targeted data reduction, aiming to distill relevant information from large datasets, and a mere 1% of studies involved regression, a task focused on predicting numerical values based on input data. For a visual representation of the distribution of these ML tasks, please refer to Figure 13. Classification accounts for 60%, indicating that while not dominant, the task is the primary consideration for aviation accident analysis with ML. This exploration of machine learning tasks elucidates the multifaceted strategies employed to glean insights from aviation accident data, contributing to the advancement of safety measures within the industry.

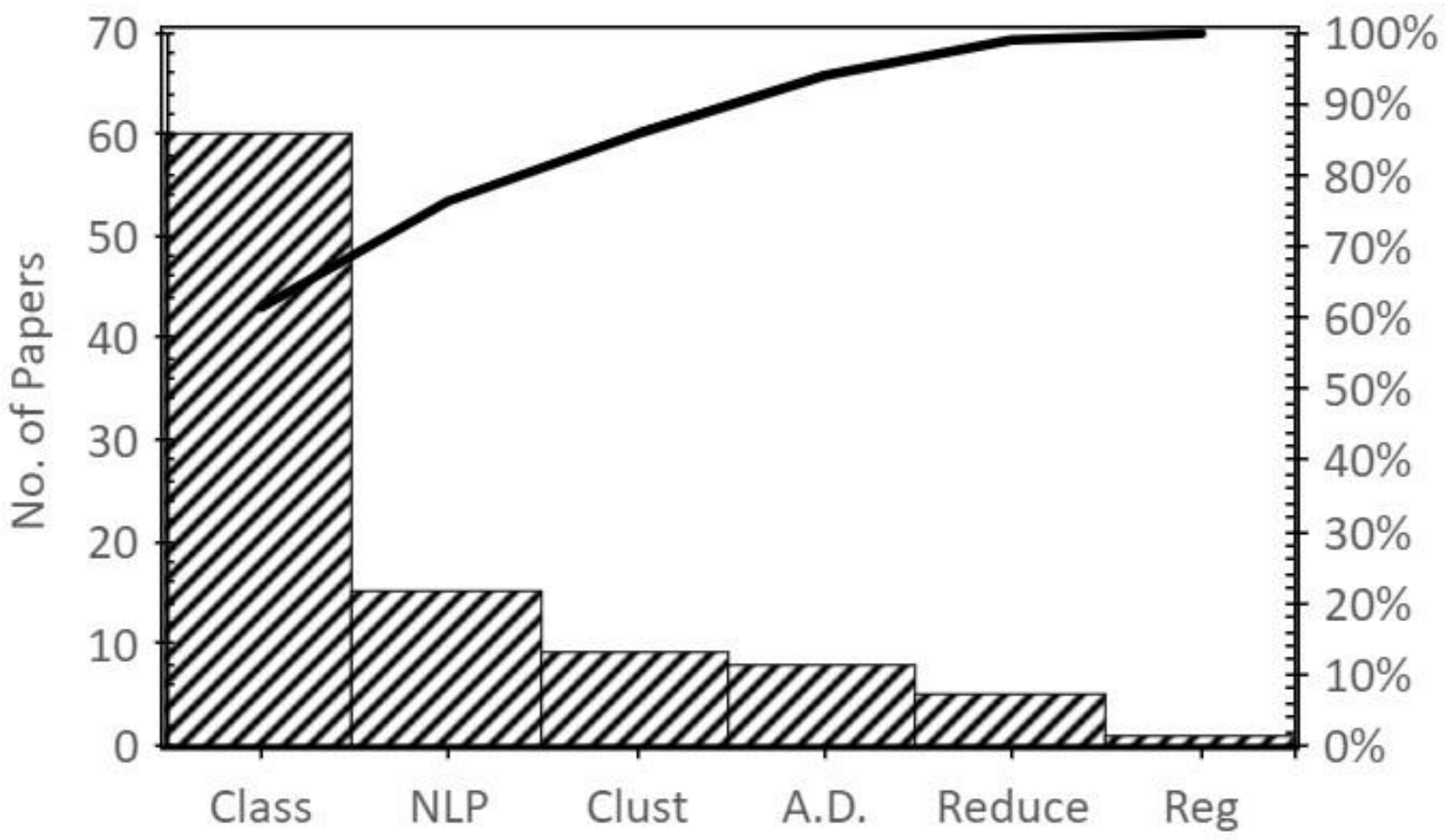

Figure 13. Pareto plot showing the publication count for each of the ML tasks utilized.

2.5.9.4 Machine Learning Algorithms

In response to Research Question 2c, researchers explored various machine learning algorithms utilized for post-accident analysis within the aviation industry. The analysis revealed diverse algorithms deployed to uncover meaningful insights from complex aviation accident-related data. The top 10 machine learning techniques used in this context are detailed in this section.

The most widespread algorithm employed was Deep Learning (DL), constituting 21% of the studies. Deep Learning's capacity to capture intricate patterns and representations in complex data has made it a favoured choice for unravelling hidden insights from aviation accident data [78, 79].

Decision Trees (DT) were utilized in 16% of studies, providing a visual representation of decision-making processes and enabling the identification of key factors in accidents. Support Vector Machines (SVM) followed closely, being adopted in 14% of studies. SVM is a robust classification and regression technique that effectively handles both linear and non-linear data. Principal Component Analysis (PCA) was applied in 4% of studies, facilitating dimensionality reduction and feature extraction from high-dimensional accident data. Linear Discriminant Analysis (LDA) played a role in 6% of studies, offering a powerful technique for dimensionality reduction and classification. Logistic Regression (LR) was employed in 6% of studies, providing insights into the relationships between accident features and outcomes.

K-Nearest Neighbors (KNN) emerged in 10% of studies, enabling the classification of accidents based on the similarity to neighbouring data points. Naive Bayes (NB) was utilized in 12% of studies, offering a probabilistic approach to classification based on Bayes' theorem. Random Forest (RF) algorithms were applied in 9% of studies, highlighting their utility in ensemble learning for classification and regression tasks. K-Means, a clustering algorithm, was used in 2% of studies, aiding in grouping similar accidents for pattern recognition [88]. It is important to note many studies tried multiple approaches and so are represented often.

For a visual representation of the prevalence of these ML algorithms, please refer to Figure 14. This analysis highlights the diverse toolkit of algorithms that researchers have harnessed to extract meaningful insights from aviation accident data, fostering advancements in safety measures and accident prevention. However, the fact that the 80% mark of the counts corresponds to half of the algorithm indicates that there is currently no primary focus on the ML algorithms utilized in aviation accident analysis. DL is an early leader, but it will be interesting to see how this aspect continues to evolve.

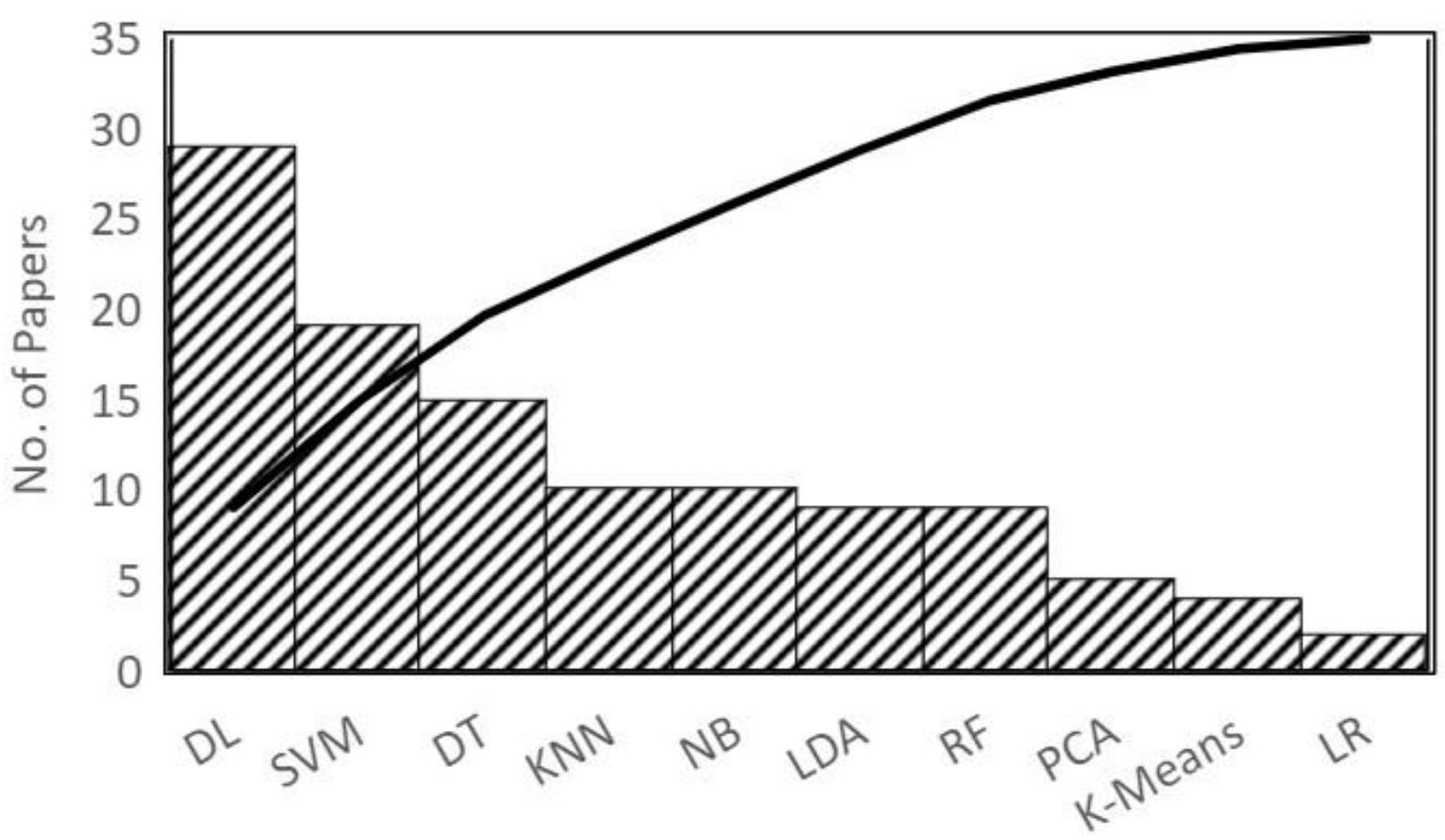

Figure 14. Pareto plot of the publication counts based on the ML algorithms utilized.

2.5.9.5 Machine Learning Summary

The systematic review unveiled ML techniques harnessed in post-accident analysis within the aviation sector. Ranging from conventional algorithms like decision trees (DT) and support vector machines (SVM) to advanced methods such as deep learning (DL) and neural networks (NN), these techniques were found to be crucial tools. They facilitated tasks like clustering, classification, and regression, contributing across various stages of accident analysis, including identifying factors that led to accidents and predicting potential outcomes. Examples include the works of [89, 90] who utilized ASRS and ECCAIRS datasets within the domain of civil aviation. Zhou et al., [79] employed ML techniques on ACARS (Aircraft Communication Addressing & Reporting Systems) data, thus offering insights into civil aviation. In parallel, [91] delved into the application of ML, using FOQA (Flight Operational Quality Assurance) data within the commercial aviation context.

#### *2.5.10 Enhancements to Safety*

The utilization of a diverse array of ML algorithms within the aviation industry has markedly advanced the domain of safety enhancement and insights extraction from aviation accident data. By harnessing these sophisticated analytical tools, researchers have been able to unearth valuable information, leading to more effective safety measures and deeper comprehension of aviation accidents. The convergence of these algorithms with aviation data has yielded the following seven multifaceted benefits:

**Risk Assessment and Prediction**: Machine learning algorithms like Support Vector Machine (SVM), Random Forest (RF), and Naive Bayes (NB) have enabled the prediction of potential risks and hazards in aviation systems [52, 78, 92]. These algorithms scrutinize historical data to identify patterns, which can be used to forecast potential threats and pre-emptively implement necessary safety measures.

**Anomaly Detection**: Algorithms such as Decision Trees (DT) and Deep Learning (DL) techniques have excelled in detecting anomalies in aviation data [77, 93]. These anomalies signify deviation from expected behaviour, which, when identified, can trigger immediate corrective actions.

**Classification of Accidents and Factors**: A wide range of classification algorithms including SVM, KNearest Neighbors (KNN), and Gradient Boosting (GB) have enabled accurate categorization of aviation accidents [85, 94, 95]. Additionally, these algorithms have allowed for the identification of contributing factors and patterns that might have otherwise gone unnoticed.

**Natural Language Processing (NLP) Applications**: NLP-based algorithms like Latent Dirichlet Allocation (LDA) and Recurrent Neural Networks (RNN) have enabled the extraction of meaningful information from textual safety reports [86, 96, 97]. By analyzing text narratives, these algorithms can uncover hidden insights and trends, thereby augmenting the understanding of safety-related incidents.

**Clustering and Pattern Identification**: Clustering algorithms such as K-Means, DBN (Deep Belief Network), and Self-Organizing Maps (SOM) have been instrumental in identifying latent patterns within aviation accident data [87, 98]. These patterns can help in understanding the root causes of accidents and guide future safety initiatives.

**Dimensionality Reduction**: Techniques like Principal Component Analysis (PCA) and Non-negative Matrix Factorization (NMF) have aided in reducing the dimensionality of complex data, thereby enabling a more manageable representation for analysis [99, 100]. This simplification facilitates the identification of crucial features and relationships within the data.

**Human Factors Analysis**: Machine learning (ML) algorithms have also facilitated the analysis of human factors in aviation accidents, enabling a better understanding of behavioural and cognitive aspects that contribute to safety incidents [81]. By identifying these factors, targeted interventions can be designed to mitigate risks.

In essence, the combination of various ML algorithms with aviation accident data has led to a comprehensive enhancement of safety measures and the generation of insights that contribute to informed decision-making within the aviation industry.

## 2.6 Discussion

By examining the contextual nuances that influence the application of ML and delving into the varied techniques employed, an understanding of the dynamic landscape of ML in aviation safety has been provided. While the study encompasses a timeframe of 25 years (1998–2023), the study has summarized all the ML used in various aviation applications (Table 1, Table 2) and discussed different software packages (Figure 9), data sources (Figure 10), country of publication (Figure 11), ML types (Figure 12), ML Tasks (Figure 13) and ML algorithms (Figure 14). This study has reviewed the ML algorithms used in the literature and summarized them in Table 4 for each chosen article. The review aimed to provide a comprehensive analysis of past research and offer a forward-looking perspective on the potential impact of ML in further enhancing aviation safety standards.

### *2.6.1. Diverse Machine Learning Techniques and Applications*

The unique challenges posed by the aviation industry necessitate a focused approach to enhancing safety through ML applications. The review underscores the significance of comprehending the application context and defining evaluation measures that align with the specific requirements of the aviation sector.

The analysis emphasizes the industry's acknowledgment of the need for ML techniques to improve safety in both general aviation and passenger transport. Beyond accident investigation, these techniques find versatile applications in air traffic control, airport management, and military aviation [101]. This versatility highlights the richness of ML techniques in addressing safety concerns within the aviation domain. The classification of these techniques according to their respective ML tasks underscores their adaptability across accident analysis, predictive modeling, and real-time incident detection. From traditional methodologies to recent advancements like deep learning, this toolkit reaffirms the efficacy of ML in addressing the multifaceted challenges of aviation safety.

### *2.6.2. Data Sources and Quality Assessment*

The foundation of effective ML-driven post-accident analysis lies in the quality and relevance of the data sources employed. Notably, the extensive utilization of datasets such as ASRS and NTSB highlights the industry's reliance on real-world aviation data to unveil intricate patterns and contribute to safety enhancement. The rigorous quality assessment protocols followed across the selected studies uphold the integrity, validity, and reliability of research findings, elevating the credibility of outcomes [35, 39].

It is pivotal to recognize the decisive role that high-quality datasets play in generating actionable insights for aviation safety. The review not only categorizes the datasets in use but also delves into their implications for research outcomes. This in-depth understanding empowers stakeholders to choose appropriate data sources and underpins data-driven decision-making, amplifying the precision and pertinence of aviation safety analyses.

#### *2.6.3. Safety Enhancement and Insights Generation*

Machine learning (ML) techniques hold promise for improving safety standards within the aviation industry. These methodologies excel at identifying concealed patterns and underlying factors that might have contributed to accidents, thereby empowering stakeholders to proactively address safety concerns. Additionally, the predictive ability of ML algorithms equips practitioners with the means to anticipate potential occurrences, facilitating the development of precise preventive strategies. Through furnishing valuable insights into aviation accidents, these techniques play a pivotal role in informed decision-making and fostering an all-encompassing enhancement of safety measures [58, 92, 102].

#### *2.6.4. Variability in Study Types and Purposes*

This review underscores the imperative of considering diverse study types and objectives to yield meaningful advancements in aviation safety. By showcasing the array of methodologies and algorithms (Figure 14) adopted, insights into the spectrum of research endeavors are highlighted.

#### *2.6.5. Future Research Directions*

The literature underscores the significance of ethical considerations and bias mitigation in ML applications. The effectiveness of aviation safety analysis in the future relies on the implementation of transparent, fair, and unbiased ML models. As the ML field evolves, seven exciting avenues for future research emerge, charting the course for further advancements in ML-driven aviation safety analysis.

##### 2.6.5.1. Interpretable and Explainable AI:

The aviation industry's emphasis on transparency and accountability necessitates the development of ML models that provide understandable explanations for their decisions, particularly when influencing safetycritical choices. Researchers should delve into methodologies that enhance the interpretability of ML algorithms, fostering trust among stakeholders and promoting wider adoption [57].

##### 2.6.5.2. Real-time Accident Prediction:

Incorporating predictive analytics and real-time data processing, future research should focus on creating models capable of real-time accident prediction. These models can serve as early warning systems, facilitating swift interventions and proactive measures to mitigate potential accidents. The synthesis highlights the importance of ML for aviation safety, demonstrating measurable progress in accident prevention. This empirical evidence highlights the effectiveness of ML applications in shaping proactive safety measures and promotes further exploration of techniques with the potential to generate more safety improvements [22].

#### 2.6.5.3. Hybrid Models:

The exploration of hybrid approaches, combining ML with expert systems or physics-based models, holds promise for enhancing the accuracy and robustness of accident analysis. Investigating these combinations will harness the strengths of different methodologies, leading to more comprehensive insights into aviation safety.

#### 2.6.5.4. Handling Unbalanced Data Sets:

Given the inherent imbalance in aviation accident datasets, where the number of accidents is relatively small compared to non-accident instances, innovative research should focus on methods that effectively address this challenge. Developing techniques that can navigate imbalanced data will enhance the models' capability to identify rare but pivotal events. One such technique that has shown promise is the use of costsensitive learning [103]. This approach involves assigning different costs to misclassification errors based on the class distribution, ensuring that the model prioritizes correctly classifying the minority class (accidents) even at the expense of higher error rates in the majority class (non-accidents). By incorporating the relative importance of each class into the learning process, cost-sensitive learning can significantly improve the model's ability to detect rare events while maintaining overall classification accuracy.

#### 2.6.5.5. Privacy and Data Security:

As the handling of sensitive aviation accident data requires a delicate balance between research needs and privacy protection, future studies should delve into privacy-preserving ML techniques and secure datasharing mechanisms. Striking this balance will ensure ethical data use and enhance collaboration while safeguarding individual privacy. An example of a privacy-preserving ML technique is differential privacy. Differential privacy aims to provide guarantees that the output of a computation does not reveal sensitive information about any individual data point in the dataset [104]. By adding noise to the data or query responses in a carefully controlled manner, differential privacy enables the analysis of sensitive datasets while protecting the privacy of individuals. Additionally, secure multi-party computation (SMPC) protocols allow multiple parties to jointly compute a function over their inputs while keeping those inputs private [105]. These techniques, among others, offer promising avenues for ensuring privacy and data security in aviation accident data analysis.

#### 2.6.5.6. Human-Machine Interfaces for Safety Professionals:

Future research directions should emphasize the integration of human factors considerations into MLdriven accident analysis. Recognizing the significant role of human errors in accidents and incorporating cognitive psychology and human factors expertise will lead to more actionable insights.

5.5.7. Regulatory Implications: The increase of ML techniques in accident analysis necessitates the formulation of regulatory frameworks that guide their ethical and effective usage. Future research should actively engage in developing these frameworks and strategies to facilitate the widespread adoption of ML-driven accident analysis across the aviation sector.

## 2.7 Conclusion

In conclusion, this comprehensive systematic literature review has shed light on the potential of ML applications in enhancing aviation safety. The integration of ML techniques into post-accident analysis and safety management has ushered in a new era of proactive and data-driven safety practices. By synthesizing a wide array of studies conducted over a quarter-century span, the impact of ML across various aviation contexts has been highlighted. The examination of diverse ML algorithms, data sources, and applications demonstrates their adaptability and effectiveness in addressing the myriad challenges of aviation safety. From classification and anomaly detection to predictive modeling and real-time incident identification, ML techniques offer a powerful toolkit for stakeholders in the aviation industry to bolster safety measures in seven key safety areas (see Section 2.5). Moreover, the discussion on contextual nuances and ethical considerations reaffirms the importance of understanding the complex interplay between ML techniques and the aviation environment. Transparency, fairness, and bias mitigation are paramount as ML becomes increasingly intertwined with safety-critical decision-making.

Looking ahead, seven future research directions are identified in key areas where further exploration can drive advancements in ML-driven aviation safety analysis (see Section 2.5). Pursuing these directions should continue to propel the aviation industry toward safer skies while fostering innovation, collaboration, and ethical compliance. In the dynamic landscape of aviation safety, ML stands as a steadfast ally, augmenting human expertise and enabling data-driven insights that were once inconceivable. As one navigates the complexities of modern aviation, it is hoped that this review serves as a compass, guiding researchers, practitioners, and policymakers toward a future where safety is not only a priority but an unwavering reality. Other important safety-monitored areas like healthcare, autonomous vehicles, and industrial manufacturing should be encouraged to follow aviation's lead in adopting AI ML for its safety and beneficence.

Table 4. General description of the 88 systematized Articles

| ID | Reference | Title | Algorithms used | Number of Accidents(period) | ML Tasks | Data Source |
|---|---|---|---|---|---|---|
| 1. | [52] | Setting up new standards in the aviation industry with the help of artificial intelligence – machine learning application | SVM, RF, NB | 3000 (20years) | Classification, Anomaly Detection | ASRS |
| 2. | [106] | Prediction of warning level in aircraft accidents using data mining techniques | DT, KNN, SVM, NN and NB, PCA | -(42years) | Classification, Data reduction | Airline databases |
| 3. | [53] | Prediction of Warning Level in Aircraft Accidents Using Classification Techniques: An Empirical Study | DT, KNN, SVM, NN, NB | 500(62years) | Classification | FAA |
| 4. | [54] | Large-scale data analysis on aviation accident database using different data mining techniques | DT, NB, SVM, KNN, NN | | Clustering | |
| 5. | [107] | Feature selection techniques for prediction of warning level in aircraft accidents | PCA, DT | 500(62years) | Classification | FAA |
| 6. | [108] | Enabling the Discovery of Recurring Anomalies in Aerospace Problem Reports using High-Dimensional Clustering Techniques | SVM, NB, LDA, LR, ADT | 62(-) | Clustering | ASRS |
| 7. | [96] | Text mining of accident reports using semisupervised keyword extraction and topic modeling | LDA | 37,678(9years) | Data Reduction | ASRS, PHMSA |
| 8. | [109] | An Ensemble Machine and Deep Learning Model for Risk Prediction in Aviation Systems | DL, SVM, NB | 6(12years) | Classification | ASRS |
| 9. | [110] | Analyzing Aviation Safety Reports: From Topic Modeling to Scalable Multi-Label Classification. | LDA, BMR | 66309(-) | Classification | ASRS |
| 10. | [111] | Knowledge Graph–Deep Learning: A Case Study in Question Answering in the Aviation Safety Domain | DT | 4000(53years) | NLP | NTSB |
| 11. | [112] | Deep learning for extracting word-level meaning from safety report narratives | DT | 186,000(39years) | NLP | ASRS |
| 12. | [113] | A state-based approach to modeling general aviation accidents | - | 6180(33years) | Classification | NTSB |
| 13. | [114] | Natural Language Processing of Aviation Safety Reports to Identify Inefficient Operational Patterns | DT | 4195(23years) | Anomaly Detection | ASRS |
| 14. | [115] | Predicting General Aviation Accidents Using Machine Learning Algorithms | DT, NN, RF, LR, GB | 27,786(20years) | Classification | NTSB |
| 15. | [97] | Augmenting topic findings in the NASA aviation safety reporting system using topic modeling | LDA | 100(3years) | NLP | ASRS |
| 16. | [116] | Incorporation of Pilot Factors into Risk Analysis of Civil Aviation Accidents from 2008 to 2020: A Data-Driven Bayesian Network Approach. | BN | 163(12years) | Classification | NTSB |
| 17. | [117] | Multi-concept document classification using a perceptron-like algorithm. | Perceptron-like algorithm | - | Classification | ASRS |
| 18. | [77] | Computational Solution to Prevent Aeronautic Accidents Cause by Wake Turbulence Using Machine Learning | KNN, NN, DT, NB | (1year) | Anomaly Detection | EUROCONTROL |
| 19. | [118] | Deep learning-based approach for civil aircraft hazard identification and prediction. | RNN, KNN, LSTM, NN | 1244(2years) | Anomaly Detection | ACARS |
| 20. | [119] | Machine learning for helicopter accident analysis using supervised classification: Inference, prediction, and implications | KNN, DT, ADT, RF, NB, DNN | 13,055(11years) | Classification | NTSB |
| 21. | [93] | Advanced text mining algorithms for aerospace anomaly identification. | NN, NB, ADT, LDA, LR, SVM | 9910(24years) | Anomaly Detection | ASRS |
| 22. | [55] | The effect of COVID-19 on self-reported safety incidents in aviation: An examination of the heterogeneous effects using causal machine learning | RF | 7,246(2years) | Anomaly Detection | ASRS |
| 23. | [120] | Hybrid safety analysis method based on SVM and RST: An application to carrier landing of aircraft | SVM, RST | 635 | Classification | NADC |
| 24. | [121] | Textual indicator extraction from aviation accident reports | SVM, DNN | 61687(35years) | NLP | NTSB |
| 25. | [100] | Civil aviation safety evaluation based on deep belief network and principal component analysis. | DBN, PCA | 0(5years) | Classification, Data Reduction |  |
| 26. | [122] | A hybrid data-driven approach to analyzing aviation incident reports | DNN, SVM | 64573(11years) | NLP, Classification | ASRS |
| 27. | [30] | Ensemble machine learning models for aviation incident risk prediction | DNN, SVM | 64573(11years) | Classification | ASRS |
| 28. | [123] | Bayesian network modeling of accident investigation reports for aviation safety assessment. | BN | 2243(24years) |  | NTSB |
| 29. | [124] | Classification of aviation safety reports using machine learning. | RF, SVM, KNN, NN, NB | 73,000(5years) | Classification | ICAO |
| 30. | [125] | The analysis of fatal aviation accidents more than 100 dead passengers: an application of machine learning | SVM, NN, PCA, LR, DT | 220 | Classification, Data Reduction | ICAO |
| 31. | [126] | Semi-supervised learning with semantic knowledge extraction for improved speech recognition in air traffic control | DNN | 6004 | NLP | ASR |
| 32. | [127] | Predicting airline crash due to bird strike using machine learning | DT, KNN, NB | - | Classification | NTSB |
| 33. | [128] | Deep learning-based Time Series Forecasting of Go-around Incidents in the National Airspace System | LSTM | 3835(24years) | Classification | ASRS |
| 34. | [78] | An Innovative Approach to Modeling Aviation Safety Incidents | CNN, LSTM, SVM, RF, NB, LR | 158,070 | Classification | ASRS |
| 35. | [88] | Flight crash investigation using data mining techniques | K-Mean | 5268(101years) | Clustering | - |
| 36. | [129] | Applying Distilled BERT for Question Answering on ASRS Reports | BERT | 1625738(43years) | NLP | ASRS |
| 37. | [130] | Application of Machine Learning to mapping Primary Causal Factors in self-reported safety narratives | LSA | 7484(4years) | NLP | ASRS |
| 38. | [131] | Visual representation of safety narratives | LSA | 4497(2years) | NLP | ASRS |
| 39. | [132] | Temporal topic modeling applied to aviation safety reports: A subject matter expert review | LDA | 64,776(14years) | Clustering | ASRS |
| 40. | [56] | Data mining application on aviation accident data for predicting topmost causes of accidents | NN, SVM, KNN, DT, NB | 1,610 (95years) | Classification | FAA |
| 41. | [92] | Application of machine learning techniques for incident-accident classification problem in aviation safety management | SVM, NB, DT, RF | 84262(57years) | Classification | NTSB |
| 42. | [81] | Learning Methods and Predictive Modeling to Identify Failure by Human Factors in the Aviation Industry. | NN, RF | 1105(10years) | Classification | ASN |
| 43. | [86] | Application of structural topic modeling to aviation safety data. | LDA | 386(8years) | NLP | ASRS, NTSB |
| 44. | [87] | Natural language processing-based method for clustering and analysis of aviation safety narratives. | PCA, K-Means | 13,336(10years) | Clustering | ASRS |
| 45. | [133] | A textual analysis of dangerous goods incidents on aircraft. | SVDA | 383(10years) | NLP | ASRS |
| 46. | [76] | Prediction of injuries and fatalities in aviation accidents through machine learning | DT, KNN, SVM, NN | 31,974 (27years) | Classification | FAA |
| 47. | [10] | Prediction of aviation accidents using logistic regression model. | LR | 7415 | Classification | ASN |
| 48. | [134] | Airline Safety Data: How Predictable Are Accidents and Fatalities? | NN | 10(61years) | Classification | FAA |
| 49. | [98] | Aircraft safety analysis using clustering algorithms | K-Means | 1500(25years) | Clustering | - |
| 50. | [135] | Analysis of General Aviation fixed-wing aircraft accidents involving inflight loss of control using a state-based approach | SBA | 5726(18years) | Clustering | NTSB |
| 51. | [136] | Apriori algorithm for association rules mining in aircraft runway excursions. | AR | 434(10years) | Classification | ASN |
| 52. | [84] | Using correlation-based subspace clustering for multi-label text data classification | KNN | 15, 000 | Classification | ASRS, Reuters, 20Newsgroups |
| 53. | [79] | A model fusion strategy for identifying aircraft risk using CNN and Att-BiLSTM | CNN, BLSTM, DNN | 32(10years) | Classification | ASRS |
| 54. | [59] | Using Neural Networks to predict HFACS unsafe acts from the pre-conditions of unsafe acts | NN | 523(24years) | Classification | ROC |
| 55. | [58] | A data-mining approach to identification of risk factors in safety management systems | NB | 168227(24years) | Classification | ASRS |
| 56. | [137] | Causes and risk factors for fatal accidents in non-commercial twin engine piston general aviation aircraft. | LR | 376(10years) | Classification | NTSB |
| 57. | [95] | Examination of Aircraft Accidents That Occurred in the Last 20 Years in the World. | KNN, NB, DT, LR, GBM | 588(20years) | Classification | - |
| 58. | [138] | Classification of aviation accidents using data mining algorithms | DT, NB, SMO | 588(20years) | Classification | - |
| 59. | [102] | Predictive safety analytics: Inferring aviation accident shaping factors and causation | BN | 315(23years) | Classification | NTSB |
| 60. | [139] | Analysis of Helicopter Accidents and Certification Categories Using Machine Learning. | RF, DT | 1,576 (10years) | Classification | NTSB |
| 61. | [94] | Application of machine learning for aviation safety risk metric | GBM, RNN, SVM | 10,634(20years) | Classification | NTSB, MOR, ASIAS |
| 62. | [140] | Descriptive and predictive analyses of data representing aviation accidents. | DT, KNN, RF | 25,000(4years) | Classification | FAA |
| 63. | [141] | Flight Accident Modeling and Predicting Based on Least Squares Support Vector Machine | SVM | 40years | Classification | NTSB |
| 64. | [67] | Research on Aviation Safety Prediction Based on Variable Selection and LSTM | LSTM, LASSO | 138(10years) | Classification | ASRS |
| 65. | [142] | On the chaos analysis and prediction of aircraft accidents based on multi-timescales | SVM | 59511(55years) | Classification | NTSB |
| 66. | [143] | Critical parameter identification for safety events in commercial aviation using machine learning. | NB, RF, DT, KNN | 70 (6years) | Classification | FOQA |
| 67. | [144] | PIA Accidents Analysis Using Naïve Bayes Classifier | NB | 22(6years) | Classification | - |
| 68. | [145] | Failing &! Falling (F&! F): Learning to Classify | DT, NN | 137,236 | Classification | FAA |
| | | Accidents and Incidents in Aircraft Data | | | | |
| 69. | [146] | Subjectivity classification and analysis of the ASRS corpus | SVM, ADT | 140599(2years) | Classification | ASRS |
| 70. | [147] | Using random forests to diagnose aviation turbulence | RF, KNN, LR | 778(2years) | Classification | NTSB |
| 71. | [148] | Understanding general aviation accidents in terms of safety systems. | - | 2303(10years) | Classification | NTSB |
| 72. | [149] | Using Machine Learning Models to Study Human Error-Related Factors in Aviation Accidents and Incidents | NB, RF, LR, SVM, NN | 90000(47years) | Classification | NTSB |
| 73. | [150] | Using structural topic modeling to identify latent topics and trends in aviation incident reports. | LDA | 25,706(5years) | NLP | ASRS |
| 74. | [151] | Automated aviation occurrences categorization | NN | 12500(6years) | Classification | ASRS |
| 75. | [99] | Understanding large text corpora via sparse machine learning. | PCA, LDA, LASSO | 20000(4years) | NLP | ASRS |
| 76. | [152] | Sparse machine learning methods for understanding large text corpora. | PCA, LDA, LASSO | 20000(4years) | NLP | ASRS |
| 77. | [65] | Text Mining Classification and Prediction of Aviation Accidents Based on TF-IDF-SVR Method. | LDA, SVM, NB, RF, LR | 20000(59years) | NLP, Classification | NTSB |
| 78. | [153] | Cause identification from aviation safety incident reports via weakly supervised semantic lexicon construction | SVM | 140,599(9years) | Classification | ASRS |
| 79. | [154] | Analysis of Aviation Accidents Data. | RF, NB, KNN, DT, GBT | 19455(14years) | Classification | NTSB |
| 80. | [155] | Document classification using nonnegative matrix factorization and underapproximation. | NMF | 21519(1year) | Clustering | ASRS |
| 81. | [156] | Towards online prediction of safety-critical landing metrics in aviation using supervised machine learning | LSTM, NN, RF | 623 | Regression | FOQA |
| 82. | [157] | Identifying Incident Causal Factors to Improve Aviation Transportation Safety: Proposing a Deep Learning Approach | LSTM | 200,000(32years) | Classification | ASRS |
| 83. | [80] | Recent Experiences with Data Mining in Aviation Safety | DT | 1256(9years) | Classification | ASRS |
| 84. | [158] | Safer Approaches and Landings: A Multivariate Analysis of Critical Factors | DT, LR | 287(16years) | Classification | NTSB, ASRS |
| 85. | [85] | Multi-label asrs dataset classification using semi-supervised subspace clustering | KNN | 10000 | Clustering | ASRS, Reuters, 20Newsg roups |
| 86. | [159] | Sequential Classification of Aviation Safety Occurrences with Natural Language Processing. | LSTM, BLSTM, GRU, RNN | 27,000(15year) | Classification | NTSB |
| 87. | [69] | Classification of Survivor/Non-Survivor Passengers in Fatal Aviation Accidents: A Machine Learning Approach | NN, DT, PCA | 100(1year) | Classification, Data Reduction | BAAA |

Note: DT means decision trees; BN means Bayesian networks; NN means neural networks; SVM means support vectors machine; DL means deep learning; KNN means k-nearest neighbor; NB means naïve Bayes; PCA means principle component analysis; STS means Semantic Text Similarity; LR means logistic regression; LSTM means Long Short-Term Memory, Bidirectional Long Short-Term Memory (BLSTM), Gated Recurrent Unit (GRU), Recurrent Neural Network (RNN), Non-Negative Matrix Factorization (NMF) and Latent Dirichlet Allocation (LDA), Latent Semantic Analysis (LSA), Convolutional Neural Networks (CNN), Deep Neural Networks (DNN), Random Forest (RF), Rough Set Theory (RST), and Adaboost Decision Tree (ADT)

# Chapter 3. Applications of natural language processing in aviation safety A review and qualitative analysis

Nanyonga, A., Joiner, K., Turhan, U., & Wild, G. (2025). Applications of natural language processing in aviation safety: A review and qualitative analysis. In AIAA SCITECH 2025 Forum (p. 2153).

3.1. Abstract

This study explores the use of Natural Language Processing (NLP) in aviation safety, focusing on machine learning algorithms designed to enhance safety measures. There are currently (May 2024), 34 Scopus results from the keyword search “natural language processing” and “aviation safety.” Analyzing these studies allows us to uncover trends in the methodologies, findings, and implications of NLP in aviation. Both qualitative and quantitative tools have been used to investigate the current state of literature on NLP for aviation safety. Qualitative analysis summarizes the research motivations, objectives, and outcomes, showing how NLP can be utilized to help identify critical safety issues and improve aviation safety. This study also identifies research gaps and suggests areas for future exploration, providing practical recommendations for the aviation industry. We discuss challenges in implementing NLP in aviation safety, such as the need for large, annotated datasets, and the difficulty in interpreting complex models. We propose solutions like active learning for data annotation and explainable AI for model interpretation. Case studies demonstrate the successful application of NLP in improving aviation safety, highlighting its potential to make aviation safer and more efficient.

3.2. Introduction

It is anticipated that the future aerospace transportation system will involve a significant increase in traffic, both conventional and urban based [160]. There are a number of barriers to the safe integration of future technologies like urban air mobility [161], with automation being a key aspect. Data is also another key aspect, with ISSA intended to “mitigates risks before they can lead to an incident or accident” [162]. The ISSA is part of the designed IASMS. Investigation is an important aspect of safety assurance and SMS in general [163]. To take advantage of automation and the “big data” available, the use of artificial intelligence and machine learning is essential. Starting with more general applications in aviation and aerospace, such as onboard fault monitoring and diagnosis [164], intelligent decision support [165], as well as planning and operations [29], the growing application to safety is key [159].

Aviation safety is critical for the global aviation industry, where the primary goal is to minimize the number and severity of safety occurrences, either accidents or incidents, that may ultimately result in a loss of life and direct economic costs (hull losses etc.) [166]. The aviation industry prioritizing safety to ensure passenger and crew well-being and to maintain public trust in air travel [167]. To achieve this, postaccident analysis is essential in identifying the causes of aviation incidents and preventing future occurrences [168]. Traditionally, safety analysis has relied heavily on manual inspection and categorization of incident reports, a time-consuming process susceptible to human error [169]. Experts often manually review narratives, findings, and recommendations to find recurring patterns and contributing factors in accident reports; this approach, though valuable, is limited by human capacity, subjectivity, and the potential for oversight [170]. In recent years, the advent of natural language processing (NLP) and machine learning (ML) has provided new opportunities to enhance the analysis of aviation safety data [159, 170] offering more efficient and accurate ways to uncover insights from large volumes of text-based reports [63, 171, 172].

#### *3.2.1 Natural Language Processing*

NLP, a subfield of artificial intelligence (AI), focuses on the interaction between computers and human language. It enables the automatic extraction and processing of information from unstructured text, making it particularly suitable for analyzing aviation safety reports, which are often narrative-based and unstructured. ML algorithms, when combined with NLP techniques, can identify patterns, classify text, and predict outcomes based on the textual data available. These capabilities have significant implications for improving aviation safety by providing deeper insights into incident causation, operational inefficiencies, and emerging safety trends.

Various studies have demonstrated the application of NLP and ML in aviation safety. For instance, a study by Rose and colleagues [87] employed Singular Value Decomposition (SVD) to analyze dangerous goods incidents reported in the Aviation Safety Reporting System (ASRS), uncovering major topics related to such incidents. Similarly, Rankin et al. (2015) utilized Latent Semantic Analysis (LSA) to map primary causal factors in self-reported safety narratives, achieving notable accuracy in categorizing these narratives compared to expert huma results. Another significant work by Kuhn applied Structural Topic Modeling (STM) to ASRS and National Transportation Safety Board (NTSB) reports, highlighting the strengths of STM in identifying themes within technical datasets [150].

Advances in deep learning have further enhanced the potential of NLP in aviation safety. A study by Kierszbaum & Lapasset et al; [129] demonstrated the use of the Bidirectional Encoder Representations from Transformers (BERT) model for question answering on ASRS reports, achieving a 70% accuracy rate in extracting relevant information from free-text narratives. Additionally, Dong et al. [157] proposed a deep learning approach using Long Short-Term Memory (LSTM) networks to identify causal factors in incident reports, showing significant improvements over traditional methods.

#### *3.2.2 Air Transport System Safety*

NASA's system safety program is the application of engineering and management principles, criteria, and techniques to achieve acceptable mishap risk within the constraints of operational effectiveness and suitability, time, and cost throughout all phases of the system life cycle [173]. It is an integral part of the interdisciplinary approach of systems engineering and its pursuit of systems that meet stakeholder expectations [173]. The methods of System Safety are diverse and are driven by many factors, including the high cost of testing, increasing system complexity, the development of systems that operate at the edge of engineering capability, and the use of unproven technology [173].

Predictive safety for the air transport system is a part of the System-Wide Safety (SWS) project [174]. The transformation of aviation to make flights more efficient and accessible, even as the global demand for air transportation services steadily increases, will require new research tools, innovative technologies, and operational methods [174]. Enabling this to be done safely at every step is the goal of the SWS project [174]. Predictive safety management in aviation involves the use of predictive (forecasting) and causal modelling methods to identify potential and possible hazards in the future, as well as their causal factors which can help define timely and efficient mitigation measures to prevent or restrain emerging hazards turning into adverse events [175] . This methodology is used in various aviation sectors (air navigation services, airport operations, airline operations) for planning purposes [176]. It is a crucial mechanism to maintain and continuously improve safety levels in aviation organizations.

*3.2.3. Motivation and Significance*

The motivation behind integrating NLP into aviation safety is multi-faceted. First, it addresses the need for more accurate, proactive, and efficient data processing techniques that can handle the growing volume of safety reports. Second, it enhances the accuracy of safety analysis by reducing the subjectivity and potential biases inherent in manual categorization [177]. Third, it facilitates the early detection of safety trends and the identification of latent issues that may not be immediately apparent through conventional methods [178].

More importantly, it is a steppingstone towards qualitative predictive safety. That is, the development of language models and qualitative analytics using AI/ML facilitates not only contemporary quantitative predictive safety but enables sources of data like voice comms and ACARS messages, etc., to be used as sources of data that can be tracked over time looking for precursors to potential future safety occurrences. While the concept of "big brother" is not the goal, looking at an accident such as Colgan Air Flight 3407 [179], precursors to safety may very well be embedded in audio data. Similarly, text in check-and-training reports, or other continuous documentation may uncover issues. However, in the context of NLP for aviation safety occurrence reports, there are precursors in prior incidents that were near misses, which can be learnt from (as is currently done through the safety culture in aviation), and when similar situations are encountered during operations, a preventative measure may be taken to prevent an accident from happening, an artificial memetic algorithm.

*3.2.4. Aim and Research Objectives*

The overarching aim of this study is to critically evaluate the role of NLP in enhancing aviation safety through a systematic review and qualitative analysis of existing literature. By examining a selection of studies indexed in Scopus, this research seeks to synthesize the current state of knowledge regarding the application of NLP in aviation safety, identify gaps in the literature, and propose directions for future research. The associated research questions are:

1) What are the machine learning algorithms used for NLP in aviation safety, and how do they contribute to the field?

2) How does NLP facilitate the identification and analysis of critical safety issues within the aviation industry?
3) What challenges are associated with implementing NLP in aviation safety, and what solutions have been proposed or implemented to address these challenges?
4) In what ways can NLP contribute to the advancement of predictive safety management within the aviation sector?
5) What are the implications of NLP applications in aviation safety for industry stakeholders, and how can these insights inform policy-making and best practices?

### 3.3 Methodology

The methodology employed in this study involved a detailed examination of 34 elected papers focusing on NLP applications in aviation safety. The selection process was guided by a systematic literature review titled "A Systematic Review of Machine Learning Analytic Methods for Aviation Accident Research," currently under review for publication [180]. This review identified relevant papers through comprehensive searches on Scopus and backward reference searches via Google Scholar, resulting in the identification of 87 papers.

From this pool, the first seventeen papers were chosen for inclusion in this study based on their relevance to NLP applications in aviation safety. These papers were further filtered based on their utilization of data sources from authoritative aviation safety organizations such as the Aviation Safety Reporting System (ASRS), International Civil Aviation Organization (ICAO), National Transportation Safety Board (NTSB), Australian Transport Safety Bureau (ATSB), and Aviation Safety Network (ASN). Papers selected for inclusion were published up to the year 2022, consistent with the timeframe of the systematic literature review. Also, to ensure the timeliness of the findings, papers from 2023 and 2024 were also included, resulting in 34 papers reviewed.

#### *3.3.1 Data Collection*

Detailed information from each paper, including the year of publication, title, data source, sample size, aviation application, machine learning algorithms employed, comments, research questions/objectives, study design, outcomes/conclusions, and keywords, was systematically extracted.

#### *3.3.2 Statistical Analysis*

Word clouds were generated using free online tools such as WordItOut and Textalyser to visualize frequently occurring terms in the selected papers. This analysis provided insights into the prominent themes and topics addressed in the literature. Additionally, keyword counts were performed to identify prevalent themes and topics across the selected papers. This quantitative analysis helped in understanding the distribution and significance of key terms within the literature. VOSviewer was utilized to generate maps and networks based on the research objectives and abstracts from all the papers, facilitating a comprehensive visualization of the relationships and connections between key concepts and topics in the research [181, 182].

### *3.3.3 Study Design Considerations*

In the study design considerations, the research questions and objectives of each paper were systematically analyzed to understand the overarching goals and motivations behind the studies. Additionally, the methodologies employed in the selected papers were examined to identify common approaches and techniques used in NLP applications for aviation safety. The outcomes and conclusions drawn from each paper were carefully reviewed to extract insights and implications for future research and practice in the field.

## *3.4 Advancing Aviation Safety*

The integration of NLP into aviation safety research is driven by several compelling factors. The aviation industry generates much safety data through various reporting systems such as the ASRS and the NTSB reports. These reports, often rich in narrative content, provide detailed accounts of incidents and accidents. However, the unstructured nature of pure text narratives poses a significant challenge for traditional data analysis methods, which are typically designed for structured data. NLP offers a powerful solution by enabling the automated extraction and processing of meaningful information from large volumes of text, thereby enhancing the analysis of safety data.

### *3.4.1 Efficiency and Scalability*

One of the primary motivations for employing NLP in aviation safety is the need for more efficient and scalable data processing techniques. Manual analysis of safety reports is not only labour-intensive but also time-consuming, limiting the ability to promptly address emerging safety concerns. NLP techniques, such as topic modelling, clustering, and semantic analysis, can process vast amounts of text data quickly and accurately. For instance, a study by Walton et al. [133] demonstrated how SVD could be used to analyze dangerous goods incidents from the ASRS database, efficiently uncovering key topics and trends over ten years.

### *3.4.2 Accuracy and Objectivity*

Traditional methods of safety data analysis often rely on subjective judgment and manual categorization, which can introduce biases and inconsistencies. NLP algorithms, by contrast, offer a more objective approach to text analysis. Machine learning models such as LSA and BERT have been shown to achieve high levels of accuracy in categorizing and extracting information from safety narratives. For example, Robinson [130] utilized LSA to map primary causal factors in safety reports, demonstrating an unsupervised categorization accuracy of 44% within an existing taxonomy. Similarly, Kierszbaum & Lapasset et al. [129] applied BERT for question answering on ASRS reports, achieving roughly 70% correct answers, other studies [111, 183-187] also applied BERT to their studies thereby highlighting the model's capability to handle the complexity of natural language narratives.

#### 3.4.3 Early Detection and Proactive Safety Management

Another significant motivation for applying NLP in aviation safety is its potential to facilitate the early detection of safety trends and latent issues. By analyzing large datasets of safety reports, NLP can identify patterns and correlations that might not be immediately apparent through manual analysis. This proactive approach allows for the identification of emerging risks and the development of preventive measures before issues escalate into more severe incidents. For instance, studies such as [101, 112, 157, 159, 160, 170, 188, 189] applied Deep Learning (DL) algorithms to aviation safety data, identifying themes and correlations within safety narratives that could inform future safety studies and interventions.

#### 3.4.4 Enhancing Human Factors Analysis

Human factors play a crucial role in aviation safety, with many incidents and accidents attributed to human error. NLP techniques can significantly enhance the analysis of human factors by extracting relevant information from narrative reports and identifying underlying causes and contributing factors [184]. Dong et al. proposed a deep learning approach using LSTM networks to identify causal factors in incident reports, focusing on human error and other contributory elements. Their model demonstrated significant improvements in accuracy and efficiency compared to traditional analysis methods [157].

#### 3.4.5 Contribution to Knowledge and Best Practices

The application of NLP in aviation safety improves immediate safety outcomes and contributes to broader knowledge in the field. By providing a deeper understanding of safety issues and their underlying causes, NLP-driven research informs best practices and policymaking. Studies such [87, 114, 146] exemplify how NLP can generate valuable insights that enhance aviation safety protocols and decisionmaking processes.

## 3.5 Navigating Aviation Safety

The use of NLP in aviation safety has been explored across various dimensions, encompassing a range of ML algorithms and techniques tailored to different safety datasets. This chapter reviews the specific applications of NLP in aviation safety, drawing on insights from 34 selected studies. We examine the methodologies used, research objectives, study designs, and key outcomes to understand how NLP has been used to enhance aviation safety.

#### 3.5.1 Methodologies and Machine Learning Algorithms

Different NLP methodologies and machine learning algorithms have been applied to aviation safety data, each with its unique strengths and applications. The following sections detail these methodologies and their applications:

1) SVD: Walton and Marion utilized SVD to analyze dangerous goods incidents reported in the ASRS. SVD, a dimensionality reduction technique, was employed to identify key topics and trends within the safety narratives. This approach enabled the extraction of meaningful patterns from a large and complex dataset, demonstrating the utility of SVD in making sense of unstructured text data [133].

2) LSA: Robinson et al. applied LSA to map primary causal factors in self-reported safety narratives. LSA is a statistical method for extracting and representing the contextual usage meaning of words by analyzing relationships between a set of documents and the terms they contain. This technique proved effective in categorizing safety narratives and identifying primary causal factors, highlighting the potential of LSA in contextual and thematic evaluation of safety reports [130].

3) STM: Rose et al. [86] employed STM to analyze safety data from the ASRS and NTSB. STM is a probabilistic model that identifies topics within a set of documents and can accommodate metadata to improve topic coherence. This study demonstrated the capability of STM to uncover latent themes within technical datasets, which can inform safety interventions and policymaking (Rose et al., 2022).

4) BERT and DL: Kierszbaum and Lapasset explored the use of the BERT model for answering questions on ASRS reports. BERT, a transformer-based model, excels in understanding the context of words in search queries and is particularly effective for tasks requiring a deep understanding of language. Their study showed that BERT could accurately extract relevant information from free-text narratives, indicating its potential for improving information retrieval in safety databases [129]. Dong et al. [157] developed a deep learning approach using LSTM networks to identify causal factors in incident reports. LSTM networks are well-suited for sequence prediction tasks and can effectively handle the complexity of narrative data. This approach enhanced the identification of primary and contributing factors, demonstrating the advantages of deep learning in the causal analysis of aviation incidents [157].

#### *3.5.2 Research Objectives and Study Designs*

The research objectives and study designs across the reviewed studies varied, reflecting the diverse applications of NLP in aviation safety. Common objectives included improving the efficiency of data processing, enhancing the accuracy of safety analysis, and identifying latent safety issues. The following sections summarize the research questions and study designs employed in the selected studies.

1) Improving Data Processing Efficiency: Several studies aimed to develop frameworks and methodologies to process safety data more efficiently. For instance, Walton and Marion focused on establishing a data-driven framework using SVD to analyze dangerous goods incidents. This study highlighted the potential of NLP techniques to handle large datasets and uncover significant trends without extensive manual intervention [133].

2) Enhancing Accuracy of Safety Analysis: Other studies aimed to improve the accuracy and objectivity of safety analysis. Robinson et al. sought to evaluate the effectiveness of LSA in categorizing safety narratives and identifying primary causal factors. Their findings demonstrated that NLP could enhance the precision of safety data analysis, reducing the reliance on subjective judgment [130].

3) Identifying Latent Safety Issues: In his study, Kuhn aimed to identify latent topics and trends within aviation safety reports using STM. This approach revealed previously unreported connections and themes, underscoring the capability of NLP to detect emerging risks and inform proactive safety management [150].

#### *3.5.3 Key Outcomes and Implications*

The reviewed studies collectively highlight the significant contributions of NLP to aviation safety. Key outcomes include improved data processing efficiency, enhanced accuracy in safety analysis, and the identification of latent safety issues. These outcomes have important implications for the aviation industry, suggesting that NLP can transform safety management practices by providing deeper insights into safety data.

For instance, the study by Kierszbaum and Lapasset [129] demonstrated that BERT could accurately extract relevant information from safety reports, suggesting its potential for improving the efficiency of information retrieval in safety databases. Similarly, the work by Dong et al. [157] showed that deep learning models could effectively identify causal factors in incident reports, providing valuable insights for preventive measures and safety interventions.

### 3.6 Quantitative Review

We used the *Satisfy* app (*https://statisty.app/)*, a publicly available online tool, for statistical analysis of the selected papers. Figure 15 shows the Receiver Operating Characteristic (ROC) curve, illustrating the sensitivity versus specificity of the selected words, with an Area Under the Curve (AUC) of 0.879, indicating a good level of discrimination between relevant and irrelevant words in the 34 NLP papers. Figure 16 presents a Sankey diagram illustrating the flow of data sources, aviation applications, and machine learning algorithms identified in the study. The ASRS (Aviation Safety Reporting System) was the most utilized data source, featuring in more than 20 studies (57% of the total), all aviation constituted 27 out of the 34 studies, followed by airline-specific studies with 6, and cargo-related studies with 1. DL and Latent Dirichlet Allocation (LDA) were the most frequently used algorithms, each appearing in 23% of the studies, with BERT used in 20% of the studies.

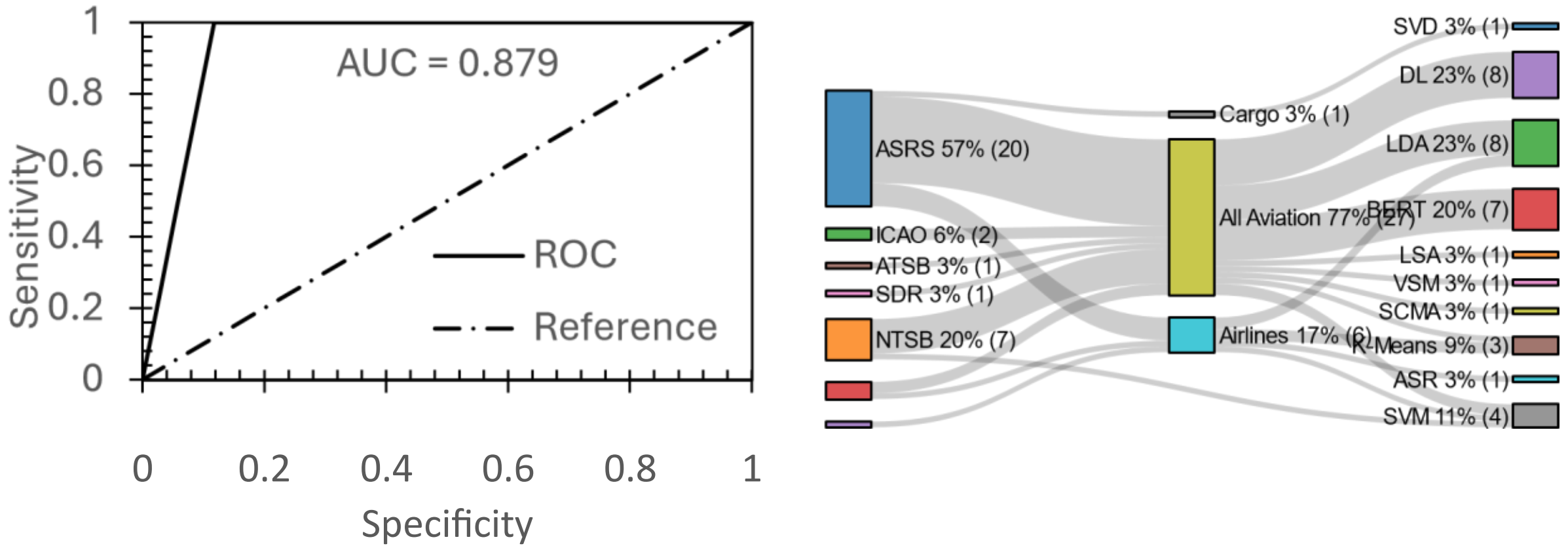

Figure 16. Sankey Diagram

Figure 15. ROC Curve

Figure 17 shows the distribution of studies applying different ML algorithms across various aviation applications, highlighting the diverse application of ML techniques in enhancing aviation safety. Figure 20 and Figure 21, illustrate the most frequently used words across the studies, with "NLP" appearing most frequently with over 17 mentions, followed by "ASRS" (12 mentions), "Text mining" (9 mentions), and "Aviation safety" (8 mentions).

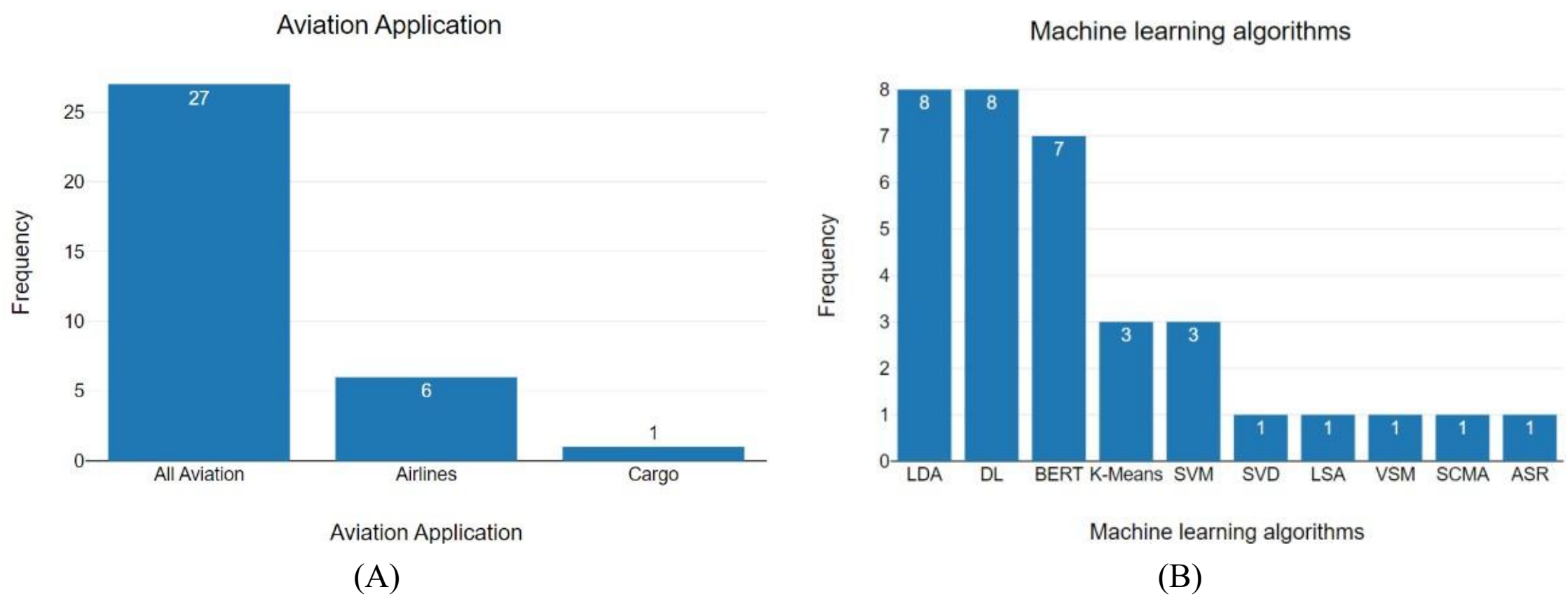

Figure 17. The number of studies applying (A) different ML algorithms, across (B) various aviation applications

Next, we employed VOSviewer for bibliometric analysis of the keywords in the selected publications related to NLP and aviation safety. Figure 18 presents the co-occurrence of keywords within the same publications. The size of the nodes represents the frequency of occurrence, while the curves between nodes indicate their co-occurrence; the heat map in Figure 18 (b) also illustrates the relative size of the nodes as a hotter colour. Keywords such as “natural language processing”, “aviation safety”, and “text mining” were the most frequent, while terms like “fires” and “nasa” were less common. The shorter distance between nodes indicates a higher number of co-occurrences, suggesting strong thematic connections between frequently co-occurring keywords. Figure 19 shows a Network visualizing the frequency of all keywords. The font size in the word cloud represents the frequency of occurrence, with terms like “natural language processing”, “aviation safety”, and “text mining” dominating, reflecting their central importance in the reviewed studies. These analyses highlight the prevalent themes and methodological trends in applying NLP and machine learning to aviation safety, emphasizing the field's complexity and the necessity for robust analytical approaches. Table 5 shows Aviation Applications across various ML algorithms while Table 6 summarizes different key findings identified from Aviation Safety Reports

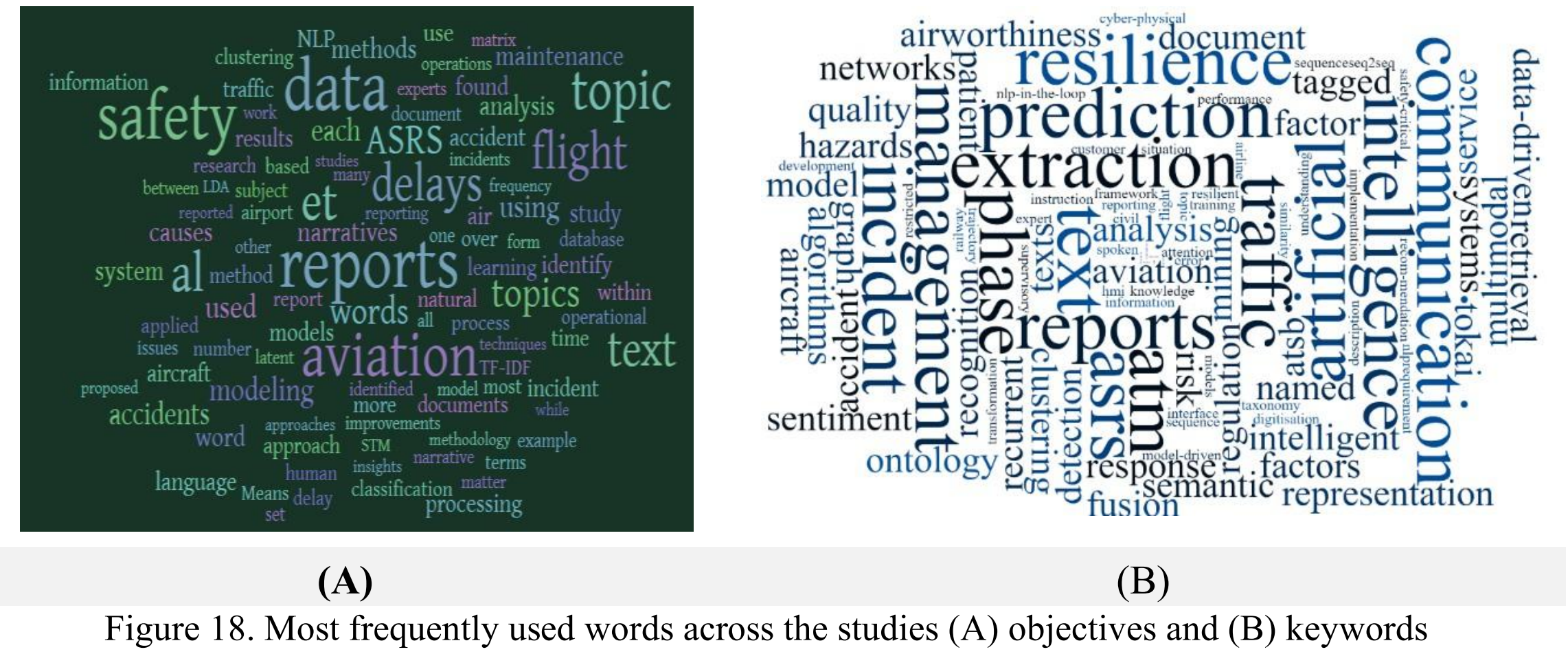

Figure 18. Most frequently used words across the studies (A) objectives and (B) keywords

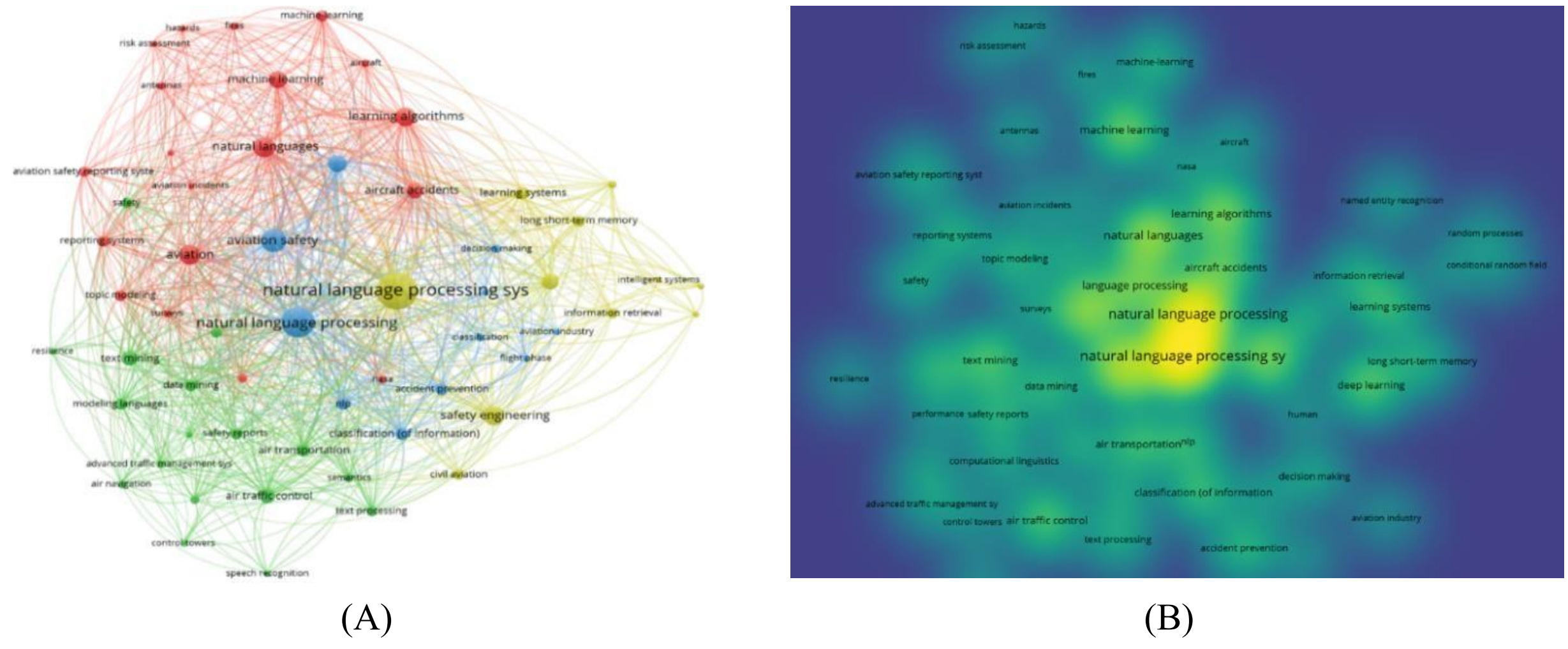

Figure 19. Shows the co-occurrence of words (A) Network and (B) mapping within the same publications.

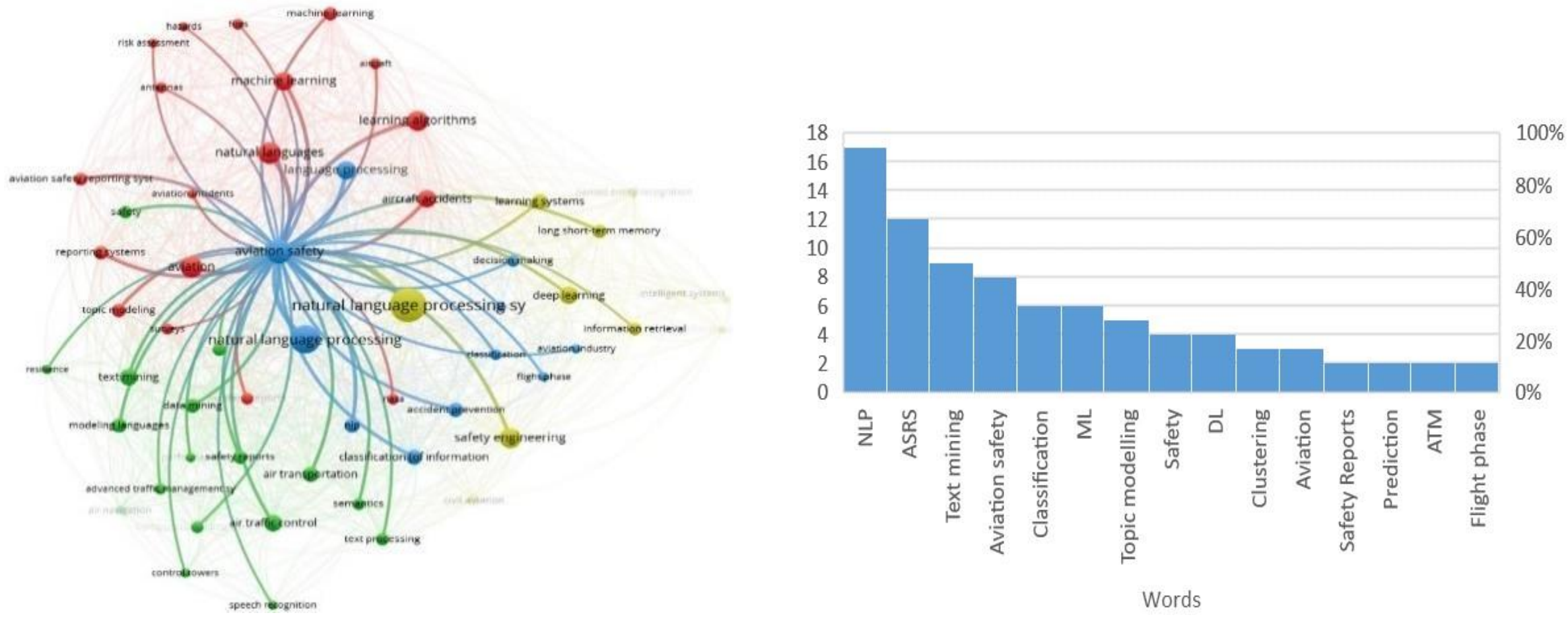

Figure 20. Network Visualization of Materials Figure 21. Word frequency in the 34 studies Keyword

Table 5. Distribution of ML algorithms and Aviation application

| | SVD | LSA | LDA | BERT | DL | K-Means | SCMA | ASR | SVM | VSM | Total |
|---|---|---|---|---|---|---|---|---|---|---|---|
| Cargo | 1 | 0 | 0 | 0 | 0 | 0 | 0 | 0 | 0 | 0 | 1 |
| All Aviation | 0 | 1 | 7 | 7 | 8 | 1 | 1 | 0 | 1 | 1 | 27 |
| Airlines | 0 | 0 | 2 | 0 | 0 | 2 | 0 | 1 | 2 | 0 | 7 |
| Total | 1 | 1 | 9 | 7 | 8 | 3 | 1 | 1 | 3 | 1 | 35 |

Table 6: ML Algorithms Applied to Aviation Safety Reports

| **Authors** | **Year** | **Data** | **Application** | **Algorithms** |
|---|---|---|---|---|
| Walton & Marion [133] | 2020 | ASRS | Cargo | SVD |
| Robinson et al; [130] | 2015 | ASRS | All Aviation | LSA |
| Rose et al; [86] | 2022 | ASRS | Airlines | LDA |
| Kierszbaum & Lapasset [129] | 2020 | ASRS | All Aviation | BERT |
| Paradis et al; [97] | 2021 | ASRS | All Aviation | LDA |
| Chanen [112] | 2016 | ASRS | All Aviation | DL |
| Dong et al; [157] | 2021 | ASRS | All Aviation | DL |
| Agarwal et al; [111] | 2022 | NTSB | All Aviation | BERT |
| Nanyonga et al; [170] | 2021 | NTSB | All Aviation | DL |
| Rose et al; [87] | 2020 | ASRS | Airlines | K-Means |
| Perboli et al; [190] | 2021 | ICAO | All Aviation | VSM |
| Miyamoto et al; [114] | 2022 | ASRS | Airlines | K-Means |
| Nanyonga et al; [159] | 2023 | NTSB | All Aviation | DL |
| Robinson, [132] | 2019 | ASRS | Airlines | LDA |
| Zhao, et al; [65] | 2022 | NTSB | All Aviation | LDA |
| Abdhul et al; [96] | 2021 | ASRS | All Aviation | LDA |
| Kuhn, [150] | 2018 | ASRS | All Aviation | LDA |
| Tanguy et al; [24] | 2016 | ASRS | All Aviation | SVM |
| Madeira et al; [191] | 2021 | ASN | Airlines | SVM |
| Nanyonga et al; [170] | 2023 | NTSB | All Aviation | DL |
| Luo & Shi [64] | 2019 | ASRS | All Aviation | LDA |
| Kierszbaum et al; [183] | 2021 | ASRS | All Aviation | BERT |
| Switzer et al; [146] | 2011 | ASRS | All Aviation | SCMA |
| Buselli et al; [192] | 2022 | ICAO | All Aviation | LDA |
| Rose et al; [87] | 2020 | ASRS | All Aviation | K-Means |
| Abdullah et al; [193] | 2017 | Airline | Airlines | ASR |
| Nanyonga et al; [188] | 2023 | ATSB | All Aviation | DL |
| Persing [194] | 2009 | ASRS | All Aviation | SVM |
| Ono & Nakanishi [184] | 2023 | ASRS | All Aviation | BERT |
| Mbaye et al; [185] | 2023 | ASRS | All Aviation | BERT |
| Nanyonga & Wild [101] | 2023 | ASN; NTSB | All Aviation | DL |
| Nanyonga et al; [189] | 2023 | ASN | All Aviation | DL |
| Niraula et al; [186] | 2023 | SDR | All Aviation | BERT; SVM |

| Gao et al; [187] | 2024 | NTSB | All Aviation | BERT |
|---|---|---|---|---|

## 3.7 Discussion

### 3.7.1 Key Trends and Methodologies

The synthesis of findings from the reviewed studies reveals several key trends and insights into the application of NLP in aviation safety. Across the studies, common methodologies and machine learning algorithms were employed to analyze textual data from various aviation safety sources, including the ASRS and the NTSB. The effectiveness of these methodologies varied based on the specific objectives of each study and the nature of the aviation applications examined.

#### *3.7.1.1 Topic Modeling Techniques*

Several studies utilized topic modelling techniques such as LDA and STM to identify latent topics and themes within aviation safety narratives. These approaches enabled researchers to uncover underlying patterns and trends in safety reports, facilitating a deeper understanding of safety-related issues and contributing to proactive risk mitigation efforts.

#### *3.7.1.2 Deep Learning Models*

The application of deep learning models, including LSTM networks and BLSTM, demonstrated promising results in extracting meaningful insights from aviation safety narratives. These models enhanced the accuracy of incident classification and causal factor identification, providing valuable support for safety analysis and decision-making processes.

#### *3.7.1.3 Semantic Analysis and Word Embeddings*

Studies leveraging semantic analysis techniques such as LSA and word embeddings (e.g., word2vec) highlighted the ability to capture word-level meaning and semantic relationships within safety reports. By quantifying semantic similarities and associations, these approaches facilitated a more nuanced analysis of safety narratives and improved the interpretability of model outputs.

#### *3.7.1.4 Implications for the Industry*

The findings from the reviewed studies have significant implications for the aviation industry, particularly in enhancing safety management practices and accident prevention strategies. By leveraging NLP technologies and machine learning algorithms, aviation stakeholders can:

- Gain deeper insights into safety-related incidents and causal factors.
- Prioritize safety interventions based on identified trends and patterns.
- Improve the efficiency of safety reporting and analysis processes.
- Enhance predictive capabilities for identifying potential safety risks and hazards.
- Facilitate evidence-based decision-making and proactive risk mitigation efforts.

### 3.7.2 Research Gaps and Future Directions

Despite significant advancements in applying NLP to aviation safety, several research gaps remain that warrant further exploration. This chapter identifies these gaps and proposes potential areas for future research, along with specific questions and topics that should be addressed.

#### *3.7.2.1 Research Gaps*

One significant gap is the limited integration of multimodal data sources. Most current studies focus exclusively on textual data from safety reports, neglecting other valuable data types such as audio recordings, sensor data, and video footage. Integrating these diverse data sources could provide a more comprehensive understanding of safety incidents and help identify underlying causes that are not apparent from text alone.

Another gap is the underutilization of real-time data. Most research deals with historical data, with minimal emphasis on real-time data analysis. Real-time processing and analysis could enable proactive safety measures and immediate response strategies, significantly enhancing aviation safety management. The ability to detect and address safety issues as they occur, rather than retrospectively, could lead to more effective interventions and reduced incident rates. The current literature also shows an insufficient focus on human factors. Although some studies have explored the role of human factors in aviation safety, this area remains underdeveloped. There is a need for a more in-depth analysis of cognitive and behavioral aspects to understand their impact on safety incidents comprehensively. Factors such as fatigue, decisionmaking processes, and communication dynamics are critical to aviation safety but are not adequately addressed by existing NLP models.

Additionally, there is a lack of research encompassing diverse geographic and cultural contexts. Most studies are concentrated in specific regions, particularly the United States and Europe, with limited research on aviation safety in other parts of the world. Expanding research to include a variety of geographic and cultural contexts could uncover unique safety challenges and solutions applicable to different regions. This diversity would help develop more universally applicable safety protocols and interventions.

Another significant gap is the lack of standardized evaluation metrics. There is a considerable inconsistency in how the effectiveness of NLP models in aviation safety applications is measured, making it challenging to compare results across different studies and identify best practices. Standardizing these metrics would facilitate better benchmarking and improvement of NLP techniques in the field.

#### *3.7.2.2 Future Directions*

Future research should focus on the integration of multimodal data. Methods for combining textual, audio, video, and sensor data should be developed to provide a holistic view of safety incidents. Techniques such as multimodal deep learning and data fusion could be employed to achieve this integration, leading to more comprehensive and accurate safety analyses.

Real-time data analysis represents another promising area for future exploration. Developing frameworks and algorithms capable of handling streaming data from various sources in real-time can enable immediate safety interventions and improve situational awareness. Research should focus on creating robust real-time data processing systems that can detect and respond to safety issues as they occur, potentially preventing incidents before they escalate.

Comprehensive studies on human factors should be prioritized. Future research should delve deeper into cognitive and behavioral aspects of aviation safety, using NLP to analyze how these factors influence safety outcomes. Investigating the impact of human factors such as fatigue, stress, communication breakdowns, and decision-making processes could provide valuable insights into preventing human errorrelated incidents.

Expanding research to include diverse geographic and cultural contexts is also essential. By conducting studies in various regions and cultural settings, researchers can uncover unique safety challenges and develop solutions that are globally applicable. This approach would help create more effective and inclusive safety protocols and interventions.

Standardizing evaluation metrics for NLP models in aviation safety applications should be another focus. Establishing consistent benchmarks for assessing the performance of these models would facilitate better comparison of research outcomes and drive improvements in NLP techniques. Developing a set of agreed-upon metrics would enhance the reliability and validity of research findings in this field.

#### *3.7.2.3 Specific Questions and Topics for Future Research*

To address these research gaps, future studies should consider the following questions and topics:

1. How can multimodal data integration improve the accuracy and comprehensiveness of aviation safety analyses?
2. What frameworks and algorithms are most effective for real-time data processing in aviation safety?
3. What are the critical human factors influencing aviation safety, and how can they be effectively analyzed using NLP?
4. How do aviation safety challenges and solutions vary across different geographic and cultural contexts?
5. What standardized metrics should be developed to evaluate the effectiveness of NLP models in aviation safety applications?

### 3.7.3 Practical Recommendations

*3.7.3.1. Integration of NLP into Safety Management Systems*

NLP can significantly enhance safety management systems by enabling the analysis of textual data from safety reports. By integrating NLP tools into existing systems, aviation stakeholders can extract valuable insights from narrative data, facilitating more informed decision-making and proactive safety measures. For instance, NLP algorithms can analyze incident reports, identify recurring patterns, and prioritize corrective actions based on risk assessment.

*3.7.3.2. Overcoming Challenges in Data Annotation and Model Interpretation*

An important challenge in implementing NLP solutions is annotating data required for model training. To address this, the industry can develop standardized annotation guidelines and leverage automated annotation tools to streamline the process. Additionally, incorporating expert review and feedback mechanisms can ensure the accuracy and relevance of annotated data, thereby enhancing the quality of NLP models. Ensuring transparency and interpretability of NLP models is essential for gaining stakeholders' trust and facilitating model adoption. To achieve this, aviation organizations should employ interpretable machine learning methods and visualization techniques to make model outputs understandable to safety analysts. Techniques such as SHAP and LIME can provide insights into the underlying mechanisms of NLP models, aiding in model interpretation and decision-making.

*3.7.3.3 Guidance on Model Integration, Deployment, and Capacity Building*

Effective integration and deployment of NLP models require careful consideration of organizational requirements and constraints. Aviation stakeholders should develop robust model integration frameworks and establish clear guidelines for model deployment. This includes defining model performance metrics, establishing protocols for model updates and maintenance, and ensuring compliance with regulatory standards and industry best practices. Continued investment in research and development is crucial for advancing NLP applications in aviation safety. Industry stakeholders should collaborate with academic institutions and research organizations to explore emerging NLP techniques and develop tailored solutions for aviation-specific challenges. This includes investigating novel approaches for multimodal data analysis, real-time data processing, and human factors modelling. Building internal capacity and expertise in NLP is essential for the successful implementation and utilization of NLP solutions within the aviation industry. Organizations should invest in training programs and workshops to upskill personnel in NLP techniques, data analytics, and machine learning. This will enable aviation professionals to effectively leverage NLP tools and extract actionable insights from textual data.

### 3.8 Case Studies

In recent years, several case studies have demonstrated the successful application of natural language processing (NLP) in enhancing aviation safety practices. These examples highlight the tangible benefits of integrating NLP technologies into safety management systems and provide valuable insights for the broader industry.

### *3.8.1 Incident Report Analysis with NLP*

One notable case study involved the application of NLP algorithms to analyze incident reports from the Aviation Safety Reporting System (ASRS) [133]. By leveraging advanced text mining techniques, researchers extracted key insights from narrative data, including identifying latent safety risks and contributing factors. The outcomes of this study demonstrated the effectiveness of NLP in uncovering hidden patterns and trends within safety reports, enabling proactive safety measures to be implemented.

### *3.8.2 Predictive Maintenance Using NLP*

Another compelling case study focused on the use of NLP for predictive maintenance in aviation [114]. By analyzing maintenance records and aircraft performance data, researchers developed NLP-based predictive models to forecast equipment failures and maintenance requirements. The application of NLP facilitated early detection of potential issues and optimized maintenance schedules, resulting in cost savings and operational efficiency improvements.

### *3.8.3 Safety Culture Assessment through Textual Data Analysis*

Additionally, researchers have utilized NLP techniques to assess safety culture within aviation organizations by analyzing textual data from employee reports and feedback [132]. By analyzing the language used in safety-related communications, NLP algorithms can identify sentiments, attitudes, and perceptions related to safety practices and organizational culture. Insights derived from these analyses can inform targeted interventions and training programs aimed at enhancing safety culture and promoting proactive safety behaviours among aviation personnel.

### *3.8.4 Relevance to the Broader Industry*

These case studies underscore the relevance and potential of NLP in addressing key challenges faced by the aviation industry in ensuring safety and operational excellence. By harnessing the power of NLP, aviation stakeholders can gain deeper insights from textual data sources, enabling more informed decisionmaking, proactive risk management, and continuous improvement initiatives. As NLP technologies continue to evolve, the opportunities for their application in aviation safety are expected to expand, driving further advancements in safety management practices, and contributing to the industry's overarching goal of enhancing safety for all stakeholders.

## 3.9 Challenges and Solutions

Implementing natural language processing (NLP) in aviation safety presents various challenges stemming from the complexity of textual data, regulatory requirements, and organizational constraints. Addressing these challenges requires innovative solutions and a multi-faceted approach that combines technological advancements with methodological refinements.

#### *3.9.1. Data Complexity and Quality*

Aviation safety data, including incident reports and maintenance records, often exhibit complexity and variability in language use, making it challenging to extract meaningful insights. To address this challenge, aviation stakeholders can leverage advanced NLP techniques such as deep learning models and semantic analysis algorithms [157]. These technologies can effectively process unstructured textual data, identify relevant information, and extract actionable insights, even from diverse and complex datasets.

#### *3.9.2. Regulatory Compliance and Privacy Concerns*

Aviation organizations must adhere to stringent regulatory requirements governing data privacy and security, complicating the implementation of NLP solutions that involve sensitive information. To mitigate regulatory risks and privacy concerns, organizations can adopt privacy-preserving NLP techniques such as federated learning and differential privacy [111]. These approaches allow for collaborative model training across distributed data sources while preserving the confidentiality of sensitive information and ensuring compliance with regulatory standards.

#### *3.9.3. Scalability and Integration*

Integrating NLP solutions into existing safety management systems and workflows requires seamless integration with legacy infrastructure and processes, posing scalability challenges. To address scalability concerns, aviation stakeholders can adopt modular and interoperable NLP platforms that facilitate seamless integration with existing systems [129]. Additionally, cloud-based NLP services offer scalability and flexibility, enabling organizations to leverage on-demand computational resources and scale NLP capabilities according to evolving needs.

#### *3.9.4. Expertise and Training*

Building internal expertise in NLP techniques and methodologies is essential for successful implementation and utilization but requires specialized skills and training. Organizations can invest in comprehensive training programs and workshops to upskill personnel in NLP techniques, data analytics, and machine learning [159]. Collaborating with academic institutions and industry experts can also facilitate knowledge transfer and skill development, enabling aviation professionals to effectively leverage NLP tools and techniques for safety management narrative.

#### *3.9.5. Model Interpretability and Transparency*

Ensuring the interpretability and transparency of NLP models is critical for gaining stakeholders' trust and facilitating model adoption but can be challenging due to the inherent complexity of deep learning models. Aviation organizations can employ interpretable machine learning techniques and visualization methods to enhance model interpretability [190]. Techniques such as SHAP (SHapley Additive exPlanations) and LIME (Local Interpretable Model-agnostic Explanations) provide insights into model predictions and help stakeholders understand the underlying factors driving model outputs, thereby improving transparency, and fostering trust.

### 3.10 Conclusion

The exploration of NLP in aviation safety has revealed a promising direction for enhancing the efficiency and effectiveness of safety measures within the industry. The initial qualitative analysis of seventeen studies indexed in Scopus underscores the transformative potential of NLP and machine learning algorithms in identifying critical safety issues, interpreting complex data, and providing actionable insights.

The work completed so far has highlighted the significant advancements in NLP methodologies, from SVD to deep learning techniques like BERT and LSTM, which have been successfully applied to aviation safety reports. These technologies have demonstrated their ability to process vast amounts of unstructured text data, offering a more objective and accurate approach to safety analysis compared to traditional manual methods.

Moreover, the study has identified key challenges such as the need for large, annotated datasets and the interpretability of complex models. It has proposed innovative solutions like active learning for data annotation and explainable AI to address these challenges, thereby paving the way for more robust and reliable safety management practices. The case studies presented provide concrete examples of NLP's impact on aviation safety, showcasing its role in the early detection of safety trends, enhancement of human factors analysis, and contribution to the development of best practices and policymaking.

Therefore, the integration of NLP into aviation safety represents a significant step forward in the pursuit of predictive safety management. It offers a path to not only analyze historical data but also to anticipate and prevent future safety occurrences. As the aviation industry continues to evolve, the continued research and application of NLP will undoubtedly play a crucial role in maintaining and improving safety standards, ensuring the well-being of passengers and the sustainability of air transport systems worldwide. The findings of this study serve as a foundation for future research and development in the field, encouraging ongoing innovation and collaboration among researchers, practitioners, and policymakers.

# Section II. Fundamentals of VAE, Explainable AI and Transformer.

## Chapter 4. Predicting the Indicative Likely Causal Factors of an Aviation Incident Using NLP and Variational Auto-encoder

The contents of this chapter, with only minor modifications, have been submitted and are currently under review in the Journal of IEEE Access. The recommended citation for the original publication is:

[387], Nanyonga, A., Wasswa, H., Joiner, K., Turhan, U., & Wild, G. (2025). Predicting Probable Causes of Aviation Incidents Using NLP and Variational Autoencoders.

## 4.1 Abstract

With the rapid technological advancement across the transport industry, air transport has continued to be safer than ever before consequently attracting an increased number of users. However, in the event of an aviation incident, establishing the likely cause has remained an expensive task in terms of finance, human experts, and time. This research attempts to address this challenge by proposing an AI-based approach that can predict the likely causal factors based on the analysis narrative of the series of events before and/or during the incident. The proposed approach deploys NLP to extract numerical encoding of the textual analysis narrative followed by the analysis of a generative AI model. The Variational Auto-encoder (VAE) is trained to generate the probable cause of an aviation incident given a textual narrative of the pre-incident events. The cosine similarity index was used to give a measure of how close the generated probable cause is to the actual probable cause provided in the test dataset. Evaluated on the NTSB aviation dataset, results show that the proposed model can competitively generate the likely cause for an aviation incident given the analysis narrative with an average similarity score of 0.786

## 4.2 Introduction

Establishing the cause of an incident, to prevent it from re-occurring in the future, is the core goal of any aviation incident investigation and analysis. Conventionally, whenever an investigation is deemed necessary in the event of an aviation incident, the primary source of information is usually the Cockpit Voice Recorder (CVR) and Flight Data Recorder (FDR) devices [195]. Data from these two devices is vital in giving an account of what was happening within the cockpit and the input to the aircraft from the pilot, respectively, minutes before and at the time of the incident. However, retrieving these two devices at times take months or even years and in the worst-case scenario they get severely damaged during or after the incident making the data irretrievable. In such cases, where the data on the devices does not give conclusive findings or is not readily available for the investigations to start, the experts often divert their attention to other sources which can include eyewitnesses, pilot reports, air traffic controllers, satellite images, radar information and weather stations readings at the time of the incident [196]. This gathered information is often prepared and presented as a narrative describing the series of events and conditions in which the incident occurred. Experts then analyze this information to establish the cause of the incident [157]. This allows the experts to suggest possible measures to prevent such incidents from happening again.

However, this entire process is time-consuming, and in the event of a design flaw, until the cause is established, and a preventative measure designed and implemented, the lives of passengers flying with such an aircraft model remain at risk.

As an example, the flaw in the design of Boeing 737 MAX's Manoeuvring Characteristics Augmentation System (MCAS) feature which counteracted the pilots' input was found caused two fatal accidents including the crash of Lion Air (JT610) [197, 198]. This accident followed, five months later, by Ethiopian Airlines flight 302 which again crashed a few minutes after take-off killing all people on board [197, 199]. If the cause of the Lion Air accident had been established quickly and acted upon appropriately, the ET302 [199] aviation incident investigation time, and allowing the quick establishment of the cause, researchers have utilized various natural language processing (NLP) and topic modeling-based approaches like latent Dirichlet Allocation (LDA), Latent Semantic Analysis (LSA), Parallel Latent Dirichlet Allocation (PLDA), among others [63, 89, 150, 171, 172, 200, 201]. These proposed schemes analyze and group aviation terms with related meanings or that are connected to a given phase of flight, flight conditions, and causes into related topics. This can help the investigation leading to quicker establishment of the likely cause.

However, no previous study has proposed a scheme for generating the probable cause given the analysis narrative of the pre-incident conditions and/or activities. To this end, this work proposes an AI-based model leveraging NLP and Variational Autoencoder (VAE) to predict the probable cause of an aviation incident from the initial analysis narrative. This work leverages the power of NLP to transform the textual narratives into feature vectors and train a VAE model that takes as input the extracted feature vectors and predicts the probable cause. To achieve this, we modified the VAE's loss function to use the cosine similarity index as the objective loss function for the reconstruction error. In other words, instead of minimizing the conventional mean square error (MSE) between the input vectors and VAE reconstructed vectors, the model was trained to maximize the cosine-similarity index between the predicted probable cause and actual probable cause provided in the test set. This way, being a generative model, VAE can generate the probable cause given the initial analysis narrative.

## 4.3 RELATED WORK

In recent years, there has been increasing interest in applying NLP and deep learning techniques in the aviation domain. These techniques have the potential to enhance safety and efficiency in the aviation industry by reducing the likelihood of human error and decreasing analysis time. VAE is a deep learning method that enables the generation of new data points by learning the underlying distribution of the data, while cosine similarity is a measure of similarity between two non-zero vectors, quantifying the cosine of the angle between them. Cosine similarity has been applied in aviation for text matching and document retrieval tasks [202, 203]. VAE has been used for various applications in aviation such as fault detection and isolation in aircraft systems [204].

In his study, M'Charrak [205] explored the use of VAE for NLP task, highlighting its application in tasks such as topic modeling, machine translation, and text classification. The study provided an overview of the potential benefits of using VAE for NLP and suggested various designs tailored to NLP tasks [205].

Similarly, Morisio and Felipe discussed the use of VAE in NLP and its potential benefits for tasks such as text generation and sentiment analysis [201].

Study [202], explored the integration of topic information into VAE for text semantic similarity. Their paper was presented in the international conference on neural Information Processing where a novel VAE based model that incorporates topic modeling techniques into the encoding and decoding processes was proposed to improve the representation of text data. Their paper also discussed the possible benefits of using the VAE for various NLP tasks such as information retrieval and question answering. Their paper gave a novel approach to enhance text semantic similarity using VAE with topic information integration.

In Study Dong et al. [157] proposed a deep learning approach for identifying causal factors of aviation incidents to enhance transportation safety. The authors emphasized the importance of analysing and identifying the factors contributing to aviation incidents to improve safety metrics. A deep learning model was developed to classify and analyze the textual description of incidents into causal categories. Their developed deep learning model classifies and analyzes textual descriptions of accidents into causal categories, demonstrating its effectiveness in enhancing aviation safety [157].

Dai and Quoc V. Le [206] introduced a semi-supervised sequence learning method combining VAE and cosine similarity across various datasets, including labelled and unlabelled data. Their study illustrates the effectiveness of the proposed approach across a range of NLP tasks, demonstrating the flexibility of VAE and cosine similarity for various applications, including aviation.

In 2015, Bowman et al. [207] presented a method for generating sentences from a continuous space using a neural network architecture, training a VAE on a dataset of sentences. Their approach outperformed existing methods in generating coherent and diverse sentences, allowing users to manipulate specific aspects of a generated text, which highlights the versatility of VAEs in natural language generation.

Study [203] reviewed a book by Christopher et al. that provided an introduction to information retrieval, covering the basics of the field, including cosine similarity, which is utilized in our study. Daniel& Martin authored a textbook on "Speech and Language Processing," discussing various NLP topics, including cosine similarity for measuring word and document similarity. Their insights are beneficial for improving safety and efficiency in the aviation industry [208]. Also, a study was carried out and a comprehensive textbook was produced covering deep learning foundations, including neural network architectures such as VAE. The book discusses sequence modelling and includes case studies on real world applications of deep learning, including NLP [209].

Sánchez et al. [210] introduced the Interval-Valued Variational Autoencoder (iVAE) to address the limitations of traditional VAEs in handling imprecise, interval-valued data and epistemic uncertainty. By incorporating a family of prior distributions and specialized neurons, the iVAE efficiently processes incomplete and noisy data, which would otherwise challenge conventional models. The effectiveness of iVAE was validated through case studies, including predicting aviation engine deterioration and remaining useful life (RUL), outperforming traditional methods like ICA and proving useful for moderate-sized datasets with high uncertainty.

Wang and Breckon [211] presented a Coupled Conditional Variational Autoencoder (CCVAE) to tackle the challenge of generalised zero-shot domain adaptation (GZSDA), where only a subset of classes in the target domain is available. This approach generates synthetic features for unseen target domain classes using source-domain data, enabling effective learning even with limited annotated target data. They tested the CCVAE on multiple datasets, including the Baggage Xray dataset designed for aviation security, demonstrating strong performance both in benchmarks and practical applications.

Wang et al. [212] explored the nonlinear dynamics of wind driven runback water film (WDRWF) flows, critical for understanding aircraft icing phenomena. They applied the Sparse Identification of Nonlinear Dynamics (SINDy) technique to analyze experimental datasets of WDRWF flows, initially using Proper Orthogonal Decomposition (POD) for data reduction, before transitioning to a more accurate VAE. Although VAE improved flow reconstruction, challenges arose in generating an accurate ordinary differential equation (ODE) system via SINDy due to temporal discontinuity in the latent space.

Chin et al., [213] focused on anomaly detection in helicopter operations, particularly during initial climb segments, a critical phase with heightened safety risks. Traditional flight data monitoring methods rely on predefined thresholds that vary across operations and aircraft types. To enhance detection, the authors proposed a data-driven sequential approach, including trajectory pattern mining and shape analysis, effectively identifying anomalies using a convolutional variational autoencoder.

Lastly, Nanyonga et al. [159, 170, 189] utilized various deep learning models such as Long Short-Term Memory (LSTM), Convolutional Neural Networks (CNN), Bidirectional LSTM (BLSTM), and simple Recurrent Neural Networks (sRNN) to classify flight phases from Australian Transport Safety Bureau (ATSB) reports. These models demonstrated high accuracy and effectiveness, with LSTM achieving the best performance in terms of accuracy, precision, recall, and F1 scores. Similarly, prior studies have explored the use of LSTM and CNN models in analyzing unstructured aviation data, focusing on enhancing safety through improved prediction accuracy and feature extraction. Additionally, the combination of NLP techniques with deep learning architectures has proven valuable in extracting thematic information from textual safety reports, further assisting in risk assessment and mitigation efforts within the aviation industry [101, 188]. These contributions underline the potential of integrating NLP and machine learning to enhance the accuracy and efficiency of aviation safety analysis.

While these studies contribute valuable insights into the application of machine learning techniques in aviation safety, our work distinguishes itself by employing NLP and VAE to predict the possible causes of an aviation incident. The novel method takes as input a numeric vector representation of the pre-and postincident textual narrative, projects into the learned low-dimensional latent space and the resulting lowdimensional space vectors are used by the decoder to predict the likely cause of the incident, an approach that has not been explored in prior studies. Our research directly focuses on incident causality prediction by integrating a generative model (VAE) with NLP techniques to create an AI-driven approach for analyzing aviation safety reports, thereby offering a more focused and potentially efficient solution for aviation safety analysis

## 4.4 PROPOSED APPROACH

The approach proposed in this work deploys a hybrid of techniques and models including NLP techniques to vectorize text input, VAE to generate vectors that are representative of words that most define the probable cause, cosine-similarity measure for model performance evaluation, and the word cloud module for output visualization [207, 214]. The subsequent subsections give a full description of the dataset and a brief account of how each of the techniques used works and why it was deployed for this study. The section ends with a detailed description of the experimental design highlighting the various model parameters and variable settings.

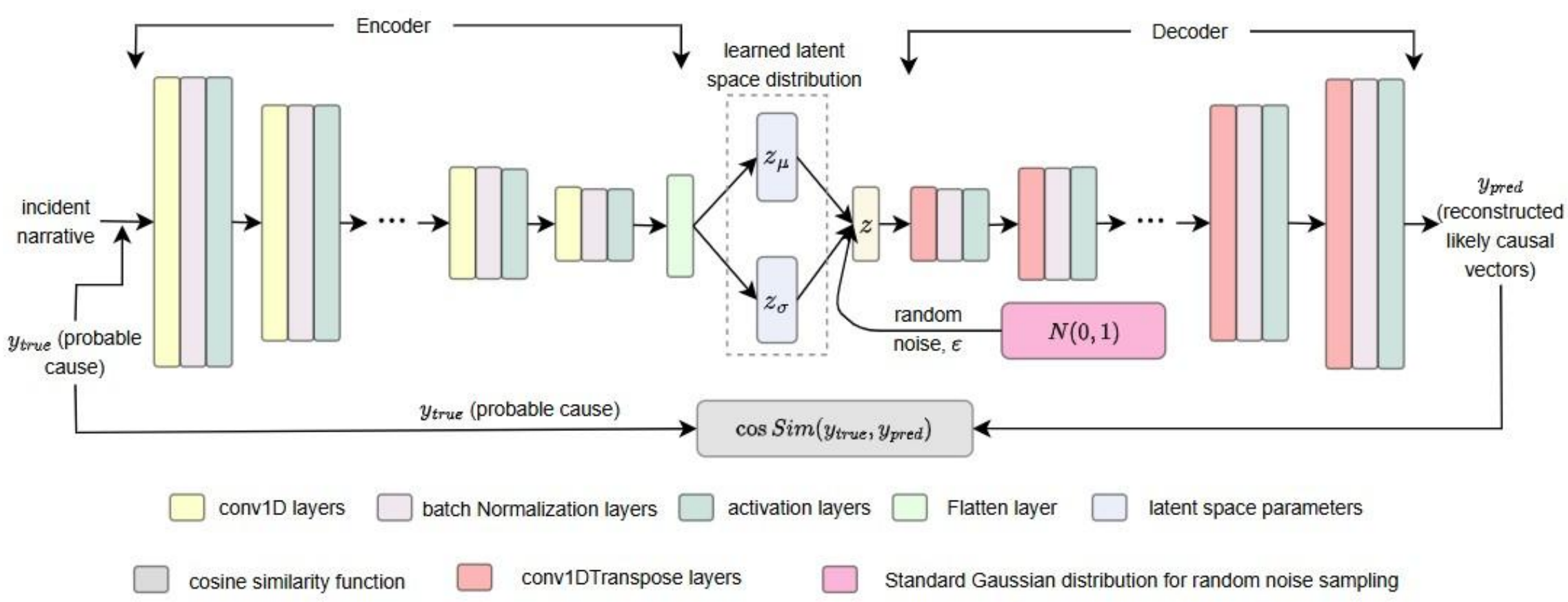

Figure 22. The proposed model design

### *4.4.1 Dataset*

Various bodies such as ATSB, ASRS (Aviation Safety Reporting System), and NTSB collect and publish aviation incident/accident investigation reports monthly. For this study, we used the NTSB aviation incident/accident investigation reports. This dataset can be accessed from the NTSB website, together with the metadata in various forms including individual monthly published .pdf reports, .json files, or by querying individual reports online or a .csv summary. In this work, we downloaded a .json file containing incident/accident investigation details for the years 2005 to 2020. Also, for this study, we considered incidents and accidents whose investigations were completed resulting in a dataset for 20919.

For each report, we extracted the ‘analysis Narrative’ and ‘probable cause’ fields for training and validation of our model.

*4.4.2 DATA PRE-PROCESSING*

Since VAE-based models are not designed to handle natural language text data training instances were transformed from the human-readable text to numeric feature vectors that the model can work with before training was conducted. This paper leverages the power of NLP for the data transformation tasks. Two popular python-based libraries, *spacy*, and *nltk*, are utilized for this purpose. With the help of these libraries, basic text data processing operations were performed including tokenization, stopword and punctuation removal, *url* and *html* tags removal, and stemming and lemmatization operations [215]. To convert the cleaned text data to the numeric form, we deployed a pre-trained word vectorization model, the Global word Vectors (GloVe) [216]. Glove gives the dictionary-like mapping of each word to a representative vector and which, for this study, was used to generate sequence matrices for both the analysis narrative and the probable cause. A comprehensive statistical analysis of text length distributions revealed the average lengths to be 1,116 and 165 words for “analysis Narrative” and “probable Cause” fields and standard deviations of 858.36 and 93.12 respectively. Considering these facts and after several rigorous experiments, the maximum sequence length for the analysis narrative was set to 1200 while that of "probable cause" was set to 200.

*4.4.3 DATA STANDARDIZATION*

Standardization aims to normalize the training features. It transforms the feature values to fall in some range. This work utilized the min-max scaling which scale the entries of a given feature to lie in the range (0,1). For a given feature $x$ of the train set with $x_{min}$ and $x_{max}$ being its minimum and maximum entries, an entry $x_i$ is scale to a value $x_i^{'}$ using the Eq. 1 such that $0 \leq x_i^{'} \leq 1$.

$$x_i^{'} = \frac{x_i^{\prime} - x_{min}}{} \quad (1) \; xmax - xmin$$

For this work the *scikit* learn *MinMaxScaler* module was used for this task. To enable 1D-based convolutional operations, each of the analysis narrative and probable cause entries was reshaped as (1200,1) and (200,1) respectively.

*4.4.4 Variational Autoencoder*

Variational Autoencoders (VAEs) have recently been used for various tasks in aviation such as [217, 218]. VAE has been known to be good in learning complex representations of aircraft systems and identifying patterns in large datasets hence helping in the improvement of aviation performance and safety.

A variation autoencoder, first introduced by [219], is an artificial neural network-based generative model that operates by learning a set of parameters, $\emptyset$, of an approximate posterior, $q_{\emptyset}(z/x)$ for the unknown true posterior (encoder), $p_{\theta}(z/x)$, where $x$ is a random variable from the unknown underlying prior dataset distribution, $p_{\theta}(x)$, of the training dataset while $z \sim N(\mu, \sigma)$ is the low-dimension latent space of the learned VAE and often approximates to a normal distribution. The decoder, $p_{\theta}(x/z)$, is trained to reconstruct the input variable, $x'$, by sampling from the learned latent space distribution, $z$, so that $x \cong x'$. The model learns by minimizing two errors, i.e, the Kullback Leibler (KL) divergence between the actual and approximate posterior and the reconstruction error between the input data points, $x$, and reconstructed data points, $x'$. Figure 23 depicts the general architecture of a VAE.

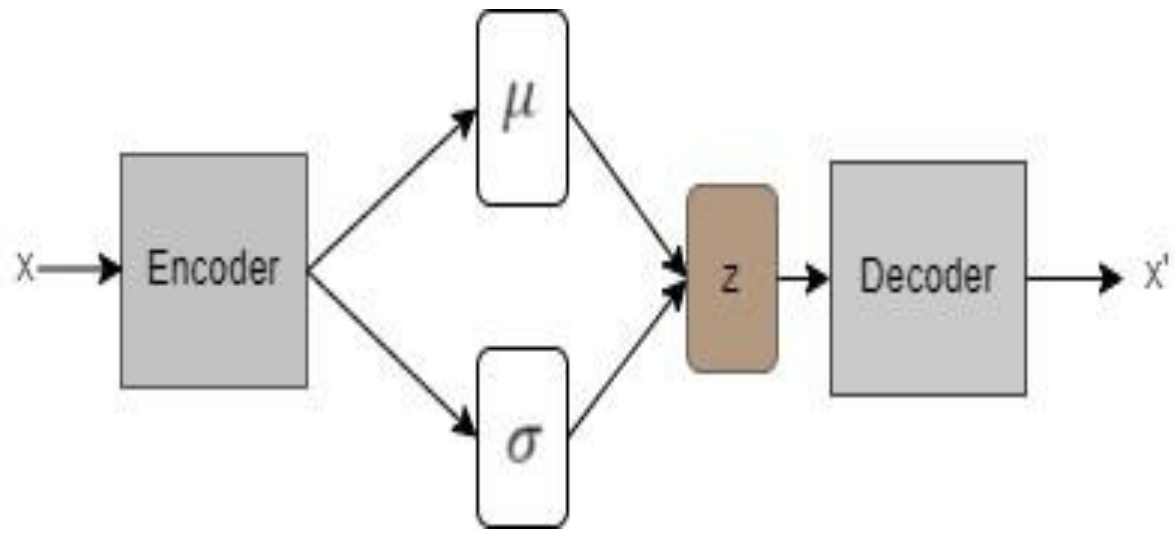

Figure 23: Variational autoencoder architecture

The latent space produced by a classic autoencoder is usually not regularized. VAEs overcome this challenge by constraining the latent distribution so that it is as close as possible to a normal distribution. This way VAEs allow generative power to the whole latent space for any sampled data point in this space.

For a given dataset $D = (x_i)^n{}_{x=1}$ of size $n$, in a classic autoencoder, the encoder deterministically maps input vectors $x \in \mathbb{R}^d$ to $z \in \mathbb{R}^k$ and the decoder, also in a deterministic approach, reconstructs $x' \in \mathbb{R}^d$ from $z \in \mathbb{R}^k$ where $k < d$ such that $x \simeq x'$.

Unlike a classic autoencoder, where a deterministic approach is learned for both the encoder and decoder operations, a variational autoencoder deploys the Baye's variational inference to, in a probabilistic manner, learn parameters for the encoder and decoder. The VAEs target to train a generative model where $x$ can be generated by sampling from the latent encoding $z$.

For the dataset, $D = (x_i)^n{}_{x=1}$, the distribution $P(D)$ can be expressed as:

$$P(D) = \prod_{i=1}^{n} P(x_i) \quad (1)$$

If the observed dataset, $D$ follows an underlying process with an unknown true distribution. For any variable, $x$, randomly sampled from $D$ we can learn an approximation of the underlying process by introducing a latent variable $z$, where the joint distribution $p_\theta(x, z)$ over variable $x$ and the latent variable $z$ can be learned by learning the parameters, $\theta$. The marginal distribution $p_\theta(x)$ can therefore be defined as:

$$p_\theta(x) = \int p_\theta(x, z)dz \quad (2)$$

However, the marginal distribution $p_\theta(x)$is intractable and therefore difficult to compute due to the integral function. Assuming the latent variable $z$, follows some simple distribution, we can assume that $p_\theta(x|z) \simeq p_\theta(z|x)$. This way we can learn a set of parameters $\phi$ for another network $q_\phi(z|x)$ which acts as a proxy to allow keeping track of $p_\theta(z \mid x)$ .

From, $p_\theta(z \mid x) = \frac{p_\theta(x,z)}{p_\theta(x)}$ , We can express $p_\theta(x)$ as:

$$p_\theta(x) = \frac{p_\theta(x,z)}{p_\theta(z|x)} = \frac{p_\theta(x,z)}{q_\phi(z|x)} \cdot \frac{q_\phi(z|x)}{p_\theta(z|x)} \tag{3}$$

Expressing $p_\theta(x)$ as an exponential function from Eq. gives:

$$\log p_\theta(x) = \log \frac{p_\theta(x,z)}{q_\phi(z|x)} + \log \frac{q_\phi(z|x)}{p_\theta(z|x)} \tag{4}$$

From Eq. (4) the expectation of $log\ p_\theta(x)$ over parameter $\phi$ can then be computed as:

$$E_{z\sim q_\phi(z|x)} \log p_\theta(x) = E_{z\sim q_\phi(z|x)} \log \frac{p_\theta(x,z)}{q_\phi(z|x)} + E_{z\sim q_\phi(z|x)} \log \frac{q_\phi(z|x)}{p_\theta(z|x)} \tag{5}$$

From Eq. (6), $\mathcal{L}(\theta, \phi, z)$ defines the Evidence Lower Bound (ELBO) also known as the variational lower bound while $D_{KL}(q_\phi(z|x)||p_\theta(z \mid x))$ is the Kullback Leibler divergence commonly known as KL divergence $D_{KL}$ between the approximate and true posterior distributions.

Since $x$ is given then $E_{z\sim q_\phi(z \mid x)} \log p_\theta(x)$ is constant and is $\geq 0$ . Therefore, in the variational autoencoder, parameters, $\phi$, which are the weight of the approximate estimator of the decoder (posterior distribution) $q_\phi(z \mid x)$ are learned in such a way that the value of the variational lower bound is optimized (maximized) since maximizing it lowers $D_{KL}$.

The encoder outputs $k$ means and $k$ variances respectively corresponding to the mean and variance of each variable in the latent space of dimension $k$.

Since sampling from a standard distribution like the Bernoulli distribution, Gaussian distribution, uniform distribution, etc. is straight forward, therefore, the latent distribution $z$ is approximated to a standard distribution, in particular, the Gaussian distribution to allow sampling by the decoder to generate $x'$. For the posterior model (decoder) to generate $x'$ from the latent space, a point $z$ is sampled from the encoder output ($\mu(z)$ and $\sigma(z)$)as:

$$z = \mu(z) + \varepsilon \cdot \sigma(z) \tag{7}$$

where $\varepsilon$ is random noise sampled from a standard normal distribution with mean 0 and standard deviation 1 defined as $\varepsilon \sim \mathcal{N}N(0, I_k)$. This approach of resampling is referred to as the reparameterization trick. The goal here is to have $x'$ as close to $x$ as possible.

*4.4.5 Variational Autoencoder structure*

The VAE constitutes an encoder component that projects the high dimensional analysis narrative word vectors to a 30-dimensional latent space of size and a decoder component that samples from the learned 30-dimensional latent space distribution to generate probable cause vectors for each input. The encoder constitutes an input layer, followed by 11 hidden Conv2D layers, followed by a flattened layer, and ends in a 30Dimensional dense layer. The decoder, on the other hand, comprises an input layer, followed by a dense layer, a reshaping layer to convert each 2D vector into a 3D tensor, and 10 Conv2DTranspose hidden layers and ends in an output layer producing a 3D tensor. The structure of the model and its corresponding parameters is summarised in Table 7. Apart from the input, output and reshape layers, each hidden convolutional layer used LeackyReLu for neuron activation and is accompanied by a batch normalization layer.

Table 7. VAE encoder and decoder summary

| | VAE Encoder | | | VAE Decoder | | | |
|---|---|---|---|---|---|---|---|
| Layer | *Filters* | Shape | Strides | Layer | Filters | Shape | Strides |
| Input | - | (30,30,1) | - | Input | 30 | - | - |
| Conv2D | 256 | (6,6) | 9 | Dense | 64 | - | - |
| Conv2D | 128 | (5,5) | 8 | reshape | - | (1,1,64) | - |
| Conv2D | 128 | (5,5) | 5 | Conv2D Transpose | 64 | (2,2) | 1 |
| Conv2D | 64 | (4,4) | 4 | Conv2D Transpose | 64 | (2,2) | 2 |
| Conv2D | 64 | (3,3) | 3 | Conv2D Transpose | 64 | (2,2) | 1 |
| Conv2D | 64 | (2,2) | 3 | Conv2D Transpose | 64 | (2,2) | 1 |
| Conv2D | 64 | (1,1) | 2 | Conv2D Transpose | 64 | (2,2) | 1 |
| Conv2D | 64 | (2,2) | 4 | Conv2D Transpose | 64 | (2,2) | 1 |
| Conv2D | 64 | (2,2) | 3 | Conv2D Transpose | 64 | (2,2) | 1 |
| Conv2D | 64 | (1,1) | 2 | Conv2D Transpose | 64 | (2,2) | 1 |
| Conv2D | 64 | (1,1) | 1 | Conv2D Transpose | 64 | (2,2) | 1 |
| Flatten | - | - | - | Conv2D Transpose | 1 | (1,1) | 1 |

| Dense | 8 | - | - | output | (11,11,1) | - | - |
|---|---|---|---|---|---|---|---|

*4.4.6. Experimental environment*

All experiments in this study were executed on an Ubuntu server with 256 CPU cores, 4 NVIDIA A100-SXM4-80GB GPUs, and 512GB of RAM. All experiments were written in the python programming language [220]. Used libraries and modules include TensorFlow, nltk, sklearn's, NumPy, and pandas.

*4.4.7. Model Training and Evaluation*

The transformed word vectors were randomly split into 85% trainset and 15% test-set. We further set aside 10% of the train set for model validation during each training epoch. The model training hyperparameters were set as 1000, 32, 5e-4, and 200 for the reconstruction error factor, batch size, learning rate with Adam optimizer, and epochs, respectively. with the learning rate, the number of epochs, and batch size are set to 1e-4, 20, and 128 respectively.

## 4.5 RESULTS AND DISCUSSION

The findings obtained from testing the proposed model on the unseen test set revealed that the model, given a textual narrative of the incident/accident, can predict a probable cause with competitive efficiency. For each unseen input narrative from the test dataset, we computed the cosine-similarity score between the actual probable cause given in the dataset and the predicted probable cause. We then computed the average score for all scores and registered an average score of 0.786 with a standard deviation of ±0.120 from the result visualization in **Figure 24**. We can see that the normal curve is skewed towards the high score and with very few scores below 0.5 which gives a visual confirmation of the model's performance.

The cosine-similarity scores from test-set instances are visualized in **Figure 25** and it can clearly be seen that our model can, considering the high similarity scores, reliably predict the probable cause. To make the model's output more intuitive to the human experts, we converted the generated vectors back to their original word representations and **Figure 26** shows word cloud plots for random samples of the predicted causes.

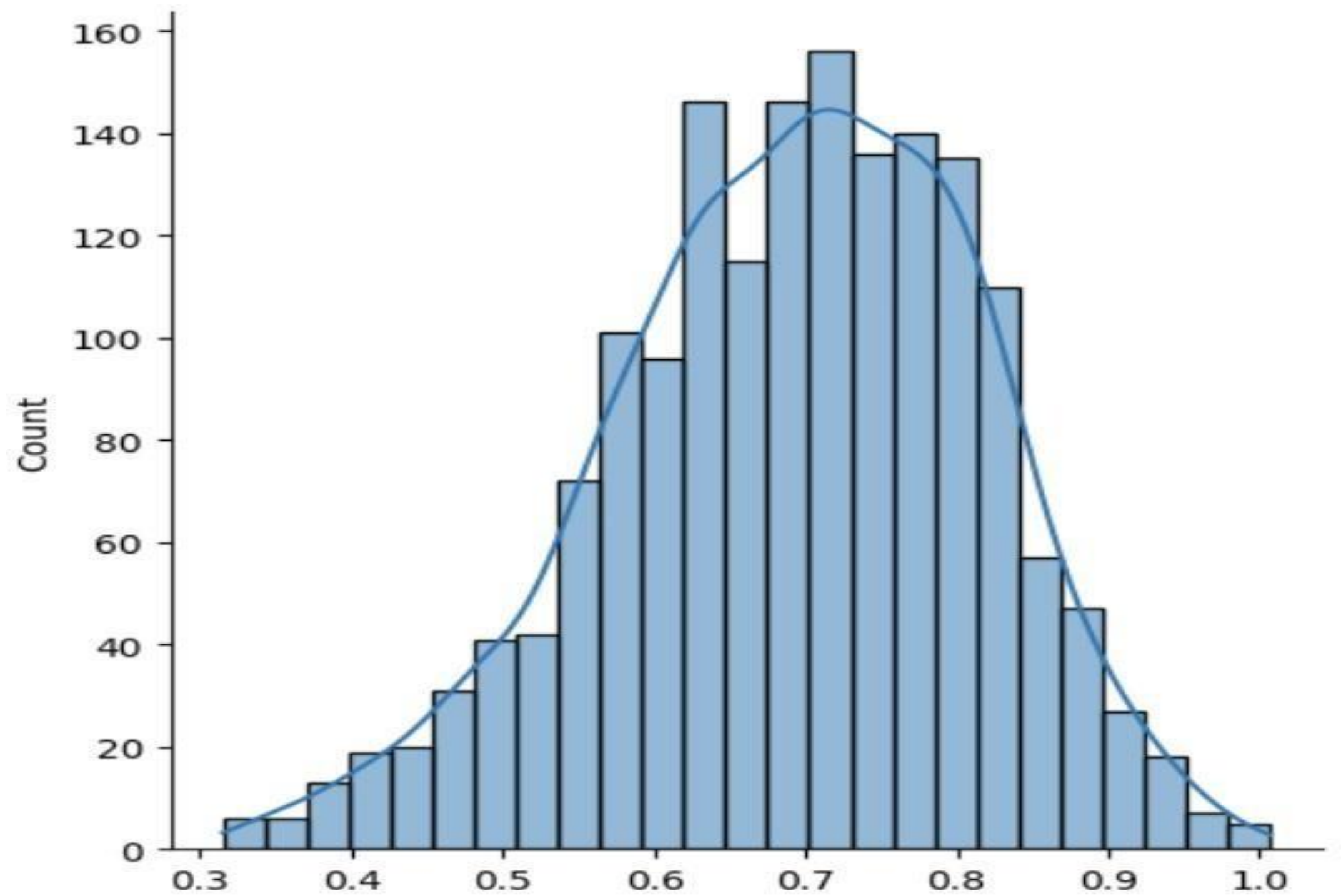

Figure 24: Normal plot for cosine-similarity scores

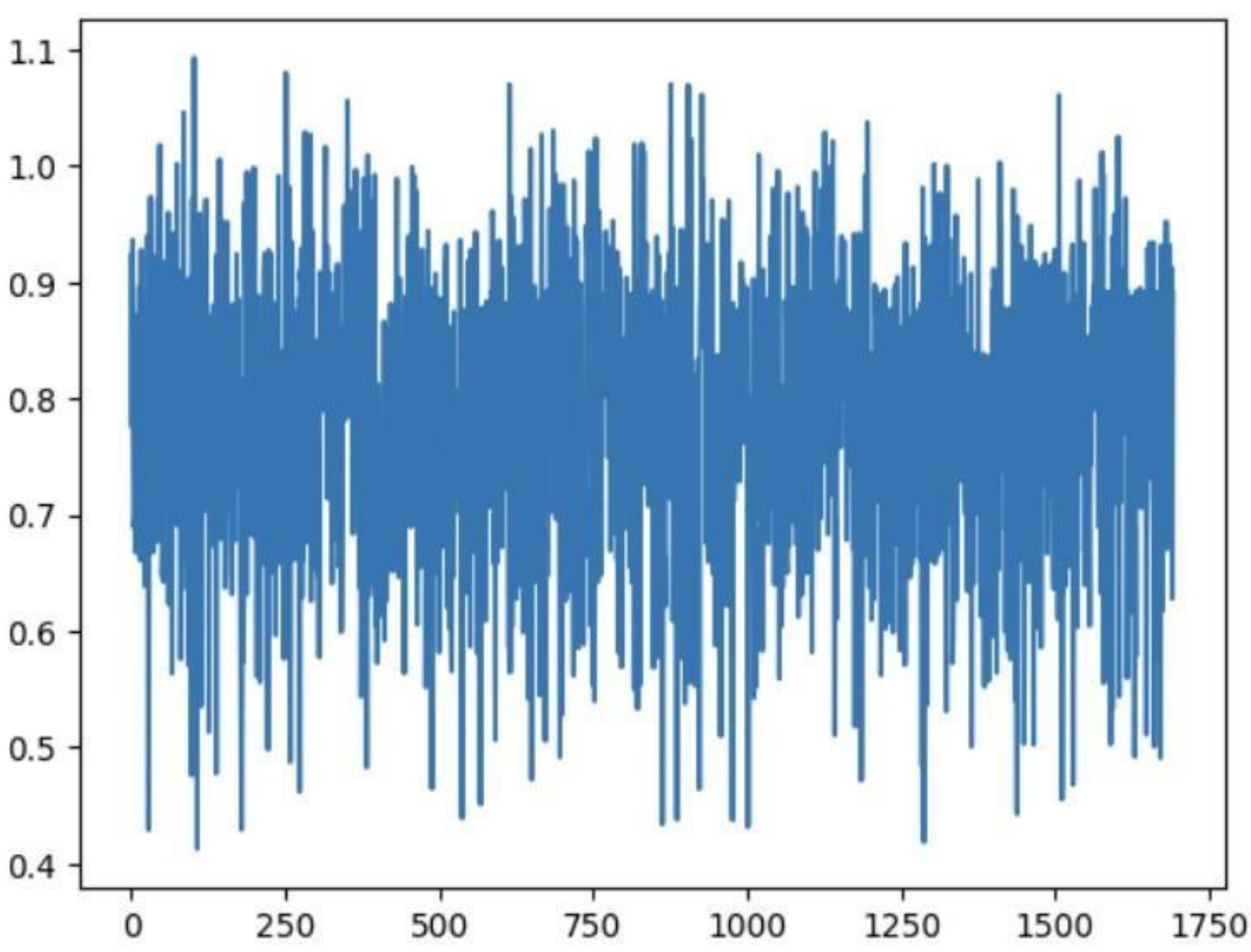

Figure 25. Cosine-similarity score plots from the unseen dataset samples

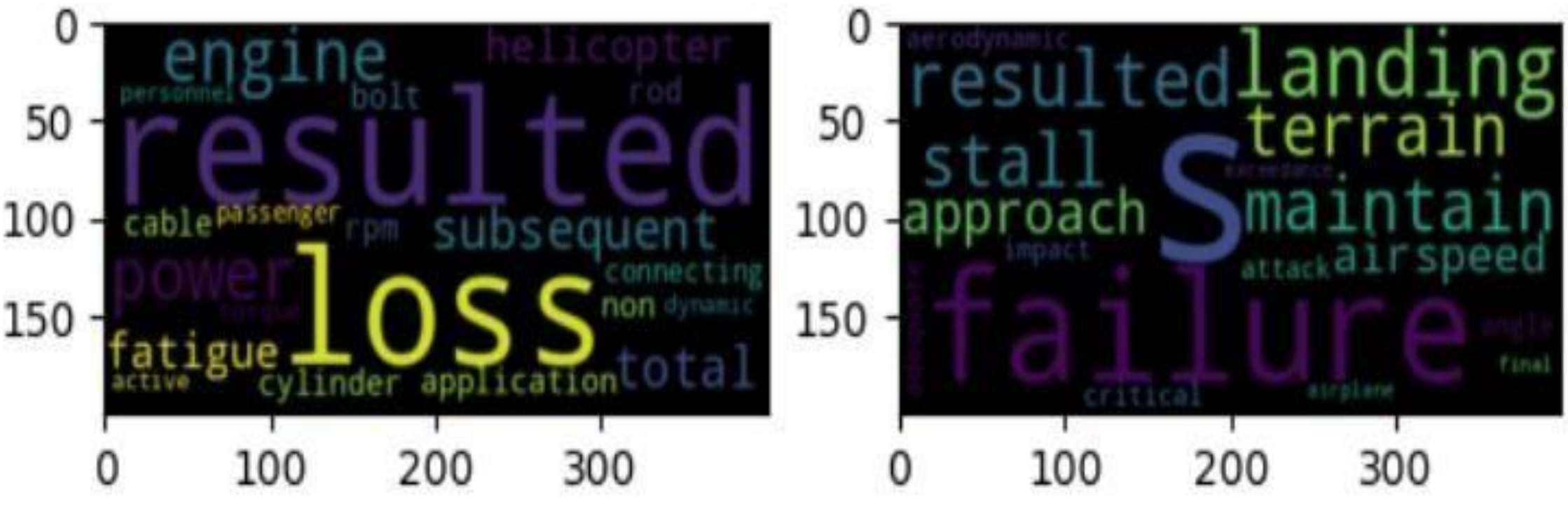

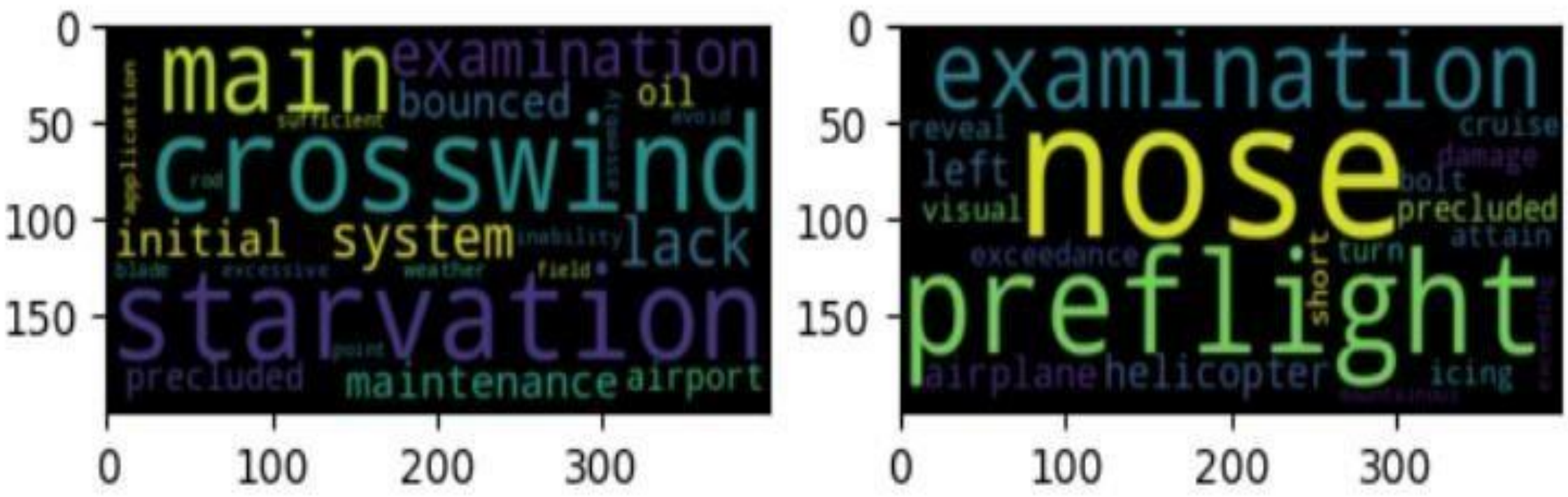

Figure 26: sample word cloud plots for the predicted probable causes

The above results show that the proposed model can efficiently generate the probable cause of any aviation incident or accident given a textual narrative of the pre- and post-incident events. This comes with the advantage of cutting resources including money, time and human experts that would be spent on establishing the cause. Also, the quick establishment of the cause would facility timely response of the various stakeholders to design and implements safety measures, including policies, protocols, guidelines, and regulations for preventing such unfortunate events from happening in future consequently enhancing safety.

## 4.6 CONCLUSION

In this work, we demonstrate the possibility of generating the probable cause of an aviation accident/incident by using natural language processing and variational autoencoder. By redefining VAE's loss function, we were able to train it to generate the probable cause by optimising the cosine similarity measure between the actual probable cause and the generated probable cause of the decoder component. The model constitutes the encoder which takes as input the numerical encoding of the textual analysis narrative obtained by deployment of NLP-based text pre-processing techniques. The Encoder projects the high dimensional numerically encoded vector to a low dimensional learned prior (latent space distribution). From the learned prior, the decoder samples and generates the probable cause for any input analysis narrative. By optimizing the cosine similarity measure, our model can with a high precision generate vectors that point in the direction of words that best describe the probable cause. This research is the first time such an approach has been researched in an aviation safety context. If applied, the approach can potentially reduce aviation incident investigation time, enable a swifter determination of the cause, and foster a safer aviation environment.

# Chapter 5. Explainable Supervised Learning Models for Aviation Predictions in Australia.

The contents of this chapter, with only minor modifications, have been accepted and are under revision in the Journal of Aerospace. It has also been made available as a preprint. The recommended citation for the original publication is:

Nanyonga, A., Wasswa, H., Joiner, K., Turhan, U., & Wild, G. (2025). Explainable supervised learning models for aviation predictions in Australia. Aerospace, 12(3), 223.

## 5.1 Abstract

Despite its recent success in various industries, artificial intelligence has not received full acceptance, hence full deployment by the aviation industry. This is partly attributed to, among other factors, the AI (Artificial Intelligence) model works as a black-box model with no clear explanations of how outputs are generated from the input samples. Aviation is an extremely sensitive application field, and this model's opaqueness makes it hard for a human user in the aviation industry to trust such a model. The work in this study examines the classification performance of various AI algorithms and then applies the SHAP (SHapley Additive exPlanations) framework to generate and visualize global-based model explanations to understand which features are learned for the decision boundary of each model and how much each model contributes to the final model output. We also deployed a variation autoencoder to handle the imbalanced class distribution nature of the ATSB (Australian Transport Safety Bureau) dataset and recorded competitive classification performance in terms of accuracy, precision, recall, and F1-score for a threeclass supervised learning-based classification problem.

## 5.2 Introduction

Recent years have seen a rapid proliferation of advanced artificial intelligence (AI) algorithms with profound performance regarding detection, prediction, and classification tasks [159, 170, 221] Consequently, this has triggered an exponential rise in the development and deployment of AI-based applications across various industries for accelerated process automation. This change in basic assumptions has further been motivated by the evolution of Internet of things (IoT) technology which equips "things" that conventionally had no computing capabilities with some degree of intelligence through sensors and actuators that monitor the world around them, collecting data about every event and transmitting it to the concerned entities for further analysis and decision-making support [222]. For instance, the Airbus A350 comes with more than 250,000 sensors that monitor every component, in and out, of the aircraft and its surroundings and collect data ranging from weather to in-flight commands issued by pilots [223]. This translates into huge volumes of data flowing into air traffic management systems that require real-time analysis to support timely decisions by human stakeholders.

In non-safety critical fields, advanced data analytics techniques coupled with advanced AI algorithms have come to the rescue in such situations. However, the inherent lack of model transparency by AI models has turned out to be a great impediment to the deployment of AI in the aviation industry due to its sensitivity to safety-threatening occurrences [221, 222, 224]. The rationale behind AI models' outputs remains opaque to human users to this day which is non-trivial to applications with high sensitivity to wrong model outputs, like the aviation industry, where even a seemingly slight misjudgment of the situation can have far-reaching catastrophic consequences ranging from loss of life to unrecoverable losses caused by accidents that often result in aircraft getting completely destroyed and aviation companies losing huge sums of money in compensating the victims' families.

To achieve AI model transparency and trustworthiness, AI researchers are shifting their focus to finding approaches that can make models explainable both at a local level where the rationale behind individual model predictions is explained and at a global level which is concerned with giving an intuition of the decision boundary learned by the model while providing a high-level insight of how the learned features impact the model's outputs [222]. This concept of making AI models explainable is known as Explainable Artificial Intelligence (XAI) [225, 226]. Various XAI-based approaches have been proposed in the last half-decade including model-agnostic and model-independent approaches. Most of the traditional machine learning algorithms like support vector machines (SVM), Decision Trees (DT), and K-Nearest Neighbor (KNN), among others, are intrinsically explainable, that is, their mode of operation is easily explainable and human users can easily visualize the rationale behind a given model output [225, 227].

On the other hand, deep learning models continue to operate as a "Blackbox" with no human-centric explanations for their output owing to the enormous amount of non-linear mathematical operations that take place in their hidden layers [221, 225, 228, 229]. Deep learning XAI approaches have been proposed to clear this barrier, including SHapley Additive exPlanations (SHAP) [230, 231], DeepSHAP [230, 231], Local Interpretable Model-agnostic Explanation (LIME) [232], DeepLIFT [233], among others.

In this paper, owing to the outstanding performance recorded by supervised machine learning algorithms in recent AI-based studies including healthcare [62], air transport [29], and computer vision [234] among others, and the labelled nature of the dataset used for this study supervised learning was deployed for all the experiments. We leveraged Variational Autoencoder (VAE) models for underrepresented class instances augmentation to ease the class imbalance distribution of the ATSB dataset.

Then we trained and evaluated four AI algorithms for a three-class classification problem. Out of the four models, three are standard machine learning algorithms (i.e., SVM, Logistic regression (LR), and Random

Forest (RF)) while the fourth is a deep neural network (DNN) constituting five hidden layers. We then deployed the SHAP technique to each model intending to understand the features pertinent to the decision boundary of each model and how they differ [235, 236].

Our contributions to this study are as follows.

- We investigate the impact of imbalanced distribution on the classification performance of various machine learning models and then propose a VAE-based data levelling approach to enhance model performance regarding minority class instances recognition.
- We evaluate and visualize the impact of each independent variable on the model output to allow model explainability and consequently make the model classifications more transparent and trustable by human stakeholders in the aviation industry.

The rest of this paper is structured as: Section II presents existing prior work related to this study followed by section III which gives a detailed description of the various techniques deployed to bring this work to existence, Section IV presents the experimental results and comprehensive discussion of their implications. Finally, Section V concludes this paper with a brief account of the current state of research on XAI concerning aviation safety and the possible direction for future research as far as this topic is concerned.

## 5.3 Related Work

Despite capturing the attention of many AI researchers across various fields of research, Explainable Artificial Intelligence (XAI) is still in its infant stage, that is, not much has been done to fully explore it for most fields, aviation not being exceptional. This section presents the prior studies whose context was the deployment of XAI to the aviation industry.

A study by [237] presented a taxonomy of the various levels of XAI relating to the aviation industry that can enhance trust in machine intelligence to human users. The study also gives an insight into the different state-of-the-art XAI approaches highlighting the pros and cons of each and suggesting the most suitable application area for each. However, no quantitative-based experiments are conducted to prove their concept, and consequently, no quantified performance comparison of the approaches for real-world scenarios is done. The work in [237] proposed a framework for generating human-centric and scientific explanations to enable human users to describe and interpret model predictions in the aviation industry. The authors contextualized XAI at four levels including interactivity with users where the XAI provides interactive explanations, level of observation where XAI is realized by either studying the observed data or the observed model, model structure where explanations are either model-agnostic or specific, and purpose of the explanations.

In a case study, Midtfjord et al. [238]highlighted the complexities involved in accurately modeling the physical dynamics of runway surface friction incurred by aircraft during landing. This challenge arises due to the complex and nonlinear interactions among the various physical factors influencing surface friction, as well as their time-dependent interdependencies. By incorporating weather data and runway condition reports, the study employed eXtreme Gradient Boosting (XGBoost) models to develop a hybrid assessment framework capable of classifying slippery runway conditions and predicting slipperiness levels. To ensure model transparency and support reliable decision-making for airport operators, the SHAP (SHapley Additive exPlanations) method was applied, offering localized explanations of the model outputs.

By using data provided by the United States Air Force (USAF) spanning from 2010 to 2018, study [239] trained and assessed several versions of interpretable models based on Bayesian networks, designed to function as decision support systems for selecting USAF candidate pilots. The authors utilized SHAP values alongside split conformal prediction techniques to provide explanations for the resulting black-box model hence facilitating a transparent and manageable assessment of each feature's impact on the model's prediction, as well as a reliable estimation of the prediction's uncertainty.

The study by Groot et al. [240] emphasized that in scenarios with high traffic density, analytical conflict resolution methods can lead to airspace instabilities. The authors also highlighted that, despite existing alternative AI-based methods, such as those leveraging deep reinforcement learning, having demonstrated promising results in these settings, they functioned as black-box models, making their decision-making processes challenging to interpret. To address this, the authors explored ways to explain the behavior of a Soft Actor-Critic model trained for vertical conflict resolution within a structured urban airspace. They employed a heat map of chosen actions as a tool to interpret and visualize the learning policy.

Study [241] introduced a green performance evaluation model for airport buildings, leveraging PCA and hierarchical clustering algorithms within an explainable, semi-supervised AI framework. This approach aimed to minimize reliance on human intervention when assessing green scores. By employing methods such as the scatter coefficient, Psi Index, variation, and permutation to reduce dataset dimensions, alongside divisive and agglomerative hierarchical clustering techniques, the authors were able to derive interpretable decision trees from the corresponding hierarchical dendrograms.

In their study, [242] developed a deep neural network model aimed at forecasting airport throughput based on both current and anticipated weather and air traffic conditions. To offer insights into which features the model relied upon to generate specific outputs, the authors proposed a method incorporating the explainable principal component analysis (EPCA) algorithm [243]. This approach enabled users to explore feature vectors derived from processing autoencoder outputs, providing a clearer understanding of the model's internal processes and decision-making behavior. Through the proposed method, users could gain insight into both what the model learned and how particular outcomes were achieved. The technique was applied to a real-world case study where deep learning was used to predict the total number of flights—either landings or takeoffs—at major US airports within a 15-minute interval.

Zeldam et al., [244] proposed and applied an automated failure diagnosis model where feasibility was tested using XAI on RNLAF (Royal Netherlands Air Force) F-16s dataset. Their approach was demonstrated using several learning models---Support Vector Machine (SVM), Random Forest (RF), Naive Bayes (NB), and Neural Network (NN)---and experimental findings showed RF to record a higher performance of $\simeq$ 80%. Their research is also based on XAI for model interpretation and understanding of the assessed diagnosis. The survey conducted in [222] revealed that LR, SVM, RF, and NN models are better in AI prediction and hence can be used on aviation datasets. Their survey was on AI and XAI in Air Traffic Management (ATM) where they identified the current and future research developments in the same domain. They indicated that LIME and SHAP explanations are now commonly used as they are acceptable to researchers and thus should be used to give more knowledge and understanding to what models give to avoid biases and trust issues among the end users. They also studied the importance of using autoencoders in AI models as explained by Dubot and Olive in their studies specifically regarding air transportation [245, 246].

In [247] a recurrent LSTM-based deep neural network was trained and evaluated on a historic flight dataset collected at the John F. Kennedy International Airport for predicting future unsafe situations. To ensure transparency of model operations, the authors leveraged a combination of input perturbation techniques and an intrinsically explainable hyperplane-based classification model to, in a global approach, explain the decision boundary of the proposed neural network model. The proposed LSTM (Long ShortTerm Memory) model recorded 9.4%,12.8%, and 42.3 seconds while the hyperplane-based model recorded 97.2%, 18.0%, and 51.8 seconds in terms of accuracy, miss-detection rate, and duration before the occurrence of the degraded state, respectively.

In study [248], Saraf et al., proposed a prototype tool for the verification and validation of AI-based aviation systems. A LIME-based technique was used with the aim to generate reliable, humanunderstandable explanations for predictions from the Multiple Kernel Anomaly Detection (MKAD) algorithm, an SMV-based algorithm developed by NASA (National Aeronautics & Space Administration) for aircraft trajectory anomaly detection. Using the enhanced input perturbation technique, the proposed approach located exact anomaly points from the aircraft trajectory images. The authors also proposed a prototype for verifying and validating AI-based aviation systems where they applied XAI for aviation safety applications. Their work gave proof of concept for an explanation as they used XAI to make reasonable interpretations that humans can understand in making decisions based on the models. And hence concluded that the application of XAI can be used by subject matter experts (SMEs) and is useful in relation to users accepting the decision made by AI-based support tools.

In study [248] a genetic algorithm-based conflict resolution algorithm for ATM was proposed to determine the best aircraft trajectories. Model explainability was evaluated by assessing the degree to which the controller understood its operations. Flight trajectory data was collected via flight simulations with nine controllers and the authors designed questionnaires to get feedback from each controller. To get fine-grained details, the authors also designed and utilized semi-structured questionnaires to assess the explanations at four levels, that is, at a Black box level where only the selected solution was presented, at a heat map level where a combination of potential solutions was demonstrated and at a storytelling level where human-centric explanations in form of storytelling were used.

Despite many of the above studies emphasizing the importance of working with explainable and interpretable models in safety-critical fields, no prior studies have explored explainable models in the classification of aviation incidents using the ATSB aviation incident reports, while the few that have studied aviation incidents using this dataset have not considered the impact of its imbalanced distribution on model performance. This study closes this knowledge gap by training and evaluating the performance of various classic machine learning models and a deep learning model in classifying aviation incidents in the Australian Aviation industry while giving an insight into how the various features influence model predictions for both balanced and imbalanced learning.

## 5.4 Methodology

The experimental execution in this study followed a three-step approach starting with data preprocessing followed by model training and evaluation and finally, the application of XAI-based techniques, SHAP and LIME, for model explainability. To study the impact of class imbalance, the investigated models were trained and evaluated on both the imbalanced and balanced datasets before and after the application of VAE.

### *5.4.1 Dataset*

This study considered Accidents, Incidents, and Serious Incidents that were recorded in Australia for the period of 20 years resulting in a dataset with 26,262 records distributed among the three classes as 3,402 (12.95%) “Accidents”, 20,894 (79.56%) “Incidents”, 1,966 (7.49%) “Serious Incidents” where the data was sourced directly from the ATSB investigation authorities.

### *5.4.2 Data preprocessing*

To ensure the dataset was in the desired format various data analytics techniques were utilized for the identification and removal of outliers and missing values, elimination of irrelevant features, handling of class imbalance, and feature encoding [249] Categorical features including “State”, “FuelType”, “RegistrationType”, “ActivityType”, and “ActivitySubType” were encoded as numerical features using 0ne-hot encoding. To handle class imbalance, additional instances were generated for the two underrepresented classes— the “Accidents” class and the “Serious incidents” class—using a variational autoencoder (VAE) [219] . The use of VAE for imbalanced learning in this study was inspired by the impressive performance recorded by VAEs in enhancing the recognition of minority class instances for IoT botnet detection in the study [250]. A detailed description of how VAE operates is presented in the Subsection below.

### *5.4.3 Variational Autoencoder for class imbalance*

As seen from the previous subsection, the dataset distribution in terms of the "occurrenceClass" variable is uneven with the incident class instances forming almost 80% of the total instances while accident and serious incident instances share the remaining 20%. Learning from imbalanced datasets can result in biased learning whereby the model learns patterns of the majority class and nothing from the underrepresented classes. In such a case accuracy can be misleading since it can be high even when no minority class is assigned to the right class. This can be seen in section IV where we present validation results of models trained on the imbalanced dataset. To mitigate this issue, we employed variational autoencoders to generate more instances for the two under-represented classes.

A variation autoencoder, first introduced [219], is an artificial neural network-based generative model that operates by learning a set of parameters, $\emptyset$, of an approximate posterior, $q_{\emptyset}(z/x)$ for the unknown true posterior (encoder), $p_{\theta}(z/x)$, where $x$ is a random variable from the unknown underlying prior dataset distribution, $p_{\theta}(x)$, of the training dataset while $z \sim N(\mu, \sigma)$ is the low-dimension latent space of the learned VAE and often approximates a normal distribution. The decoder, $p_{\theta}(x/z)$ , is trained to reconstruct the input variable, $x'$, by sampling from the learned latent space distribution, $zz$ so that $x \cong x'$. The model learns by minimizing two errors, i.e, the Kullback Leibler (KL) divergence between the actual and approximate posterior and the reconstruction error between the input data points, $x,$ and reconstructed data points, $x'$. Figure 27 depicts the general architecture of a VAE.

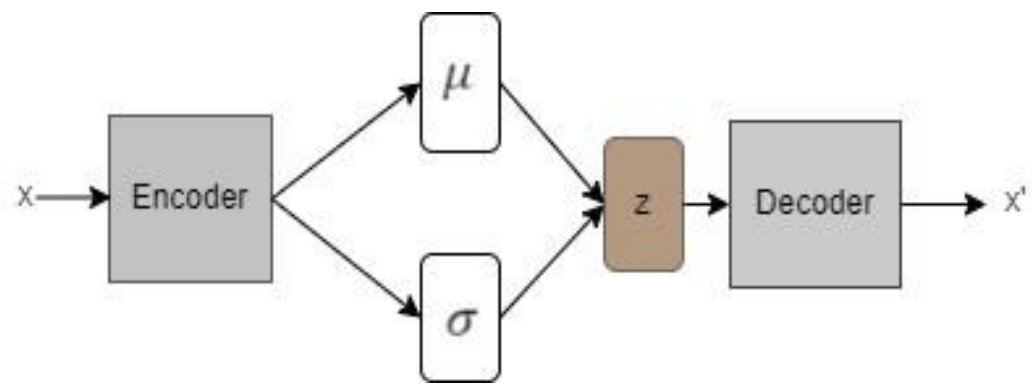

Figure 27. Variational Autoencoder architecture

In this work, we developed two VAE models for the two under-represented occurrence class instances of the ATSB dataset. For instance, to generate more samples for the serious incident class, we trained a VAE using only samples of the serious class and then used the decoder model of the trained VAE to generate additional samples for that class. The same procedure was followed to generate additional "accident" class instances to balance the distribution between the three classes of instances of the "occurrence" class variables. 16000 and 17000 instances were generated for the "Accident" and "Serious Incident" classes respectively.

*5.4.4 Model performance Evaluation.*

This section presents a description of the evaluation metrics that were used in this study to assess the performance of the model. This work focused on multi-class classification and measured the performance by considering how well the predictions were distributed among classes based on precision, recall, F1 score, and accuracy[159, 251-254] . F1-score as one of the performance metrics used in this study is simply the weighted combination of both recall and precision and its range is between 0 and 1 and the higher value is more preferred [252]. Table 8 shows the equations and explains how these performance metrics are calculated.

Where.

**TP** denotes the True Positive value where the actual value is similar/equal to the predicted value (i.e;

the value of accident predictions that is correct according to the labeled data).

**TN:** The True Negative value for a class where it is the sum of all columns and rows except the class's values being calculated.

**FP**: False Positive value where the class is the sum of values of the corresponding column except for the True Positive value (i.e; the value of accident predictions which is not correct according to the labeled data).

**FN**: False Negative value where the class is equal to the number of wrongly predicted values or not detected by the model (i.e; the value of accident labeled in the original data but not/wrongly detected by the model)

Table 8. Summary of different evaluation metrics used

| **Metrics used** | **Formula** | **Evaluation focus** |
|---|---|---|
| **Precision (p)** | $\frac{TP}{TP+FP}$ | Precision measures the correctly predicted positives from the total predicted patterns in a positive class |
| **Recall (r)** | $\frac{TP}{TP+TN}$ | This recall measures the fraction of positive patterns that are correctly classified |
| **F1-score (F)** | $\frac{2*precision*recaal}{precision+recall}$ | F-score measures the weighted average score of precision and recall |
| **Accuracy (acc)** | $\frac{TP+TN}{TP+FP+TN+FN}$ | Accuracy simply measures the total number of instances evaluated on the ratio predicted correctly. |

**i) Confusion matrix**

A confusion matrix is a square matrix in the $\mathcal{R}^m$ dimensional space, where $m$ is the number of unique entries of the dependent variable instances distributed among the class labels by the trained AI model during the testing phase. It is a tool used to give visual information of how well the model performed on the test dataset and is often used as a yardstick to measure the goodness of the model. Table 9 demonstrates a simple confusion matrix for a binary classification problem discriminating between positive and negative instances of the test dataset. The dark cells show the correctly predicted value i.e., true positives (TP) and true negatives (TN), while the lighter cells show the wrong predictions i.e., false negatives (FN) and false positives (FP) [254].

Table 9. Confusion matrix

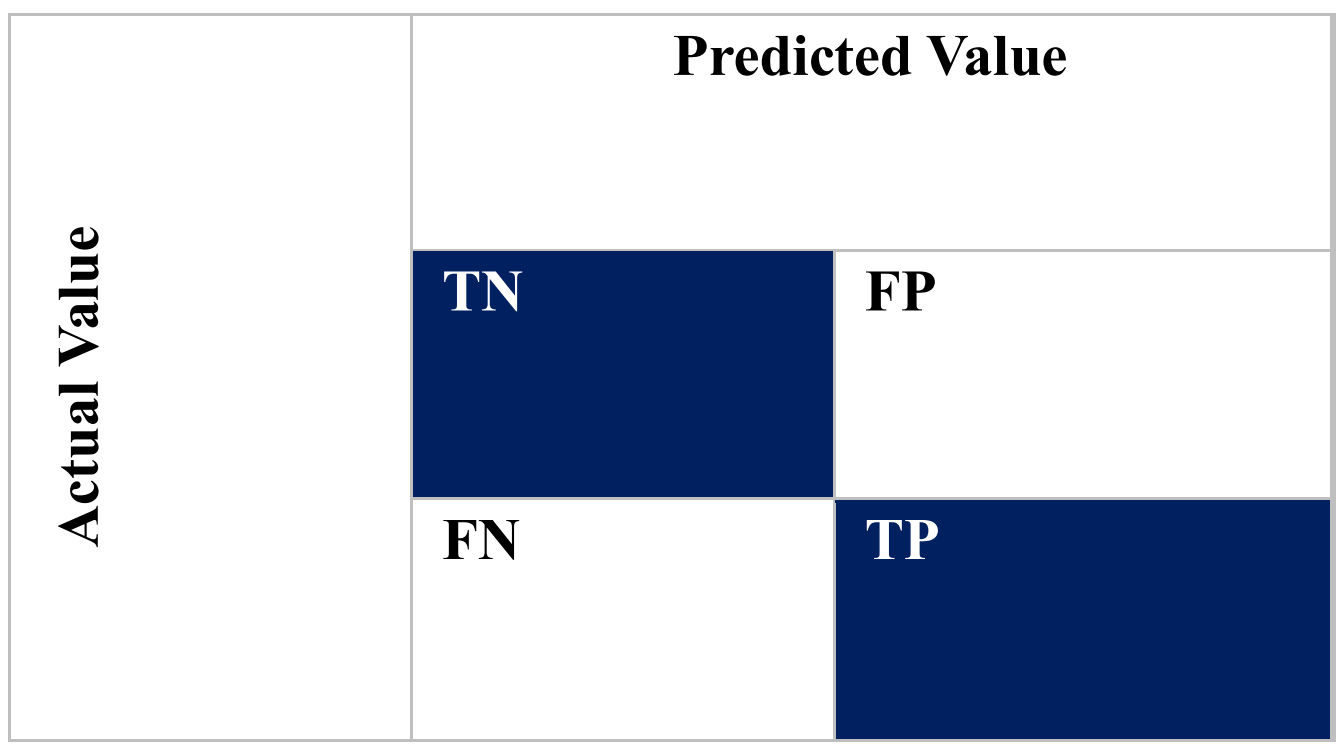

| | Predicted Value | |
|---|---|---|
| **Actual Value** | **TN** | **FP** |
| | **FN** | **TP** |

*5.4.5 Model Training and validation*

As indicated earlier, four classifiers including Logistic Regression, (LR), Random Forest (RF), Support Vector Machine (SVM), and the conventional feedforward Deep Neural Network (DNN) with backpropagation, were trained and evaluated for discriminating between instances of the occurrence class of the ATSB dataset. For both imbalanced and balanced dataset scenarios, python's scikit-learn library was utilized to randomly split the datasets into train and test sets in a 4:1 ratio. The same library was used for instantiating the three standard machine learning classifiers (LR, RF, and SVM). It is worth noting that the three models' hyper-parameter were set to sklearn's defaults except the max_ depth on RF, which was set to 30, and the SVM kernel which was set to linear.

Also, we maintained the common architecture for the DNN which conventionally constitutes an input layer followed by a finite number of hidden dense layers (five hidden layers for the DNN in this study) and an output layer. For all hidden layers, a Rectified Linear Unit (ReLU) was deployed for activation while the output layer deployed softmax as the activation function. To prevent the model from overfitting, each of the first three hidden layers is followed by a dropout layer that in a stochastic manner selects 20% of nodes in each weight cycle [255]. To generate the predicted class, the argmax function was used. This returns the index corresponding to the entry with the highest probability from the softmax output.

Python libraries including *Shap*, *pandas*, and *Numpy* were respectively used for model explainability, loading and managing dataframes, numerical dataset transformation and label encoding while Matplotlib and Seaborn served to generate visual plots of model scores. All experiments were implemented in a Jupyter notebook on a Linux server with 256 CPU cores and 256GB RAM running Ubuntu OS.

## 5.5 Explanation methods

To assert model transparency and trust by human users, we deployed XAI to generate model explanations. XAI is a technique applied in AI such that the results of a specific decision can be understood by humans. Various algorithms have been proposed in the recent past for the generation of individual prediction-based explanations, also known as local explanations. Local Interpretable Model-agnostic Explanation (LIME) proposed by [232] is a good example of such a technique which has seen application in numerous studies [256, 257]. However, LIME and similar models have the drawback of the inability to explain the model's predictions at a global scale, therefore, we utilized the SHAP module to generate and visualize global explanations for each of the classifiers used in this work. Despite the structures and prediction processes of the three classic ML (Machine Learning) algorithms (LR, RF, and SVM) being inherently explainable, we still applied SHAP to them to allow for a visual comparison between the features they learn and those learned by the DNN to define the decision boundary.

*5.5.1 SHapley Additive exPlanations (SHAP)*

This technique was proposed by [230] as a way of transforming "black box" AI models into transparent "white box" models to enhance their trustworthiness by generating and visualizing global explanations for the model's learned decision boundary. SHAP was proposed to enable the explainability of complex AI models including ensembles and deep neural networks which were hard for even AI experts to explain. It works by assigning a numerical value (known as the shap_value) to each feature in the train set that defines its degree of importance to the model's outputs. SHAP deploys the feature importance additive measure to generate a unified solution that combines the model explanation benefits of the previous local explanation XAI-based schemes including LIME [232], DeepLIFT [233, 258] Layer-wise relevance propagation [259] and Classic Shapley Value Estimation [260262], to generate global explanations. Figure 28 gives a high-level visual overview of the scope of XAI deployment in this work.

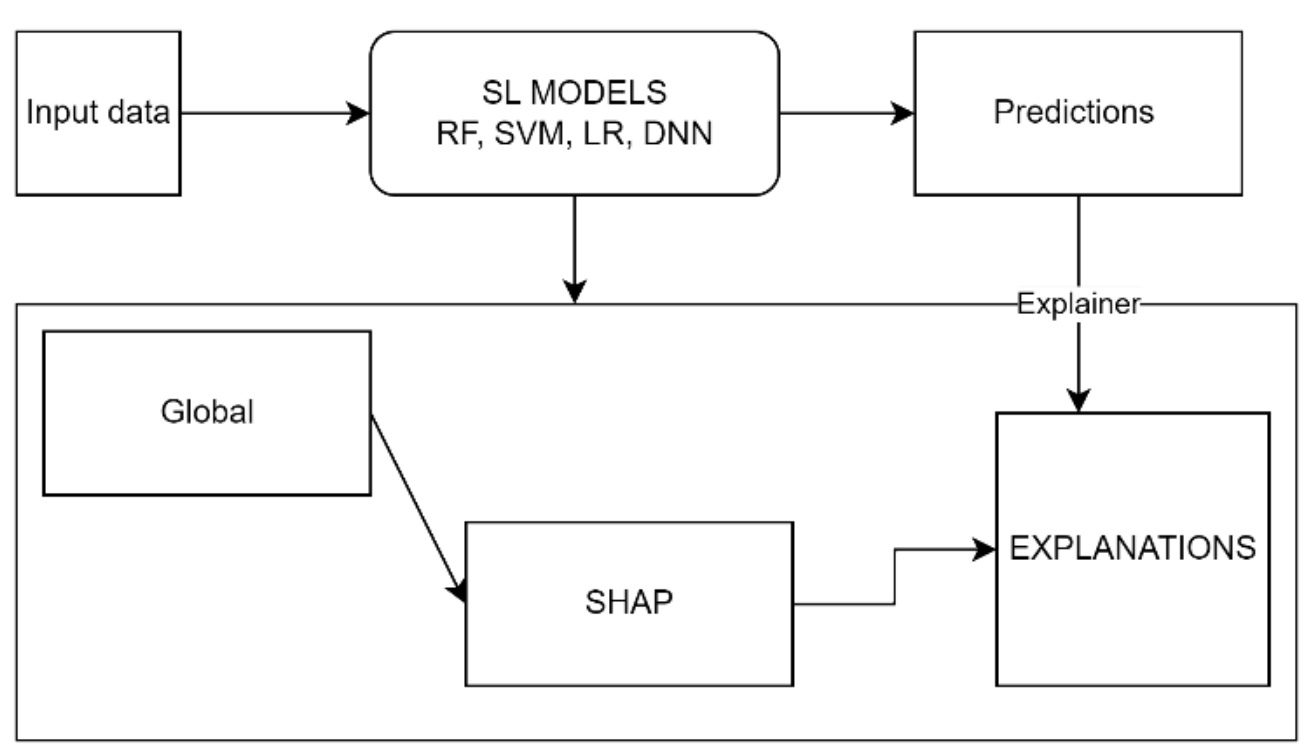

Figure 28. Overview of the scope of explainability methods used

## 5.6 Results and Discussion

In this section, the findings of this study are presented and discussed in the context of classification performance and XAI, clearly highlighting the impact of imbalanced and balanced dataset distribution in both cases.

*5.6.1 Machine learning model performance*

The performance of the four classification models studied in this work was evaluated in terms of accuracy and Table 10, clearly shows how the models perform on both the balanced and imbalanced datasets. These results clearly show that the application of VAE improved the classification performance for all models. However, in many cases, accuracy alone is not enough to judge the goodness of a model, especially for classification problems where the training dataset is highly imbalanced. For this matter, we investigated model performance in terms of other metrics and results are presented in Table 11 and Table 12 for the imbalanced and balanced train datasets, respectively.

Table 10. Comparison of classification accuracy for the various models

| Algorithm | Accuracy | |
|---|---|---|
| | **Imbalanced dataset** | **Balanced dataset** |
| Random forest | 74.4% | 99.9% |
| Support Vector Machine | 74.5% | 98.7% |
| Deep Neural Networks | 73.9% | 99.2% |
| Logistic Regression | 75.0% | 95.0% |

It is clear from Table 11 that high accuracy can be misleading when working with class imbalance situations. That is, despite all models giving a classification accuracy far greater than the random guess value of 33.33% for a three-class classification problem, they learn nothing from the serious incidents instances as depicted from the recorded precision, recall, and F1-scores of 0.0 from all of them. This is further depicted by the confusion matrix of each model in Figure 28. However, the application of VAE for imbalanced learning, as seen in Table 12, not only improved the classification accuracy but also greatly enhanced the recognition of minority class instances, with great improvements in precision, recall and F1-score for both the accident and serious incident class. This is further evidenced in Figure 29, by the confusion matrix plots.

Table 11. Model performance on an imbalanced dataset

| **Model** | **Class** | Performance | | |
|---|---|---|---|---|
| | | **PRECISION** | **RECALL** | **F1-SCORE** |
| **SVM** | ACCIDENT | 0.67 | 0.04 | 0.07 |
| | INCIDENT | 0.75 | 1.00 | 0.87 |
| | SERIOUS-INCIDENT | 0.00 | 0.00 | 0.00 |
| **RF** | ACCIDENT | 0.84 | 0.03 | 0.07 |
| | INCIDENT | 0.74 | 1.00 | 0.85 |
| | SERIOUS-INCIDENT | 0.00 | 0.00 | 0.00 |
| **LR** | ACCIDENT | 0.66 | 0.09 | 0.16 |
| | INCIDENT | 0.75 | 0.99 | 0.86 |
| | SERIOUS-INCIDENT | 0.00 | 0.00 | 0.00 |
| **DNN** | ACCIDENT | 0.00 | 0.00 | 0.00 |
| | INCIDENT | 0.74 | 1.00 | 0.85 |
| | SERIOUS-INCIDENT | 0.00 | 0.00 | 0.00 |

Table 12. Model Performance on a Balanced Dataset

| **Model** | **Class** | Performance | | |
|---|---|---|---|---|
| | | **PRECISION** | **RECALL** | **F1-SCORE** |
| **SVM** | ACCIDENT | 0.91 | 0.95 | 0.93 |
| | INCIDENT | 0.99 | 1.00 | 1.00 |
| | SERIOUS-INCIDENT | 0.95 | 0.90 | 0.93 |
| **RF** | ACCIDENT | 1.00 | 1.00 | 1.00 |
| | INCIDENT | 1.00 | 1.00 | 1.00 |
| | SERIOUS-INCIDENT | 1.00 | 1.00 | 1.00 |
| **LR** | ACCIDENT | 0.91 | 0.94 | 0.93 |
| | INCIDENT | 1.00 | 1.00 | 1.00 |
| | SERIOUS-INCIDENT | 0.94 | 0.91 | 0.92 |
| **DNN** | ACCIDENT | 0.99 | 0.98 | 0.99 |
| | INCIDENT | 1.00 | 1.00 | 1.00 |
| | SERIOUS-INCIDENT | 0.98 | 0.99 | 0.99 |

Figures 29 and 30 are the confusion matrix plots from the test sets of imbalanced and balanced datasets, respectively. The plots give a visual insight of how the different models perform on the test showing how the test instances are distributed amongst each class for each trained model.

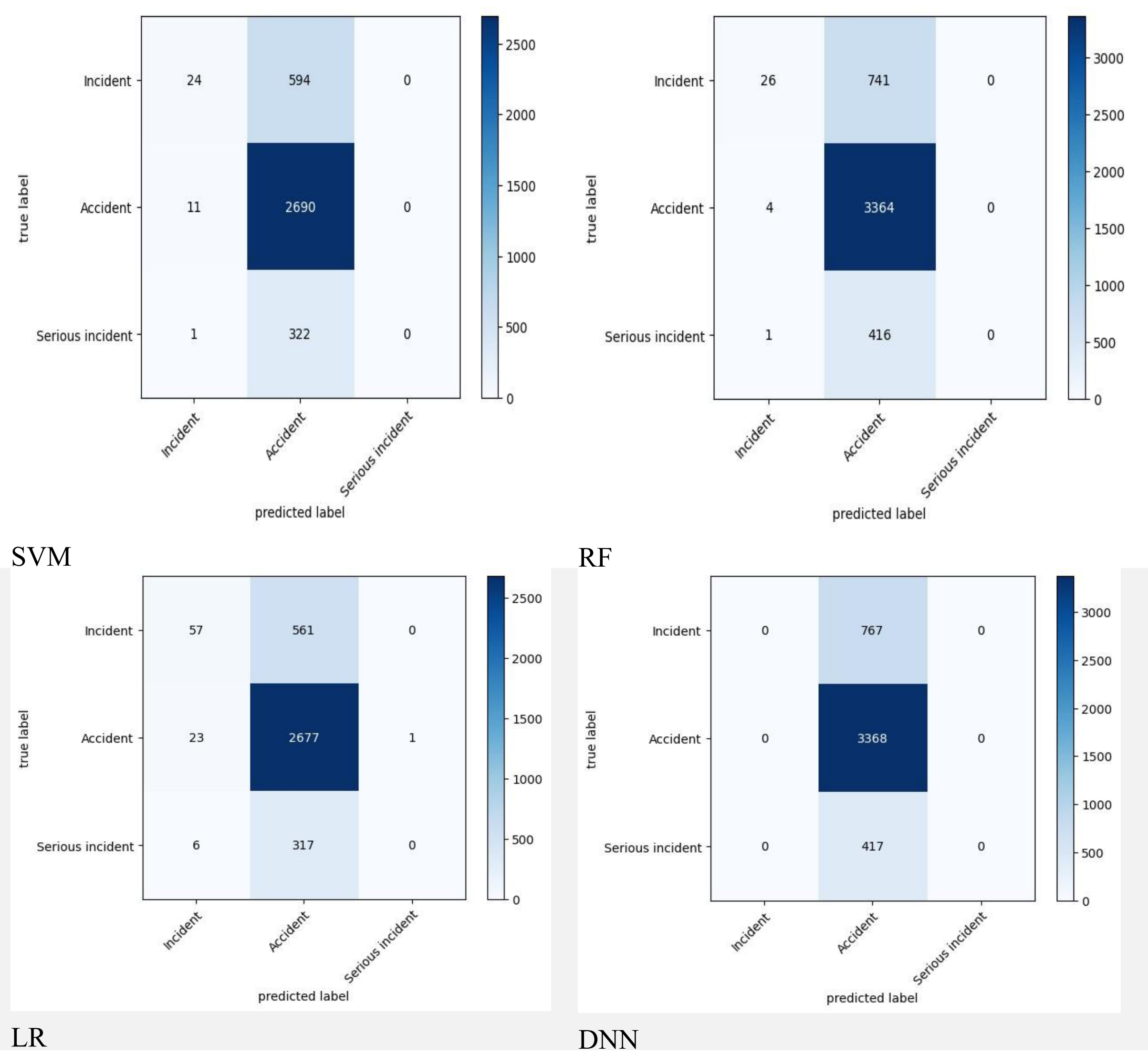

Figure 29. *Confusion Matrix for models trained on imbalanced Dataset*

SVM

| true label \ predicted label | Incident | Accident | Serious incident |
|---|---|---|---|
| Incident | 2639 | 10 | 31 |
| Accident | 0 | 2703 | 0 |
| Serious incident | 53 | 6 | 2655 |

RF

| true label \ predicted label | Incident | Accident | Serious incident |
|---|---|---|---|
| Incident | 3361 | 0 | 1 |
| Accident | 0 | 3378 | 0 |
| Serious incident | 0 | 0 | 3381 |

LR

| true label \ predicted label | Incident | Accident | Serious incident |
|---|---|---|---|
| Incident | 2529 | 0 | 151 |
| Accident | 0 | 2703 | 0 |
| Serious incident | 248 | 2 | 2464 |

DNN

| true label \ predicted label | Incident | Accident | Serious incident |
|---|---|---|---|
| Incident | 3303 | 0 | 56 |
| Accident | 0 | 3322 | 0 |
| Serious incident | 22 | 1 | 3417 |

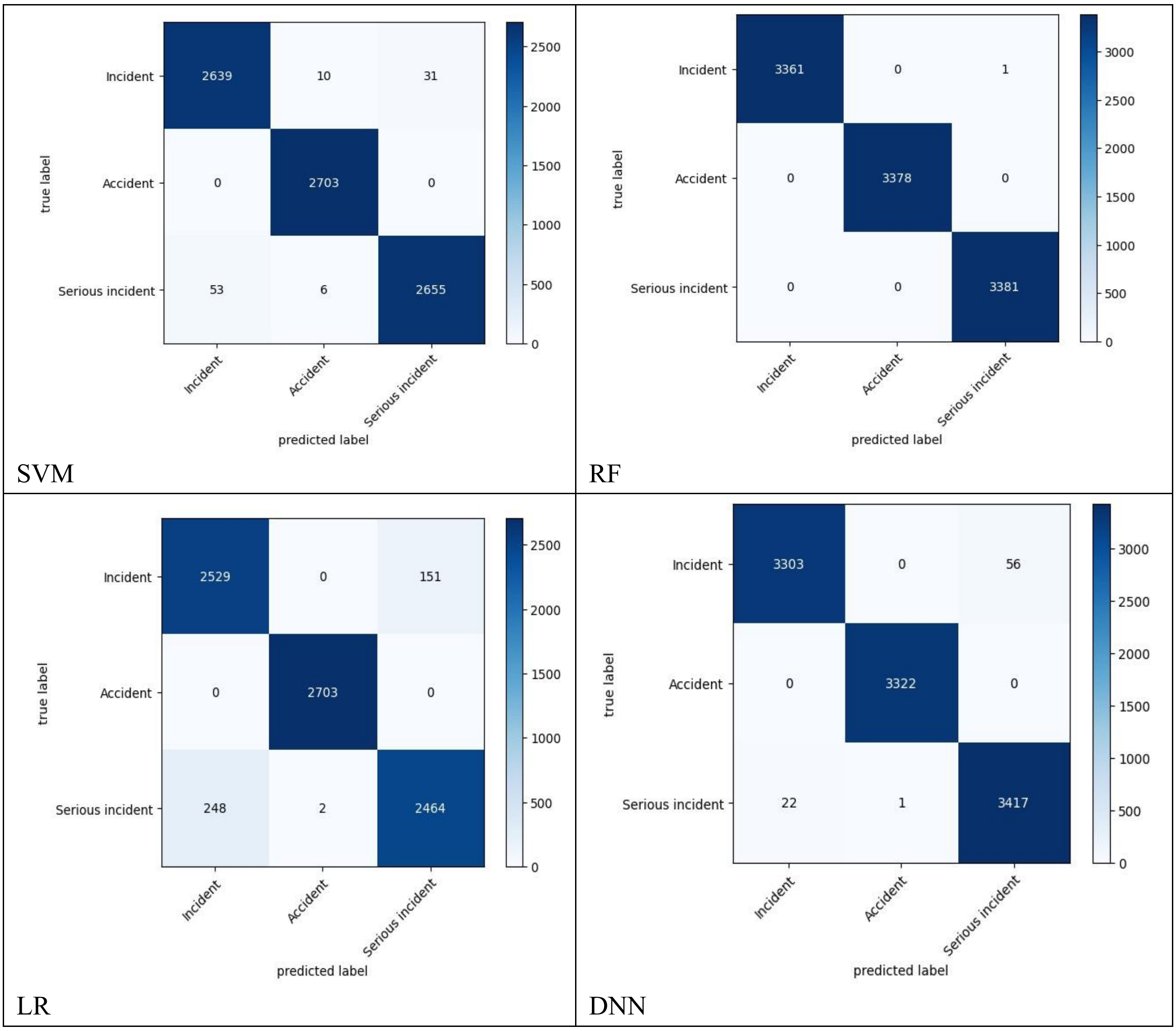

Figure 30*: Confusion Matrix for models trained on balanced Dataset*

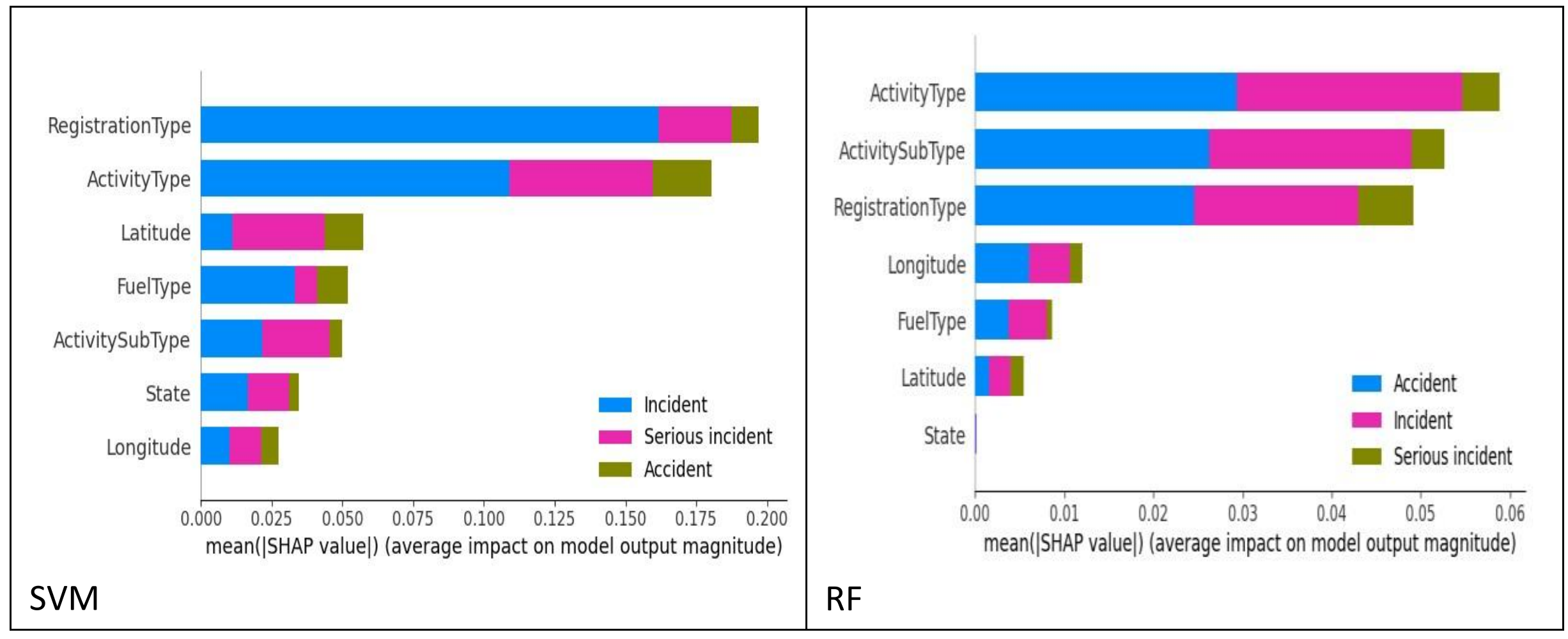

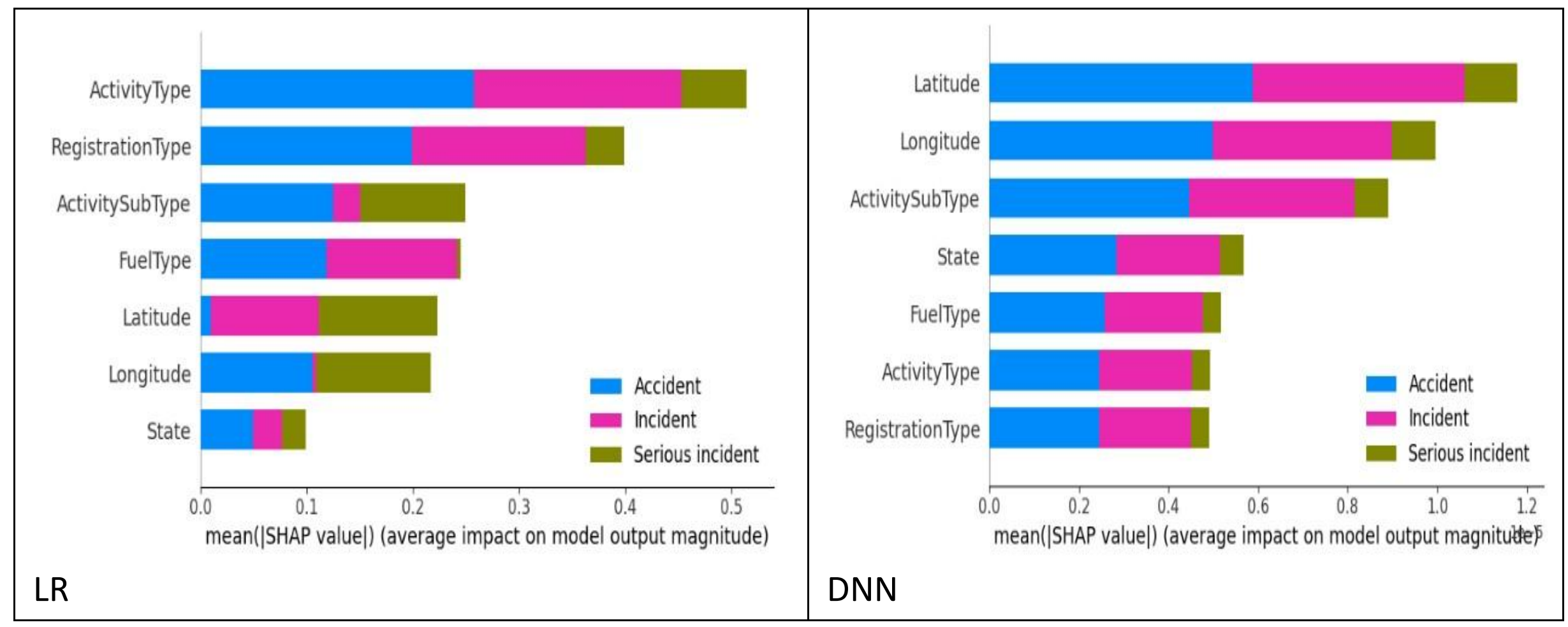

Figure 31*: SHAP plots for models trained on imbalanced dataset*

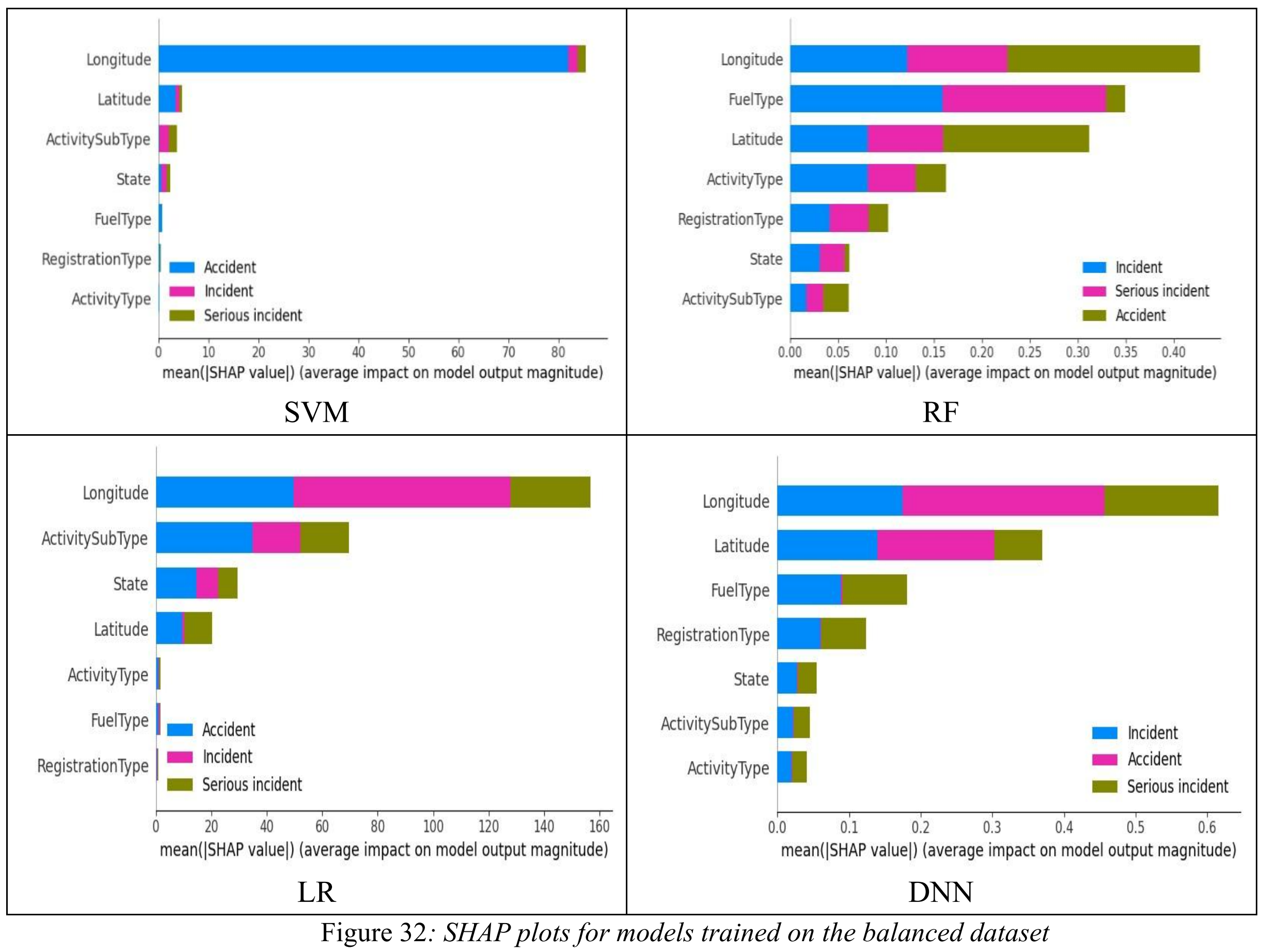

Figure 32*: SHAP plots for models trained on the balanced dataset*

## 5.7 Conclusion

Artificial Intelligence has been firmly established as a core for process automation in various industries, especially in areas where the cost of wrong predictions is considered less harmful. However, this technology has faced retarded acceptance rates with high sensitivity to misclassification such as in the aviation industry. This is attributed to a lack of model transparency and trustworthiness by human users. This work studied the impact of class imbalance on model performance to improve AI-based model performance in the aviation industry. Various experiments were conducted on both balanced and imbalanced datasets and an in-depth discussion of the results was presented. To assert explainability and model transparency, intending to enhance AI acceptance by stakeholders in the aviation industry, we explored the XAI of various AI algorithms using SHAP. Using different SHAP-based model explainers, we were able to generate, visualize and discuss explanations for all models studied in this work.

## Chapter 6. A Multi-head Attention-Based Transformer Model for Predicting Causes in Aviation Incident

The contents of this chapter, with only minor modifications and is under review in the Journal of Modelling. It has also been made available as a preprint. The recommended citation for the original publication is:

[386], Nanyonga, A., Wasswa, H., Joiner, K., Turhan, U., & Wild, G. (2025). A multi-head attention-based transformer model for predicting causes in aviation incidents. Modelling, 6(2), 27.

### 6.1 Abstract

The timely identification of probable causes in aviation incidents is crucial for averting future tragedies and safeguarding passengers. Typically, investigators rely on flight data recorders yet delays in data retrieval or damage to the devices can impede progress. In such instances, experts resort to supplementary sources like eyewitness testimonies and radar data to construct analytical narratives. Delays in this process have tangible consequences, as evidenced by the Boeing 737 MAX accidents involving Lion Air and Ethiopian Airlines, where the same design flaw resulted in catastrophic outcomes. To streamline investigations, scholars advocate for natural language processing (NLP) and topic modeling methodologies, which organize pertinent aviation terms for rapid analysis. However, existing techniques lack a direct mechanism for deducing probable causes. Bridging this gap, this study proposed a transformer-based model for predicting likely causes from raw text narrative inputs, leveraging advancements in long-input transformers. By training the model on comprehensive aviation incident investigation reports like those from the National Transportation Safety Board (NTSB), the proposed approach exhibits promising performance across key evaluation metrics, including Bilingual Evaluation Understudy (BLEU) with *(M=0.727, SD=0.33)*, Latent Semantic Analysis (LSA similarity) with *(M=0.696, SD=0.152)*, and Recall Oriented Understudy for Gisting Evaluation (ROUGE) with a precision, recall and F-measure scores of *(M=0.666, SD=0.217), (M=0.610, SD=0.211), (M=0.618, SD=0.192)* for rouge-1, *(M=0.488, SD=0.264), (M=0.448, SD=0.257), M=0.452, SD=0.248)* for rouge-2 and *(M=0.602, SD=0.241), (M=0.553, SD=0.235), (M=0.5560, SD=0.220)* for rouge-L, respectively. This demonstrates its potential to expedite investigations by promptly identifying probable causes from analysis narratives, thus bolstering aviation safety protocols.

## 6.2. Introduction

Establishing the cause of an aviation incident or accident, to prevent it from re-occurring in the future, is the core goal of any aviation safety occurrence investigation and analysis. Hereafter, aviation accidents will be considered a subset of aviation incidents. Conventionally, whenever an investigation is deemed necessary in the event of an aviation incident or accident, the primary source of information is usually the Cockpit Voice Recorder (CVR) and Flight Data Recorder (FDR) devices [195]. Data from these two devices is vital in giving an account of what was happening within the cockpit and the input to the aircraft received from the pilot, respectively, minutes before and at the time of the incident. However, retrieving these two devices can take months or even years and in the worst case, the devices get severely damaged during or after the incident making the data irretrievable [198]. In such cases, where the data on the devices does not give conclusive findings or is not readily available for the investigations to start, the experts often divert their attention to other sources which can include eyewitnesses, pilot reports, air traffic controllers, satellite images, radar information, damaged aircraft components, and weather stations readings at the time of the incident [196]. This gathered information is often prepared and presented as a narrative describing the series of events and conditions under which the incident/accident occurred. This information is then analysed by experts to establish the likely cause of the incident [157], allowing them to suggest possible measures that can deter such incidents from happening again.

However, this entire process is time-consuming, and in the event of a design flaw, until the cause is established, and a preventative measure designed and implemented, the lives of passengers flying with such an aircraft model remain at risk. As an example, the flaw in the design of Boeing 737 MAX's Manoeuvring Characteristics Augmentation System (MCAS) feature which in certain circumstances counteract the pilots' input caused two fatal accidents including the crash of Lion Air (JT610) [197] flight followed, five months later, by Ethiopian Airlines Flight 302. Both aircraft crashed a few minutes after taking-off killing all 189 and 157 people on board respectively [197, 199]. If the cause of the Lion Air accident had been established quickly and acted upon appropriately, the ET-302 [199] accident would likely have been avoided. With the aim of shortening aviation incident/accident investigation time, and allowing the quick establishment of the cause, researchers have proposed various natural language processing (NLP) and topic modelling-based approaches like Latent Dirichlet Allocation (LDA), Latent Semantic Analysis (LSA), Parallel Latent Dirichlet Allocation (PLDA), among others [200, 214, 263, 264]. These proposed schemes analyse and group aviation terms with related meanings or that are connected to a given phase of flight, field of aviation, flight conditions, and/or causes into related topics. Such approaches could help the investigation team to establish the area of concentration and consequently, lead to quick establishment of the causes.

However, no previous study has proposed a scheme for generating the probable causes given the analysis narrative of the pre-incident conditions and activities. To this end, this work builds and deploys a transformer-based model to predict the probable cause of an aviation incident from the initial analysis narrative. Transformer models like Bidirectional Encoder Representations from Transformers (BERT) [214] and its variants [201, 265] have demonstrated cutting-edge performance across various challenging NLP tasks. Since the analysis narratives often contain long textual paragraphs, we hypothesized that the resulting model would produce enhanced performance if based upon recent studies on long-input transformers [265-268] which have revealed that increasing the Transformer's input length positively correlates with model performance.

The training approach for the transformer model deployed in this work aligns with the fundamental principles of a language translation transformer. However, it deviates in that, instead of setting the masked input to the transformer's decoder as the target language during training, it utilizes the target probable cause. Upon evaluation on the NTSB dataset, the model showcased the potential for transformers to accelerate aviation incident investigations by generating the probable causes based on the analysis narrative. The contribution of this study is two-fold:

1. A new generative model based on multi-head attention transformer is proposed and trained for generating the probable cause of an aviation incident when given as input the raw text narrative of series events before, during or after the accident. The model accepts both long and short input narratives which should expedite the investigation process enhance air transport safety.

2. Many aviation incident dataset have instances with analysis narratives but with no corresponding entry for the probable causes. This leads to eliminating many instances during model training and consequently leads to poor model performance in terms of generalization to new instances. With the ability to generate missing probable cause, instances with missing values can be retained, likely leading to better model performance and generalization.

The rest of this paper is organised as follows: Section 2 presents a review of prior related work followed by Section 3 where a detailed description of the our approach is presented. In Section 4 we present the findings of this study. In Section 5 a detailed discussion of the findings is presented, highlighting the contributions and limitations of our study and finally section 6 gives concluding remarks, highlighting the direction of future work.

## 6.3. Related Work

The utilization of machine learning and deep learning methods and techniques in aviation analysis and prediction has garnered increasing attention from aviation safety researchers. This interest is driven by objectives such as expediting aviation incident investigations, promptly determining the causes of incidents for swift mitigation of future occurrences, predicting incidents, and extracting knowledge to enhance air transport safety. This section delves into key prior studies that have employed AI-based techniques in alignment with aviation safety.

Burnett et al.,[76] trained four conventional ML classifiers, including Decision Trees, KNN, SVM, and ANN with back propagation for prediction of aviation injuries and fatalities. The authors employed a cross validation training approach with 10 folds and looked at how factors like pilots' accumulated flight hours and age impacted the rate of injuries and fatalities. Experimental results revealed ANN to be superior for the task when evaluated on datasets sourced from Federal Aviation Administration (FAA) between 1975 and 2002 inclusive.

Nanyonga et al., [159] utilized NLP and other AI to analyze text narratives, aiming to determine aircraft damage levels from safety incidents. Four learning models: Long Short-Term Memory (LSTM), Bidirectional LSTM (BLSTM), and Gated Recurrent Units (GRU), Simple recurrent Neural Network (sRNN) and hybrid architecture models including GRU+LSTM, sRNN+BLSTM+GRU, etc, were assessed on 27,000 NTSB reports. Results indicated all models achieved over 87.9% accuracy, surpassing random guessing (25%) for a four-class problem.

[30], assessed the risk created by various anomalies in aviation events using of a hybrid classifier constituting proposed a hybrid model comprising a SVM and and several neural networks. The four-step method involved all events being categorized into five risk-level groups, followed by application of a SVM model to determine the link between textual event synopses and the resulting consequences. Next the hybrid model was trained to capture the correlations between contextual event attributes and risk-level groups. A fusion rule was then proposed to combine outcomes from the two models and finally, a stochastic-base decision tree was used to predict the risk level.

Both [123, 269] deployed Bayesian inference-based techniques for aviation incident modeling and analysis. Study [269] aimed to forecast aircraft safety incidents by employing an inventive statistical method. This method utilized Bayesian inferences and hierarchical structures to build learning models of varying complexities and goals. In contrast, [123] focused on analyzing commercial aviation accidents spanning the period between 1982 and 2006, as documented by the NTSB. This second study proposed a four-phase approach to build a Bayesian network capable of capturing the relationship between the sequence of events that led to the accidents. The methodology encompassed creating a graphical representation for visualizing aviation accident events, forming a Bayesian network representation by amalgamating the graphical representations of all accidents, while accounting for the causal and dependent relationships between aircraft damage and personnel injury.

In their study [170], trained and evaluated two models, *ResNet* and simple RNN, to classify the phase of flight during which the incident happened. Various NLP-based techniques were sequentially deployed including word tokenization, punctuation, unwanted characters and stopword removal, lemmatization operations and *word2vec* transformation of the unstructured textual analysis narratives extracted from the NTSB aviation incident investigation reports. The models recorded a classification accuracy of more than 68% on a 7-class classification problem.

In study [63], Nanyonga et al., carried out a comparative study of two topic modeling analysis techniques: LDA and Non-negative Matrix Factorization (NMF) regarding aviation accident reports. Using Coherence Value for performance evaluation the quality of generated topics was evaluated with LDA, displaying superior topic coherence and indicating its robustness in extracting semantic connections among words within topics. NMF, on the other hand, showcased exceptional performance in line with generating unique and detailed topics, facilitating a more targeted examination of particular aspects of aviation accidents.

[124] showcased an automated text classification approach, utilizing machine learning, that could enhance analysts' efficiency by accurately categorizing "Occurrence" in aviation incident reports, thereby enabling more precise querying of reporting databases. Using a Random Forest algorithm to classify more than 45,000 textural reports, an accuracy of 80-93% was recorded based on the ICAO "Occurrence" Category. The authors also conducted text cleaning that encompassed use of standard NLP techniques including stemming, removal of irrelevant words and symbols like stop words, punctuation characters and other special symbols, and then deployed the *n-gram* techniques including bi-gram, tri-gram, etc for feature extraction prior to passing the reports to the ML algorithm for classification.

Studies including [24, 49, 87, 90] deployed NLP-based techniques including topic modelling, and text classification, for information extraction from, and analysis of, aviation incident reports and have reported competitive results regarding causal factor analysis like human factors analysis, and aviation incident risk classification, aircraft damage classification, aviation report clustering and grouping, and many other AIbased tasks.

One research gap revealed in our literature review concerns attention-based transformers. Despite the attention based transformer models achieving outstanding performance on various NLP tasks, including machine translation [17, 183, 267, 270], text summarization [270-272], text simplification [272, 273], grammatical error correction [274, 275] and question answering [276], little-to-no attention has been paid to their deployment in the field of aviation safety to establish the likely causes of an aviation incident given the raw text analysis narrative. The work in this study aims to close this knowledge gap by proposing and training a transformer-based model for such tasks.

## 6.4. Proposed Approach

### *6.4.1. Dataset*

Several aviation, and transport safety agencies, such as the Australian Transport Safety Bureau (ATSB), Aviation Safety Reporting System (ASRS), and the NTSB, actively gather and release reports detailing aviation incident investigations. This research utilized aviation incident reports provided by the NTSB. These reports, along with accompanying metadata, are available on the NTSB's website in a variety of formats, such as monthly-published *.pdf* documents, *.json* files, or by querying individual reports through their online platform. A summarized version in *.csv* format can also be obtained. For our study, we focused on *.json* files containing detailed incident investigations from the years 2001 to 2020. Importantly, we only included incidents where investigations had been concluded, resulting in a dataset comprising 29,676 cases. From each report, we extracted the "analysis Narrative" and "probable Cause" sections to facilitate model training and validation processes. Additionally, a comprehensive statistical analysis was carried out to assess the distribution of text lengths within these fields. It was found that the average length of the "analysis Narrative" was 1,116 words, while the "probable Cause" field averaged 165 words, with standard deviations of 858.36 and 93.12, respectively. Further examination revealed that the shortest "analysis Narrative" entry contained only 4 words, while the longest reached 36,544 words. In comparison, the "probable Cause" field ranged from 7 to 1,600 words. **Figures 33 and 34** visually represent the distribution of text lengths for both the "analysis Narrative" and "probable Cause" fields.

*6.4.2.Data pre-processing*

Data pre-processing involved removing HTML tags and urls, transforming wrongly encoded characters; that is, characters encoded with the ASCII equivalent codes were decoded to their natural language characters. Also, reports whose "analysis Narrative" entries were longer than 10,000 words, and "probable Cause" entries longer than 1,000 words, were treated as outliers and discarded for this study.

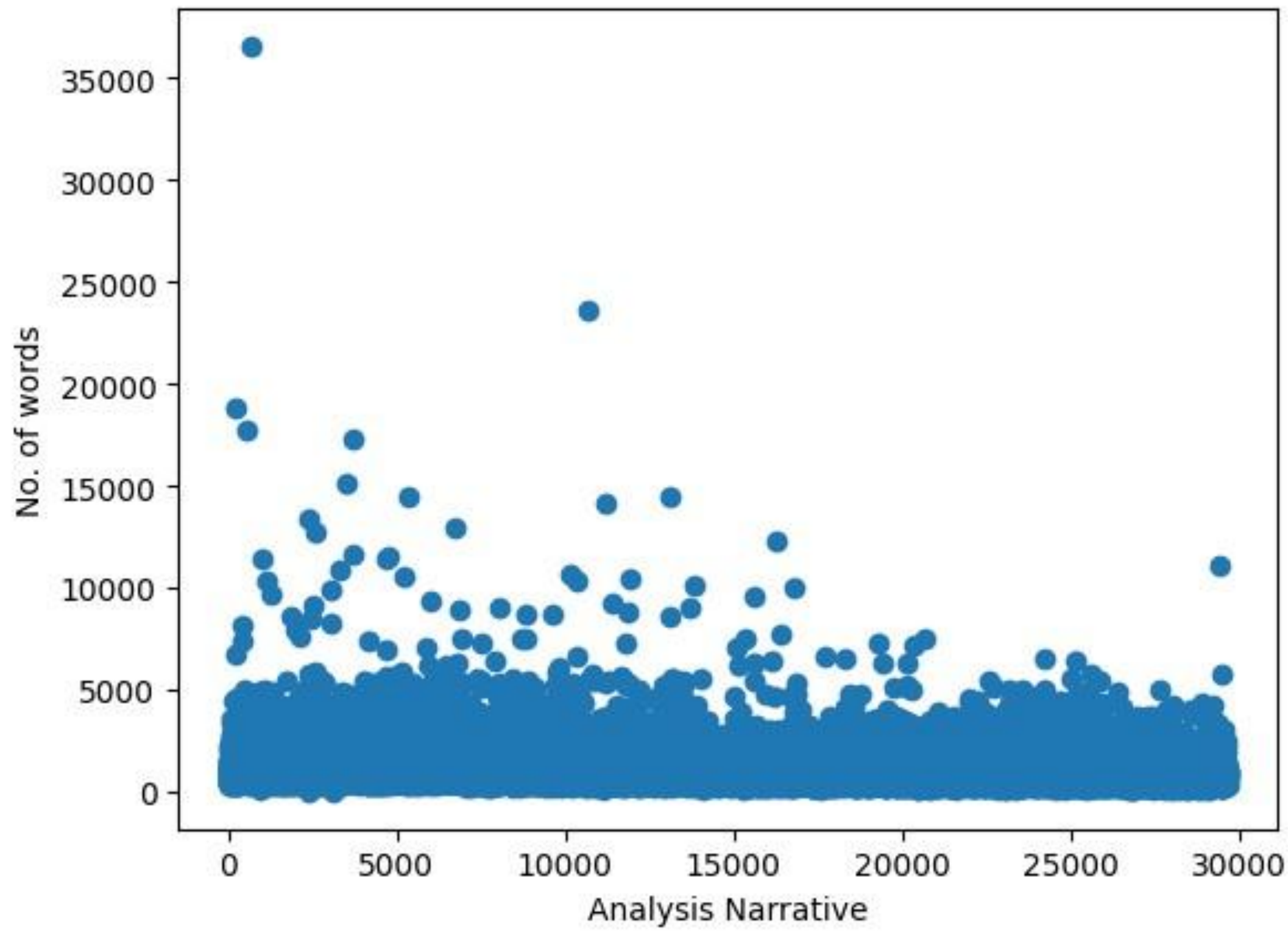

Figure 33. Text length distribution of the "analysis Narrative" field entries

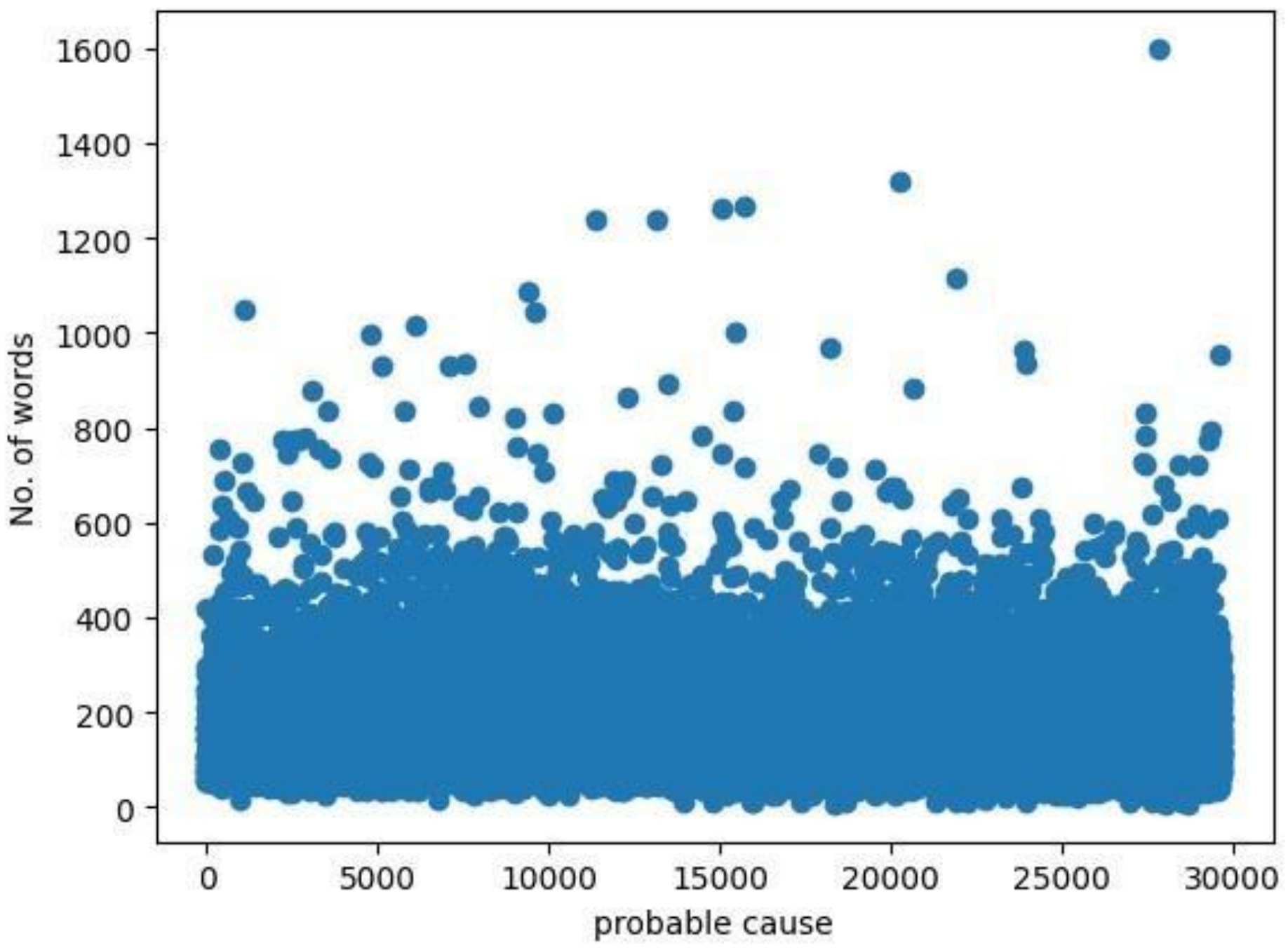

Figure 34. Text length distribution of the “probable Cause” field entries

*6.4.3.The Transformer*

The Transformer architecture, introduced in study [277], represented a revolutionary advancement in the NLP domain, yielding remarkable outcomes. Departing from traditional RNN models, the Transformer employs multi-head self-attention, enabling parallel processing and overcoming the limitations of sequential training inherent in conventional RNNs. This self-attention mechanism not only enhances computational efficiency but also captures intricate dependencies among various text components. As described by the authors, the attention process involves associating a given query, Q, with key(K)-value(V) pairs for sequence generation. Within this framework, Q, keys K, V, and the prediction are expressed as vectors. The resulting sequence is computed through a weighted summation of the V entries, with each value’s weight determined by a passing a scaled-dot product of Q and K vectors through a softmax function as depicted in Eq.(1). **Figure 35** shows the architecture of the transformer model and the architectural components of its encoder, decoder and output blocks.

$QK^T$

$Attention(Q, K, V) = softmax(\sqrt{__})V$ (1) $d_k$ Where; $d_k$ is the dimensional size of $K$.

$T$ transposes $K$ to allow matrix multiplication.

.

In order to facilitate concurrent processing, the multi-head self-attention mechanism utilizes several linear projections of Q, K, and V, each mapped to dimensions $d_k$, $d_k$ and $d_v$ respectively. These parallel operations generate outputs within the $d_v$-dimensional space, which are combined and mapped again to derive the ultimate V entries. This approach results in a model capable of simultaneously attending to information across many representational vector subspaces at various locations. The multi-head self-attention mechanism, featuring $p$ heads, is defined as presented in Eq.(2).

$$Multi\text{-}Head(Q, K, V) = cancat(head_i, \ldots, head_p)W^a \quad (2)$$

Given that $Wa \in \mathrm{R}pdv \times dmodel$

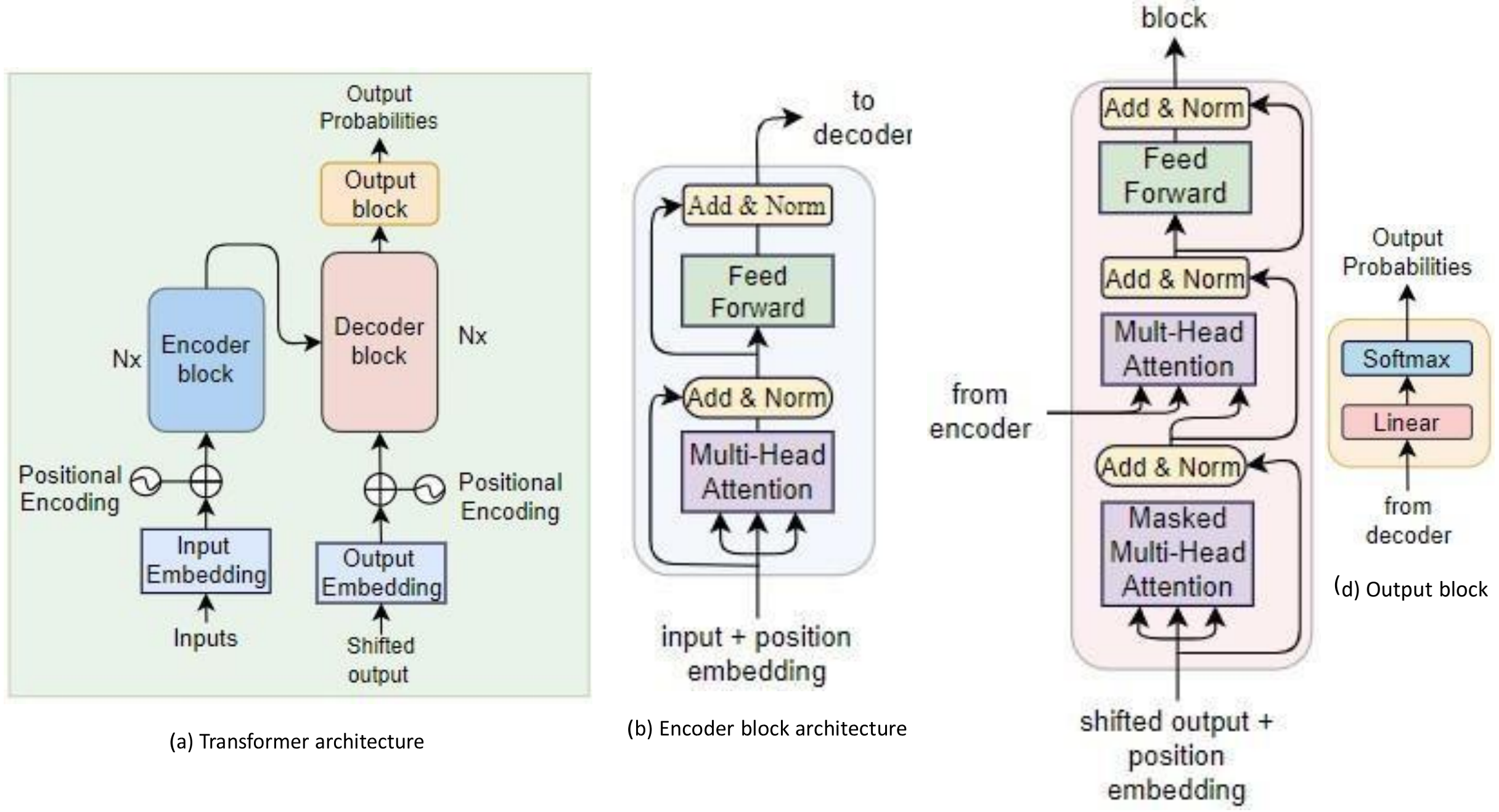

(a) Transformer architecture

(b) Encoder block architecture

(d) Output block

Figure 35.Transformer model architecture

Figure 35: Transformer model architecture and a break down of its architectural components-(a): Full transformer block architecture; (b) Architectural components of the encoder block; (c) Architectural components of the decoder block; (d) Components of the transformer output block

*6.4.4.Experimental setup*

For our experiments, the model constituted 8 encoder and decoder layers, and its embedding dimension was set to 1024 to enable long inputs. For multi-head attention, 8 attention heads were employed, while a dropout of 0.1 was used at each Batch normalization layer, and Feed-forward layer for regularisation and to prevent model over-fitting. In addition, the input sequence length was set to 1024, and the dimension of the Feed-Forward network's inner-layers was set to 2048, both the "analysisNarrative" and "probableCause's vocabulary sizes were set to 100,000.

The model was trained on 90% of the dataset while the remaining 10% was used for testing following study [278] in which this split ratio produced the best prediction results. Training was done for 50 epochs using a learning rate of 0.001 with Adam optimizer, betas were set to (0.95,0.96), epsilon set to $1e{-}10$, batch-size set to 64, and cross-entropy as the loss function.

*6.4.5.Performance metrics*

To evaluate the quality of generated probable cause, three metrics commonly used tasks involving natural language generation problems such as text summarizing, machine translation, question answering and grammatical error correction are used in this study.

a) Bilingual Evaluation Understudy (BLEU)

BLEU [279] deploys an *n-gram* based evaluation metric approach that is extensively utilized in Machine Translation assessment. It is precision-centric and assesses the degree of overlap between n-grams from the target and generated texts. This overlap is insensitive to word position, except for *n-gram* term associations. However, BLEU imposes a brevity penalty when the generated text is substantially shorter than the reference text. Besides Machine Translation, BLEU finds application in problems where the input and output use the same natural language, including grammatical error correction [280, 281], summarization [282, 283], and text simplification [273, 284], which involves rewriting a sentence into one or more simpler sentences. The BLEU score can be computed using Eq.(3) [279].

$$BLEU = BP \cdot \exp\left(\sum_{i}^{N} (w_i \cdot \ln p_i)\right) \quad (3)$$

where;

$BP \rightarrow$ Brevity Penalty, calculated using Eq.(4) $w_i \rightarrow$ order $i$ *n-gram* precision's weight. $p_i \rightarrow$ *ngram's* modified precision score of order $i$ $N \rightarrow$ maximum *n-gram* order to consider

$$BP = exp\left(1 - \frac{lp}{lravg}\right) \quad (4)$$

$l_p \rightarrow$ length of predicted cause $lr_{avg} \rightarrow$ average length of reference cause.

*6.4.5.2. Recall Oriented Understudy for Gisting Evaluation (ROUGE)*

ROUGE [285] applies a definition similar to that of BLEU. However, unlike BLEU which emphasizes precision, ROUGE's emphasis is on recall. ROUGE comes in three main versions [286, 287]: *n-rouge*, primarily examining *n-gram* overlap (such as 2-rouge and 1-rouge for 2-grams, and 1-gram respectively); *L-rouge*, which evaluates the Longest Common Text Sub-sequence; and *s-rouge*, emphasizing skip grams. Like BLEU, ROUGE finds application in both machine translation and in problems where the input and output use the same natural language, including summarizing [288-290], grammatical error correction [287, 291], and text simplification [292-294], which involves rewriting a sentence into one or more simpler sentences. For each of *rouge-1*, *rouge-2* and *rouge-L*, the precision, recall, and F-measure are calculated using Eqs.(5), (6), (7) [285].

$$Precision = \frac{Count_{mn-gram-ap}}{Count_{n-gram-p}} \quad (5)$$

$$Recall = \frac{Count_{mn-gram-ap}}{Count_{n-gram-a}} \quad (6)$$

$$F-Measure = 2 \times \frac{precison \times recall}{precison + recall} \quad (7)$$

where'

$Count_{mn-gram-ap}$ is the number of n-grams from the target probable cause matching with the predicated probable cause.

$Count_{n-gram-p}$ is the count of n-grams in predicted probable cause

$Count_{n-gram-a}$ is the count of n-grams in actual probable cause

b) Latent Semantic Analysis (LSA)

LSA [295], presented in 1997 by Landauer and Dumais in [296], calculates the semantic similarity between a reference sentence and the model's generated sentence. It relies on pre-computed word cooccurrence counts from a large corpus. Employing the bag of words (BOW) approach, it treats word order as irrelevant. Unlike ROUGE and BLEU, LSA is lenient on variations in word choice, such as" hard" versus" difficult." In essence, LSA encodes sentences or documents into vectors using a bag of words technique. These vectors enable the computation of similarity metrics, such as cosine similarity, to assess the likeness between generated and target texts. Like BLEU and ROUGE, LSA has seen application in measuring the output quality of various natural language generation models including text summarizing, grammatical correction, translation, and text simplification [297-301]. The cosine similarity between sequences, $s_1$ and $s_2$, can be obtained by converting the sequences to numeric vectors, $v_1$ and $v_2$, and then using Eq.(8) for similarity calculation [302].

$$Similarity(v_1, v_2) = \frac{dot(v_1,v_2)}{||v_1|| \times ||v_2||} \quad (8)$$

```
---------------------------------------------------------------------------
    SOURCE: According to the pilot, he made a precautionary landing on a road to investigate a possible airplane malfunctio
n. During the landing, he noted that the wind was out of the north and about 25 to 35 mph. After troubleshooting, the pilot
attempted to take off from the road to the east. According to the airplane manufacturer, the airplane's maximum demonstrated
crosswind is 20 knots at 90 degrees. During the takeoff roll, the pilot lost directional control of the airplane, and the ai
rplane departed the road. Subsequently the right main landing gear collapsed, and the right wing impacted the ground. The pi
lot reported no preimpact mechanical malfunctions or failures with the airplane or engine that would have precluded normal o
peration.
    TARGET: The pilot’s decision to take off in a crosswind that exceeded the airplane’s maximum demonstrated crosswind capa
bility and his subsequent inability to maintain directional control.
 PREDICTED: The pilot ' s failure to maintain directional control during takeoff in crosswind conditions .
---------------------------------------------------------------------------
    SOURCE: The pilot reported a loss of engine power while on an instrument approach to an airport. He was unable to reach
the airport, and collided with a tree short of the runway. Postaccident inspection disclosed no evidence of any preimpact me
chanical anomalies with the airplane or its engine. Initial responders to the accident site noted an absence of fuel smell o
r visible fuel at the site, although one wing containing fuel tanks had separated during the accident. No evidence of any fu
el stains were discovered anywhere on the airplane.
    TARGET: The pilot’s inadequate in-flight planning/decision, which resulted in a loss of engine power during an instrumen
t approach to land, and an in-flight collision with a tree and terrain.
 PREDICTED: The pilot ' s failure to follow the published instrument approach procedure , which resulted in a collision with
trees .
---------------------------------------------------------------------------

    SOURCE: The airplane experienced a total loss of engine power during cruise flight, and struck a tree during a forced
landing. After the accident, the passenger reported the engine driven fuel pump failed. Review of a flight manual for the
make and model airplane revealed that a fuel gauge was incorporated into the fuselage (header) fuel tank cap. Fuel flowed
from the wing tanks, via the engine driven fuel pump, to the fuselage tank. Once in the fuselage tank, fuel was gravity fe
d into the engine. Further review of the manual revealed that in case of engine driven fuel pump failure, the engine would
continue to function until the 6-gallon fuselage tank was drained. During an examination of the wreckage, the fuselage tan
k was observed to be empty, except for some residual fuel. The fuel was clear, consistent with 100LL aviation gasoline, an
d no contamination was observed. The engine driven fuel pump was removed from the engine for inspection. When the pump was
actuated by hand, there was no movement or continuity to its mechanical linkage that connected into the engine. The inoper
ative pump was re-attached to the engine to prevent oil leakage, and 5 gallons of gasoline was added to the fuselage tank.
The engine then started on the first attempt, without hesitation, and ran continuously at idle power and partial power.
    TARGET: A total failure of the engine driven fuel pump, and the pilot's failure to recognize a diminishing fuel supply
in the fuselage (header) fuel tank, which resulted in a total loss of engine power during cruise flight and subsequent col
lision with trees during a forced landing.
 PREDICTED: A loss of engine power due to fuel starvation as a result of the pilot ' s improper fuel management . Contribu
ting to the accident was the pilot ' s failure to comply with the airplane manufacturer ' s recommended flight procedures
for the fuel system .
---------------------------------------------------------------------------

    SOURCE: The pilot said that he normally flew the airplane with the fuel selector positioned to the right main fuel tan
k during skydiving operations. However, on the day of the accident, maintenance was performed on the airplane, and three e
ngine run-ups were performed using the left main fuel tank. The pilot ferried the airplane back to its home base uneventfu
lly with the left main fuel tank selected. Before the accident flight, the pilot verified that there was adequate fuel in
the right main fuel tank; however, he did not reposition the fuel selector to the right main fuel tank. During climb, abou
t 800 feet above ground level, the airplane experienced a total loss of engine power. The pilot was unable to restart the
engine and performed a forced landing. Subsequent examination revealed that the airplane’s right main fuel tank had been c
ompromised and was leaking fuel, whereas the left main fuel tank was intact and devoid of fuel. Additionally, data downloa
ded from the airplane's engine monitor revealed that the engine power loss was preceded by a loss of fuel flow. Postaccide
nt examination did not reveal any preimpact mechanical malfunctions or failures that would have precluded normal operatio
n.
    TARGET: The pilot's inadequate preflight preparation and fuel management, which resulted in a total loss of engine pow
er due to fuel starvation.
 PREDICTED: The pilot ' s improper fuel management , which resulted in a total loss of engine power due to fuel starvation
.
---------------------------------------------------------------------------
```

Figure 36. Some examples of analysis narratives with corresponding probable causes as presented in the original report and the model’s predicted probable causes

## 6.5. Results

For model inference, instances from the test were fed into the model, generating a probable main cause for each analysis narrative. The example of training cases is shown in **Figure 36**, where random samples of analysis narratives from the test set are passed to the model. The model generated almost semantically perfect probable causes for each input narrative.

### *6.5.1. Model performance based on the BLEU score*

The BLEU Score was used to measure how closely the predicted probable cause matched the reference probable cause. For each pair of sentences, BLEU gives a value between 0 and 1, with 1 indicating a perfect match. The minimum n-gram order was set to 1 while N was set to 4 for this work. After a series of evaluations with various random samples of size 500 from the test set, in comparison with results from other metrics, the weight vector, w was set to (0.1,0.1,0,0).

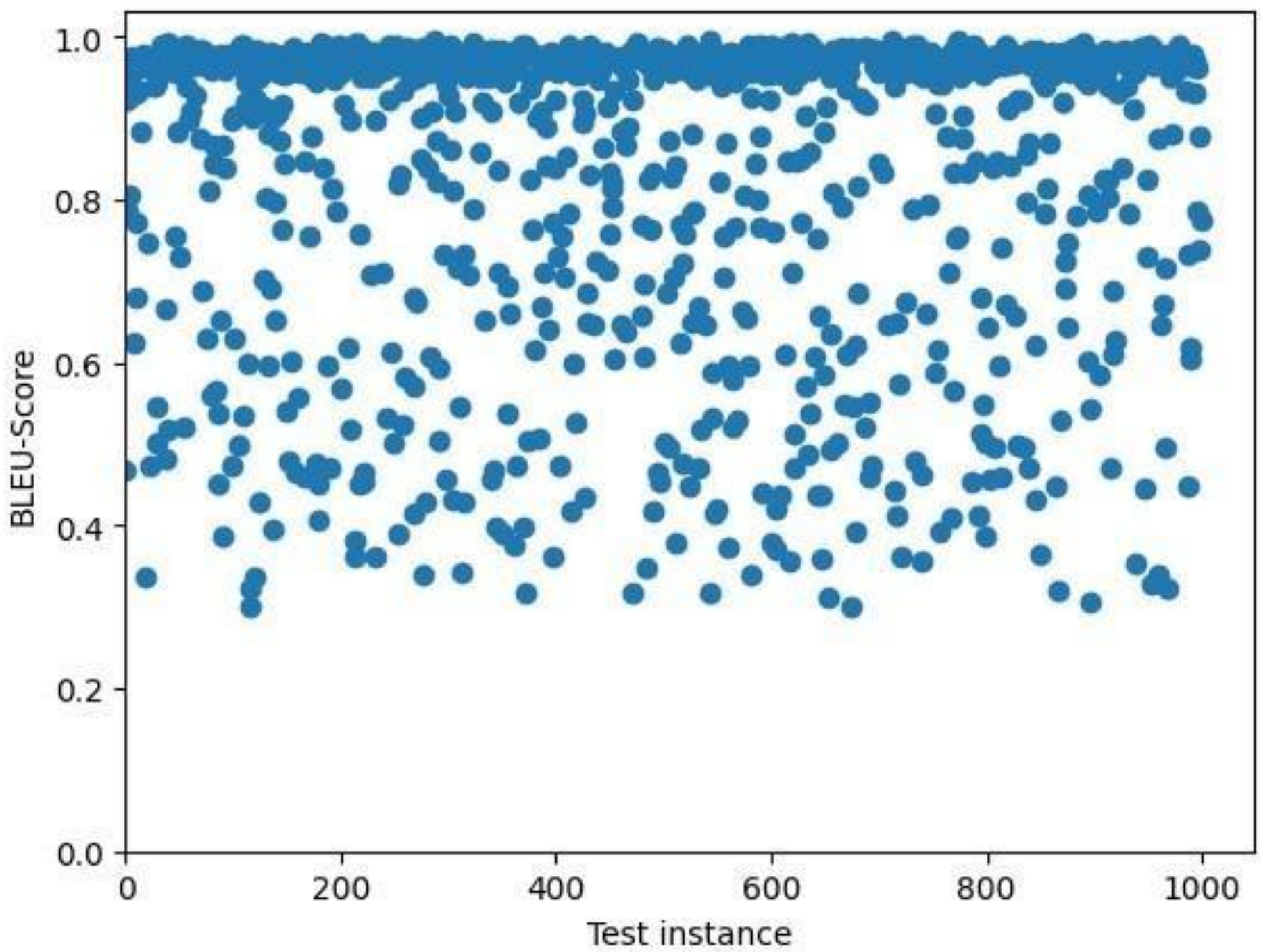

Figure 37. BLEU scores for the first 1000 test instances

For each instance in our test set, the BLEU score was computed, recording a mean score of 0.727 with a standard deviation of −/ + 0.330. A scatter distribution of the obtained BLEU scores between the first 1000 (probable cause, predicted probable cause) pairs is shown in **Figure 37**.

*6.5.2. Model performance based on the LSA Similarity score*

LSA Similarity gives the semantic similarity between vector representations of the output probable cause and target probable cause. It represents the semantic similarity rather than lexical similarity. A high similarity score implies that the sequences have closer meanings. Like the case of BLEU scores, for each instance in our test set, we obtained the (probable cause, predicted probable cause) pair. Each component of the pair was then converted into its numeric vector representation using Google's pretrained UniversalSentence-Encoder Version 4, which is the latest version at the time of writing this paper. Universalsentence-Encoder models were introduced by Google Researchers in study [303] where the cosine similarity was deployed consequently placing vector embeddings of semantically similar words close to each other. The pretrained Universal-Sentence-Encoder model used in this work can be downloaded from the TensorFlow hub [1]. Our model recorded a mean LSA similarity score of 0.697 with a standard deviation of −/ + 0.153. A distribution of the obtained Similarity scores is visualized in **Figure 38**.

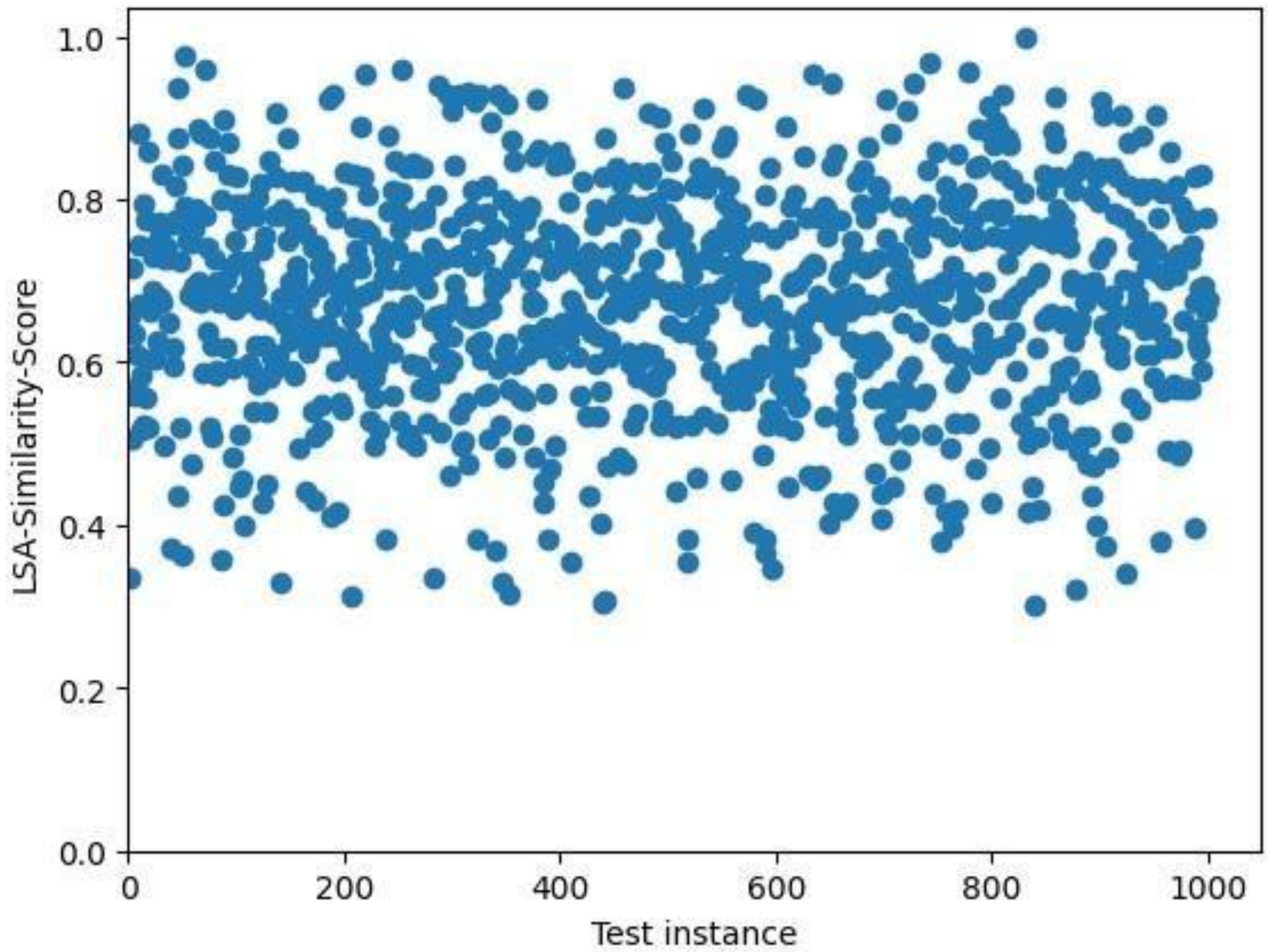

Figure 38. LSA-similarity scores for the first 1000 test instances

Table 13. ROUGE Results: Precision, Recall, and F-measure from Rouge-1, Rouge-2, and Rouge-L

| Metric | Precision | Recall | F-measure |
|---|---|---|---|
| | Mean Stddev | Mean Stddev | Mean Stddev |
| rouge-1 | 0.666 0.217 | 0.610 0.211 | 0.618 0.192 |
| rouge-2 | 0.488 0.264 | 0.448 0.257 | 0.452 0.248 |
| rouge-L | 0.602 0.241 | 0.553 0.235 | 0.560 0.220 |

*6.5.3. Model performance based on the ROUGE scores*

For ROUGE Scores, this study considered *n-rouge* (rouge-1, rouge-2) and *L-rouge*(rouge-L). These scores measure the overlap of *n-grams* between the candidate and reference sentences. *Rouge-1* gives score from

1 https://tfhub.dev/google/universal-sentence-encoder/4

unigrams, *rouge-2* gives score from *bi-grams*, while *rouge-L* gives score from the longest common subsequence. Higher scores indicate better overlap between the sentences.

*6.5.4. Analysis narrative length Vs BLEU/LSA scores*

We further investigated how the length of the input analysis narrative impacted the model's output in terms of the BLUE and LSA similarity scores. The results revealed that the analysis narrative length had no direct correlation with the model's BLUE score as shown in **Figure 39**. On the other hand, the LSA similarity score shows no correlation with the length of the input analysis narrative for shorter inputs. However, it tends to converge to the mean score as the length of the analysis pattern increases as shown in **Figure 40**. This finding emphasizes our hypothesis where we stated that working with long input sequences would enhance the model's predictive performance. This also emphasises the finding of prior studies on long-input transformers including [265-268] which revealed that increasing the Transformer's input length positively correlates with model performance.

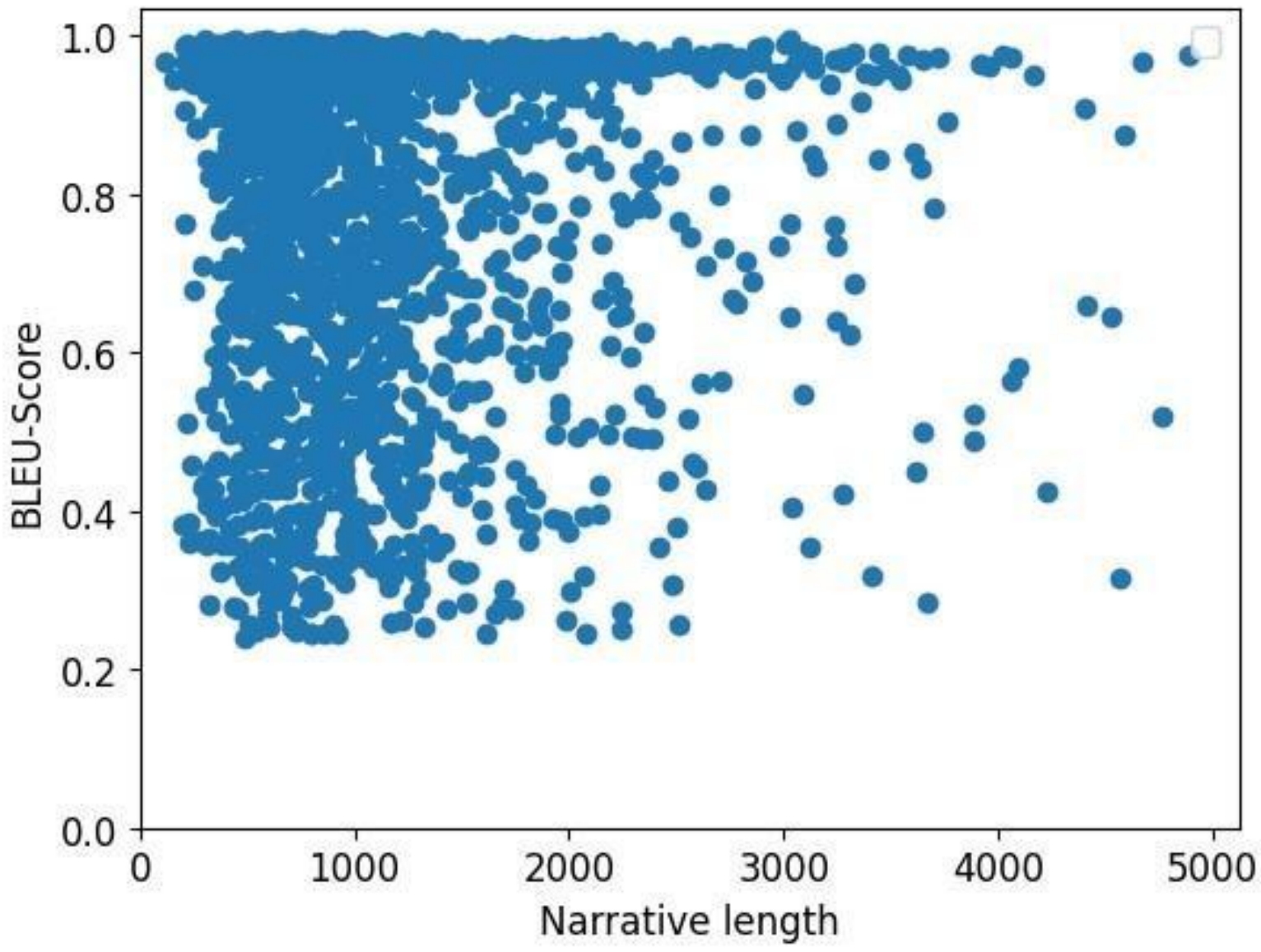

Figure 39. Impact of Analysis narrative's length on the model's BLEU score.

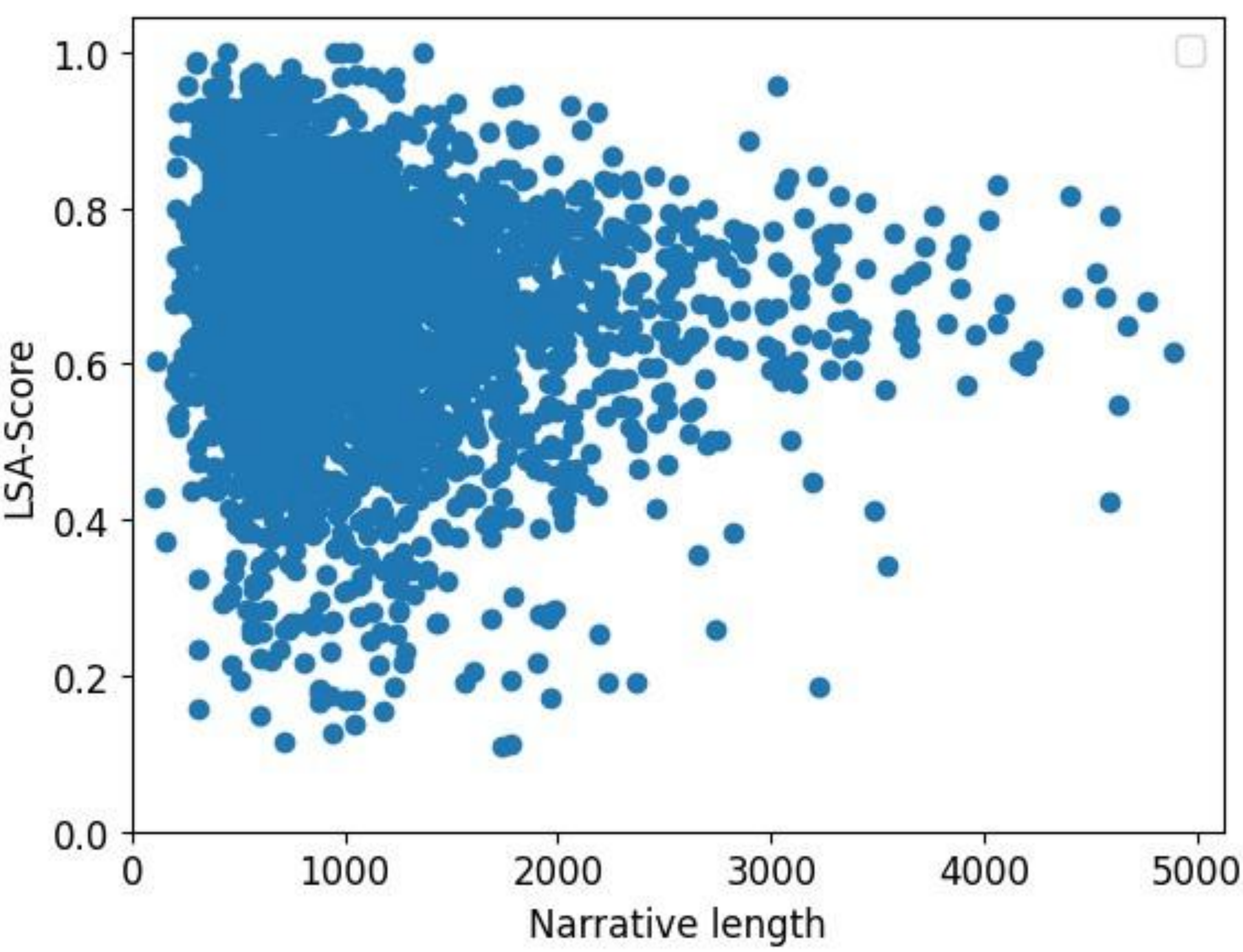

Figure 40. Impact of Analysis narrative's length on the model's LSA similarity score.

## 6.6. Discussion

Having the ability to predict the probable causes of an aviation incident can greatly expedite the investigation process. The results from this study revealed that a multi-head attention-based transformer model is a tool for solving this problem. However, although the model recorded commendable results across all metrics, by their formula, the LSA similarity score is more reliable compared to BLEU and ROUGE metrics. This is because the model's output and reference sentence can constitute a different set of word for the same semantic content. Since the LSA similarity score computes the overall semantic similarity between the sentences, it will more likely produce a high score if the two sentences are similar and vice-versa. On the other hand, the BLEU score requires that the weight vector, *w* for each n-gram is manually determined. This means that the final BLEU score greatly depends on the accuracy of the values of *w* which requires a human expert and if wrongly determined can lead to misleading results. Also, the computation of BLUE and ROUGE scores like the unigram, bi-gram, etc, depends on the overlap of words between the reference and predicted sentence, that is, observed probable cause and predicted probable cause for this study.

```
-----------------------------------------------------------------------------
    SOURCE: The pilot reported that, while on short final for landing, he heard a "high pitched whine" and that the helicopt
er's tail rotor authority subsequently degraded. He immediately initiated a full autorotation to landing. During the landing
in a field with uneven terrain, the helicopter was substantially damaged.Postaccident examination revealed that, during the
flight, the tail rotor driveshaft (TRDS) had become uncoupled from the drive assembly and main rotor belt transmission. The
forward TRDS retention nut was found safety wired; however, it was less than hand tight. The main rotor transmission input a
ft pinion nut was found backed off the pinion; the nut and pieces of its fractured cotter pin were found captured in grease
contained in the TRDS fitting grease cavity. The nut was intact, but it exhibited thread wear. The cotter pin was shorter an
d thinner than the cotter pin specified in the helicopter manufacturer's illustrated parts catalog. The grease in the aft pi
nion splines was discolored, caked, and dry, indicating that it had not been serviced in a long time.The main rotor transmis
sion input pinion exhibited significant wear on the forward spline set and the aft spline set that engaged the driving splin
e, which was normally retained by the aft pinion nut. The aft upper H-frame bearing was displaced on the driving spline. The
driving splines' interior spline teeth exhibited severe wear matching the wear on the pinion, and rotation damage was found
on the spline ends. Other areas of the pinion exhibited fretting corrosion.According to the helicopter manufacturer's Handbo
ok of Maintenance Instructions, during the 1,200-hour inspection, the mechanic was required to remove the upper pulley, insp
ect the pinion, and torque and safety wire the aft pinion nut during reassembly. The mechanic who performed the 1,200-hour i
nspection (and others) stated that, during the inspection, he had "never gotten into" the drive assembly, had not checked th
e torque on the aft pinion nut, and had not inspected the cotter pin.
    TARGET: The mechanic's improper maintenance of the main transmission aft pinion nut and belt drive system, which resulte
d in the uncoupling of the tail rotor driveshaft and the subsequent loss of helicopter control.
 PREDICTED: The failure of the main rotor drive belts due to a loss of belt tension on the main rotor drive system as a resu
lt of maintenance personnel s failure to properly secure the - nut and the helicopter s main rotor drive belts .
```

Figure 41. Reference model output screenshot for discussing the BLEU, ROUGE and LSA Similarity scores: The model's output constitutes a slightly different word set from the reference probable cause For instance, considering the output in the screenshot in **Figure 41**, the reference probable cause as given in the dataset is:

"*The mechanic's improper maintenance of the main transmission aft pinion nut and belt drive system, which resulted in the uncoupling of the tail rotor driveshaft and the subsequent loss of helicopter control*".

While the model's prediction given the same analysis narrative, is:

"The failure of the main rotor drive belts due to a loss of belt tension on the main rotor drive system as a result of maintenance personnel's failure to properly secure the - nut and the helicopter's main rotor drive belts."

Although the semantic meanings of the two narratives are close and would both draw the incident investigator's attention to the same component and attribute the failure to the maintenance personnel's not properly securing the nut and belt drive system, BLEU scores differed across different weight vector values as shown in Table 14.

On the other hand, the ROUGE scores were rouge-1: precision=0.476, recall=0.606, Fmeasure=0.533, rouge-2: precision=0.146, recall=0.188, Fmeasure=0.164 and rouge-L: precision=0.310, recall=0.394, Fmeasure=0.347. As can be seen, the results from the BLEU score largely depend on the values of vector $w$. It is also clear that the score greatly degrades when $w$ contains entries for the tri-gram and quad-gram which correspond to the third and fourth entries of $w$ respectively. The value is also, misleading for very small entries of the uni-gram and bi-gram as seen when $w$ is set to (0.01,0.01,0,0).

Table 14. BLEU-scores for various weight vector values

| Weight vector | BLUE-Score |
|---|---|
| [0.1,0.1,0.1,0.1)] | $8.67 \times 10^{-32}$ |
| [0.01,0.01,0.01,0.01] | $7.83 \times 10^{-4}$ |
| [0.25,0.25,0,0] | 0.459 |
| [0.1,0.1,0,0] | 0.732 |
| [0.01,0.01,0,0] | 0.969 |

Generally, the recorded scores in the case of ROUGE metrics are relative more reliable for the *rouge-1* and *rougeL*. The *rouge-2* has recorded poor performance due to the fact that the word sequence in the reference text does not always overlap with the word sequence in the model's output. For the example output in **Figure 40**, the recorded ROUGE scores are poor in terms of *precision*, *recall*, and *F-measure* for all the three n-grams used in this study despite the semantic meaning being very similar. On the other hand, because the LSA returns the semantic similarity between two text sequences, its output is considerably high (0.757) for this particular example indicating that despite the discrepancies in the used set of words, the semantic meaning is greatly similar.

Finally, the LSA Similarity score's input length-model performance analysis indicated that training the model with long inputs can result in stable model performance as the score converged to the mean score with increasing input length **(See Fig41).** It is worth noting that training a highly efficient transformer model requires huge amounts of training data which was a great limitation for this study.

## *6.7.* Conclusion

Identifying potential causes of aviation incidents quickly is crucial for preventing future tragedies. While flight data recorders are commonly used, delays or damage can obstruct their effectiveness. The Boeing 737 MAX accidents with Lion Air and Ethiopian Airlines highlight the impact of such delays. To improve investigation efficiency, this study developed a transformer-based model for predicting the probable cause of an aviation incident given an analysis narrative of the pre/post incident series of events that can be collected from sources including eyewitnesses, radar systems, Air traffic controllers that oversaw the flight under investigation, maintenance history/logs, etc. The model was trained on extensive aviation incident reports from the NTSB and allows both short and long-input narratives. This approach shows promise in expediting investigations and enhancing aviation safety through key metrics like BLEU, ROUGE, and LSA.

We assume that, with a larger training dataset, the model's output can improve. Therefore, as direction of future work, we intend to combine analysis narratives from other Aviation investigation bureaus like the ATB, with the NTSB narratives and retrain our model on a larger dataset for improved predictions.

# Section III. Deep Learning

### Chapter 7. Sequential Classification of Aviation Safety Occurrences with Natural Language Processing

Nanyonga, A., Wasswa, H., Turhan, U., Molloy, O., & Wild, G. (2023). Sequential classification of aviation safety occurrences with natural language processing. In AIAA AVIATION 2023 Forum (p. 4325).

7.1 Abstract.

Safety is a critical aspect of the air transport system given even slight operational anomalies can result in serious consequences. To reduce the chances of aviation safety occurrences, accidents and incidents are reported to establish the root cause, propose safety recommendations etc. However, analysis narratives of the pre-accident events are presented using human-understandable, raw, unstructured, text that a computer system cannot understand. The ability to classify and categorise safety occurrences from their textual narratives would help aviation industry stakeholders make informed safety-critical decisions. To classify and categorise safety occurrences, we applied natural language processing (NLP) and AI (Artificial Intelligence) models to process text narratives. The study aimed to answer the question, "How well can the damage level caused to the aircraft in a safety occurrence be inferred from the text narrative using natural language processing?" The classification performance of various deep learning models including LSTM, BLSTM, GRU, sRNN, and combinations of these models including LSTM+GRU, BLSTM+GRU, sRNN+LSTM, sRNN+BLSTM, sRNN+GRU, sRNN+BLSTM+GRU, and sRNN+LSTM+GRU was evaluated on a set of 27,000 safety occurrence reports from the NTSB. The results of this study indicate that all models investigated performed competitively well recording an accuracy of over 87.9% which is well above the random guess of 25% for a four-class classification problem. Also, the models recorded high precision, recall, and F1 scores above 80%, 88%, and 85%, respectively. sRNN slightly outperformed other single models in terms of recall (90%) and accuracy (90%) while LSTM reported slightly better performance in terms of precision (87%). Further, GRU+LSTM and sRNN+BLSTM+GRU recorded the best performance in terms of recall (90%), and accuracy (90%) for joint models. These results suggest that the damage level can be inferred from the raw text narratives using NLP and deep learning models.

## 7.2 Introduction

Air transport is among the most sensitive fields to operational errors in the transport industry. It is a highly safetycritical industry where even the slightest mistake or minor system misconfiguration can result in a serious disaster whose consequences may include, though not limited to, loss of lives, huge financial losses, loss of customer trust, and property among others [22]. When an aviation safety incident occurs, be it minor, serious, or fatal, reporting and investigations are required. The main goal of this reporting on safety occurrences is not to assign blame but to ensure that similar events are prevented.

The report from aviation safety occurrence, either investigations or self-reports, containing unstructured text narratives, in a human-understandable language [24, 29, 87, 190], highlighting the series of events that can be considered probable causes of the incident/accident. Also, these reports are freely available to the public. The goal of publishing these reports is partly to enable various stakeholders in the aviation industry ranging from the maintenance teams, operational manager to researchers in the aviation safety field access, analyze, critique, and implement/follow the recommended safety standard operating procedures, rules, and regulations.

However, in the event of a similar occurrence, investigation teams tend to revisit the reports from the previous occurrences. This can be extremely time-consuming if the conditions in which the incident in question occurred relate to conditions from many previous incidents. This is because, despite most reports including a tabular summary of the conditions in which the event occurred to make them more readable to humans, the actual detailed series of events that tend to be the probable incident/accident causes are usually given as a narrative using some natural language understood by the human stakeholders.

In addition, when reporting such narratives, humans often use non-standard terms that can be understood by fellow humans but not computers. This implies that, in case of a need to revisit these reports, humans will have to search through a large database of reports and sequentially read each of the related reports in a one-by-one approach. Consequently, this manual approach delays the investigation process. To mitigate this challenge, various models have been proposed to accelerate aviation data processing and aid in safety-critical decisions by the various aviation authorities and stakeholders. Approaches such as the famous multiple linear regression (MLR) have seen a wide deployment by numerous studies [304, 305] to propose models that predict, and consequently plan for, the air transport market to maximize its financial and economic related benefits.

In addition, state-of-the-art models have been proposed to predict the same, using both traditional and modern/advanced machine learning algorithms such as Artificial Neural Networks (ANN) [29, 32] Random Forest (RF), [306] Support Vector Machine (SVM) [91, 109] and many more. However, most of these studies have focused on structured data, paying little or no attention to the unstructured text narrative. A few studies that have deployed natural language processing to aviation incident reports, have mostly used models like Latent Dirichlet Allocation (LDA) to execute topic modelling tasks in the aviation industry [171, 172]. Though topic modeling can be used to perform classification of safety data report into several topics, its efficient falls with increasing topic overlap/ambiguity as a set of many words can be assigned to more than one topic/class making it highly unreliable. Although, deep learning models have gained increased popularity and are embraced across aviation field [63]. NLP for spam text email detection [307] and in bioinformatics for breast cancer detection, myocardial infarction survival prediction [306], and many more, it is not yet fully embraced by the aviation industry vis-a-vis big data analysis.

In this work, we apply NLP on aviation incident/accident reports for text analysis and advanced deep learning algorithms including Long Short-Term Memory (LSTM), Bidirectional-LSTM (BLSTM), Gated Recurrent Unit (GRU), and Simple Recurrent Neural Network (SimpleRNN) for classification [308]. We compare the classification performance of each algorithm against the rest of the algorithms used in this study. The underlying research question is, given a text narrative describing the series of safety occurrence events, can we predict the resultant aircraft damage level? We based our study on the categories given in the NTSB investigation reports where for any aviation accident/incident the damage level to the aircraft is classified as either destroyed, substantial, minor or none. This can help the aircraft maintenance team to make informed decisions e.g., whether the aircraft can be put back to operation or not and if yes, how much repair does it need and consequently, anticipate the require budget, and other resource requirements and financial implications to the aviation company. However, this can have profound consequences if a wrong classification is made and can lead to unrecoverable losses or even forcing the company out of business. The rest of this work is organized as follows. Section III gives an account of the existing literature regarding aviation safety and machine learning. Section IV gives details of the proposed approach, and the implementation procedure followed to realize it. In section V we present the results of the study and give a detailed discussion and their implication to aviation safety research. Section VI presents the conclusion highlighting the direction of future work.

## 7.3 Related Work

According to [309] machine learning enables computers to learn and make rational decisions based on experience, action, and reaction. Machine Learning (ML) has been successfully used in many fields of aviation, medicine, bioinformatics, biology, and many others. The most significant application of machine learning is data mining and different NLP and machine learning (ML) techniques have been applied to aviation safety reports. This part looks at different recent studies on text classification and text mining techniques that have been applied on various occasions.

Different studies applied different RNN architectures on sentence modelling where CNN and their results shows that these models are more suitable when applied sequential dataset such text mining, [310-312] they applied RNN technique for predicting weather related task and they reported that with their results they were able to regulate information recorded before the flight take off, [89] looked at different NLP techniques on aviation reports his main focus was to examine all the existing NLP that has previously been used on specifically civil aviation corpus though the performance of those techniques were not clearly reported. His study recommended the use of NLP, especially RNN, on text mining of time series data.

In his study, [112] proposed a deep learning approach for obtaining meaningful narratives from aviation safety reports. His study analyzed 186,000 ASRS (Aviation Safety Reporting System) reports using word2vec model for identifying a set of similar terms and find semantic terms within those reports. His proposed model showed that it can help in improving the set of terms used by reporting experts if used in aviation reports and hence reduce the uncertainty of existing safety reports thus making it easy for experts to understand. Though his study focused on only one NLP tool hence and hence a need for an improvement using different NLP approaches.

In their study, [313] used sequential deep learning for aviation safety prognosis on NTSB reports. They developed classification models using LSTM and word embedding and reported that these models were good and would help aviation experts in reviewing and analyzing all safety investigation reports. Their proposed method was done in two parts where they first transformed data extract on NTSB dataset into label and DL models were later developed for prediction regarding accident and other events.

In another study in predicting excess Events in Aircraft engines [314], applied an LSTM model to prediction task using data from 85 flights with a total of 79557 seconds (about 22 hours) of data. They aimed to examine suitable RNN using LSTM architecture neurons in predicting vibrations of aircraft engine. After model training and testing from aircraft engine vibration dataset, the proposed model was able to predict vibration values. LSTM architecture was compared with traditional RNN, and it showed that LSTM RNNs (recurrent neural networks) can correctly predict the vibration on flight dataset.

The authors in [138] evaluated the ability of different ML techniques to determine critical causes of accidents in air transport and its impact. To be precise, their study used decision tree, Naïve Bayes, and Sequential minimal optimization (SMO) to classify aviation accidents. Their study classified several airplane accidents and after the analysis, the results showed that the decision tree algorithm was better in predicting the cause of accident. It showed that loss of attention (human factor) as the most cause of accident in aviation and this would help air transport experts to minimize the loss of lives. The focus of their study is based on structured datasets which unstructured datasets also need to be considered based on different ML algorithms.

In a partnership study between CLLE-ERSS research lab and CFH safety company, [24] carried out a description of several NLP tools and text mining methods that can appropriately be used on aviation report unstructured data. The techniques and experiments conducted in their research only based on natural language processing tools and they based on [315] that suggested that these NLP techniques are more important when it comes to accident report analysis hence reported that NLP can be so helpful at any level of accident and incident report starting from reporting the incident to the final stage of analysis. Their study compared NLP on different kinds of dataset from different databases including ASRS, ECCAIRS (European Coordination Centre for Accident and Incident Reporting Systems), the Accident/ Incident Data Reporting (ADREP), DGAC and CFH / Safety Data. These are found in the public domain for use with safety reports. These studies did not report the exact NLP tools they used, and it is not clear if some tools could be used for better performance.

In their work, [308] produced and tested different RNN models on a dataset from NASA's Ames and Langley Research Centers. Their research was examining and predicting F-16 fighter jet data driven modeling and the fault sensor condition. They used different RNN architectures, and this include simpleRNN (sRNN), gated recurrent unit and long-short-term memory (LSTM) and a combination of these RNN architectures was later carried out where GRU-LSTM, GRU-sRNN, and LSTM-sRNN was combined to each other to determine the best performing layer. The study based on the performance of these models to predict the angel of attach of F-16 flight jet, faulty of angel of attach measurements and many others. The study showed that sRNN using trained estimated algorithm (Adam) achieved better performance in predicting as compared to all other algorithms used and sRNN combined with GRU performed better than all other networks that were used in their study.

Their study [171] used BLSTM, CNN and RNN to analyze aviation accident narratives and their main aim was to improve the text classification accuracy with a better understanding of accident records. Their study showed that these models can increase the accuracy on accident dataset and a recommendation on using more models on accident narrative dataset on was made. Although the proposed model performed better but its performance was less than 80% and if different parameters were used it would yield better performance of these models.

According to reference [123] proposed a 4-step approach to construct a network based on Bayesian probability distribution that would infer the causal relationships between a series of aviation accidents. Their approach utilized aviation data collected from 1982 to 2006 as reported in the National Transportation Safety Board (NTSB) aviation accident database. However, although their approach captured most of the causal relationships and recorded commendable performance, the dataset used for their study is incredibly old and does not capture the current trends and features embedded in today's aviation data. This is attributed to the fact there has been great advancement in aviation technology which translates into datasets with different distribution patterns

In another study, [316] proposed a data management scheme that deployed ontologies and conceptual models to enhance aviation safety data management. The main aim was to improve civil aviation management, repair, and overhaul maintenance. However, their approach focused on modelling conceptual frameworks and does not deploy any advanced data analytics or machine learning techniques. Though the proposed scheme recorded commendable performance, it does not adapt to the dynamic nature of the current aviation environment with big data being the denominator of most operations.

In their work, [100] used PCA (Principal Components Analysis), a flight incident prediction model based on deep brief networks and principal component analysis. The deep brief network architecture has the benefit that each layer learns a separate set of complex features than the layer before. PCA on the other hand, that has proved to be handy in dimensional reduction, it is blind to data classes and often leads to reduced classification performance. This would degrade aviation safety whose consequences may be extremely catastrophic.

## 7.4 Methodology

To realize the proposed approach, phases including training data set identification, text processing, and classification were executed as shown in **Figure 42**. The next subsection gives a brief description of the data sets used in this study.

### *7.4.1 Data Acquisition*

Various bodies, including ATSB (Australian Transport Safety Bureau), ASRS, and NTSB, among others collect and publish aviation incident/accident investigation reports. For this study, we used the NTSB aviation incident/accident investigation reports. Depending on the nature of the problem in question, this dataset can be accessed from NTSB website, together with the meta data in various forms including individual monthly published .pdf reports, .json files, or by querying individual reports online or a .csv summary. In this work we downloaded a .json file containing incident/accident investigation details for the years 2005 to 2020. Also, for the purpose of this study we considered incidents whose investigations were completed resulting in a dataset with 16919 records distributed among the four classes as 1409 destroyed, 15163 substantial, 195 minor and 152 with no damage to the aircraft. For each report we extracted the 'analysis Narrative' and 'damage Level' fields for training and validation of our deep learning models.

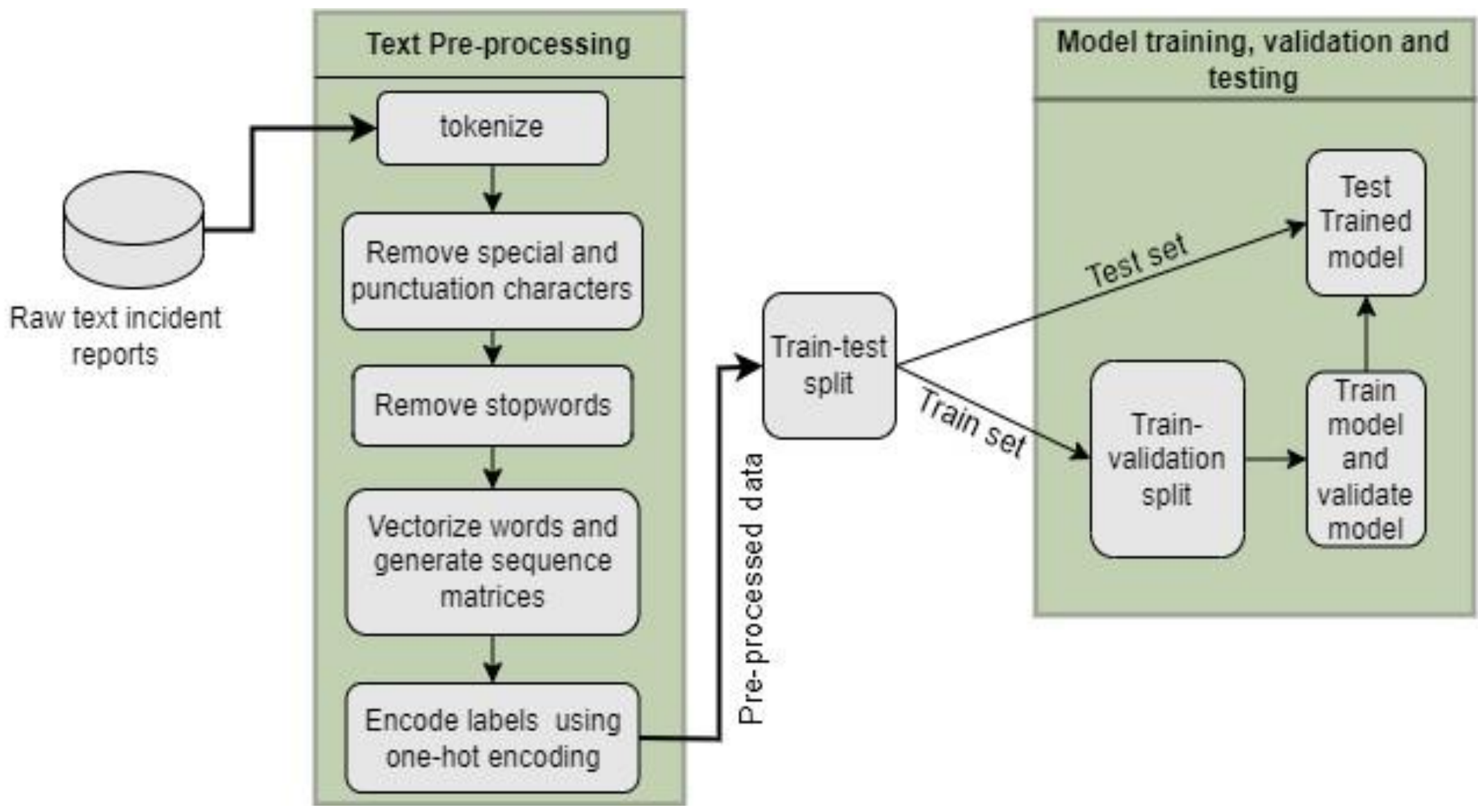

Figure 42. Methodological Framework

*7.4.2 Text Processing*

Machine leaning models are not designed to understand raw text data and therefore text data must be transformed from the human readable form to a numeric form that can be understood by the models. The Keras deep learning library was used since it hosts a whole bunch of deep learning models, and the various variants of model layers used in in this study. In addition, it provides some advanced modules for text preprocessing including the Tokenizer module which generates tokens and the sequence vectors for the input text, the categorical module which maps each of the unique categorical entries of the damage level variable (i.e., destroyed, Substantial, Minor and None) to a unique numerical entry or as a one-hot encoded entry for every data instance. For unwanted special and punctuation characters, and stop-words removal, and word lemmatization, the spacy library was used. Spacy is a Python library designed specifically to perform text processing tasks like named entity recognition, and word tagging and also incorporates a comprehensive list of special characters, punctuation and stop-words and is regularly updated whenever need arises. Using the above tools, each of the input text narrative was processed and transformed into a representative sequence/vector of length 2000. Numeric sequences generated from text narratives with number of words less than 2000 are padded with zeros while those whose length exceeds 2000 are truncated. The corpus vocabulary was set to 100,000. Scikit-learn's train-test-split module is deployed for splitting the data set into training, validation, and testing sets. All experiments are conducted using Python as the programing language and Jupyter notebook as a code editor.

*7.4.3 Text Classification*

The processed data set is then randomly split as 80% train, and 20% test sets. In addition, 10% of the train set is held out for each epoch for model validation during training. Deep learning models including CNN, LSTM, BLSTM, GRU, SimpleRNN, and combinations of these models including LSTM+GRU, BLSTM+GRU, SimpleRNN+LSTM, SimpleRNN+BLSTM, and SimpleRNN+GRU are trained, and their performance evaluated and compared against other models. For model optimization, the Adam optimizer was deployed. However, this study was not concerned about finding the best optimizer and therefore any other optimization can be used.

*7.4.4 Deep learning model Architecture*

To maintain uniformity, with the exception of slight variations for combined models, all models used a common architecture constituting an embedding layer, hidden layers and output layer. For all hidden layers, Rectified Linear Unit (ReLU) was deployed for activation while the output layer deployed SoftMax as the activation function. To generate the predicted class, the argmax function was used. This returns the index corresponding to the entry with highest probability from the SoftMax output. **Figure 43** shows the general deep learning architectures used in this study for a single RNN-based deep learning algorithm (a) and when two different RNN-based deep learning algorithms are combined (b).

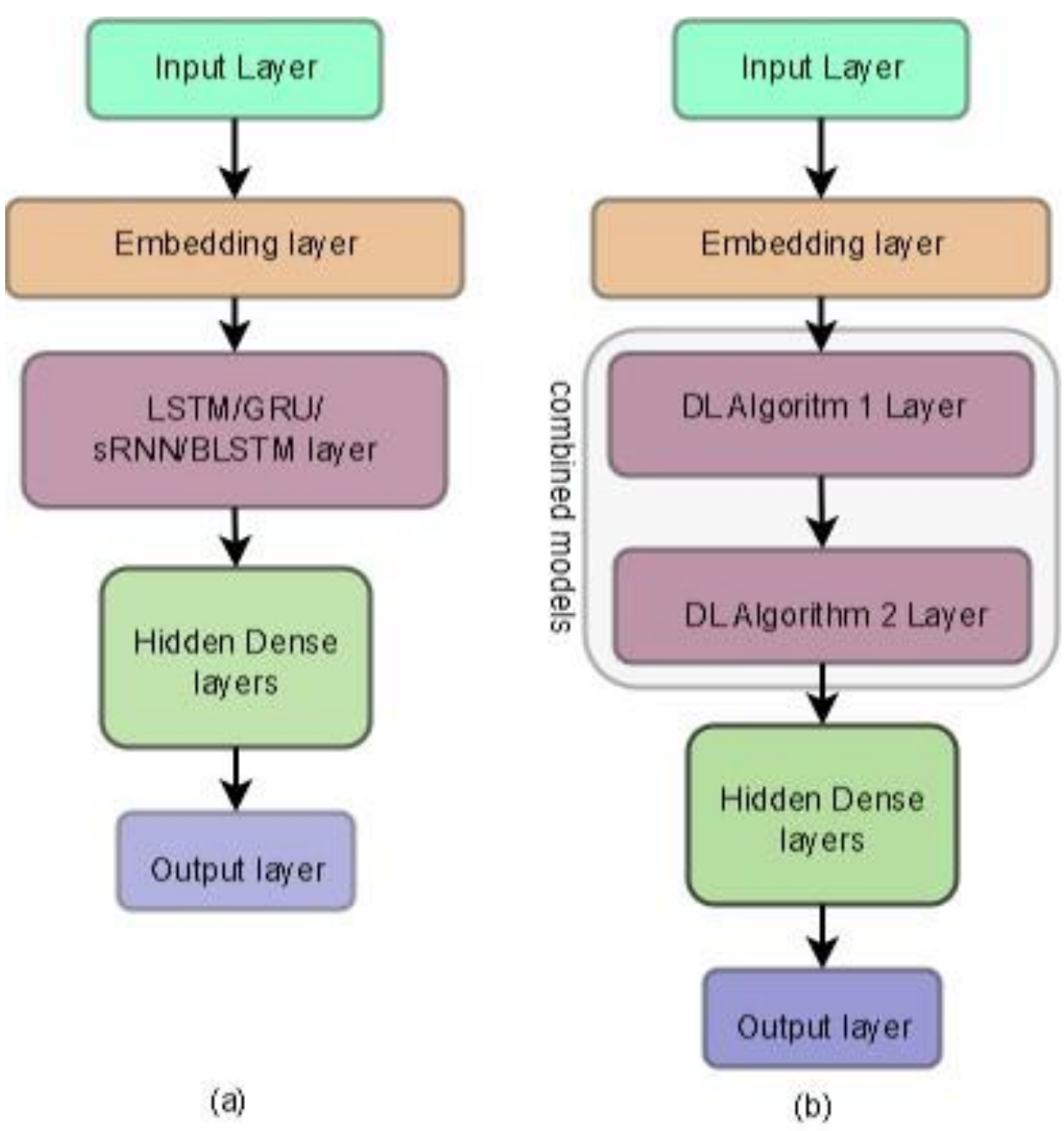

Figure 43. Deep learning architectures: (a) using single deep learning algorithm and (b) using two different Deep learning algorithms.

## 7.5 Results and Discussion

The results from the experiments conducted in this study showed that, using natural language processing together with advanced/modern deep learning models, one can, with a high degree of accuracy and competitive precision, predict degree of impact to the aircraft in terms of damage level given the unstructured text narrative of the pre-accident series of events. In this section we present and discuss the experimental results obtained for each deep learning model used accompanied with a precise description of their implications. As already indicated in section, we compared the performance of different deep models for classification of accident narrative reports. The architectures were evaluated in two perspectives, i.e., single model architecture where a single recurrent neural network layer was used for instances GRU, sRNN, LSTM, and BLSTM and a joint RNN architecture where the model constituted layers of two or more RNN models for instance GRU with LSTM, BLSTM, sRNN as GRU-LSTM, GRULSTM, GRU-sRNN and sRNN with LSTM, BLSTM as BLSM-sRNN, LSTM-sRNN.

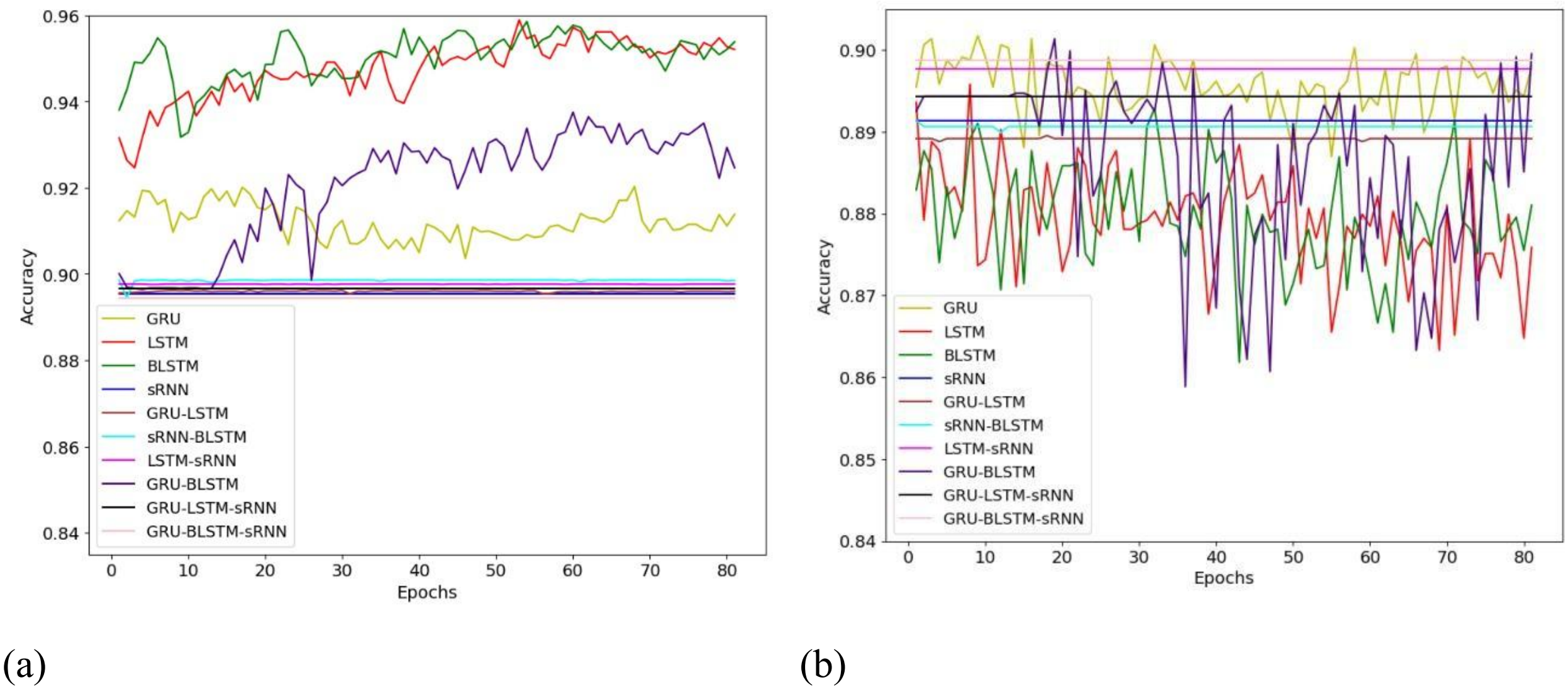

Figure 44. Single RNN-based Model performance in terms of training (a) and validation (b) accuracy

### 7.5.1 Detection performance of single RNN-based deep learning models

In this section, the individual performance of each model is presented and compared with others in terms of training accuracy, validation accuracy, testing accuracy, precision, recall and F1 measure. It is worth noting that in the context of this study the term single RNN-based deep learning model refers to a deep learning model that embeds only one RNN-based layer (see Figure 43.( a)) while joint RNN-based deep learning model refers to a deep learning model that embeds two or more RNN-based layers with the output of one layer serving as input for the subsequent layer of a different RNN architecture (see Figure 43.(b)). Table 15 shows the performance of each of the four single RNN-based deep learning models used in this study. Apart from testing performance the training and validation results are analyzed and presented in Figure 44.

Table 15. Single RNN-based Deep learning model performance

| Modals | Precision (%) | Recall (%) | F1-Score (%) | Accuracy (%) |
|---|---|---|---|---|
| LSTM | **87** | 89 | **88** | 88.9 |
| BLSTM | 85 | 88 | 86 | 87.9 |
| sRNN | 82 | **90** | 86 | **90.0** |
| GRU | 85 | 89 | **88** | 89.0 |

It can be seen from Table 15 that all models perform competitively well with simple recurrent neural network (sRNN) slightly outperforming others in terms of recall and accuracy while LSTM reported slightly better performance in terms of precision. This result, indicates that, the damage level caused to the aircraft can, with an accuracy far above random guess, be predicted from the narrative of pre-accident series of events given that the unstructured narrative is appropriately processed and transformed into a correct state for machine learning models.

7.5.2 Detection performance of joint RNN-based deep learning models

Aside from single RNN-based deep learning models, joint RNN-based models were evaluated on the same data set. Table. 16 presents the prediction results from the six different combinations of the four RNN-based models evaluated in this study. The training and validation results of joint models are visualized in Figure 45. A comparison of the test performances between single models and joint models in terms of accuracy, precision, recall and F1-score is shown in Figure 46.

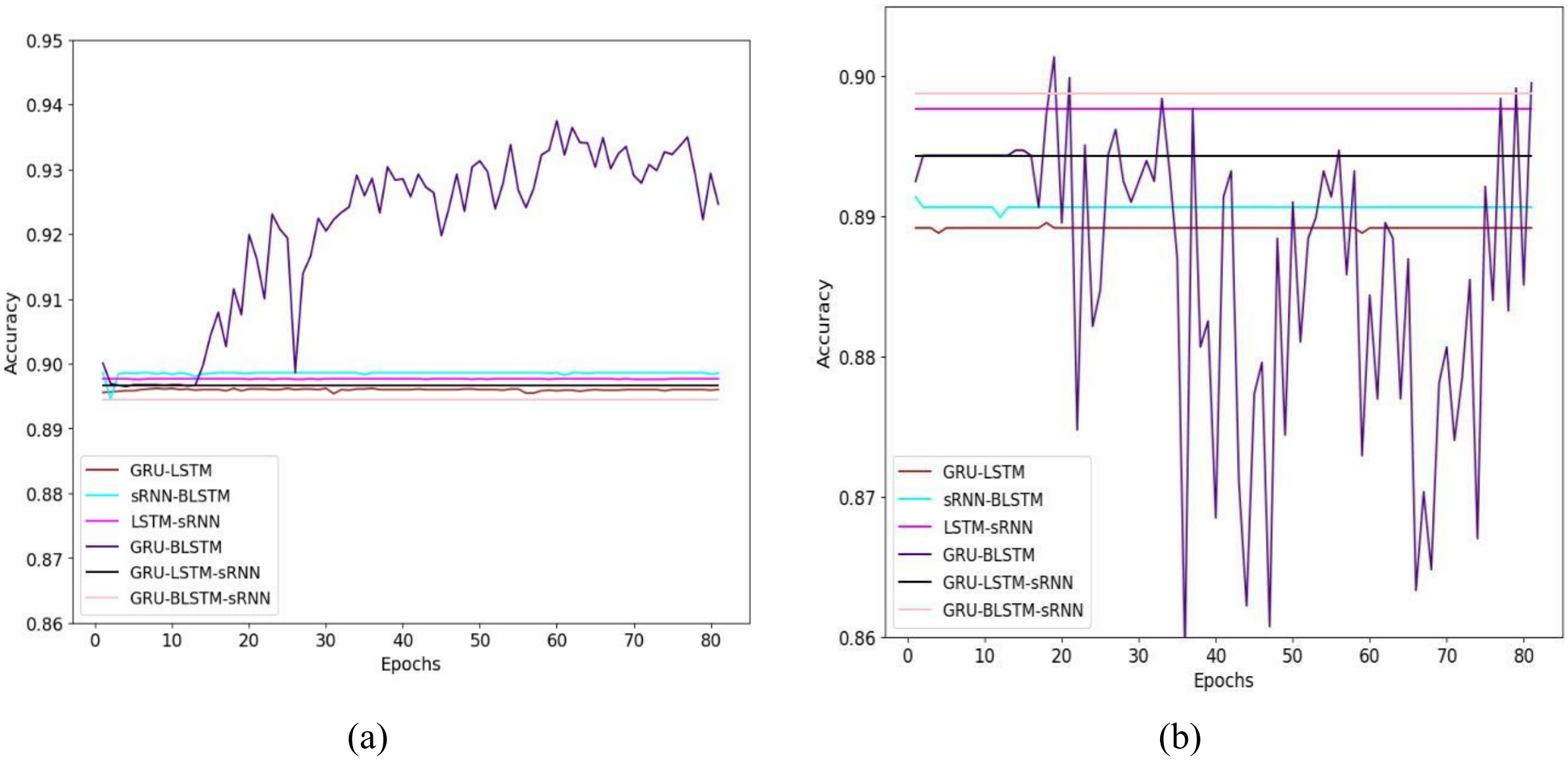

Figure 45. Joint RNN-based Model performance in terms of training (a) and validation (b) accuracy.

Table 16. Joint RNN-based Deep learning model performance

| Modals | Precision (%) | Recall (%) | F1-Score (%) | Accuracy (%) |
|---|---|---|---|---|
| GRU- LSTM | 81 | 90 | 86 | **90.0** |
| GRU- BLSTM | **88** | 88 | 88 | 88.1 |
| sRNN-BLSTM | 81 | 90 | 85 | 89.8 |
| sRNN-LSTM | **88** | 89 | **89** | 89.0 |
| GRU-BLSTM- sRNN | 81 | 90 | 85 | **90.0** |
| GRU-LSTM-sRNN | 80 | 90 | 85 | 89.6 |

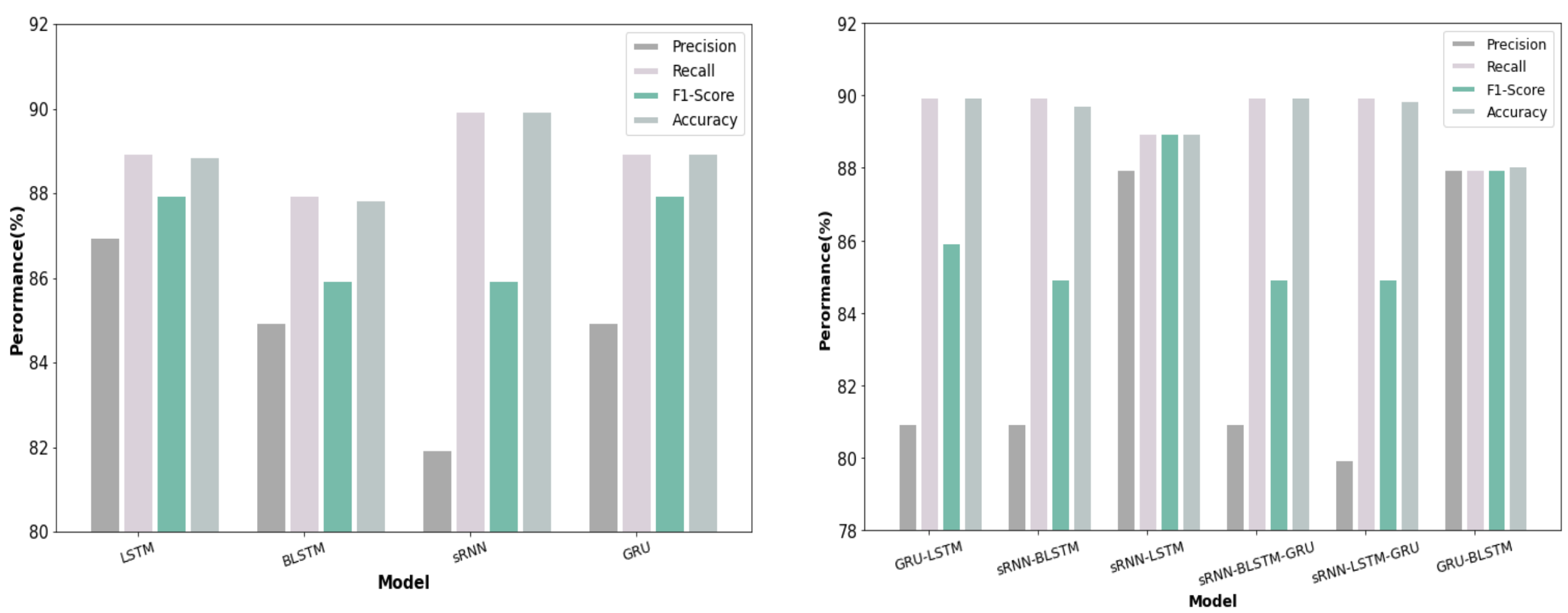

Figure 46. Test Performance Visualization of all models

### 7.5.3 Single RNN-based versus Joint RNN-based deep learning models

In this section we present the analysis of the two classes of models studied in this work. Table 17 presents the performance for all the models evaluated in this study. Although the individual performance for each of the studied models is satisfactorily competitive, generally, the joint RNN-based models outperformed the single RNN-based models with, considering general performance across metrics, sRNNLSTM showing the overall best performance.

Table 17. Model performance

| Methods | Precision (%) | Recall (%) | F1-Score (%) | Accuracy (%) |
|---|---|---|---|---|
| LSTM | 87 | 89 | 88 | 88.9 |
| BLSTM | 85 | 88 | 86 | 87.9 |
| sRNN | 82 | **90** | 86 | **90.0** |
| GRU | 85 | 89 | 88 | 89.0 |
| GRU- LSTM | 81 | **90** | 86 | **90.0** |
| GRU- BLSTM | **88** | 88 | 88 | 88.1 |
| sRNN-BLSTM | 81 | **90** | 85 | 89.8 |
| sRNN-LSTM | **88** | 89 | **89** | 89.0 |
| GRU-BLSTM- sRNN | 81 | **90** | 85 | **90.0** |
| GRU-LSTM-sRNN | 80 | **90** | 85 | 89.9 |

## 7.6 Conclusion

Safety is one of the core components in the aviation industry and therefore requires dire monitoring accompanied with critical observation of the standard operating procedure, safety rules, regulation, and recommendations from the field experts. Often the reports from investigations about aviation anomalies that threaten the safety of stakeholders at all levels, are given with a textural narrative describing the series of events. In this paper we leveraged various NLP techniques and various deep learning models to classify the damage level of the aviation accident to the aircraft based on textual accident/incident narratives from the NTSB investigation reports. With considerably competitive results from our experiments, we answered the question of whether the damage level can be inferred from the raw text narratives.

# Chapter 8. Natural Language Processing and Deep Learning Models to Classify Phase of Flight in Aviation Safety Occurrences

Nanyonga, A., Wasswa, H., Molloy, O., Turhan, U., & Wild, G. (2023, September). Natural language processing and deep learning models to classify phase of flight in aviation safety occurrences. In *2023 IEEE Region 10 Symposium (TENSYMP)* (pp. 1-6). IEEE.

## 8.1 Abstract

The air transport system recognizes the criticality of safety, as even minor anomalies can have severe consequences. Reporting accidents and incidents play a vital role in identifying their causes and proposing safety recommendations. However, the narratives describing pre-accident events are presented in unstructured text that is not easily understood by computer systems. Classifying and categorizing safety occurrences based on these narratives can support informed decision-making by aviation industry stakeholders. In this study, researchers applied natural language processing (NLP) and artificial intelligence (AI) models to process text narratives to classify the flight phases of safety occurrences. The classification performance of two deep learning models, ResNet and sRNN was evaluated, using an initial dataset of 27,000 safety occurrence reports from the NTSB. The results demonstrated good performance, with both models achieving an accuracy exceeding 68%, well above the random guess rate of 14% for a seven-class classification problem. The models also exhibited high precision, recall, and F1 scores. The sRNN model greatly outperformed the simplified ResNet model architecture used in this study. These findings indicate that NLP and deep learning models can infer the flight phase from raw text narratives, enabling effective analysis of safety occurrences.

## 8.2 Introduction

Air transportation is a highly safety-critical industry that is susceptible to operational errors, making it one of the most sensitive fields in the transport sector. Even the smallest mistake or minor system misconfiguration can lead to catastrophic disasters, resulting in various consequences such as loss of lives, significant financial losses, erosion of customer trust, and property damage, among other impacts [22]. . In the event of an aviation safety incident, regardless of its severity, reporting, and investigations are essential.

The primary objective of these safety reports is not to assign blame but rather to ensure that similar incidents are prevented from occurring again. Aviation safety reports, which encompass both formal investigations and voluntary self-reports, consist of unstructured text narratives presented in a language understandable to humans [24, 29, 87, 190]. These narratives highlight the sequence of events that are potentially contributing or causal factors of the incident or accident. Moreover, these reports are made publicly available in some jurisdictions. The publication of these reports serves multiple purposes, including facilitating access for various stakeholders within the aviation industry. This includes maintenance teams, operational managers, and researchers in the field of aviation safety, who can utilize the reports to access, analyze, critique, and implement or adhere to the recommended safety standard operating procedures, rules, and regulations.

However, when similar incidents occur, investigation teams often need to refer to previous reports, which can be a time-consuming process, especially if the conditions of the current incident are related to multiple past incidents. The challenge arises because, although most reports provide a tabular summary of the conditions surrounding the event to enhance readability for humans, the detailed sequence of events that likely caused the incident or accident is typically described in narrative form using natural language understood by humans. Furthermore, when humans report such narratives, they often use non-standard terms that are comprehensible to other humans but not to computers. Consequently, if there is a need to revisit these reports, humans are required to search through a large database of reports and read them one by one in a sequential manner. This manual approach significantly delays the investigation process. To overcome this challenge, various models have been proposed to expedite the processing of aviation data and assist aviation authorities and stakeholders in making safety-critical decisions. Typical tools of decision making, and risk assessment (operations research) can be utilized for modelling and forecasting in aviation safety, such as multi-attribute utility modelling [317]. More fundamental statistical or probability methods can be utilized as well, such as Bayesian networks [32]. The resultant models enable faster data analysis and decision-making, providing valuable insights to support the industry.

Furthermore, advanced machine learning algorithms such as Artificial Neural Networks (ANN) [29], Random Forest (RF) [306], and Support Vector Machine (SVM)[91, 109] , among others, have been proposed as state-of-the-art models for predicting similar outcomes. However, most of these studies have primarily focused on structured data, neglecting the unstructured text narrative. Some studies have used models like Latent Dirichlet Allocation (LDA) [175] in natural language processing applications for aviation incident reports, primarily for topic modeling task [171, 172]. Although topic modeling can classify safety data reports into different topics, its effectiveness diminishes as the overlap and ambiguity between topics increase, as many words can be assigned to multiple topics, rendering it highly unreliable. Despite the increasing popularity and adoption of deep learning models in various domains, including health and NLP for spam email detection [307], the aviation industry has not fully embraced deep learning models for big data analysis.

In this study, an approach that utilizes natural language processing (NLP) and artificial intelligence (AI) models is presented to process text narratives and classify the flight phases of safety occurrences. While previous studies have explored NLP techniques for text analysis in aviation safety, this research extends the use of deep learning models, namely ResNet and sRNN, to aviation safety for classifying the Flight phase at which the incident occurred. This application of deep learning algorithms to infer flight phases from raw text narratives in safety occurrences represents a unique contribution to the field.

The motivation behind this study lies in the potential benefits it offers to the aviation industry. By accurately classifying safety occurrences into specific flight phases, such as take-off, landing, approach, enroute, maneuvering, and more, we can identify stressors or patterns that may lead to component failures. This proactive identification allows airlines and maintenance crews to address issues and reduce unplanned maintenance events. Furthermore, our approach enables aviation safety professionals to identify potential risks and hazards associated with different phases of flight, particularly during the safetycritical phases at the beginning and end of an operation (take-off through initial climb, and final approach to landing).

Therefore, the main objective is to compare the performance of ResNet and sRNN algorithms in classifying the flight phase during which the safety occurrence event took place. The classification can utilize supervised learning for the flight phase categories, as this is provided in the NTSB aviation accident database. Specifically, aviation safety occurrences (accidents and incidents) are classified with the relevant phase of flight, such as Landing, Approach, Take-off, Enroute, Maneuvering, etc. This enables airlines and maintenance crews to proactively address issues and reduce unplanned maintenance events.

The rest of this work is organized as follows. Section III gives an account of the existing literature regarding aviation safety and machine learning. Section IV gives details of the proposed approach and the implementation procedure followed to realize it. In section V the results of the study are presented and give a detailed discussion and their implication for aviation safety research. Section VI presents the conclusion highlighting the direction of future work.

## 8.3 Related Work

Machine learning has been widely applied in various fields, including aviation, medicine, bioinformatics, and biology [309]. In the context of aviation safety reports, machine-learning techniques have been utilized for data mining and text classification. Recent studies have explored different approaches for text mining and classification in aviation safety reports.

Another study examined the application of various recurrent neural network (RNN) architectures for sentence modelling, showing that these models are suitable for sequential datasets such as text mining [310-312] . Another study [312] used RNN techniques to predict weather-related tasks and regulate preflight information. Similarly, a study by Paul [89] focused on different natural language processing (NLP) techniques applied to civil aviation corpus, suggesting the use of NLP, particularly RNN, for time series data mining.

In a study analyzing aviation safety reports, Chanen [112] proposed a deep learning approach using word2vec models to identify similar terms and semantic relationships within the reports. The model aimed to improve the understanding of reporting experts and reduce uncertainty in safety reports. However, the study only focused on one NLP tool, leaving room for improvement using different NLP approaches.

Another related study employed sequential deep learning [159], specifically LSTM and word embedding, for aviation safety prognosis using NTSB reports. Their classification models showed promise in reviewing and analyzing safety investigation reports. Similarly, ElSaid et al. [314] used LSTM models to predict aircraft engine vibrations, demonstrating that LSTM recurrent neural networks can accurately predict vibrations in flight datasets.

In the evaluation of ML techniques for determining critical causes of accidents in air transport [138], it was found that the decision tree algorithm performed better in predicting the cause of accidents. Their study highlighted the human factor, particularly loss of attention, as a significant cause of accidents in aviation.

A study from CLLE-ERSS research lab and CFH safety company [24] , explored NLP tools and textmining methods for analyzing unstructured data in aviation reports. The study emphasized the importance of NLP techniques in accident report analysis and compared various datasets from different databases. However, specific NLP tools used were not reported, indicating potential for better performance with different tools.

In 2022, Ya et al. conducted a study on the application of deep learning, particularly ResNet, for radio signal recognition in real-world scenarios [25]. Their comprehensive research showcased the efficacy of deep learning models in accurately identifying and categorizing radio signals. Notably, ResNet demonstrated exceptional performance in managing extensive datasets and achieving substantial enhancements in recognition accuracy. These findings contribute to the progress of deep learning techniques for radio signal recognition, with potential applications across different fields, including the aviation industry. The study emphasizes the significance of this technology in advancing signal recognition capabilities.

In the context of F-16 fighter jet data modeling, Mersha et al. [308] tested different RNN architectures and found that sRNN using the trained estimated algorithm achieved the best performance in predicting certain parameters. Similarly, in analyzing railroad accident narratives, Song et al.[309] used BLSTM, CNN, and RNN models to improve text classification accuracy, although their performance was below 80%.

A proposed approach [123] constructed a network based on Bayesian probability distribution to infer causal relationships between aviation accidents. However, the study used an outdated dataset that does not capture current trends in aviation data. Another study [316] proposed a data management scheme using ontologies and conceptual models to enhance aviation safety data management, but it lacked advanced data analytics and machine learning techniques.

Lastly, [100] utilized principal component analysis (PCA) and deep brief networks for flight incident prediction. Although the deep brief network architecture showed benefits in learning complex features, PCA alone may lead to reduced classification performance, potentially compromising aviation safety.

Overall, these studies demonstrate the application of machine learning and NLP techniques in analyzing aviation safety reports and improving safety measures. However, there is still room for further exploration and improvement in utilizing advanced data analytics and adapting to the dynamic nature of the aviation industry.

## 8.4 Methodology

To implement the proposed approach, the research involved several distinct steps, including the identification of the training dataset, text processing, and classification, as depicted in **Figure 47**. A brief description of the utilized datasets is provided in the following subsection.

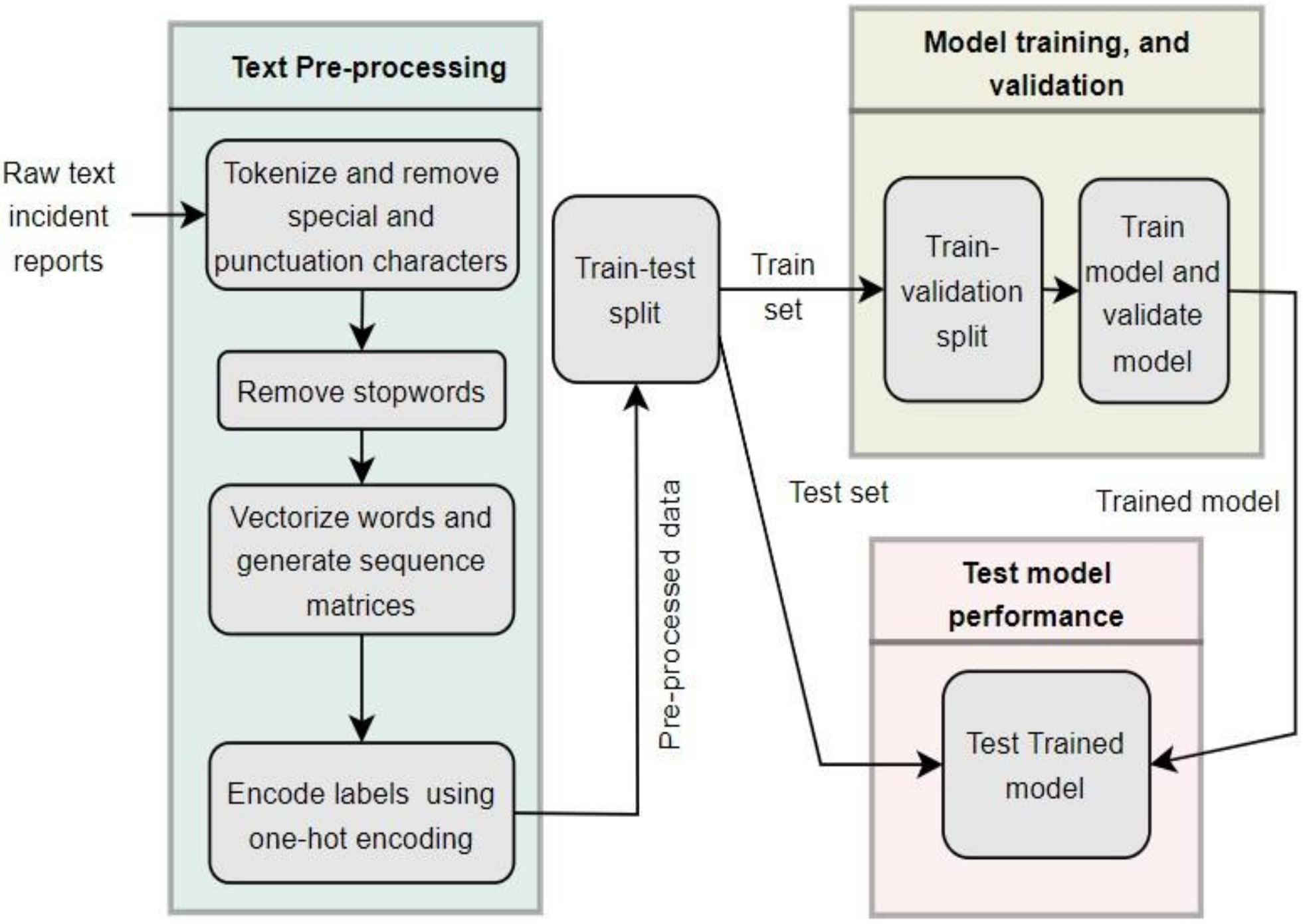

Figure 47. Methodological framework

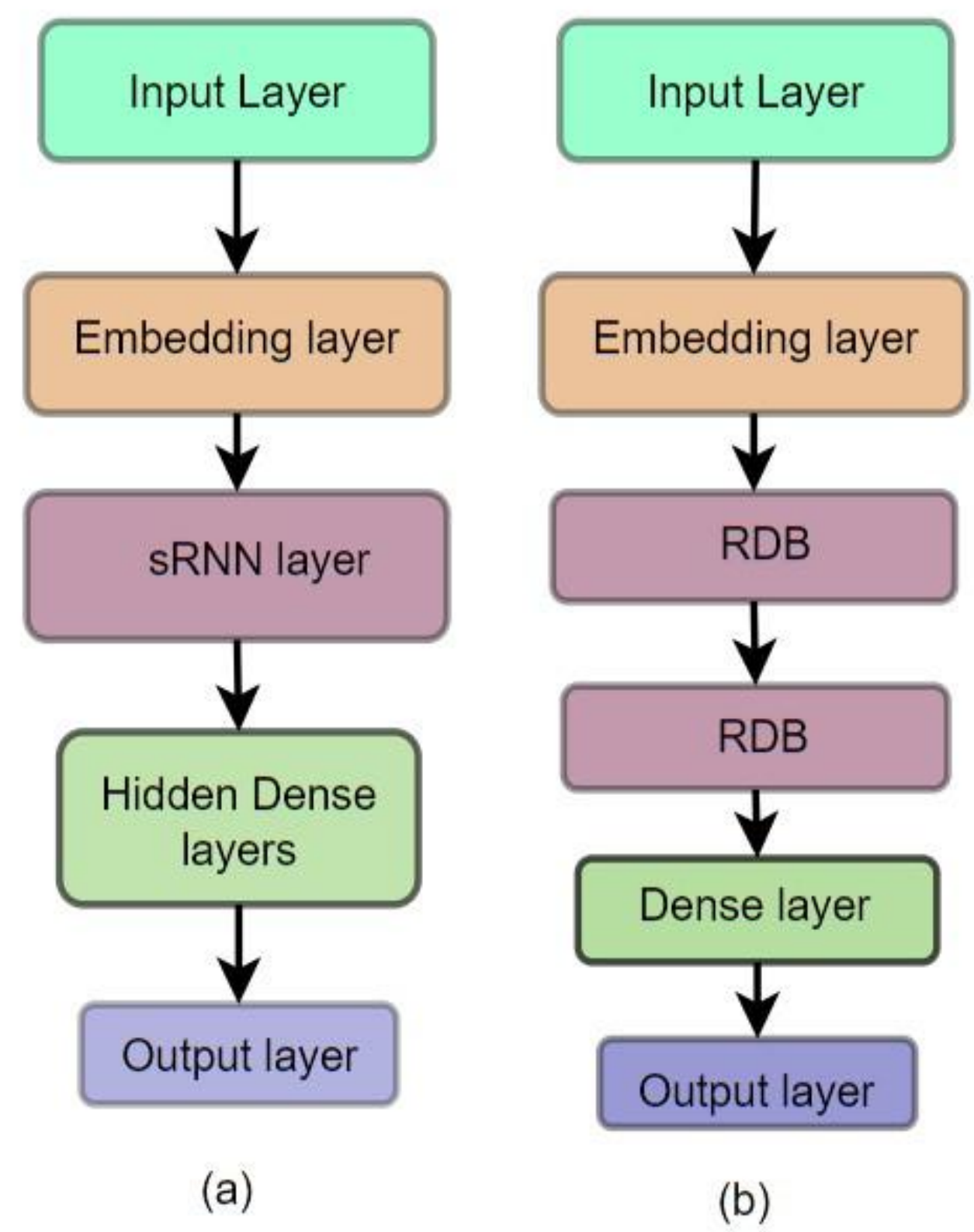

Figure 48. Deep learning architectures: (a) simple RNN and (b) ResNet with two residual dense blocks

*8.4.1 Data Acquisition*

Aviation incident/accident investigation reports are collected and published by various organizations such as ATSB (Australian Transport Safety Bureau), ASRS, and NTSB. For this study, the researchers utilized the NTSB aviation incident/accident investigation reports[2] . Depending on the specific problem being addressed, this dataset can be obtained from the NTSB website, which offers different forms of metadata including individual monthly published PDF reports, and JSON files, or by querying individual reports online or through a CSV summary. In this research, researchers downloaded a JSON file containing incident/accident investigation details spanning from 2005 to 2020. Moreover, the researcher focused on incidents where investigations were completed, resulting in a dataset comprising 16,919 records following data preprocessing and cleaning. From each report, the 'analysis Narrative' and 'flight phase' fields were extracted for training and validating our deep learning models.

*8.4.2 Text Processing*

Machine learning models are not inherently capable of understanding text data in its raw form. Therefore, the text data needs to be converted into a numeric format that can be interpreted by the models. In this study, the Karas deep learning library was employed due to its extensive collection of deep learning models and different types of model layers. This library also offers advanced modules for text preprocessing. The Tokenizer module, for instance, generates tokens and sequence vectors for the input text.

---

The categorical module is used to map unique categorical entries of the damage level variable (e.g., destroyed, substantial, minor, and none) to corresponding numerical values or as one-hot encoded entries for each data instance.

To handle unwanted special and punctuation characters, remove stop-words, and perform word lemmatization, the spacy library was utilized. Spacy is a Python library specifically designed for textprocessing tasks, including named entity recognition and word tagging. It incorporates a comprehensive list of special characters, punctuation, and stop-words, and is regularly updated when necessary.

With the a fore mentioned tools, each input text narrative was processed and transformed into a representative sequence or vector with a length of 2000. Numeric sequences derived from text narratives with fewer than 2000 words were padded with zeros, while those exceeding 2000 words were truncated. The vocabulary size of the corpus was set to 100,000.

To split the dataset into training, validation, and testing sets, the train-test-split module from scikitlearn was employed. All experiments in this study were conducted using Python as the programming language, with Jupyter Notebook serving as the code editor.

*8.4.3 Text Classification*

The data set is pre-processed and divided randomly into an 80% training set and a 20% test set. Furthermore, during the training process, 10% of the training set is set aside for model validation in each epoch. Two deep learning models, ResNet and sRNN, are trained using this data, and their performance is assessed. The Adam optimizer is employed for model optimization. However, it is worth noting that this study did not focus on identifying the best optimizer, so alternative optimization techniques can be utilized if desired.

*8.4.4 Deep learning model Architecture*

To ensure consistency, a shared architecture was utilized for all models, with minor variations in the case of combined models. This common architecture consisted of an embedding layer, hidden layers, and an output layer. ReLU (Rectified Linear Unit) activation function was employed for all hidden layers, while the SoftMax activation function was used for the output layer. The predicted class was generated using the argmax function, which identifies the index associated with the highest probability in the SoftMax output.

Figure 39 provides an overview of the deep learning architectures employed in this research.

8.4.4.1 ResNet Architecture:

The ResNet architecture consists of several layers organized into blocks. Each block contains multiple residual units, which are the basic building blocks of ResNet. In a ResNet, the input passes through several blocks, and each block contains a set of residual units. The output of each block is fed into the next block until the final output is obtained. The basic residual unit comprises two or three convolutional layers, followed by element-wise addition with the input. The key concept behind ResNet is residual learning, which involves learning the residual mapping instead of learning the entire mapping from input to output directly. This is achieved by introducing skip connections or shortcut connections that allow the network to learn the residual mapping more effectively [318].

In this work, a simplified ResNet model was built constituting only two residual dense blocks (RDB) as shown in **Figure 48** above, and with each RDB having only 3 skip connections as shown in **Figure 49**. The skip connection enables the gradient to flow directly from the later layers to the earlier layers during backpropagation. This alleviates the vanishing gradient problem and facilitates the training of deeper networks. Additionally, it helps to preserve the learned information from earlier layers, making it easier for the network to capture fine-grained details and avoid degradation in performance as the network depth increases.

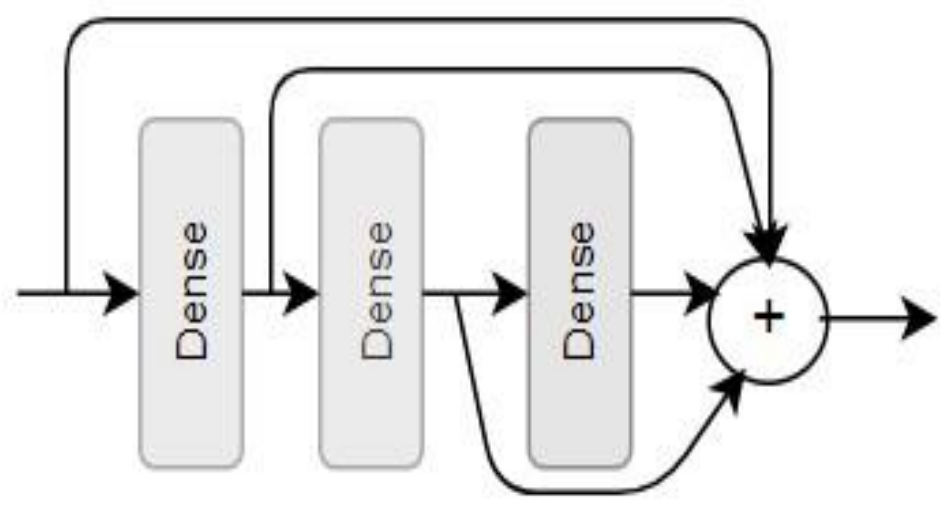

Figure 49. Residual Dense Block

The formulas for the forward pass of each residual unit within the residual dense blocks can be summarized as follows:

Given an input x, the output y of a residual unit can be computed as:

$y = f(x, W) + x$, where $f(x, W)$ represents the transformation performed by the convolutional layers in the residual unit

with learnable weights W.

The "+" operator denotes the element-wise addition of the transformed input ($f(x, W)$) and the original input (x). This additionally combines the residual or the difference between the input and the desired output with the original input [319].

**8.4.4.2 Simple Recurrent Neural Networks (RNNs).**

Simple RNNs, introduced by Elman in 1990, are a fundamental type of RNN architecture. They consist of recurrent connections that allow information to flow in a loop, creating a feedback mechanism. This loop enables Simple RNNs to retain and utilize information from previous time steps, making them suitable for sequential data analysis.

Embedding layers are essential components in many sequence analysis tasks, such as natural language processing. They convert discrete inputs, such as words, into continuous, dense vector representations. Embeddings capture semantic relationships between inputs, enabling the network to better understand the underlying patterns in the data.

A simple Recurrent Neural Network (RNN) is an architecture that processes sequential data by maintaining a hidden state that captures information from previous inputs. It is a type of neural network designed to handle sequential or time-series data, where the order of the data points matters.

Hidden layers are the primary components responsible for learning and extracting meaningful representations from the sequential input. They consist of neurons that process the input at each time step and pass information to subsequent time steps. The depth and width of hidden layers determine the model's capacity to capture complex patterns and relationships in the sequential data.

The concept of a simple RNN can be explained using the following formulas:

Hidden state calculation: $h_t = activation(W_{hh} * h_{t-1} + W_{xh} * x_t + b_h)$

In this formula, $h_t$ represents the hidden state at time step t, $x_t$ is the input at time step t, and activation is an activation function like the hyperbolic tangent (tanh) or the rectified linear unit (ReLU). $W_{hh}$ is the weight matrix connecting the previous hidden state to the current hidden state, $W_{xh}$ is the weight matrix connecting the input to the hidden state, and $b_h$ is the bias term for the hidden state. $y_t = activation(W_{hy} * h_t + b_y)$

Here, $y_t$ is the output at time step t, $W_{hy}$ is the weight matrix connecting the hidden state to the output, and $b_y$ is the bias term for the output.

During training, the RNN processes the input sequence one-time step at a time, updating the hidden state based on the previous hidden state and the current input. The output is then computed using the updated hidden state. To train an RNN, a loss function is defined, such as mean squared error (MSE) or cross-entropy, comparing the predicted output to the target output. The gradients are then calculated through backpropagation through time (BPTT), and optimization algorithms like gradient descent or Adam are used to update the weights and biases of the network [320].

## 8.5 Results and Discussion

In this section, the performance of the two models is presented and compared with each other in terms of accuracy, precision, recall, and F1 measure. **Figure 50** illustrates a visual comparison of the validation accuracy achieved using the two models. It is clear from **Figure 50** that sRNN captures more patterns for discrimination between the various flight phases. Also, model performance was evaluated on the unseen test set samples and the results in Table 18 reveal that sRNN outperformed ResNet in terms of all performance evaluation metrics.

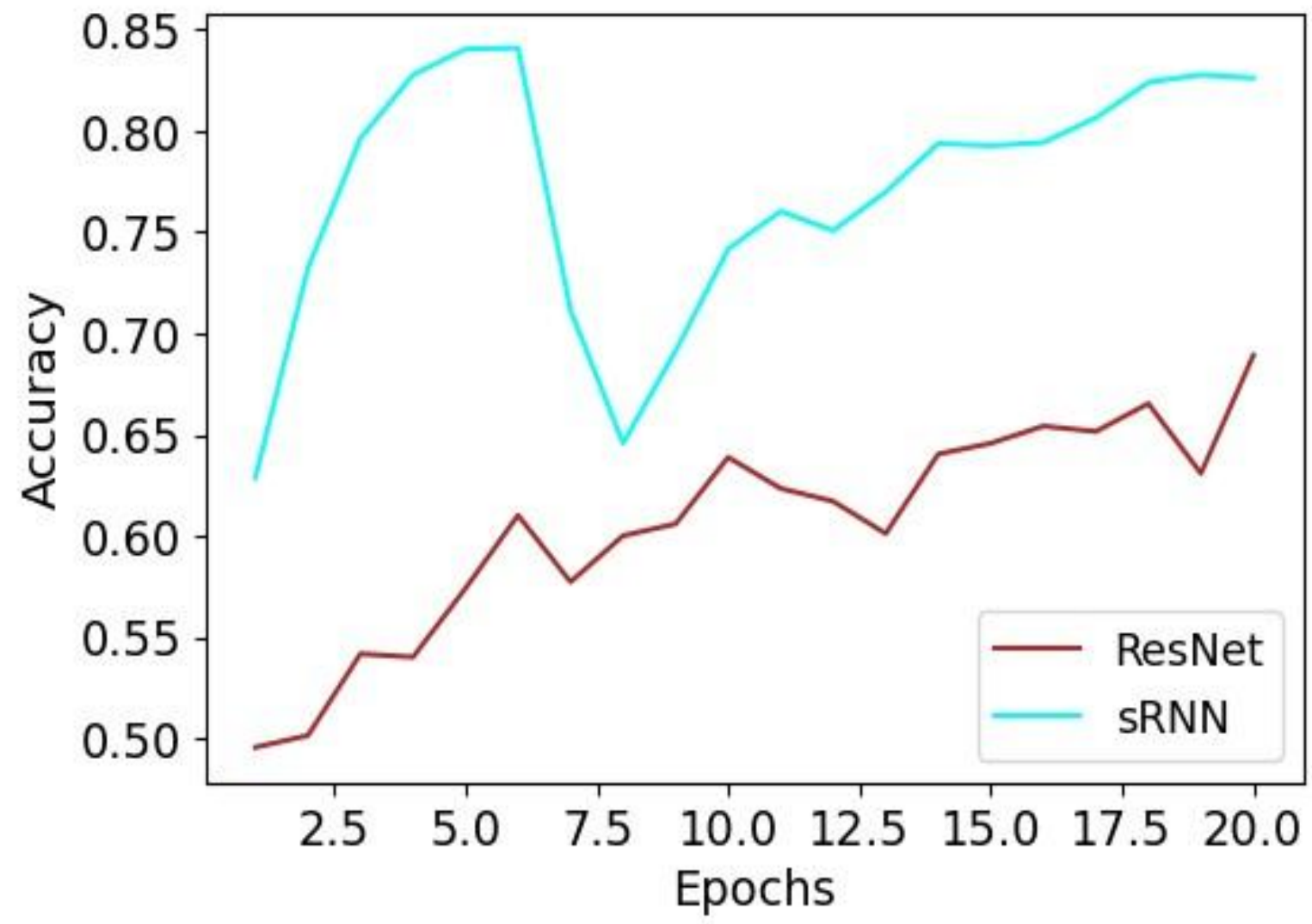

Figure 50. Validation accuracy performance

Table 18. DEEP LEARNING MODEL PERFORMANCE

| **Models** | Precision (%) | Recall (%) | F1 (%) | Accuracy (%) |
|---|---|---|---|---|
| **ResNet** | 69 | 69 | 68 | 68.9 |
| **sRNN** | 84 | 83 | 83 | 83.2 |

The findings of this study demonstrate that by employing natural language processing techniques in conjunction with advanced deep learning models, it is possible to predict the phase of flight at which an incident occurred based on the unstructured text narrative describing the events leading up to the accident.

## 8.6 Discussion

Our previous work focused on applying natural language processing (NLP) and deep learning models, such as LSTM, BLSTM, GRU, and sRNN, to classify and categorize safety occurrences in aviation based on text narratives. The objective was to determine the extent to which the damage level to an aircraft in a safety occurrence can be inferred from the text narrative using NLP techniques. The study evaluated the performance of various deep learning models individually and in combination, including sRNN. The models achieved competitive results, with an accuracy of over 87.9% and high precision, recall, and F1 scores. In the current work, NLP is also applied to aviation incident/accident reports for text analysis, and advanced deep learning algorithms including ResNet and SimpleRNN are used for classification. The research question revolves around predicting the resultant flight phase based on the text narrative describing the series of safety occurrence events. The classification is based on the categories provided in the NTSB investigation reports, classifying the flight phase. The classification of the aircraft damage level and flight phase has significant implications for the decision-making process of the aircraft maintenance team. It helps determine whether the aircraft can be put back into operation, the extent of repairs required, and the associated budget, resource, and financial implications for the aviation company.

Overall, both the previous and current works demonstrate the utilization of NLP and deep learning techniques to analyze safety occurrences in the aviation industry using the NTSB dataset. While the previous work focuses on the classification of damage level in safety occurrences, the current work extends the analysis to aviation incident/accident reports with the aim of predicting the resultant flight phase based on textual narratives [159].

## 8.7 Conclusion

Safety is a fundamental aspect of the aviation industry and requires continuous monitoring and adherence to standard operating procedures, safety rules, regulations, and recommendations. Reports from investigations regarding aviation anomalies that pose a threat to stakeholders' safety often include textual narratives describing the sequence of events. In this study, different NLP techniques and deep learning models were utilized to classify the flight phase during which the incident occurred based on textual narratives of accidents/incidents from NTSB investigation reports. Through our preliminary experiments, the DL models achieved encouraging results, addressing the question of whether the flight phase can be inferred from the raw text narratives.

While deep learning models exhibit strong performance, they are often considered black-box models and further research should focus on enhancing interpretability and explainability of deep learning models used in aviation safety. This will foster stakeholder trust by enabling an understanding of the classification decisions and addressing concerns regarding the black-box nature of these models.

# Chapter 9. Phase of Flight Classification in Aviation Safety using LSTM, GRU, and BiLSTM: A Case Study with ASN Dataset

## 9.1. Abstract

Safety is the main concern in the aviation industry, where even minor operational issues can lead to serious consequences. This study addresses the need for comprehensive aviation accident analysis by leveraging natural language processing (NLP) and advanced AI models to classify the phase of flight from unstructured aviation accident analysis narratives. The research aims to determine whether the phase of flight can be inferred from narratives of post-accident events using NLP techniques. The classification performance of various deep learning models was evaluated. For single RNN-based models, LSTM achieved an accuracy of 63%, precision 60%, and recall 61%. BiLSTM recorded an accuracy of 64%, precision 63%, and a recall of 64%. GRU exhibited balanced performance with an accuracy and recall of 60% and a precision of 63%. Joint RNN-based models further enhanced predictive capabilities. GRU-LSTM, LSTM-BiLSTM, and GRU-BiLSTM demonstrated accuracy rates of 62%, 67%, and 60%, respectively, showcasing the benefits of combining these architectures. To provide a comprehensive overview of model performance, single and combined models were compared in terms of the various metrics. These results underscore the models' capacity to classify the phase of flight from raw text narratives, equipping aviation industry stakeholders with valuable insights for proactive decision-making. Therefore, this research signifies a substantial advancement in the application of NLP and deep learning models to enhance aviation safety.

## 9.2. Introduction

Aviation safety is of paramount importance in the modern era of air travel [22, 321]. Accurate and timely decision-making is vital to ensure the safety and reliability of air transportation systems [322]. One critical aspect of aviation safety is the precise classification of an aircraft's phase of flight, which encompasses taxiing, take-off, cruising, descending, and landing [323]. This classification enables informed and proactive decision-making, ultimately contributing to the prevention of accidents and incidents.

The significance of phase of flight classification lies in its ability to provide a real-time understanding of an aircraft's operational state [96]. For instance, during take-off and landing phases, aircraft are particularly susceptible to various risks, including runway incursions, wind shear, and bird strikes. Timely identification of these phases allows air traffic controllers to implement appropriate safety measures and ensures that pilots receive the relevant information to make informed decisions [147, 170].

The Aviation Safety Network (ASN) has been a cornerstone resource for collecting, documenting, and disseminating aviation safety data from across the globe. The ASN dataset is a comprehensive set of transport category aircraft accidents. This data presents an opportunity to conduct research and analysis to enhance aviation safety measures [81].

In recent years, the field of deep learning, particularly recurrent neural networks (RNNs), has made significant strides in handling sequential data and time-series analysis [312, 324]. Long Short-Term Memory (LSTM) network [324], Gated Recurrent Unit (GRU) networks [325], and Bidirectional LSTM (BiLSTM) networks [326] have emerged as powerful tools for various sequential data classification tasks. The motivation behind this research is two-fold. First, it arises from the pressing need to leverage cuttingedge machine learning techniques to enhance aviation safety protocols. Accurate and automated classification of an aircraft's phase of flight can significantly contribute to minimizing the risk of accidents and improving the overall safety of air travel. Second, the motivation stems from the potential of deep learning techniques to address complex and dynamic challenges in aviation safety [198]. While traditional rule-based methods have been employed for phase-of-flight classification, they may lack the adaptability and scalability needed to handle the increasing volume and complexity of aviation data. Deep learning models, with their ability to capture intricate patterns in data, offer a promising alternative [327].

The primary objective of this study is to assess the efficacy of NLP and Deep Learning techniques in categorizing flight phases within safety occurrence reports sourced from the ASN database. To fulfill this objective, we harnessed advanced deep learning architectures, including LSTM, BiLSTM, and GRU models, and combinations thereof, as suggested by [159]. These models underwent extensive training to deduce flight phase information from the unstructured text narratives present in the safety occurrence reports. To gauge model performance, we employed evaluation metrics encompassing accuracy, precision, recall, and F1-score. Our overarching goal is to showcase that NLP and Deep Learning models possess the capability to reliably infer flight phase information from raw text narratives, thereby establishing a robust foundation for more comprehensive safety occurrence analysis.

The structure of this paper is as follows: Section II provides a review of the existing literature on flight phase classification in aviation safety research, Section III gives an account of the methodology employed in this study, including data preprocessing, model selection, training, and evaluation, Section IV presents the results of our experiments showcasing the performance of the deep learning models in flight phase classification and Finally, Section V concludes the paper by summarizing the key findings and their significance in enhancing aviation safety.

## 9.3.Related Work.

Research in the field of aviation safety and phase of flight classification has seen notable developments, with an increasing focus on leveraging NLP techniques for accurate and automated classification. In this section, we review relevant studies and categorize them based on their contributions to this domain.

Early efforts in the phase of flight classification predominantly relied on rule-based approaches [328, 329]. These methods involved the formulation of specific rules and heuristics based on expert knowledge and domain-specific information. While these approaches were effective to some extent, they often struggled to adapt to evolving aviation environments and handle the complexity of unstructured text narratives from the Aviation Safety Network (ASN) dataset.

In recent years, machine learning techniques have gained prominence for phase-of-flight classification tasks [170, 323, 330]. These studies have demonstrated the potential of supervised learning algorithms, such as Support Vector Machines (SVM) and Random Forests, in automatically categorizing phases of flight based on textual descriptions. While machine learning models exhibited improved performance overrule-based methods, they faced challenges in capturing nuanced patterns and context from text narratives.

With the advent of deep learning and NLP, there has been a shift towards more sophisticated methods for phase-of-flight classification. Recurrent Neural Networks (RNNs) have shown promise in processing sequential textual data [331]. Specifically, Long Short-Term Memory (LSTM) networks and Gated Recurrent Unit (GRU) networks have been applied to extract temporal dependencies and semantics from ASN narratives [81]. Bidirectional LSTM (BiLSTM) architectures, which consider both past and future context, have further improved the accuracy of the phase of flight classification [332]. These models have the ability to capture complex dependencies in text data and learn contextual information that can be crucial for distinguishing between phases of flight.

Some recent studies have explored the combination of textual data with other modalities, such as flight data and audio transcripts, to improve the phase of flight classification [333]. These multimodal approaches aim to harness complementary information sources to enhance the accuracy and robustness of classification models.

In their study, Nanyonga et al [170] investigated the classification of safety occurrences in air transport using NLP and artificial intelligence (AI) models. The study utilized ResNet and sRNN deep learning models to classify flight phases based on unstructured text narratives of safety occurrence reports from the National Transportation Safety Board (NTSB). The research found that both models achieved an accuracy exceeding 68%, surpassing the random guess rate for the seven-class classification problem. Notably, the sRNN model outperformed the simplified ResNet architecture, suggesting the effectiveness of NLP and deep learning models in extracting flight phase information from raw text narratives.

Puranik et al [156] highlighted the application of supervised machine learning techniques in aviation safety, particularly for the prediction of safety-critical landing metrics. The study addressed the challenges of real-time risk identification by developing an offline-online framework using flight data from the approach phase. The framework leveraged Random Forest regression to predict landing true airspeed and ground speed, achieving robust and fast predictions suitable for online applications.

Fala et al [323] addressed the accurate identification of phases of flight in general aviation. The study explored dimensionality reduction algorithms and clustering techniques for phase identification. The research found that combinations of dimensionality reduction methods did not significantly impact phase identification and that both K-means and Gaussian mixture models were suitable for identifying flight phases in general aviation flights.

Our study extends the existing work in several ways. We explore the use of advanced Deep Learning models in flight phase classification, including LSTM, BiLSTM, and GRU, and a combination of these models specifically on ASN datasets that have not been extensively studied in this context.

## 9.4.Methodology

In this section, we outline the methodology employed in this research to classify flight phases within safety occurrence reports from the ASN using NLP and Deep Learning techniques. To implement the proposed approach, our research encompassed several distinct phases, including data acquisition, text processing, and classification, as illustrated in **Figure 51**.

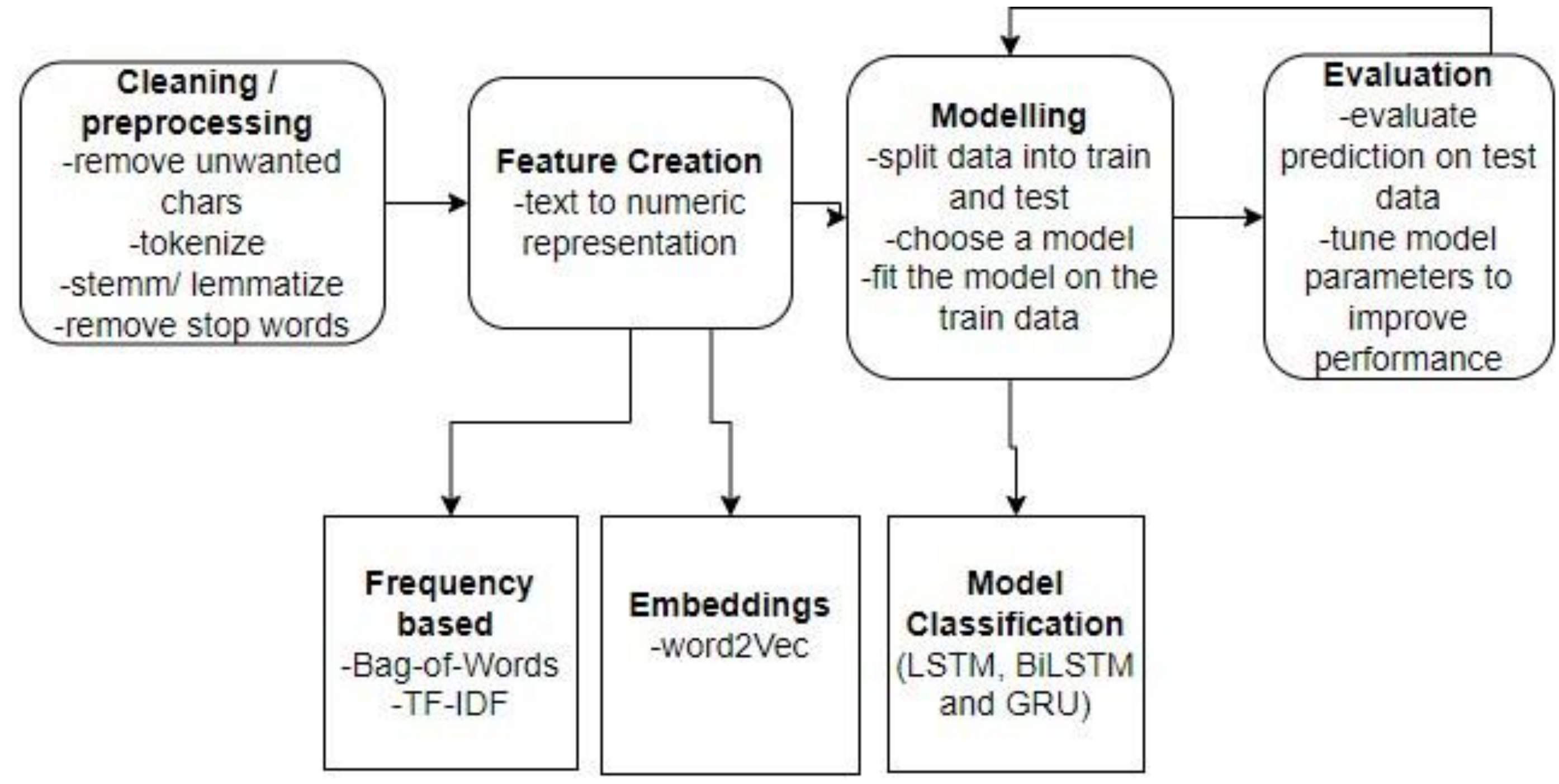

Figure 51. Methodological framework

### *9.4.1 Data Collection*

Aviation incident and accident investigation reports are regularly compiled and disseminated by various organizations, such as the Australian Transport Safety Bureau (ATSB), the National Transportation Safety Board (NTSB), and the Aviation Safety Network (ASN). In this study, we focused on utilizing aviation incident and accident investigation reports from the ASN website spanning from 01/01/2000 to 12/31/2020 which is publicly available from the source: https://aviation-safety.net/. Specifically, our analysis concentrated on incidents where investigations had been completed, resulting in a dataset comprising 4,372 records following data preprocessing and cleaning. From each report, we extracted the 'Narrative' and ' phase of flight' (POF) fields, which were used for training and validating our deep learning models.

### *9.4.2 Text Pre-processing*

Machine learning models are not inherently designed to comprehend raw text data, necessitating the transformation of human-readable text into a numerical format that can be interpreted by the models. To facilitate this transformation, we employed the Keras deep learning library, which offers a comprehensive suite of deep learning models and various model layers. Additionally, Keras provides advanced modules for text preprocessing, including the Tokenizer module, which generates tokens and sequence vectors for input text. To encode categorical data, such as phase (s) of flight labels (e.g., take-off, initial climb, approach, landing, maneuvering, cruise, descent, standing, taxiing, and others), we employed the categorical module in Keras which uses one-hot to map categorical entries to numerical values encoding. Special characters, punctuation, stop-words, and word lemmatization were handled using the spacy library, specifically designed for text processing tasks, including named entity recognition and word tagging. Spacy incorporates a comprehensive list of special characters, punctuation marks, and stop-words and is regularly updated to accommodate evolving requirements. Leveraging these tools, we processed each input text narrative, transforming it into a representative sequence/vector with a fixed length of 2000. For narratives with a word count less than 2000, numeric sequences were padded with zeros, while those exceeding 2000 were truncated. The corpus vocabulary was limited to 100,000 terms. To split the dataset into train, and test subsets, we utilized Scikit-learn's train-test-split module. All experiments were conducted using Python as the programming language and Jupyter Notebook as the code editor.

### *9.4.3 Text Classification*

The processed dataset was randomly partitioned into 80% train and 20% test sets. Additionally, 10% of the train set was held out for model validation during each epoch of training. Various deep learning models, including LSTM, BLSTM, GRU, and combinations of (LSTM+GRU, BLSTM+GRU, LSTM+BLSTM, and LSTM+BiLSTM+ GRU), were trained and evaluated, with their performance compared against other models. The Adam optimizer was employed for model optimization.

### *9.4.4 Deep Learning Model Architecture*

To maintain consistency across models, except for minor variations in combined models, a standardized architecture was adopted. This architecture consisted of an embedding layer, hidden layers, and an output layer. Rectified Linear Unit (ReLU) activation functions were employed for all hidden layers, while the output layer utilized SoftMax as the activation function. The predicted class was generated using the argmax function, which identifies the index corresponding to the entry with the highest probability in the SoftMax output. Figure 52 illustrates the general deep learning architectures employed in this study for a single RNN-based deep learning algorithm (a) and when two distinct RNN-based deep learning algorithms are combined (b).

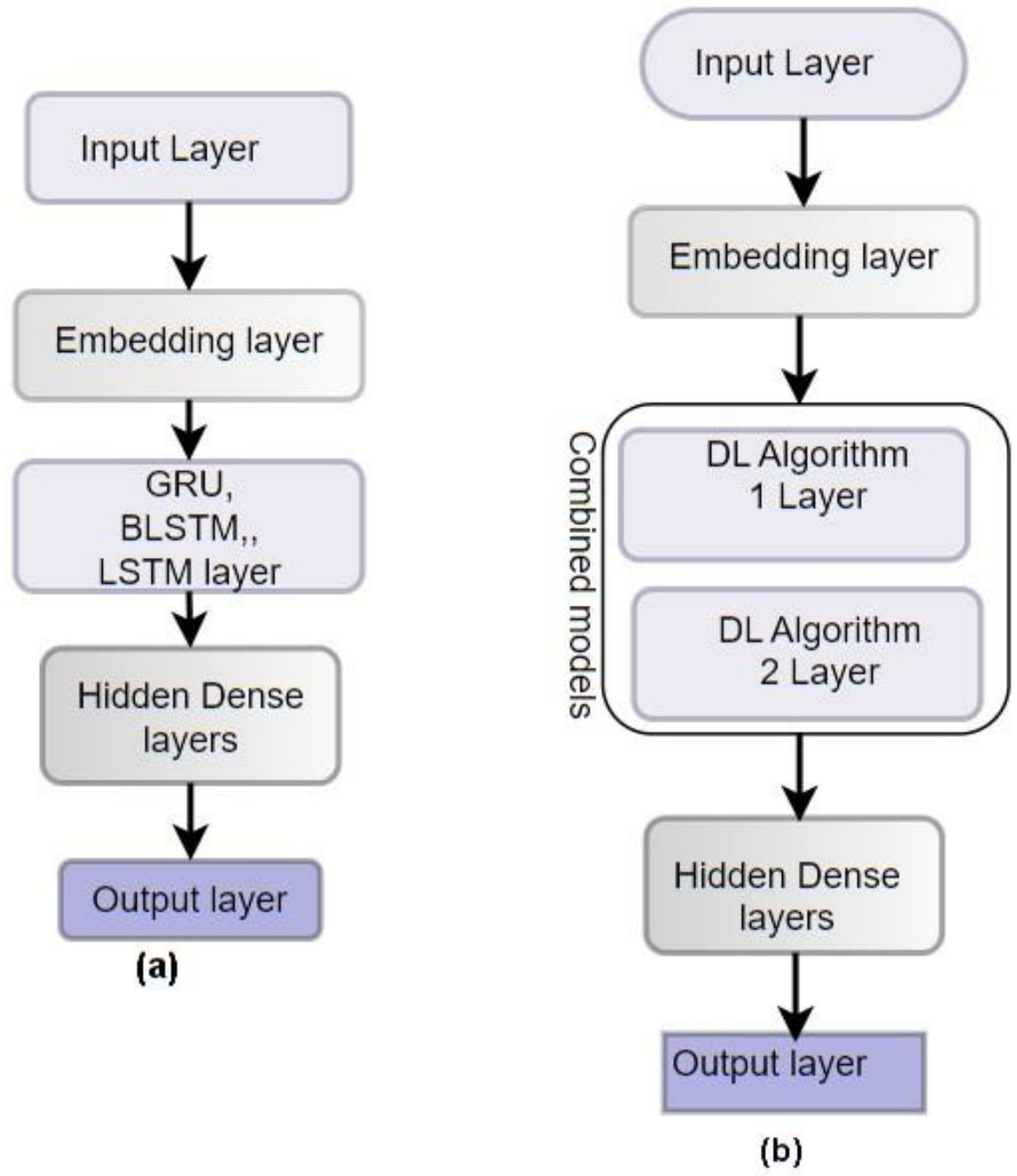

Figure 52. Deep learning architectures: (a) single DL algorithm and (b) using two different DL Algorithm

This consistency of model architectures provided a solid foundation for training and evaluation of the deep learning models, enabling a fair comparison of their performance and the accurate classification of flight phases based on unstructured safety occurrence narratives.

### *9.4.5 Model Performance Evaluation*

This section outlines the assessment criteria adopted for evaluating the models' performance in this study. Our primary focus is on multi-class classification, with performance evaluation centered on the accuracy of predictions across multiple classes. To ensure a comprehensive evaluation, we applied a set of wellestablished prediction performance metrics, including precision, recall, F1-score, and accuracy, as referenced in prior literature. These metrics, elaborated upon in Table 19, offer a thorough evaluation of the models' classification performance, with TP representing true positives, TN signifying true negatives, FN denoting false negatives, and FP indicating false positives.

TABLE 19. PERFORMANCE EVALUATION METRICS

| Metrics | Formula | Evaluation focus |
|---|---|---|
| **Precision (p)** | $\frac{TP}{TP+FP}$ | Correctly predicted positives in a positive class |
| **Recall (r)** | $\frac{TP}{TP+TN}$ | Fraction of positive patterns correctly classified |
| **F1-score (F)** | $\frac{2 * precision * recaal}{precision + recall}$ | Weighted average score of precision and recall |
| **Accuracy (acc)** | $\frac{TP+TN}{TP+FP+TN+FN}$ | Total number of instances predicted correctly |

## 9.5 RESULTS AND DISCUSSION

The experiments conducted in this study have showcased the remarkable capacity of NLP and advanced deep learning models to extract essential information from unstructured textual narratives of pre- and postaccident events, enabling the accurate classification of flight phases within safety occurrence reports. Our evaluation encompassed a diverse array of deep learning architectures, including GRU, LSTM, and BiLSTM, and joint RNN architecture, where the model incorporated multiple RNN layers of different architectures (e.g., GRU with LSTM, as GRU+LSTM, LSTM with BiLSTM as LSTM+BiLSTM and GRU with LSTM and BiLSTM as LSTM+BiLSTM+GRU). These models demonstrated exceptional power in learning intricate patterns from text narratives, which would lead to significant advancement in aviation safety analysis.

### *9.5.1 Performance of Single RNN-Based Deep Learning Models*

This section delves into the individual performance metrics of each model, comprising training accuracy, validation accuracy, testing accuracy, precision, recall, and F1 measure. It is worth noting that, in this context, a "single RNN-based deep learning model" refers to a model containing just one RNNbased layer (as depicted in Figure 53). Conversely, a "joint RNN-based deep learning model" pertains to a model that incorporates multiple RNN-based layers of different architectures (as visualized in Figure 54). Table 20 provides a record of the performance achieved by each of the three single RNN-based deep learning models employed in this study, which are further visualized in Figure 55.

Table 20. Single RNN-based model performance

| **Modals** | **Precision (%)** | **Recall (%)** | **F1-Score (%)** | **Accuracy (%)** |
|---|---|---|---|---|
| LSTM | 0.60 | 0.61 | 0.59 | 0.63 |
| BiLSTM | **0.63** | **0.64** | **0.63** | **0.64** |
| GRU | 0.63 | 0.60 | 0.61 | 0.60 |

### *9.5.2 Performance of Joint RNN-Based Models*

In addition to the single RNN-based deep learning models, we evaluated joint RNN-based models using the same dataset. Table 21 presents the prediction results from four different combinations of the three RNN-based models assessed in this study. The performance results of these joint models are visually presented in Figure 54.

Table 21. Joint RNN model performance

| **Modals** | **Precision (%)** | **Recall (%)** | **F1-Score (%)** | **Accuracy (%)** |
|---|---|---|---|---|
| GRU- LSTM | 0.61 | 0.62 | 0.60 | 0.62 |
| GRU+ BiLSTM+ LSTM | 0.67 | 0.60 | 0.60 | 0.60 |
| LSTM+BiLSTM | **0.66** | **0.67** | **0.66** | **0.67** |

| GRU+BiLSTM | 0.70 | 0.62 | 0.64 | 0.62 |
|---|---|---|---|---|

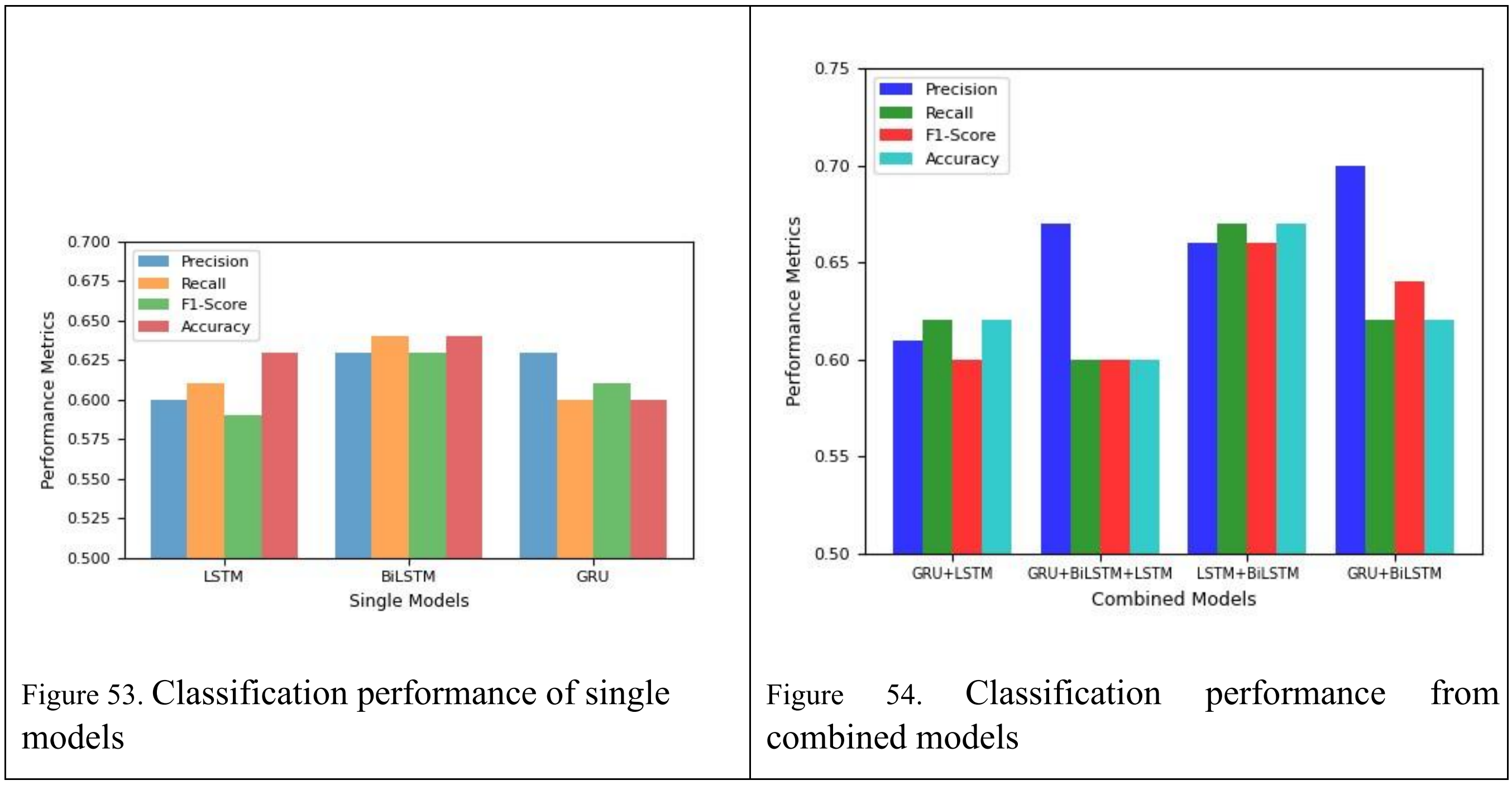

Figure 53. Classification performance of single models

Figure 54. Classification performance from combined models

These results underscore the potency of NLP and Deep Learning models in managing the complexity of aviation safety reports and extracting vital flight phase information. Notably, the LSTM+BiLSTM model exhibited the highest accuracy and precision, signifying their suitability for this task. Visualizations in Figures 55 and 56 offer insights into the training behavior and performance of different deep learning models from both single and combined architectures, in regard to validation accuracy and validation loss, over various training epochs.

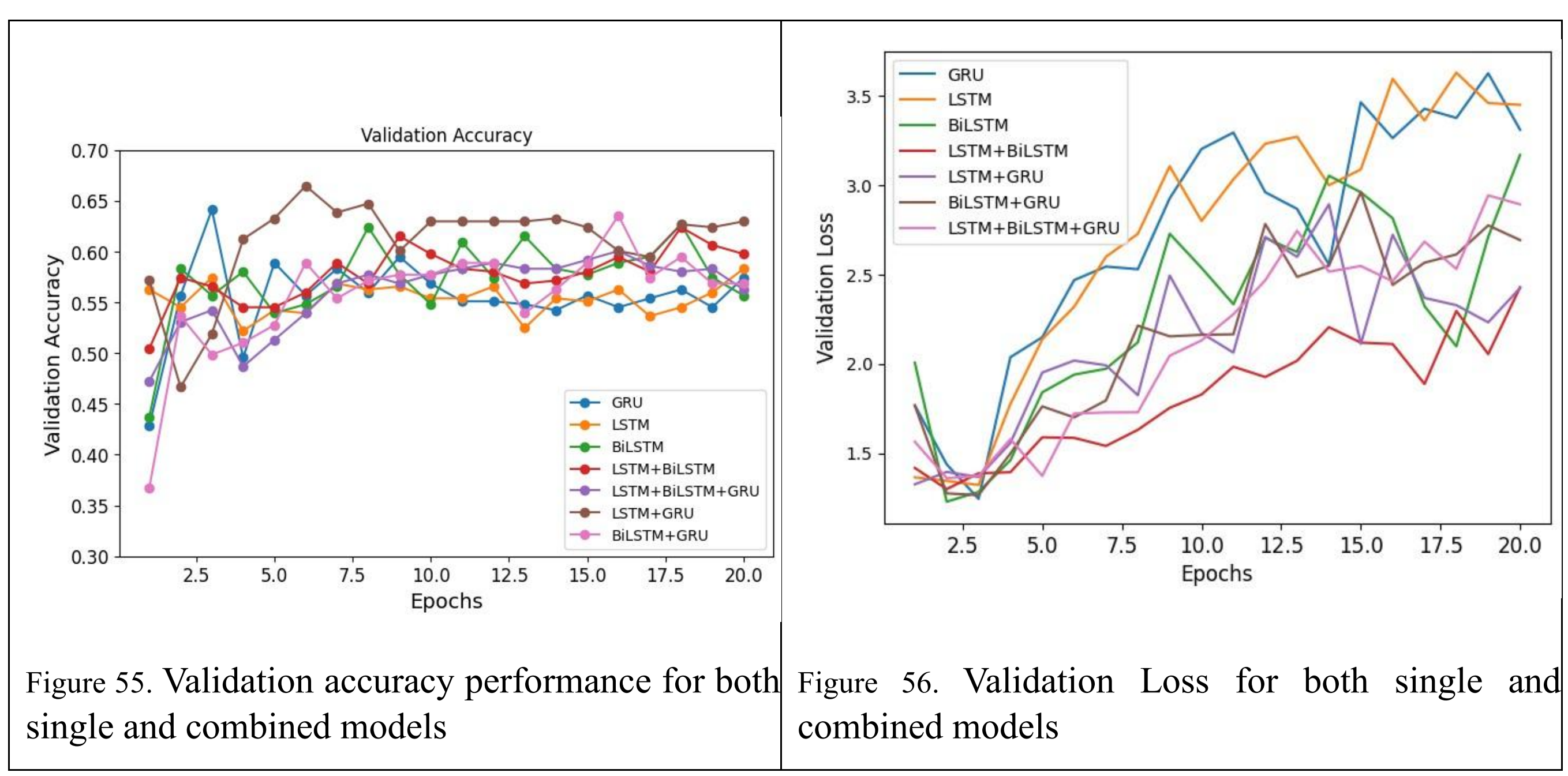

Figure 55. Validation accuracy performance for both single and combined models

Figure 56. Validation Loss for both single and combined models

Classification Report

| | precision | recall | f1-score | support |
|---|---|---|---|---|
| Approach | 0.45 | 0.47 | 0.46 | 89 |
| Enroute | 0.59 | 0.47 | 0.52 | 149 |
| Landing | 0.82 | 0.85 | 0.84 | 338 |
| Standing | 0.65 | 0.64 | 0.65 | 81 |
| Takeoff | 0.53 | 0.70 | 0.60 | 87 |
| Taxi | 0.46 | 0.41 | 0.44 | 41 |
| Unknown | 0.24 | 0.21 | 0.22 | 72 |
| accuracy | | | 0.67 | |
| macro avg | 0.56 | 0.57 | 0.56 | 122 |
| weighted avg | 0.66 | 0.67 | 0.66 | 122 |

Figure 57. Classification report for the best model LSTM+BiLSTM

Figure 57 provides an extract depicting the classification report, encompassing accuracy, precision, recall, and F1-score, for the best-performing model, LSTM+BiLSTM. This extract also elucidates the distribution of test instances among distinct phases of flight (POF) entries, as evident in the support column. Our research showcases the potential of deep learning models in effectively classifying flight phases based on unstructured text narratives. Joint models that amalgamate different RNN architectures show promise in enhancing classification accuracy. The model selection should be tailored to the specific application's requirements, considering the trade-offs between precision and recall.

## 9.6 CONCLUSION

In an era where aviation safety is paramount, the demand for automated and precise safety occurrence analysis is escalating. This study has unequivocally unveiled the remarkable potential of NLP and Deep Learning methodologies in revolutionizing the classification of flight phases within safety occurrence reports. Leveraging a dataset comprising 4,327 reports sourced from ASN, we harnessed the prowess of advanced Deep Learning architectures, including LSTM, Bidirectional LSTM (BiLSTM), and GRU. These models exhibited remarkable efficiency in extracting crucial flight phase information from the intricate realm of unstructured textual narratives, marking a substantial leap forward in aviation safety analysis.

While our models achieved commendable performance in the context of seven-class classification, surpassing the 14% random guess threshold, it is essential to acknowledge the potential for further enhancement. The limited dataset size undoubtedly played a role, and future research endeavours should encompass the utilization of more extensive datasets, including the amalgamation of data from various sources such as the NTSB and ATSB. This approach holds the promise of refining model performance and elucidating the superiority of hybrid models in phase of flight classification. Notably, BiLSTM outperformed all single models, while the combination of LSTM and BiLSTM showcased superior performance among the joint models. This study underscores the transformative potential of NLP and Deep Learning in advancing aviation safety analysis, with ample opportunities for further exploration and refinement in the quest for safer skies.

## Chapter 10. Aviation Safety Enhancement via NLP & Deep Learning: Classifying Flight Phases in ATSB Safety Reports

### *10.1* Abstract

Aviation safety is paramount, demanding precise analysis of safety occurrences during different flight phases. This study employs Natural Language Processing (NLP) and Deep Learning models, including LSTM, CNN, Bidirectional LSTM (BLSTM), and simple Recurrent Neural Networks (sRNN), to classify flight phases in safety reports from the Australian Transport Safety Bureau (ATSB). The models exhibited high accuracy, precision, recall, and F1 scores, with LSTM achieving the highest performance of 87%, 88%, 87%, and 88%, respectively. This performance highlights their effectiveness in automating safety occurrence analysis. The integration of NLP and Deep Learning technologies promises transformative enhancements in aviation safety analysis, enabling targeted safety measures and streamlined report handling.

### *10.2* Introduction

The aviation industry is one of the most regulated and safety-conscious sectors globally, driven by a commitment to ensuring passenger safety and operational excellence. To maintain and improve aviation safety, it is crucial to investigate and analyze safety occurrences systematically [22, 159, 172, 198, 334]. These occurrences range from technical malfunctions to human errors, and they occur throughout different phases of flight, including pre-flight, take-off, climb, cruise, descent, approach, and landing. Understanding when and where these occurrences are more likely to happen is vital for targeted safety measures [189]. Historically, safety occurrence reports have been a valuable resource for aviation safety analysis. They provide detailed narratives of incidents, accidents, and near-miss events, offering valuable context and information about what transpired during these events. However, analyzing these reports manually is a resource-intensive task, often limited by human capacity, subjectivity, and the potential for oversight [63].

The motivation behind this research stems from the need to automate and streamline the analysis of safety occurrence reports, specifically in the context of flight phase classification. Effective classification of safety occurrences into their corresponding flight phases can lead to a more precise understanding of when and where these incidents are likely to occur. This knowledge, in turn, enables the development of targeted safety protocols and preventative measures to reduce the frequency and severity of incidents. Moreover, by automating this process, aviation safety authorities can handle a higher volume of reports efficiently, ensuring that critical information is not overlooked, and safety enhancements are implemented promptly [198]. This research harnesses the potential of NLP and Deep Learning to achieve this goal, promising to revolutionize aviation safety analysis [335].

The primary objective of this study is to investigate the effectiveness of NLP and Deep Learning techniques in classifying flight phases within safety occurrence reports obtained from the ATSB. To achieve this, we employed advanced deep learning architectures, including LSTM, CNN, BLSTM, and sRNN models [101, 159]. These models are trained to infer flight phase information from the unstructured text narratives in the safety occurrence reports. Evaluation metrics including accuracy, precision, recall, and F1-score were used for measuring model performance. The goal is to demonstrate that NLP and Deep Learning models can effectively infer flight phase information from raw text narratives, providing a foundation for more comprehensive safety occurrence analysis.

The structure of this paper is as follows: Section II provides a review of the existing literature, highlighting the significance of flight phase classification in aviation safety research and discussing relevant prior work. Section III gives an account of the methodology employed in this study, including data preprocessing, model selection, training, and evaluation. Section IV presents the results of our experiments, showcasing the performance of the deep learning models in flight phase classification and discussing the interpretation of results, potential limitations, and implications for aviation safety. Finally, Section V concludes the paper by summarizing the key findings and their significance in enhancing aviation safety.

## *10.3* RELATED WORK

The classification of flight phases within safety occurrence reports using NLP and Deep Learning techniques represents a significant advancement in aviation safety analysis [109, 112, 159, 334, 336]. This section explores prior work related to flight phase classification, emphasizing the growing importance of automated methods.

Understanding the significance of flight phase classification within aviation safety analysis is critical. Researchers have long recognized the importance of associating safety occurrences with specific flight phases to target preventive measures effectively [172, 337].

Historically, the classification of safety occurrences into flight phases has been a labor-intensive, manual process. Aviation safety experts and investigators typically reviewed incident narratives and categorized them based on their expertise and experience. This approach, while valuable, is limited by subjectivity and resource constraints [63]. Comparative studies have been conducted to evaluate the performance of various classification methods. Study [328] compared traditional rule-based classification with machine learning approaches, finding that machine learning outperformed rule-based methods in accuracy and efficiency. However, the study did not explore the full potential of Deep Learning techniques.

The International Civil Aviation Organization (ICAO) introduced a taxonomy system for flight phase classification, providing a standardized framework for manual categorization. While this system has been widely adopted, its effectiveness depends on the availability of expert human resources [338]. The emergence of NLP and Deep Learning has revolutionized the way safety occurrences are classified by enabling automated and scalable solutions. Researchers have explored various approaches to automate flight phase classification [160].

In research [339] a study was conducted on aviation safety and emphasized the need for precise classification of incidents by flight phases based on visual scanning strategies using SVM. The study found that incidents occurring during take-off and landing phases tend to have different causation factors than those during cruising, highlighting the importance of context-aware analysis.

Also, [160] employed two deep learning techniques, ResNet and sRNN, to classify flight phases based on textual data extracted from the NTSB dataset. Impressively, their models achieved an accuracy exceeding 68%, significantly surpassing the random guess rate of 14% for a seven-class classification problem. Moreover, these models demonstrated exceptional precision, recall, and F1 scores, with sRNN notably outperforming the simplified ResNet architecture.

Our study extends the existing work in several ways. Firstly, we explore the use of advanced Deep Learning models in flight phase classification, including LSTM, CNN, BLSTM, and sRNN [159], which have not been extensively studied in this context. Secondly, we employ a substantial dataset of 53,275 safety occurrence reports from the ATSB to evaluate the performance of these models rigorously. Thirdly, we focus on comprehensive evaluation metrics, including accuracy, precision, recall, and F1-score, to provide a holistic assessment of the models' performance.

## *10.4* METHODOLOGY

In this section, we outline the methodology employed in this research to classify flight phases within safety occurrence reports from the ATSB using NLP and Deep Learning techniques. Our approach involves data preprocessing, model selection, training, and evaluation as depicted in Figure 58.

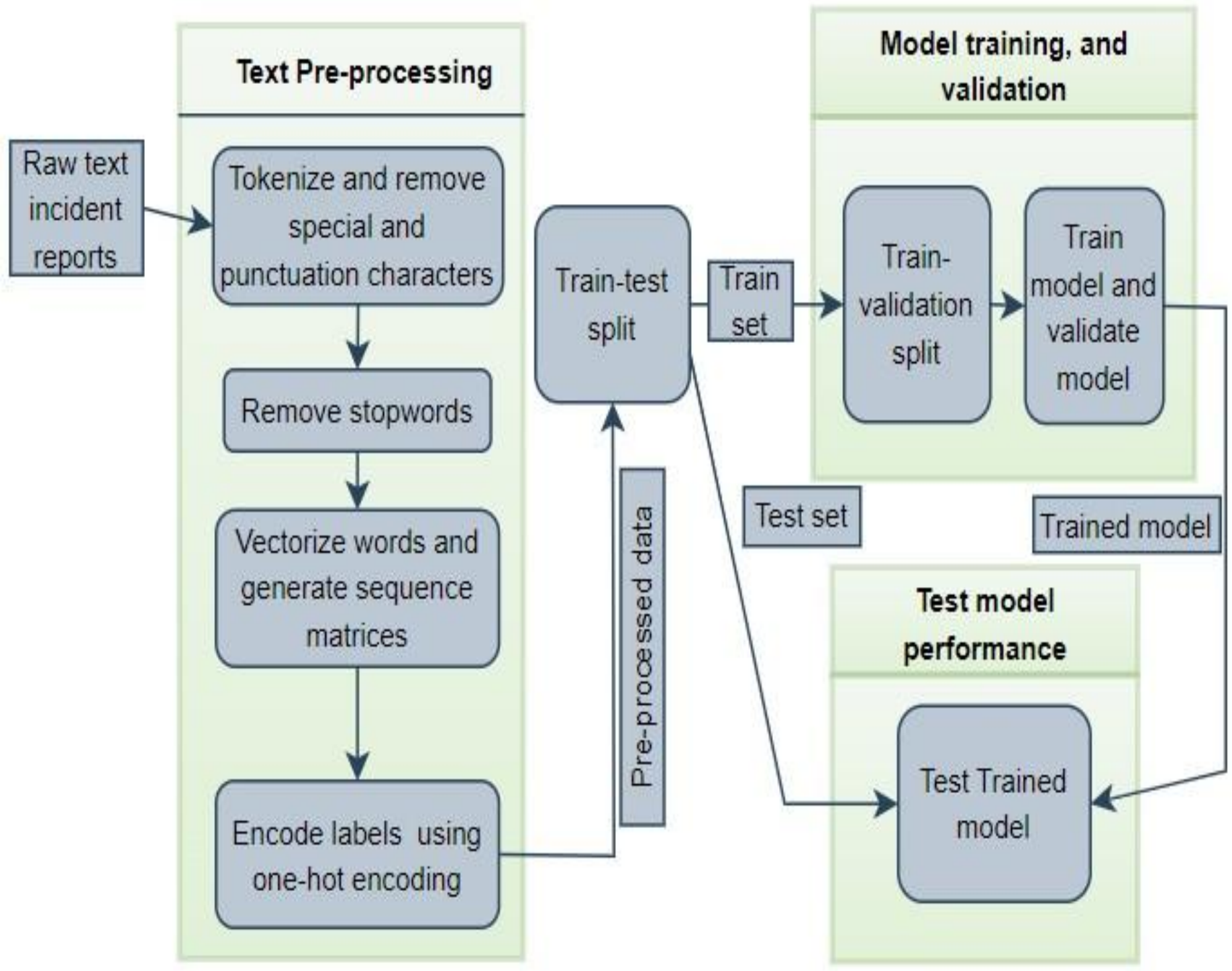

Figure 58. Methodological framework

*10.4.1 Data Acquisition*

Aviation incident/accident investigation reports are collected and published by various organizations such as ATSB, the Aviation Safety Reporting System (ASRS), and the National Transportation Safety Board (NTSB). For this study, the researchers utilized the ATSB aviation incident/accident investigation reports. Depending on the nature of the problem, we considered text narratives and phase(s) of flight that were recorded in Australia for the period of 10 years resulting in a dataset with 53,275 records where the data was sourced directly from the ATSB investigation authorities spanning from 1/01/2013 to 12/31/2022. Moreover, the researchers focused on phase (s) of flight, resulting in a dataset comprising 50,778 records following data preprocessing and cleaning. From each report, the 'Summary' and 'phase (s) of flight' fields were extracted for training and validation of deep learning models.

***10.4.2*** *Text Processing*

Text preprocessing plays a crucial role in preparing unstructured text data for machine learning models. In this study, we leveraged the Keras deep learning library for its extensive collection of deep learning models and model layers. Additionally, Keras provides advanced modules for text preprocessing. Specifically, we utilized the Tokenizer module, which efficiently generates tokens and sequence vectors from input text. To encode categorical data, such as phase (s) of flight labels (e.g., take-off, initial climb, approach, landing, manoeuvring, cruise, descent, standing, taxiing, and others), we employed the to_categorical module in Keras, mapping these categorical entries to numerical values using one-hot encoding for each data instance.

To address challenges related to special characters, punctuation, and stop words, as well as to perform word lemmatization, we harnessed the capabilities of the spacy library. Spacy is a Python library tailored for text-processing tasks, encompassing functionalities like named entity recognition and word tagging. It maintains an extensive list of special characters, punctuation marks, and stop words and undergoes regular updates whenever necessary to remain current.

With the aforementioned tools at our disposal, each input text narrative underwent a comprehensive preprocessing pipeline, ultimately being transformed into a representative sequence or vector with a fixed length of 2000. For narratives with fewer than 2000 words, we padded the numeric sequences with zeros, ensuring uniformity. In contrast, narratives exceeding 2000 words were truncated to meet this standardized length. The vocabulary size of the text corpus was set to 100,000, accommodating a broad range of terms.

To partition the dataset into training (80%), and testing (20%) sets, we utilized the train-test-split module from scikit-learn. All experiments conducted in this study were implemented using the Python programming language, with Jupyter Notebook serving as the chosen code editor. This rigorous text preprocessing framework laid the foundation for subsequent model training and evaluation, enabling the accurate classification of phase (s) of flight based on unstructured safety occurrence narratives from the ATSB dataset.

#### *10.4.3 Text Classification*

To ensure model robustness and prevent overfitting during the training, 10% of the train-set was set aside for model validation in each training epoch. This practice facilitated continuous evaluation and refinement of the models. Four distinct deep learning architectures, namely LSTM, BLSTM, CNN, and sRNN, were trained on this data, each offering unique capabilities for text classification tasks. Model optimization was accomplished using the Adam optimizer, chosen for its efficiency in gradient-based optimization. It is noteworthy that this study did not focus explicitly on identifying the best optimizer, thus allowing for the exploration of alternative optimization techniques in future research endeavours.

#### *10.4.4 Deep Learning Model Architecture*

For consistency and comparability across all models, a shared architecture served as the foundation, with slight adjustments for each model. This standardized architecture comprised three key components: an embedding layer, hidden layers, and an output layer. To introduce non-linearity and capture complex relationships in the data, the Rectified Linear Unit (ReLU) activation function was applied to all hidden layers. Meanwhile, the SoftMax activation function was adopted for the output layer, facilitating multiclass classification. The final predicted class was determined using the argmax function, which identifies the index associated with the highest probability in the SoftMax output. For a visual representation of the deep learning architectures employed in this study, please refer to **Figure 59**.

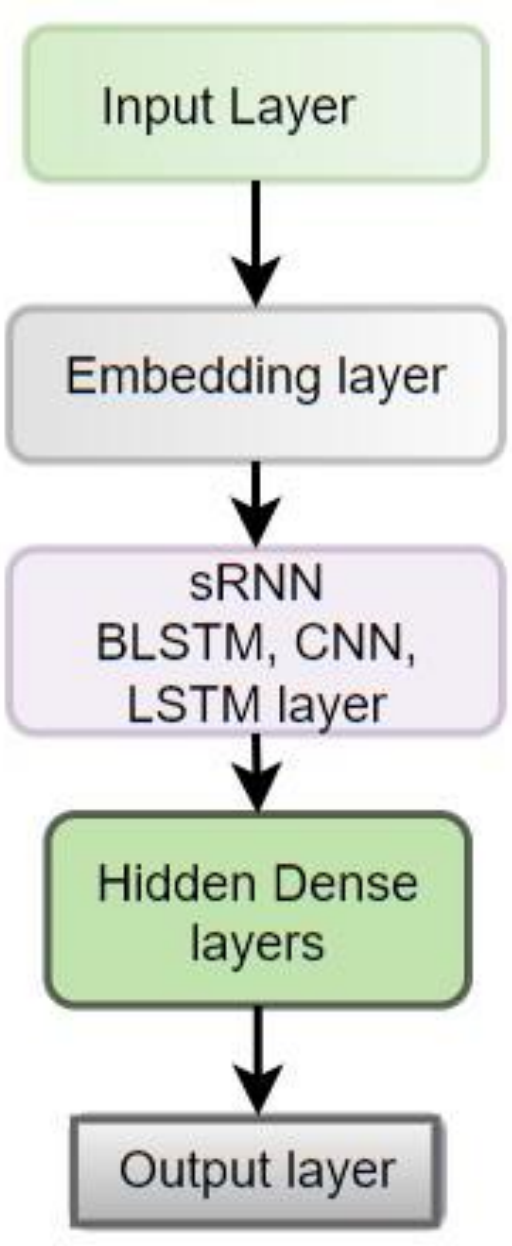

Figure 59. Deep learning architecture

This consistent architecture provided a solid foundation for training and evaluation of the deep learning models, enabling a fair comparison of their performance and the accurate classification of flight phases based on unstructured safety occurrence narratives.

#### *10.4.5 Model Performance Evaluation*

This section elucidates the evaluation criteria utilized in this study to assess the models' performance. The primary focus of this research is multi-class classification, and as such, performance was gauged based on the accuracy of predictions across various classes. To comprehensively evaluate model performance, we employed a suite of standard prediction performance metrics, including precision, recall, F1-score, and accuracy, as cited in prior literature [251, 313]. These metrics, explained in Table 22, provide a comprehensive assessment of the models' classification performance.

Table 22. SUMMARY OF EVALUATION METRICS

| Metrics | Formula | Evaluation focus |
|---|---|---|
| **Precision (p)** | $\frac{TP}{TP+FP}$ | Correctly predicted positives in a positive class |
| **Recall (r)** | $\frac{TP}{TP+TN}$ | Fraction of positive patterns correctly classified |
| **F1-score (F)** | $\frac{2*precision*recaal}{precision+recall}$ | Weighted average score of precision and recall |
| **Accuracy (acc)** | $\frac{TP+TN}{TP+FP+TN+FN}$ | Total number of instances predicted correctly |

#### *10.4.6 Confusion matrix*

A confusion matrix is an invaluable tool for visually assessing model performance in a classification task. As illustrated in Table 22, it is a square matrix in the dimensional space, m, where m is the number of unique entries in the dependent variable, essentially reflecting instances distributed among class labels during the testing phase of the AI model. The confusion matrix provides a clear visualization of the models' performance, serving as a yardstick to gauge their effectiveness. In the matrix shown in Table 23, diagonal cells denote correct predictions, including true positives (TP) and true negatives (TN), while off-diagonal cells indicate incorrect predictions, encompassing false negatives (FN) and false positives (FP) [254].

Table 23. Confusion matrix

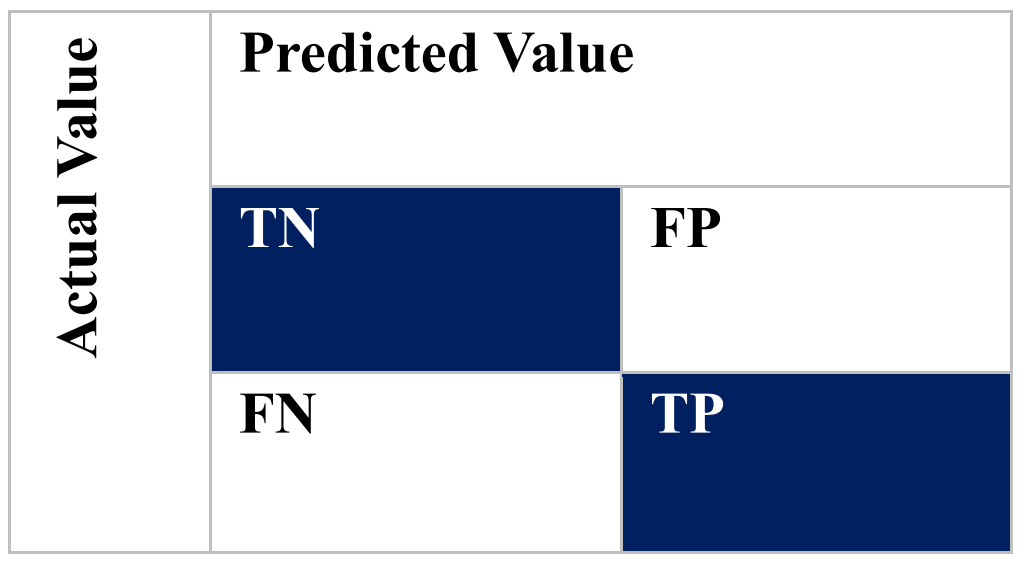

This robust evaluation approach enabled us to comprehensively assess the models' classification accuracy, precision, recall, and F1 score, providing a holistic understanding of its performance in classifying flight phases [160].

## 10.5 RESULTS AND DISCUSSION

In this study, we have explored the application of NLP and Deep Learning techniques to enhance aviation safety analysis by classifying flight phases within safety occurrence reports. Leveraging a substantial dataset of 50,778 safety occurrence reports provided by ATSB, we evaluated the performance of advanced Deep Learning models, including LSTM, CNN, BLSTM, and sRNN, using a variety of performance metrics.

### *10.5.1 Model Performance*

Our findings reveal the remarkable capabilities of these models in accurately classifying flight phases within unstructured text narratives. Table 24 shows the key performance results for each model.

Table 24. DEEP LEARNING MODEL PERFORMANCE

| **Models** | Precision (%) | Recall (%) | F1 (%) | Accuracy (%) |
|---|---|---|---|---|
| **LSTM** | **88** | 87 | **88** | **87.4** |
| **sRNN** | 78 | 77 | 77 | 77.0 |
| **BLSTM** | **88** | 87 | 87 | **87.3** |
| **CNN** | 87 | 87 | 87 | 86.5 |

These results clearly demonstrate the effectiveness of NLP and Deep Learning models in handling the complexity of aviation safety reports and inferring crucial flight phase information. Notably, the LSTM and BLSTM models exhibited the highest accuracy and precision, indicating their suitability for this task.

Both Figures 60 and 61, illustrate the training behavior and performance of the four deep learning models (SRNN, BLTM, CNN, and LSTM) in terms of validation accuracy and validation loss, respectively, over a range of training epochs. These visualizations provide valuable information for assessing and comparing the models' capabilities in solving the classification problem at hand.

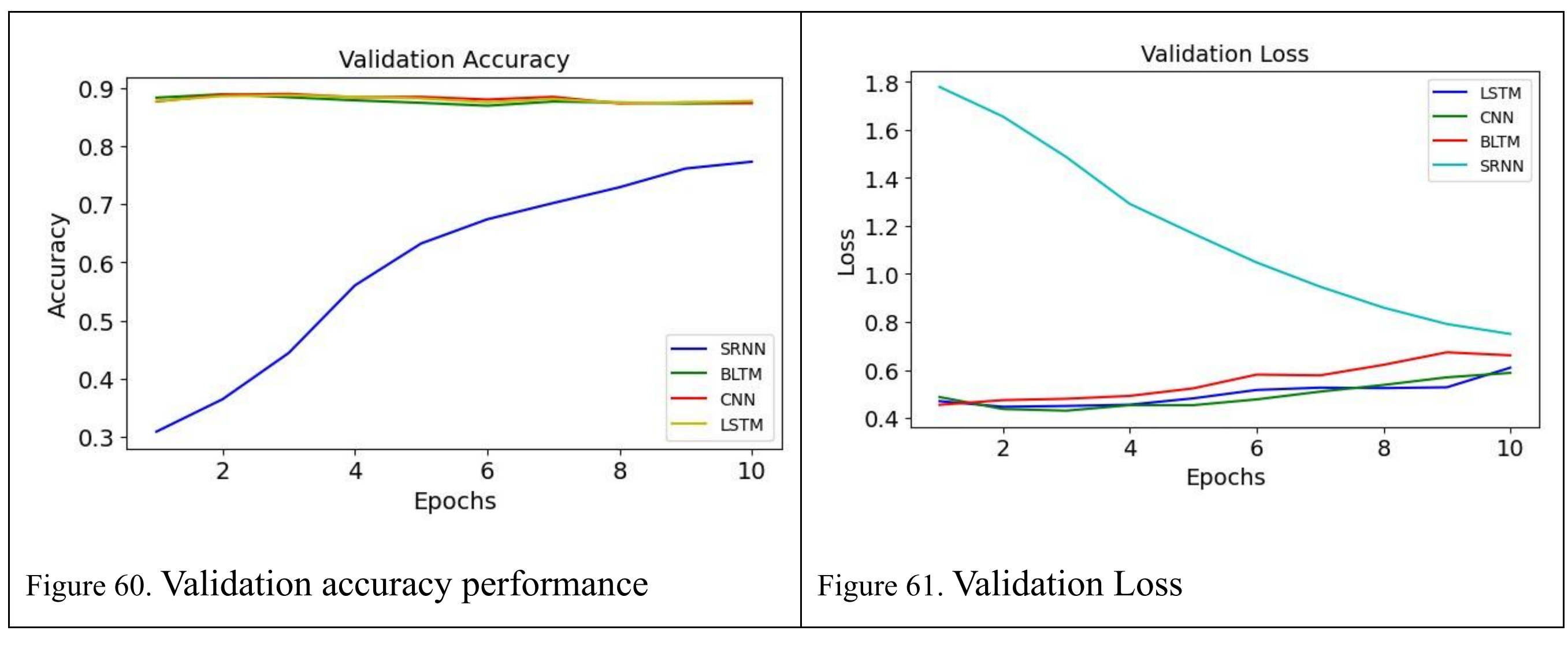

Figure 60. Validation accuracy performance

Figure 61. Validation Loss

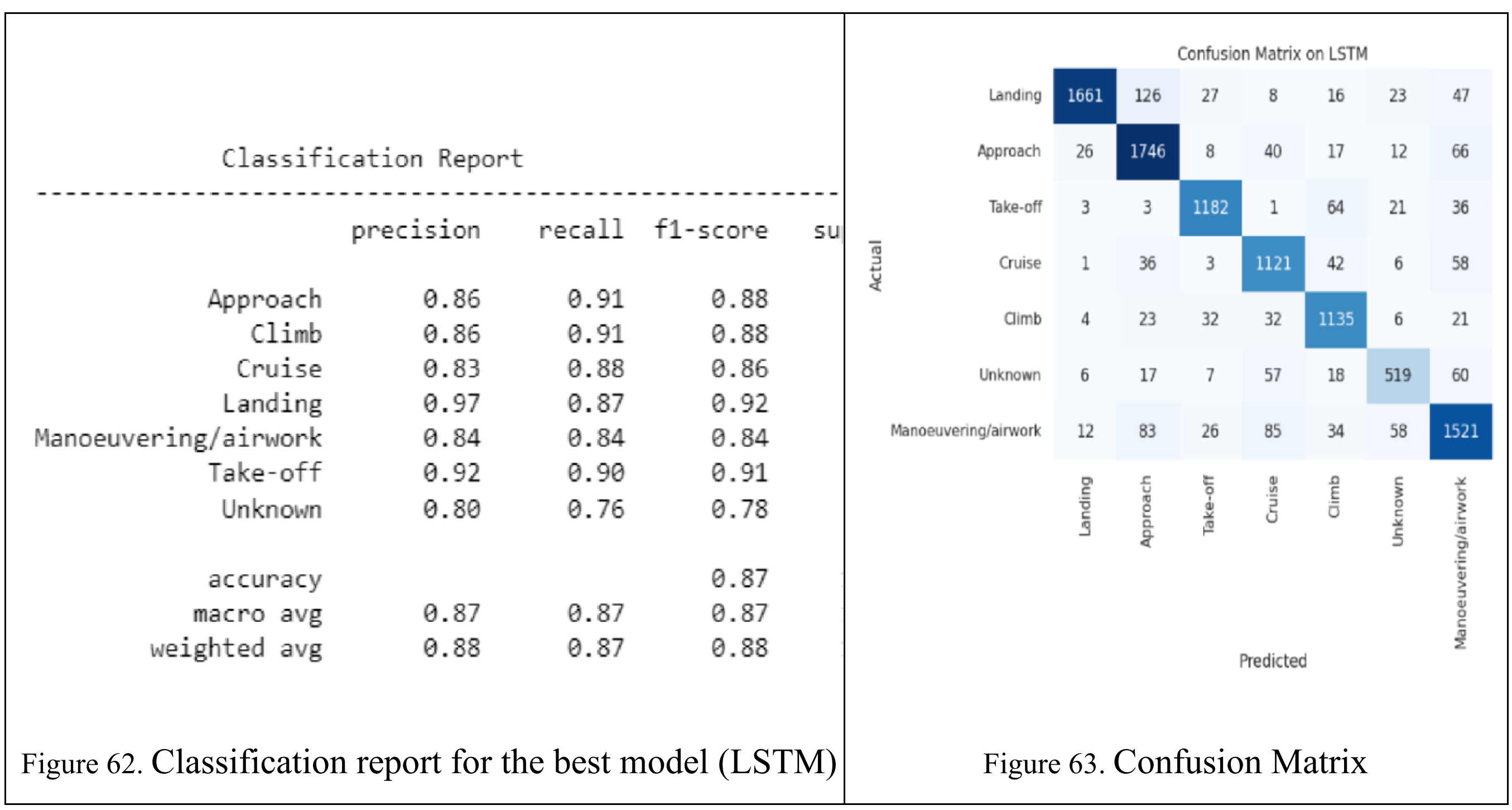

Classification Report

| | precision | recall | f1-score | su |
|---|---|---|---|---|
| Approach | 0.86 | 0.91 | 0.88 | |
| Climb | 0.86 | 0.91 | 0.88 | |
| Cruise | 0.83 | 0.88 | 0.86 | |
| Landing | 0.97 | 0.87 | 0.92 | |
| Manoeuvering/airwork | 0.84 | 0.84 | 0.84 | |
| Take-off | 0.92 | 0.90 | 0.91 | |
| Unknown | 0.80 | 0.76 | 0.78 | |
| accuracy | | | 0.87 | |
| macro avg | 0.87 | 0.87 | 0.87 | |
| weighted avg | 0.88 | 0.87 | 0.88 | |

Confusion Matrix on LSTM

| Actual \ Predicted | Landing | Approach | Take-off | Cruise | Climb | Unknown | Manoeuvering/airwork |
|---|---|---|---|---|---|---|---|
| Landing | 1661 | 126 | 27 | 8 | 16 | 23 | 47 |
| Approach | 26 | 1746 | 8 | 40 | 17 | 12 | 66 |
| Take-off | 3 | 3 | 1182 | 1 | 64 | 21 | 36 |
| Cruise | 1 | 36 | 3 | 1121 | 42 | 6 | 58 |
| Climb | 4 | 23 | 32 | 32 | 1135 | 6 | 21 |
| Unknown | 6 | 17 | 7 | 57 | 18 | 519 | 60 |
| Manoeuvering/airwork | 12 | 83 | 26 | 85 | 34 | 58 | 1521 |

Figure 62. Classification report for the best model (LSTM)

Figure 63. Confusion Matrix

The extract in Figure 62 shows the classification report in terms of accuracy, precision, recall, and F1 score for the LSTM model. The extract also gives an account of the test instance distribution among distinct phase(s) of flight entries as evidenced in the support column. On the other hand, Figure 63 gives a visual account of how the LSTM model distributes test instances in the form of a confusion matrix

## 10.6 CONCLUSION

In an era where aviation safety is paramount, the automation and precision of safety occurrence analysis take centre stage. This research has showcased the immense potential of Natural Language

Processing (NLP) and Deep Learning techniques in revolutionizing the classification of flight phases within safety occurrence reports. Leveraging a dataset of 50,778 reports from the Australian Transport Safety Bureau (ATSB), we employed advanced Deep Learning architectures, including Long Short-Term Memory (LSTM), Convolutional Neural Network (CNN), Bidirectional LSTM (BLSTM), and simple Recurrent Neural Network (sRNN), to infer critical flight phase information from unstructured text narratives. Our findings are not only promising but transformative. The LSTM, BLSTM, and CNN models exhibited remarkable performance, as evidenced by high precision, recall, and F1-score values. This exceptional performance underscores the capacity of NLP and Deep Learning models to handle the complexity of aviation safety reports, facilitating precise flight phase classification.

The implications of this research extend beyond the academic realm. The integration of NLP and Deep Learning technologies into aviation safety analysis promises to streamline the identification of safety trends, enhance contextual awareness, and inform targeted safety measures. The ability to associate safety occurrences with specific flight phases allows regulatory authorities and aviation industry stakeholders to develop proactive and context-aware safety protocols. Furthermore, the scalability and automation of this approach empower aviation safety experts to handle a higher volume of reports efficiently. This ensures that critical safety information is not overlooked and that safety enhancements can be implemented promptly, contributing to a safer aviation environment.

Therefore, the findings presented in this research signify a transformative shift in aviation safety analysis. The synergy of NLP and Deep Learning technologies not only augments the efficiency of safety occurrence analysis but also elevates the precision and contextual understanding of aviation safety incidents. Future work in this field holds great promise, including further model refinement, the integration of multimodal data sources, real-time analysis capabilities, cross-dataset validation, improved model interpretability, and the development of human-in-the-loop systems. As we continue to explore these avenues, we are poised to strengthen aviation safety measures and contribute to the ongoing mission of safer skies for all.

# Chapter 11. Comparative Study of Deep Learning Architectures for Textual Damage Level Classification

## *11.1* Abstract

Given the paramount importance of safety in the aviation industry, even minor operational anomalies can have significant consequences. Comprehensive documentation of incidents and accidents serves to identify root causes and propose safety measures. However, the unstructured nature of incident event narratives poses a challenge for computer systems to interpret. Our study aimed to leverage Natural Language Processing (NLP) and deep learning models to analyze these narratives and classify the aircraft damage level incurred during safety occurrences. Through the implementation of LSTM, BLSTM, GRU, and sRNN deep learning models, our research yielded promising results, with all models showcasing competitive performance, achieving an accuracy of over 88% significantly surpassing the 25% random guess threshold for a four-class classification problem. Notably, the sRNN model emerged as the top performer in terms of recall and accuracy, boasting a remarkable 89%. These findings underscore the potential of NLP and deep learning models in extracting actionable insights from unstructured text narratives, particularly in evaluating the extent of aircraft damage within the realm of aviation safety occurrences.

## 11.2 Introduction

In aviation safety, timely and accurate classification of aircraft damage levels is critical for effective incident analysis and the development of preventive measures [159] . Accurate damage level classification facilitates the assessment of safety risks associated with aviation incidents and accidents, aiding in the enhancement of proactive safety protocols and measures [340]. To address the challenges of accurately categorizing damage levels, we leverage the power of deep learning models in processing and analyzing textual narratives related to aviation incidents. Our study focuses on the classification of damage levels using the prominent ATSB (Air Transport Safety Bureau) dataset, a comprehensive collection of textual reports detailing aviation incidents and associated damage levels.

The accurate classification of damage levels in aviation incidents is essential for various stakeholders, including aviation regulatory authorities, airlines, manufacturers, and safety analysts. Understanding the severity of damage incurred during aviation incidents facilitates the development of robust safety protocols, maintenance practices, and incident response strategies [341]. Additionally, the efficient categorization of damage levels enables the identification of recurrent patterns and potential areas of improvement in aircraft design, maintenance procedures, and operational protocols [342, 343]. Consequently, our study contributes to the advancement of aviation safety practices and plays a pivotal role in fostering a culture of proactive safety measures within the aviation industry.

The motivation behind this research stems from the necessity to improve the current methods of damage level classification in the aviation industry. While traditional approaches rely heavily on manual inspection and analysis, our study aims to streamline and automate the classification process using advanced deeplearning models. By leveraging natural language processing (NLP) techniques and robust deep learning architectures, we seek to enhance the accuracy and efficiency of damage-level classification, enabling a more precise understanding of the severity of aviation incidents. This, in turn, empowers aviation stakeholders to make data-driven decisions and implement targeted safety measures, leading to an overall reduction in aviation risks and enhanced passenger safety.

The primary objective of this study is to evaluate the efficacy of various deep learning models, including Long Short-Term Memory (LSTM), Simple Recurrent Neural Network (sRNN), Bidirectional LSTM (BLSTM), and Gated Recurrent Unit (GRU), in accurately classifying textual narratives associated with different levels of aircraft damage. Through a comparative analysis of these models, we aim to identify the most effective and efficient approach for classifying damage levels, considering factors such as precision, recall, F1-score, and accuracy. Additionally, our study seeks to provide valuable insights into the performance and applicability of different deep-learning architectures in the domain of aviation safety, thereby contributing to the advancement of automated incident analysis and proactive safety management practices in the aviation industry.

The structure of this paper is as follows: Section II provides a review of the existing literature, highlighting the significance of aircraft damage level classification in aviation safety research and discussing relevant prior work. Section III gives an account of the methodology employed in this study, including data preprocessing, model selection, training, and evaluation. Section IV presents the results of our experiments, showcasing the performance of the deep learning models in damage level classification and discussing the interpretation of results, potential limitations, and implications for aviation safety. Finally, Section V concludes the paper by summarizing the key findings with suggestions for potential future research directions in the field of aviation safety management and incident analysis.

## 11.3 RELATED WORK

The study of damage level classification in the context of aviation incidents has garnered considerable attention in recent years, with a growing emphasis on leveraging NLP techniques and machine learning approaches for effective analysis. Notably, various studies have focused on the application of machine learning algorithms for incident severity classification, enabling the identification of critical patterns and trends in aviation safety data [89, 170].

In the specific domain of NLP-based incident analysis, researchers have explored the utilization of different text-processing methodologies, including sentiment analysis, topic modelling, and classification techniques, to extract valuable insights from textual data [331, 344]. Moreover, the application of deep learning models, such as LSTM, GRU, and other recurrent neural networks, has gained prominence in various NLP tasks, showcasing promising results in text classification and sentiment analysis [170, 324]. Another study [311] conducted an extensive exploration of various RNN architectures for sentence modelling, affirming the suitability of these models for sequential datasets like text mining. Additionally,

Pang et al. [312] applied RNN techniques to predict weather-related tasks, effectively regulating pre-flight information. Paul's comprehensive review of NLP tools in civil aviation emphasized the potential of RNN, suggesting its efficacy in mining time series data [89]. Chanen [112] proposed a deep learning approach to extract meaningful narratives from aviation safety reports, utilizing a word2vec model for semantic analysis in 186,000 ASRS reports, thereby enhancing the interpretability of safety documentation for experts. Zhang et al. [313] focused on aviation safety prognosis, employing LSTM and word embeddings to classify NTSB reports, highlighting the utility of deep learning models in improving safety analysis. Furthermore, ElSaid et al. [314] addressed the prediction of excess events in aircraft engines using LSTM recurrent neural networks, demonstrating the superior predictive capabilities of LSTM over traditional RNN architectures, particularly in the context of flight vibration datasets.

Research in aviation safety has employed various techniques to understand safety occurrences. Nanyonga et al. focused on the classification of aviation safety occurrences using natural language processing (NLP) and AI models. The study aimed to infer the damage level to the aircraft from text narratives. Evaluating the performance of various deep learning models including LSTM, BLSTM, GRU, and sRNN, they analyzed a dataset of 27,000 safety occurrence reports from the NTSB. Their results indicated competitive performance across all models. Meanwhile, Inan, [340] study delved into aircraft damage classification using machine learning methods, highlighting the impact of various factors such as zones, weather, time, and historical context on civil aviation incidents. They employed a set of machine learning algorithms, including logistic regression (LR), artificial neural networks (ANN), and decision trees (DT), to assess the significance of different parameters in classifying aircraft damage.

Additionally, Nick et al. [343] utilized an agent-based structural health monitoring system, employing unsupervised learning for identifying the existence and location of damage and supervised learning for identifying the type and severity of damage. The supervised learning techniques included support vector machines (SVM), naive Bayes classifiers (NB), and feed-forward neural networks (FFNN), while unsupervised learning techniques encompass k-means and self-organizing maps (SOM).

Furthermore, recent studies have highlighted the potential of deep learning architectures in enhancing the accuracy and efficiency of incident severity classification, demonstrating their capability to handle complex textual data and capture nuanced patterns in incident narratives [32, 122]. However, a comparative analysis of multiple deep learning models for damage level classification in the aviation domain, especially using the ATSB dataset, remains relatively scarce in the current literature.

Despite the existing research efforts, a critical gap persists in the identification of the most effective deep learning approach for damage level classification, particularly in the context of aviation incident narratives. This study aims to bridge this gap by conducting a detailed comparative analysis of LSTM, SRNN, BLSTM, and GRU models, providing valuable insights into the performance and applicability of these models in aviation incident analysis.

## 11.4 METHODOLOGY

In this section, we outline the methodology employed in this research to classify aircraft damage levels within safety occurrence reports from the ATSB using NLP and Deep Learning techniques. Our approach involves data preprocessing, model selection, training, and evaluation as depicted in **Figure 64**.

### *11.4.1 Data Collection*

Aviation incident investigation reports are collected and published by various organizations such as the Aviation Safety Reporting System (ASRS), the National Transportation Safety Board (NTSB) and ATSB. For this study, the researchers utilized the ATSB aviation incident investigation reports that were recorded in Australia for the period of 10 years were considered resulting in a dataset with 53,275 records where the data was sourced directly from the ATSB investigation authorities spanning from 1/01/2013 to 12/31/2022. Moreover, the dataset comprised 50,778 records following data preprocessing and cleaning. From each report, the 'Summary' and 'damageLevel' fields were extracted for training and validation of deep learning models.

### *11.4.2 Text Processing*

Text preprocessing plays a crucial role in preparing unstructured text data for machine learning models. In this study, we leveraged the Keras deep learning library for its extensive collection of deep learning models and model layers. The Tokenizer module was utilized to efficiently generate tokens and sequence vectors from input text. To encode categorical data, such as DamageLevel labels (i.e., destroyed, Substantial, Minor, and None), we employed the to_categorical module in Keras, mapping these categorical entries to numerical values using one-hot encoding for each data instance.

To address challenges related to special characters, punctuation, and stop words, as well as to perform word lemmatization, we harnessed the capabilities of the spacy library. Spacy is a Python library tailored for text-processing tasks, encompassing functionalities like named entity recognition and word tagging. It maintains an extensive list of special characters, punctuation marks, and stop words and undergoes regular updates whenever necessary to remain current.

With the aforementioned tools at our disposal, each input text narrative underwent a comprehensive preprocessing pipeline, ultimately being transformed into a representative sequence or vector with a fixed length of 2000. For narratives with fewer than 2000 words, we padded the numeric sequences with zeros, ensuring uniformity. In contrast, narratives exceeding 2000 words were truncated to meet this standardized length. The vocabulary size of the text corpus was set to 100,000, accommodating a broad range of terms.

To partition the dataset into training (80%), and testing (20%) sets, we utilized the train-test-split module from scikit-learn. All experiments conducted in this study were implemented using the Python programming language, with Jupyter Notebook serving as the chosen code editor. This rigorous text preprocessing framework laid the foundation for subsequent model training and evaluation, enabling the accurate classification of aircraft damage levels based on unstructured safety occurrence narratives from the ATSB dataset.

### *11.4.3 Text Classification*

To ensure model robustness and prevent overfitting during the training, 10% of the train-set was set aside for model validation in each training epoch. This practice facilitated continuous evaluation and refinement of the models. Deep learning models including sRNN, LSTM, BLSTM, and GRU are trained on this data, each offering unique capabilities for text classification tasks. Model optimization was accomplished using the Adam optimizer, chosen for its efficiency in gradient-based optimization [345]. It is noteworthy that this study did not focus explicitly on identifying the best optimizer, thus allowing for the exploration of alternative optimization techniques in future research endeavors.

### *11.4.4 Deep Learning Model Architecture*

For consistency and comparability across all models, a shared architecture served as the foundation, with slight adjustments for each model. This standardized architecture comprised three key components: an embedding layer, hidden layers, and an output layer. To introduce non-linearity and capture complex relationships in the data, the Rectified Linear Unit (ReLU) activation function was applied to all hidden layers. Meanwhile, the SoftMax activation function was adopted for the output layer, facilitating multiclass classification. The final predicted class was determined using the argmax function, which identifies the index associated with the highest probability in the SoftMax output. For a visual representation of the deep learning architectures employed in this study, please refer to Figure 65. This consistent architecture provided a solid foundation for training and evaluation of the deep learning models, enabling a fair comparison of their performance and the accurate classification of aircraft damage levels based on unstructured safety occurrence narratives.

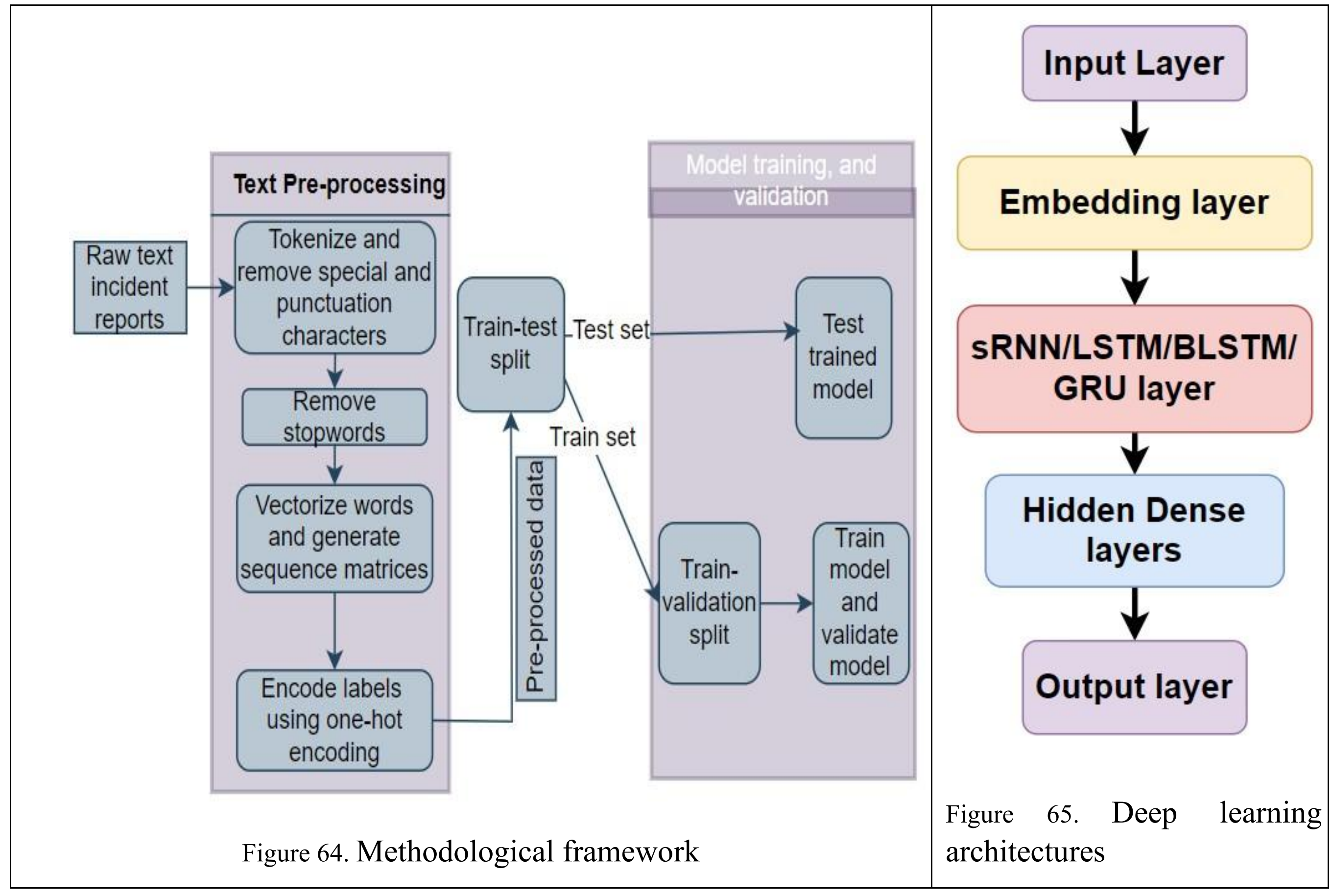

Figure 64. Methodological framework

Figure 65. Deep learning architectures

a) Simple Recurrent Neural Networks (RNNs).

The Simple Recurrent Neural Network (sRNN) is a fundamental type of RNN architecture that processes sequential data by feeding the output of the previous time step as input to the current time step. Its architecture is relatively basic, consisting of a single hidden layer that facilitates the flow of information from one step to the next [346]. The mathematical formulation for an sRNN can be described as follows:

$h_t = \sigma(W_h \cdot [h_{t-1}, x_t] + b_h)$, where $h_t$ represents the hidden state at time step $t$, $x_t$ denotes the input at time step $t$, and $\sigma$ is the activation function. Despite its simplicity, the sRNN may struggle to capture long-range dependencies and complex temporal patterns, making it more suitable for simple sequential tasks that do not involve intricate contextual relationships.

b) GRU (Gated Recurrent Unit)

The Gated Recurrent Unit (GRU) is a type of recurrent neural network (RNN) that operates on sequential data. It was designed to address the limitations of the vanishing gradient problem and the computational cost associated with the more complex LSTM architecture. GRU achieves this through a simplified architecture that consists of two gates: the update gate and the reset gate. These gates control the flow of information within the network and regulate the retention or discarding of information from the previous time step. The update gate determines how much of the past information needs to be carried forward, while the reset gate decides which parts of the past information should be forgotten [347]. The architecture of the GRU can be expressed mathematically through the following formulas:

- Update gate: $z_t = \sigma(W_z \cdot [h_{t-1}, x_t])$, where $z_t$ represents the update gate output at time step $t$, $h_{t-1}$ is the previous hidden state, $x_t$ is the input at time step $t$, and $W_z$ is the weight matrix associated with the update gate.
- Reset gate: $r_t = \sigma(W_r \cdot [h_{t-1}, x_t])$, where $r_t$ represents the reset gate output at time step $t$, and $W_r$ is the weight matrix associated with the reset gate.
- New memory content: $h_t' = tanh(W \cdot [r_t \odot h_{t-1}, x_t])$, where $h_t'$ represents the new candidate hidden state at time step $t$, and $W$ is the weight matrix associated with the new memory content.
- Final memory content: $h_t = (1 - z_t) \odot h_{t-1} + z_t \odot h_t'$, where $h_t$ represents the updated hidden state at time step $t$, combining the previous hidden state with the new memory content based on the update gate output.

c) LSTM (Long Short-Term Memory)

The Long Short-Term Memory (LSTM) network is a type of RNN that was developed to address the issue of capturing long-range dependencies in sequential data. It is designed to store information over long periods, making it well-suited for tasks that involve understanding context and temporal patterns. The architecture of LSTM is more complex compared to that of a basic RNN, featuring memory cells and various gates, including the input gate, output gate, and forget gate [324]. These gates regulate the flow of information and determine which information to keep or discard. The key formulas involved in the LSTM architecture are as follows:

- Forget gate: $f_t = \sigma(W_f \cdot [h_{t-1}, x_t] + b_f)$, where $f_t$ represents the output of the forget gate at time step $t$, $h_{t-1}$ is the previous hidden state, $x_t$ is the input at time step $t$, and $W_f$ is the weight matrix associated with the forget gate, and $b_f$ is the bias.

- Input gate: $i_t = \sigma(W_i \cdot [h_{t-1}, x_t] + b_i)$, where $i_t$ represents the output of the input gate at time step $t$, $W_i$ is the weight matrix associated with the input gate, and $b_i$ is the bias.

- New candidate values: $g_t = tanh(W_g \cdot [h_{t-1}, x_t] + b_g)$, where $g_t$ represents the new candidate values at time step $t$, $W_g$ is the weight matrix associated with the candidate values, and $b_g$ is the bias.

- Cell state: $C_t = f_t \odot C_{t-1} + i_t \odot g_t$, where $C_t$ represents the cell state at time step $t$, combining the previous cell state with the input and forget gate outputs.

- Output gate: $o_t = \sigma(W_o \cdot [h_{t-1}, x_t] + b_o)$, where $o_t$ represents the output of the output gate at time step $t$, $W_o$ is the weight matrix associated with the output gate, and $b_o$ is the bias.

- Hidden state: $h_t = o_t \odot tanh\,(C_t), where\ h_t$ represents the hidden state at time step $t$, combining the cell state with the output gate output.

d) BLSTM (Bidirectional Long Short-Term Memory)

The Bidirectional Long Short-Term Memory (BLSTM) model is an extension of the traditional LSTM architecture that processes input sequences in both forward and backward directions. By incorporating bidirectional processing, BLSTM can capture information from both past and future contexts simultaneously. This allows the model to understand the context and dependencies in a more comprehensive manner, making it well-suited for tasks that require a deep understanding of the sequence context. The mathematical formulation for a BLSTM includes the combination of forward and backward LSTM operations, allowing the model to capture long-term dependencies effectively [326]. BLSTM is commonly used in tasks such as speech recognition, language translation, and named entity recognition, where understanding both past and future context is critical for accurate predictions and analyses.

### *11.4.5 Model Performance Evaluation*

This section elucidates the evaluation criteria utilized in this study to assess the models' performance. The primary focus of this research is multi-class classification, and as such, performance was gauged based on the accuracy of predictions across various classes. To comprehensively evaluate model performance, we employed a suite of standard prediction performance metrics, including recall, F1-score, precision, and accuracy.

## 11.5 RESULTS AND DISCUSSION

In this study, we have explored the application of NLP and Deep Learning techniques to enhance aviation safety analysis by classifying aircraft damage level within safety occurrence reports. Leveraging a substantial dataset of 50,778 safety occurrence reports provided by ATSB, we evaluated the performance of advanced Deep Learning models, including sRNN, LSTM, BLSTM, and, GRU using a variety of performance metrics.

### *11.5.1 Model Performance*

Our findings reveal the remarkable capabilities of these models in accurately classifying damage level of aircraft within unstructured text narratives. Table 25 shows the key performance results for each model.

Table 25. DEEP LEARNING MODEL PERFORMANCE

| **Models** | Precision (%) | Recall (%) | F1 (%) | Accuracy (%) |
|---|---|---|---|---|
| **sRNN** | 0.87 | **0.89** | 0.87 | **0.89** |
| **LSTM** | 0.87 | 0.88 | 0.87 | 0.88 |
| **BLSTM** | 0.86 | 0.88 | 0.87 | 0.88 |
| **GRU** | 0.87 | 0.88 | 0.87 | 0.88 |

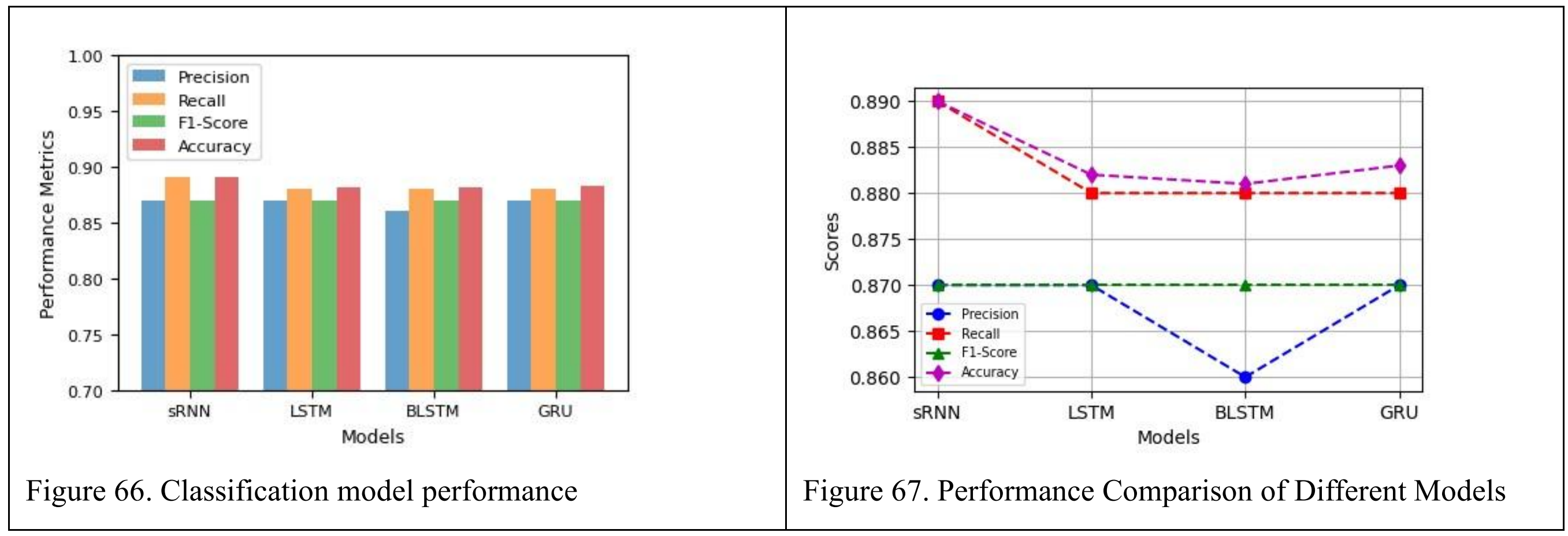

Figure 66. Classification model performance

Figure 67. Performance Comparison of Different Models

As illustrated in Figures 66 and 67 the results clearly demonstrate the effectiveness of NLP and Deep Learning models in handling the complexity of aviation safety reports and inferring aircraft damage level information from the textual narratives. Notably, the sRNN model exhibited the highest accuracy and precision, indicating its suitability for this task.

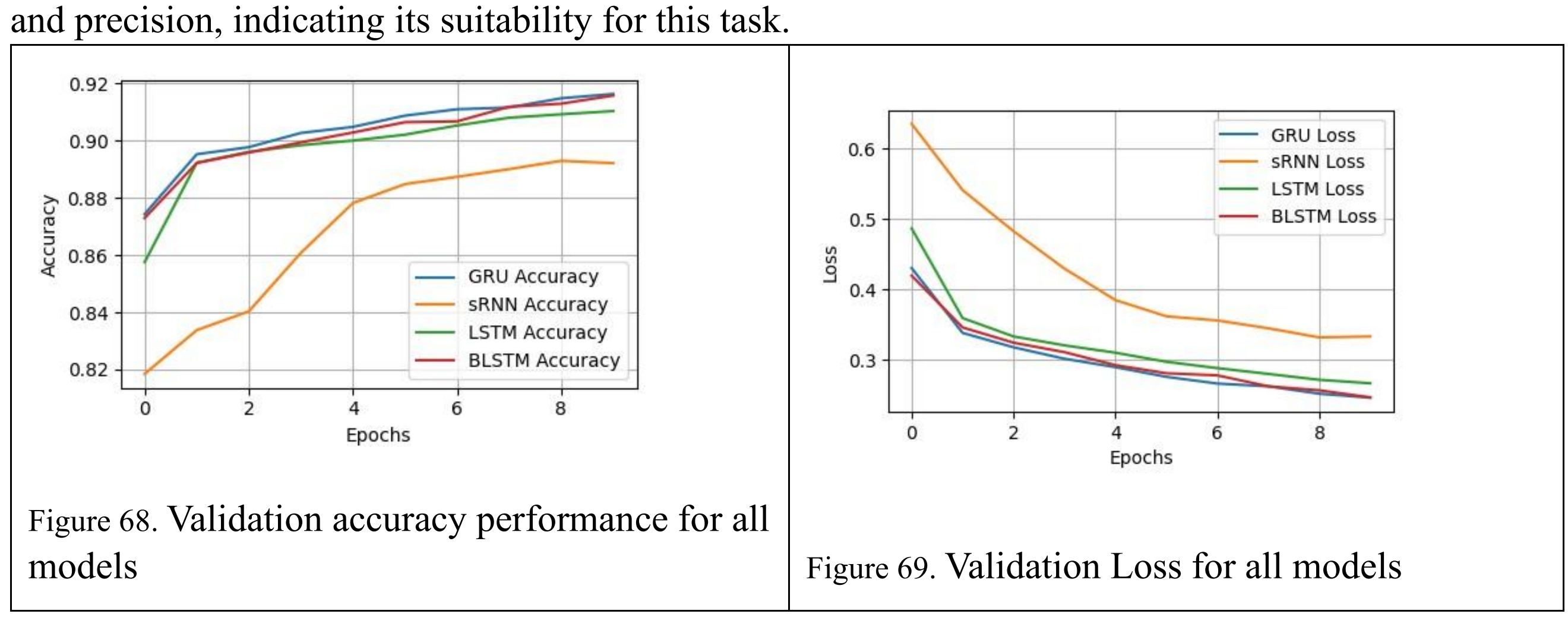

Figure 68. Validation accuracy performance for all models

Figure 69. Validation Loss for all models

Both Figures 68 and 69, illustrate the training behavior and performance of the four deep learning models (sRNN, BLTM, LSTM, and GRU) in terms of validation accuracy and validation loss, respectively, over a range of training epochs. These visualizations provide valuable information for assessing and comparing the models' capabilities in solving the classification problem at hand.

Classification Report

| | precision | recall | f1-score | support |
|---|---|---|---|---|
| Minor | 0.69 | 0.56 | 0.61 | 675 |
| None | 0.91 | 0.98 | 0.94 | 8737 |
| Substantial | 0.93 | 0.84 | 0.88 | 416 |
| Destroyed | 0.64 | 0.24 | 0.35 | 801 |
| accuracy | | | 0.89 | 10629 |
| macro avg | 0.79 | 0.65 | 0.69 | 10629 |
| weighted avg | 0.87 | 0.89 | 0.87 | 10629 |

Figure 70. Classification report for the best model (sRNN)

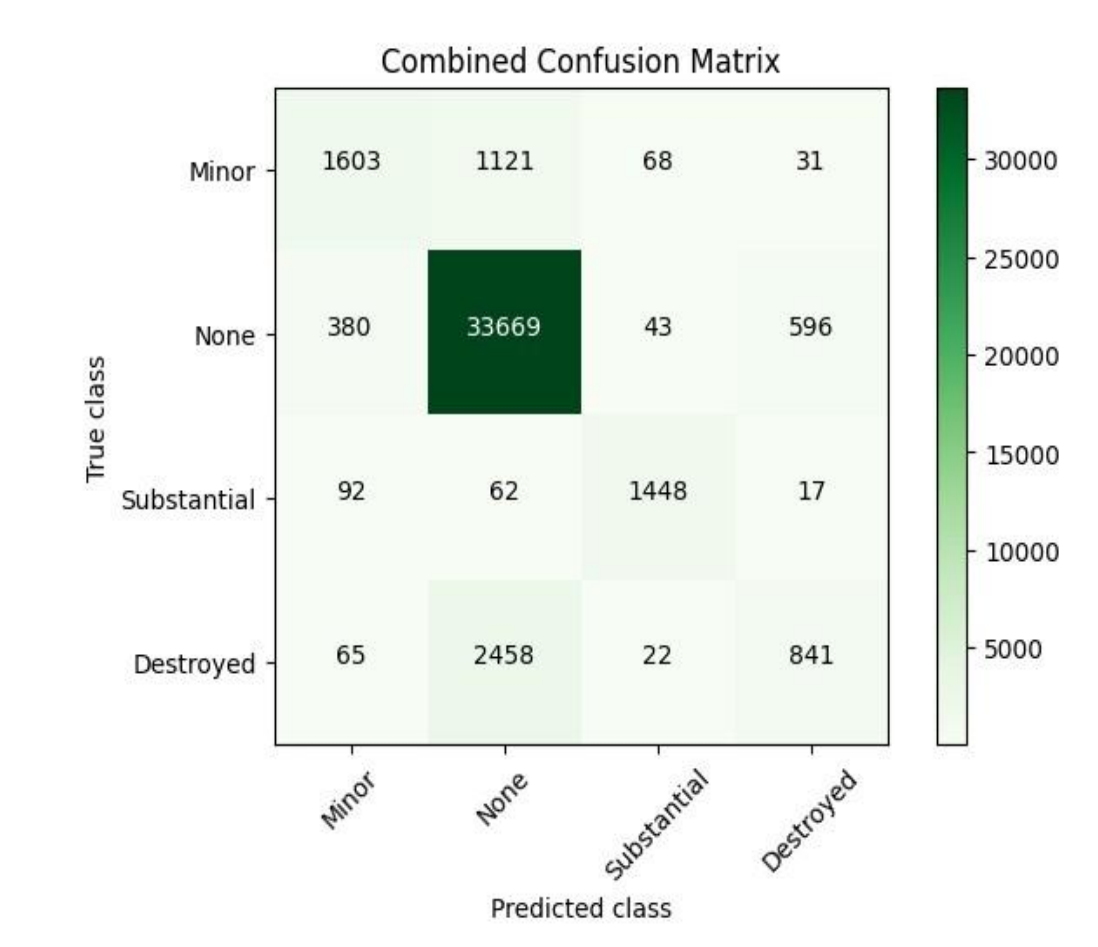

Figure 71. Confusion Matrix for all of the four models

The extract in Figure 70 shows the classification report in terms of accuracy, precision, recall, and F1 score for the sRNN model. The extract also gives an account of the test instance distribution among distinct damage level entries as evidenced in the support column. On the other hand, Figure 71 gives a visual account of how the models distribute test instances in the form of a confusion matrix.

## 11.6 CONCLUSION

This study examined four deep-learning architectures for the classification of damage levels within aircraft incident narratives sourced from the ATSB dataset. The comparative analysis of sRNN, LSTM, BLSTM, and GRU models for textual damage level classification highlighted the superior performance of the sRNN model in terms of both accuracy and recall. With precision and F1 scores on par with the other models, the notable lead in accuracy and recall underscores the effectiveness of the sRNN architecture in accurately predicting damage levels. Despite the competitive performance of LSTM, BLSTM, and GRU, the dominance of the sRNN model in key evaluation metrics signifies its potential as a reliable solution for textual damage level classification tasks.

The findings presented in Table 25, Figures 66, and 67 clearly demonstrate the efficiency and accuracy of the deep learning models in handling the complexity of aviation safety reports and inferring aircraft damage level information from the textual narratives. The performance evaluation presented in Figures 68 and 69 provides valuable insights into the training behavior and capabilities of the models, showcasing their ability for solving the classification problem. Additionally, the classification report for the sRNN model, as depicted in Figure 70, offers a comprehensive view of the model's precision, recall, and F1 score, shedding light on the test instance distribution across different damage level categories. The corresponding confusion matrix shown in Figure 71 further visualizes how the models distribute test instances, providing a holistic perspective on their classification performance.

Looking ahead, the exploration of additional research directions holds significant promise. Integration of advanced textual features and semantic analysis techniques could further enhance the models' understanding of nuanced language structures and contextual information within the incident narratives. Also, the incorporation of multimodal data sources, such as image and video data from incident reports, can enrich the classification models' comprehension of the damage severity, enabling a more comprehensive and holistic analysis of the incidents.

# Chapter 12. Impact of Dataset Size & Data Source on Aviation Safety Incident Prediction Models with Natural Language Processing

## 12.1 Abstract

This study investigates the influence of dataset size on machine learning model performance in aviation safety incident prediction. We compare two datasets: the Aviation Safety Network (ASN) dataset and the National Transport Safety Board (NTSB) dataset. These datasets cover the years 2000 to 2020. Employing neural network architectures (LSTM, GRU, CNN, and RNN) and with advanced natural language processing techniques (NLP), we process textual narratives. Our findings underscore the role of dataset size in shaping predictive model performance. Larger datasets consistently yield higher predictive accuracy. Notably, despite having the same number of instances, the NTSB dataset outperforms the ASN dataset. Additionally, CNN stands out with better performance across various evaluation metrics. Furthermore, we examine the performance of these models on datasets of different sizes, affirming that larger datasets lead to better model performance, further highlighting the significance of dataset size in aviation safety prediction. This research emphasizes dataset size as a determinant of predictive model efficacy in aviation safety and suggests improvements in the data collection process for the ASN dataset. These insights resonate within the industry, underlining the need for substantial data collection efforts and the potential benefits of advanced NLP techniques. This study contributes to enhancing safety protocols and guides the broader machine-learning community toward boundless possibilities at the intersection of data, technology, and innovation.

## 12.2 Introduction

Aviation safety is an issue of paramount importance in the modern world, as it underpins countless lives and substantial economic interests that rely on the reliability and security of air travel [341]. In an era where the aviation industry generates a wealth of textual data, encompassing safety occurrence reports, safety narratives, and accident investigations, the utilization of Natural Language Processing (NLP) and Machine Learning (ML) has become essential for extracting valuable insights from this extensive textual corpus [24], [189]. This research embarks on a mission to delve into the implications of dataset size on the performance of ML models applied to aviation text narratives. We undertake this exploration using two distinctive datasets from the Aviation Safety Network (ASN) and the National Transport Safety Board (NTSB), spanning the years 2000 to 2020. Specifically, our focus is to discern the effect of varying dataset sizes on the predictive accuracy of four popular deep learning models: Gated Recurrent Units (GRU), Long Short-Term Memory networks (LSTM), Convolutional Neural Networks (CNN), and Recurrent Neural Networks (RNN).

The aviation industry has undergone continuous evolution [29], marked by relentless efforts to enhance safety standards [198]. Data-driven approaches have played an indispensable role in this pursuit. While traditional methods of analyzing aviation safety data have been effective [198, 341], the sheer volume and complexity of textual narratives stemming from aviation incidents and accidents have necessitated the development of more sophisticated analytical tools. NLP and ML, driven by the capabilities of deep learning models, now present a promising avenue for transforming unstructured textual data into actionable insights. These innovations hold the potential to revolutionize aviation safety analysis. In recent years, the world of deep learning has seen a surge in interest and application across various domains. Models like GRU, LSTM, CNN, and RNN, initially designed to tackle a wide array of Natural Language Processing (NLP) tasks, have garnered attention for their remarkable performance in endeavors such as sentiment analysis, language translation, and text classification. However, in the realm of aviation safety narratives, their potential remains relatively untapped.

Textual data has been at the heart of aviation safety analysis for a considerable time. These narratives, encompassing safety occurrence reports, other safety narratives, and accident investigations, provide indispensable context and insights into the intricate web of events surrounding aviation incidents and accidents according to the National Research Council [348]. However, extracting meaningful information from these narratives manually is a time-consuming and intricate endeavor. Traditional methods for analyzing aviation safety data involve statistical techniques and expert assessments [198, 341]. These methods have been effective in identifying trends and contributing factors in aviation incidents [349]. However, they may not fully leverage the rich information embedded in textual narratives. In contrast, machine learning techniques have garnered prominence in aviation safety analysis. Researchers have applied a diverse array of ML algorithms to predict safety-related events, including accidents and incidents [94{Zhang, 2019 #602{Zhang, 2019 #602]}. These studies have illuminated the immense potential of machine learning in advancing aviation safety.

In our pursuit of understanding the impact of dataset size on aviation safety incident prediction, we enlist the support of four formidable deep-learning models: a) Long Short-Term Memory (LSTM) and Gated Recurrent Units (GRU), these recurrent neural networks (RNN) variants, tailored to tackle the vanishing gradient problem, have exhibited unparalleled prowess in modeling sequential data [324]. Their innate capacity to capture long-range dependencies positions them as suitable candidates for dissecting aviation safety narratives. b) Recurrent Neural Networks (RNN), as foundational elements in NLP, RNNs possess sequential processing capabilities ideal for tasks where temporal order holds significance [350]. This feature makes them valuable tools for unravelling the temporal aspects embedded within aviation safety incidents. Lastly, Convolutional Neural Networks (CNN), Although renowned for their excellence in image processing, CNN architectures have also demonstrated their adaptability to text classification and sentiment analysis [351]. By treating textual data as a one-dimensional signal, these models reveal a talent for capturing local patterns in textual narratives a skill with potential applications in aviation safety analysis.

Dataset size is an ever-present variable in the world of machine learning. Extensive research has explored its impact on model performance, revealing a general trend wherein larger datasets tend to improve model performance and generalization [352]. However, the relationship between dataset size and performance is not always linear. Diminishing returns may manifest with excessively large datasets, warranting a nuanced understanding of this dynamic. Practical constraints often dictate the availability of more limited datasets. In the context of this study, we are not focusing on data augmentation or transfer learning techniques as done by other researchers [353]. Rather, our focus is to investigate the fundamental impact of dataset size on the performance of machine learning models applied to aviation safety narratives.

Therefore, this study assumes several core objectives: a) To evaluate the predictive prowess of prominent models (LSTM, GRU, CNN, and RNN) in foreseeing aviation safety incidents across datasets of varying sizes. b) To engage in a comparative analysis of these models' predictive capabilities when applied to both a fixed-size dataset (ASN, boasting 4,526 instances) and datasets of varying sizes (NTSB, ranging from 4,526 to 24,137 instances). c) To dissect the intricate trade-offs that emerge between dataset size and model performance within the realm of aviation safety prediction and d) To distill insights and best practices that can empower the effective utilization of machine learning models in aviation safety analysis, all while considering the practical constraints tied to data collection and model deployment.

Understanding the nuanced interplay between dataset size and predictive model performance is an imperative milestone in the journey to deploy effective aviation safety incident prediction systems. These systems, capable of proactively addressing safety concerns, mitigating accident risks, and elevating industry-wide safety standards, hold the potential to revolutionize aviation safety. Moreover, this study extends its reach beyond the confines of aviation safety. By delivering empirical evidence on the influence of dataset size on model performance across a diverse array of neural network architectures, it equips data scientists, researchers, and practitioners in various domains with the tools needed to optimize their predictive models when confronted with datasets of variable sizes.

The structure of this paper aligns with the research journey we have embarked upon. After this introduction, Section II delves into the related work encompassing aviation safety and ML models (LSTM,

GRU, CNN, RNN). Section III serves as the gateway to our methodology, offering insights into dataset sizes and performance metrics. In Section IV, we unveil the results of our study, fostering a comprehensive discussion of their implications for aviation safety research. Lastly, Section V ushers in the conclusion, illuminating the path forward for future research endeavors in this dynamic field.

## 12.3 RELATED WORK

In the domain of aviation safety and machine learning, a substantial body of work has emerged, setting the stage for our study investigating the influence of dataset size on predictive model performance. This section provides a comprehensive review of relevant research endeavors, with a specific emphasis on studies that inform our investigation. We narrow our focus to the four deep learning models central to our study: Long Short-Term Memory Networks (LSTM), Gated Recurrent Units (GRU), Convolutional Neural Networks (CNN), and Recurrent Neural Networks (RNN).

Aviation safety has long been a subject of rigorous examination and research, driven by the imperative need to mitigate accidents and incidents within air travel. A notable contribution in this direction comes from Zhang and Mahadevan [30], who explored the application of machine learning (ML) techniques for aviation safety predictions. Their study delved into the predictive capabilities of various models, including decision trees and support vector machines, in identifying safety-related incidents. However, it primarily focused on structured data and did not emphasize the analysis of text narratives, which often contain rich contextual information absent in structured datasets. Our study builds upon this foundation by concentrating on these textual narratives, offering a wealth of contextual insights.

The utilization of Natural Language Processing (NLP) techniques has gained traction in aviation safety analysis, allowing for the examination of unstructured textual data prevalent in safety narratives and incident reports. In the realm of aviation safety narratives, another previous study delved into the classification of safety occurrences [160]. They also harnessed NLP and artificial intelligence (AI) models, specifically ResNet and sRNN, to categorize flight phases based on unstructured text narratives from safety occurrence reports within the National Transport Safety Board (NTSB) dataset. The study underscored the importance of extracting critical information from unstructured textual data in aviation safety narratives. Furthermore, another more recent study employed NLP methods, specifically word embeddings, RNNs, LSTM, GRU, and CNN, to analyze safety narratives, with a focus on identifying aircraft damage levels and enhancing safety measures [188]. While these studies made significant strides in leveraging NLP for aviation safety, they did not delve deeply into the effects of dataset size on model performance. This study extends this by explicitly exploring how varying dataset sizes influence the performance of deep learning models, seeking to uncover whether larger datasets lead to more accurate predictions, a question largely unexplored in the context of aviation safety narratives. Also, a comprehensive review offers an overview of deep learning models in Natural Language Processing [334]. While the focus extends to broader applications, including sentiment analysis and language modeling, they underscore the significance of models like LSTM, GRU, CNN, and RNN in capturing nuanced patterns within textual data. Our study aligns directly with this perspective, employing these very models to extract meaningful insights from aviation safety narratives.

Understanding the influence of dataset size on machine learning model performance is a foundational concern in the field. While larger datasets often yield improved results, the relationship is intricate and influenced by various factors. Halevy et al; [352] conducted an influential study on the impact of data scale on machine learning, exploring how data size influences model performance. They found that larger datasets generally lead to improved model generalization but acknowledged that the relationship is not strictly linear and can exhibit diminishing returns with exceedingly large datasets. Our investigation extends this understanding to the domain of aviation safety, aiming to determine whether the observed trends in dataset size and model performance hold true when applied to safety narratives in the aviation industry. Specifically, we seek to discern whether the benefits of larger datasets outweigh the practical constraints often associated with data collection and processing.

Several noteworthy studies have harnessed ML techniques for predictive applications, predominantly relying on the National Transport Safety Board (NTSB) dataset and the Aviation Safety Reporting System (ASRS) dataset. Heinrich [158] conducted an in-depth multivariate analysis of critical factors in aviation safety, employing classification methods with a focus on the NTSB and ASRS datasets. Similarly, Verma et al [32] provided a comprehensive comparative overview of accident forecasting approaches, incorporating neural networks (NN), support vector machines (SVM), and Bayesian networks (BN), with an emphasis on the NTSB and ASRS datasets. Additionally, another study [150] explored the application of structural topic modeling to aviation safety data, utilizing Latent Dirichlet Allocation (LDA) techniques, predominantly with the NTSB and ASRS datasets. While these studies made substantial contributions to aviation safety prediction using NTSB and ASRS data, there exists a noticeable gap in the literature concerning research that simultaneously utilizes the Aviation Safety Network (ASN) dataset and the NTSB dataset or compares the performance of machine learning models on these distinct datasets.

In a prior study conducted by Mersha et al [308], the researchers delved into the realm of aviation safety by producing and evaluating various Recurrent Neural Network (RNN) models. Their investigation centered on NASA's Ames and Langley Research Centers' dataset, focusing on predictive modeling for F16 fighter jet data and fault sensor conditions. Within their research, they explored different RNN architectures, including Simple Recurrent Neural Networks (sRNN), Gated Recurrent Units (GRU), and Long Short-Term Memory (LSTM). Furthermore, they conducted a comparative analysis by combining these RNN architectures, leading to the creation of hybrid models such as GRU-LSTM, GRU-sRNN, and LSTM-sRNN, with the objective of identifying the most effective configuration. Their study primarily assessed the performance of these models in predicting critical parameters such as the angle of attack of F-16 flight jets and the accuracy of the angle of attack measurements, among others. The research outcomes revealed that sRNN, particularly when coupled with the trained Adam optimization algorithm, exhibited superior predictive capabilities compared to other algorithms employed. Notably, the combination of sRNN and GRU outperformed alternative neural network configurations investigated in their study.

Furthermore, Song et al [309] carried out an investigation centered on the analysis of narratives related to railroad accidents. The objective was to enhance text classification accuracy and achieve a better understanding of accident records through the utilization of Bidirectional Long Short-Term Memory (BLSTM), Convolutional Neural Network (CNN), and RNN models. Their findings demonstrated the potential of these models to improve classification accuracy on accident datasets, leading to a recommendation for further exploration of various model architectures when working with accident narratives. While their proposed model exhibited enhanced performance, with results exceeding a certain threshold, the study suggested that fine-tuning model parameters could potentially yield even better results.

In the context of aircraft engine excess events prediction, ElSaid et al [314] employed an LSTM model to forecast vibrations in aircraft engines. Their investigation utilized data from 85 flights, encompassing approximately 22 hours of recorded data. The primary objective was to determine the suitability of RNNs, specifically employing LSTM architecture, in predicting aircraft engine vibrations. Comparative analysis with traditional RNNs highlighted the LSTM RNN's capacity to accurately predict vibration patterns in flight datasets. Moreover, Tutubalina and Nikolenko [311] focused on sentence modeling using various RNN architectures and observed that these models displayed superior suitability for sequential datasets, particularly in the context of text mining. Pang et al [312] harnessed RNN techniques for weather-related predictions, achieving successful outcomes for regulating pre-flight information. Lastly, de Vries [124] explored diverse Natural Language Processing (NLP) techniques applied to aviation reports, with a specific emphasis on civil aviation corpus. While the study did not extensively report the performance metrics of these techniques, it recommended the utilization of NLP, particularly RNNs, for mining timeseries data in the aviation domain.

Existing research has illuminated various aspects of aviation safety, machine learning, NLP, and the impact of dataset size on model performance. Our study stands at the convergence of these domains, contributing by specifically examining the effects of dataset size on the performance of LSTM, GRU, CNN, and RNN models when applied to aviation safety narratives. Our work addresses a critical gap in the literature, investigating the interplay between dataset size and predictive model performance within the aviation safety context. This inquiry is vital for optimizing incident prediction systems in the aviation industry, where safety remains paramount.

## 12.4 Methodology

In this section, we explain the methodology employed to compare the performance of machine learning (ML) models in aviation safety incident prediction. Our study leverages two distinct datasets, the Aviation

Safety Network (ASN) and the National Transport Safety Board (NTSB), both rich sources of textual narratives related to aviation safety incidents. Given the unstructured nature of these narratives, we rely heavily on natural language processing (NLP) techniques to facilitate the predictive analysis. Our methodology encompasses several key phases, illustrated in Figure 72, and is expounded upon in the subsequent sections.

*12.4.1 Data Acquisition*

Aviation incident and accident investigation reports are sourced from various organizations, including the Australian Transport Safety Bureau (ATSB), Aviation Safety Reporting System (ASRS), Aviation Safety Network (ASN), and National Transport Safety Board (NTSB). For our study, we specifically utilize the aviation incident and accident investigation reports from NTSB and ASN. The NTSB dataset, encompassing reports from the year 2000 to 2020, was obtained in JSON format, yielding a comprehensive dataset of 24,137 records after cleansing. The data is available from: https://www.ntsb.gov/Pages/AviationQuery.aspx. Likewise, the ASN dataset, spanning the same time frame, was collected and pre-processed, ultimately resulting in a dataset comprising 4,526 records. The data is also publicly available from: https://aviation-safety.net/database/. From each report in both datasets, we extract the 'analysis Narrative' and 'damage level' fields for the training and validation of our deep learning models on four classes destroyed, substantial, minor, and none.

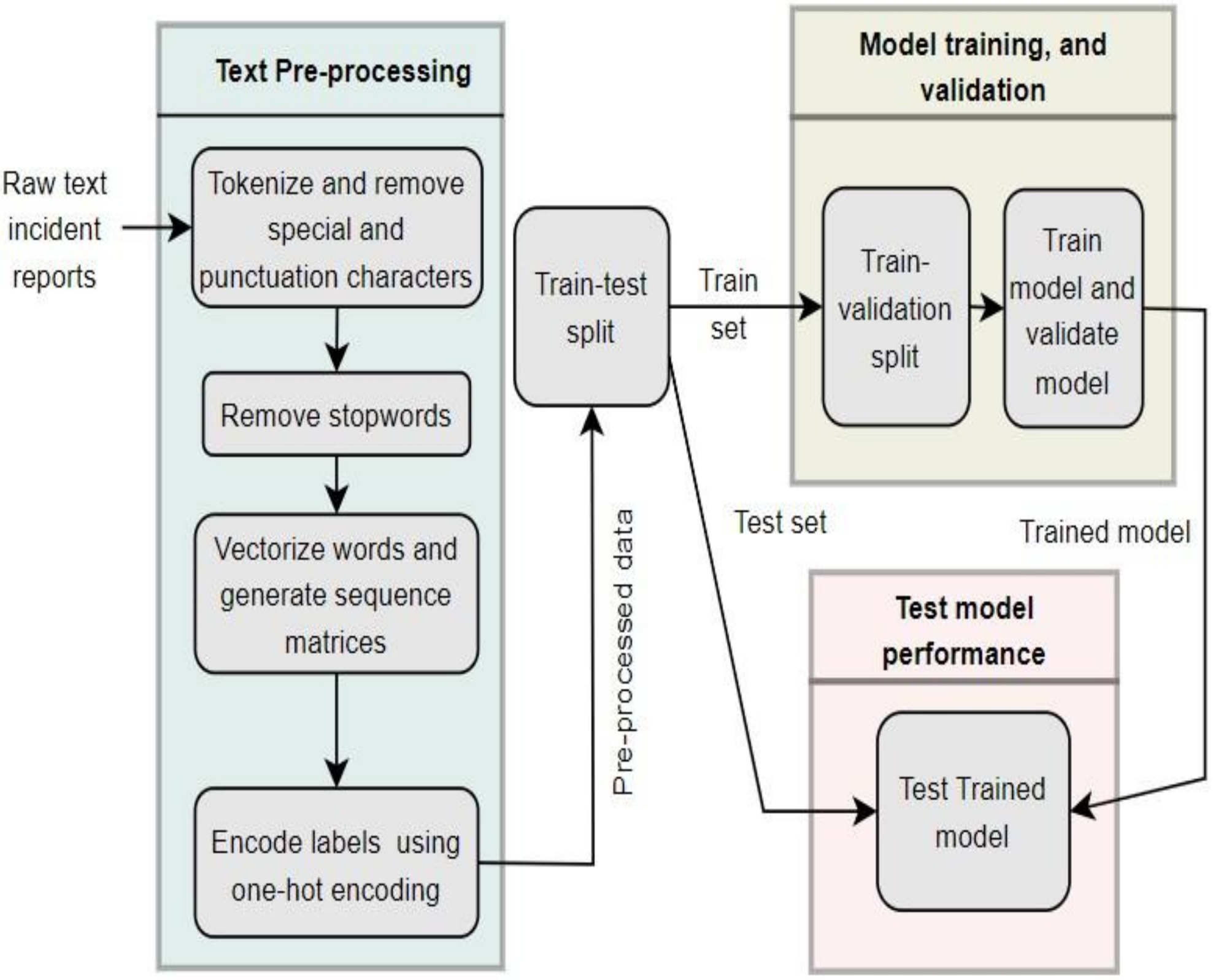

Figure 72. Methodological framework, adapted from [188].

*12.4.2 Text Processing*

Machine learning models inherently lack the ability to comprehend raw textual data; hence, text preprocessing is a pivotal step to ensure data quality and model performance. Our text preprocessing pipeline encompasses various stages, including lowercasing, punctuation removal, tokenization, stop word removal, and lemmatization. The cleaned narratives are then poised for feature extraction, a crucial step in transforming textual data into numerical features suitable for machine learning models. In this regard, we employ two distinct techniques: Term Frequency-Inverse Document Frequency (TF-IDF) and word embeddings, such as Word2Vec. These techniques effectively capture the semantic meaning of words, augmenting the models' comprehension of the narratives.

We then harness Kera's deep learning library for its extensive repertoire of deep learning models and preprocessing modules designed for text. Specifically, the Tokenizer module is instrumental in generating tokens and sequence vectors from the input text. The categorical module is then used to map unique categorical entries of the damage level variable (destroyed, substantial, minor, and none) to corresponding numerical values or as one-hot encoded entries for each data instance. To handle unwanted special and punctuation characters, stop-word removal, and word lemmatization, the spacy library is employed. Spacy is a Python library tailored for text-processing tasks and encompasses a comprehensive list of special characters, punctuation, and stop words. It serves as a robust tool for maintaining data quality and consistency.

Subsequently, each input text narrative undergoes processing, culminating in a representative sequence or vector, standardized at a length of 2000. Sequences derived from narratives with fewer than 2,000 words are zero-padded, while those exceeding 2,000 words are truncated. The vocabulary size of the corpus is set at 100,000. To partition the dataset into training, validation, and testing subsets, we leverage the traintest-split module from scikit-learn. The entirety of our experiments is conducted in Python, with Jupyter Notebook serving as the code editor.

#### *12.4.3 Text Classification, Training, and Validation*

With the preprocessed data in hand, the dataset is randomly split into an 80% training set and a 20% test set. Within the training process, an additional 10% of the training set is reserved for model validation in each epoch. Four deep learning models: Long Short-Term Memory (LSTM), Gated Recurrent Unit (GRU), Convolutional Neural Network (CNN), and Simple Recurrent Neural Network (sRNN) comprise our ensemble of models. These models are meticulously trained using datasets of varying sizes derived from the NTSB dataset, ranging from 4,526 to 24,137 instances, and the ASN dataset containing 4,526 instances.

Our primary training objective is to minimize the categorical cross-entropy loss function, a suitable choice for multiclass classification problems, employing the Adam optimization algorithm. It's essential to note that our study does not concentrate on identifying the best optimizer, and alternative optimization techniques can be explored as per specific preferences. To counteract overfitting, we implement early stopping based on validation loss, thereby terminating training when validation loss ceases to improve over a defined number of epochs.

### *12.4.4 Neural Network Model Architecture*

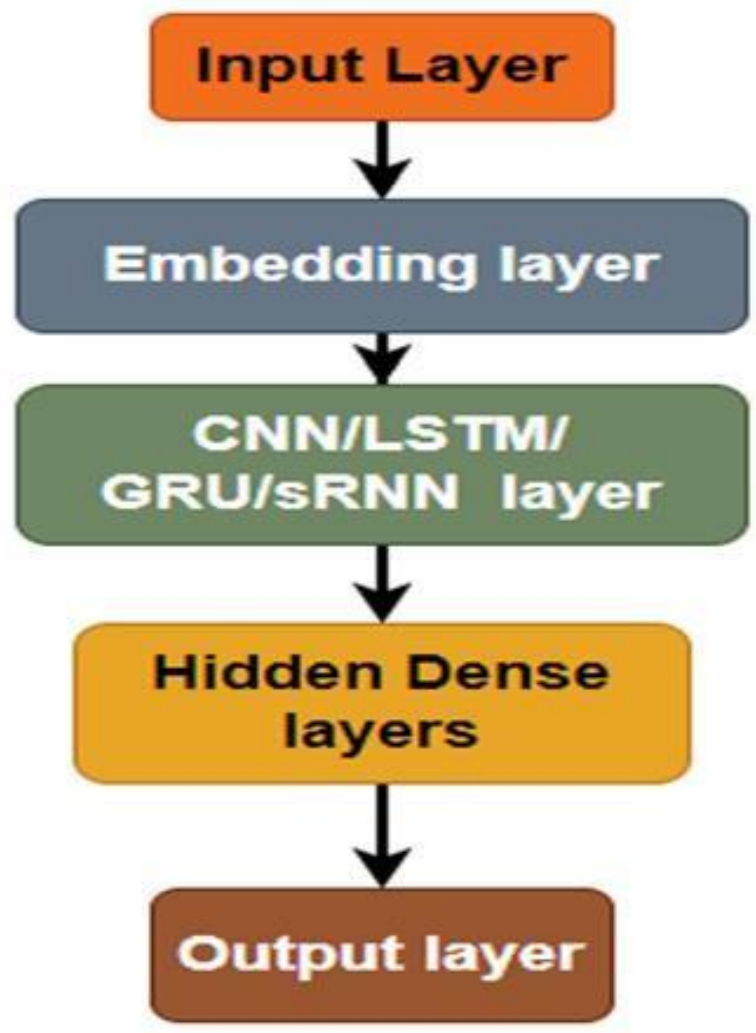

Figure 73. Neural Network architecture

To ensure uniformity and fair comparison, we employ a shared architecture across all models. This common architectural framework consists of an embedding layer, hidden layers, and an output layer. Rectified Linear Unit (ReLU) activation functions are applied to all hidden layers, while the SoftMax activation function is adopted for the output layer. Predicted class labels are determined using the argmax function, which identifies the index associated with the highest probability in the SoftMax output. Figure 73 provides a visual representation of the deep learning architectures employed in our study.

## 12.5 RESULTS AND DISCUSSION

In this section, we delve into the empirical results of our study, which centers on examining the influence of dataset size on the performance of machine learning models for aviation safety incident prediction. We conducted a thorough comparison of four neural network architectures, namely Long Short-Term Memory (LSTM), Gated Recurrent Unit (GRU), Convolutional Neural Network (CNN), and Recurrent Neural Network (RNN), across datasets of different sizes. Our particular focus is on contrasting the Aviation Safety Network (ASN) dataset, fixed at 4,526 instances, with the National Transport Safety Board (NTSB) dataset, which ranges from 4,526 to 24,137 instances.

### *12.5.1 Performance Evaluation of Neural Network Models.*

We begin our exploration by presenting and comparing the performance of the four selected models in terms of accuracy, precision, recall, and F1 measure. Figures 74 and 75 provide a visual representation of the validation accuracy achieved by these models. As depicted, CNN exhibits a pronounced ability to discern intricate patterns among various aircraft damage levels. Furthermore, when evaluating model performance on unseen test set samples, the results in Table 26 clearly indicate CNN's superiority over the other models across all performance evaluation metrics.

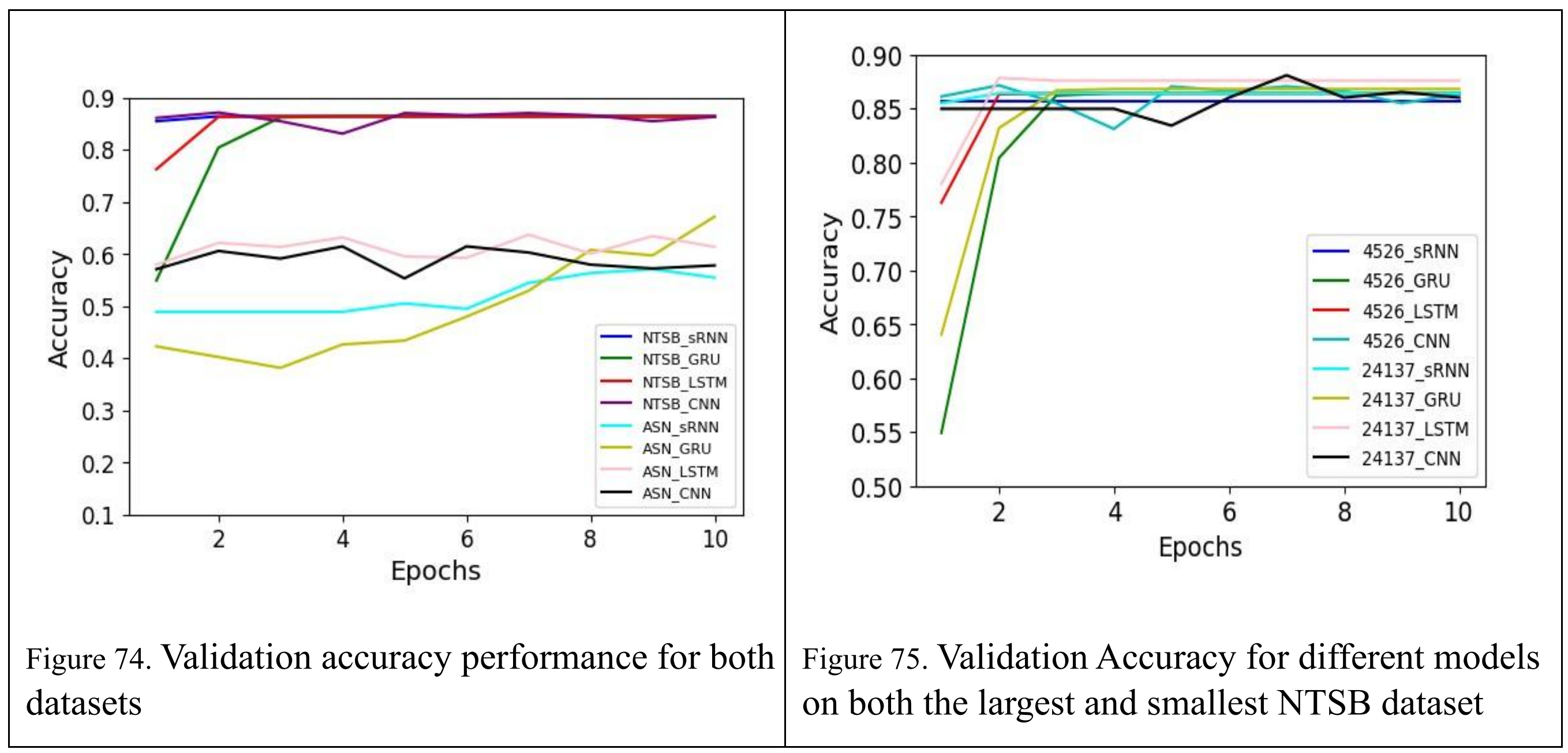

Figure 74. Validation accuracy performance for both datasets

Figure 75. Validation Accuracy for different models on both the largest and smallest NTSB dataset

*12.5.2 Confusion Matrices Analysis*

To gain deeper insights into the classification performance of the CNN model, we visualize the confusion matrices for both the NTSB and ASN datasets of the same size as seen in Figure 76. These matrices offer a comprehensive view of the model's strengths and areas requiring improvement. While the model demonstrates proficiency in certain incident categories, the matrices also highlight specific instances where it faces challenges, emphasizing the necessity for further refinement to enhance accuracy in aviation safety incident prediction.

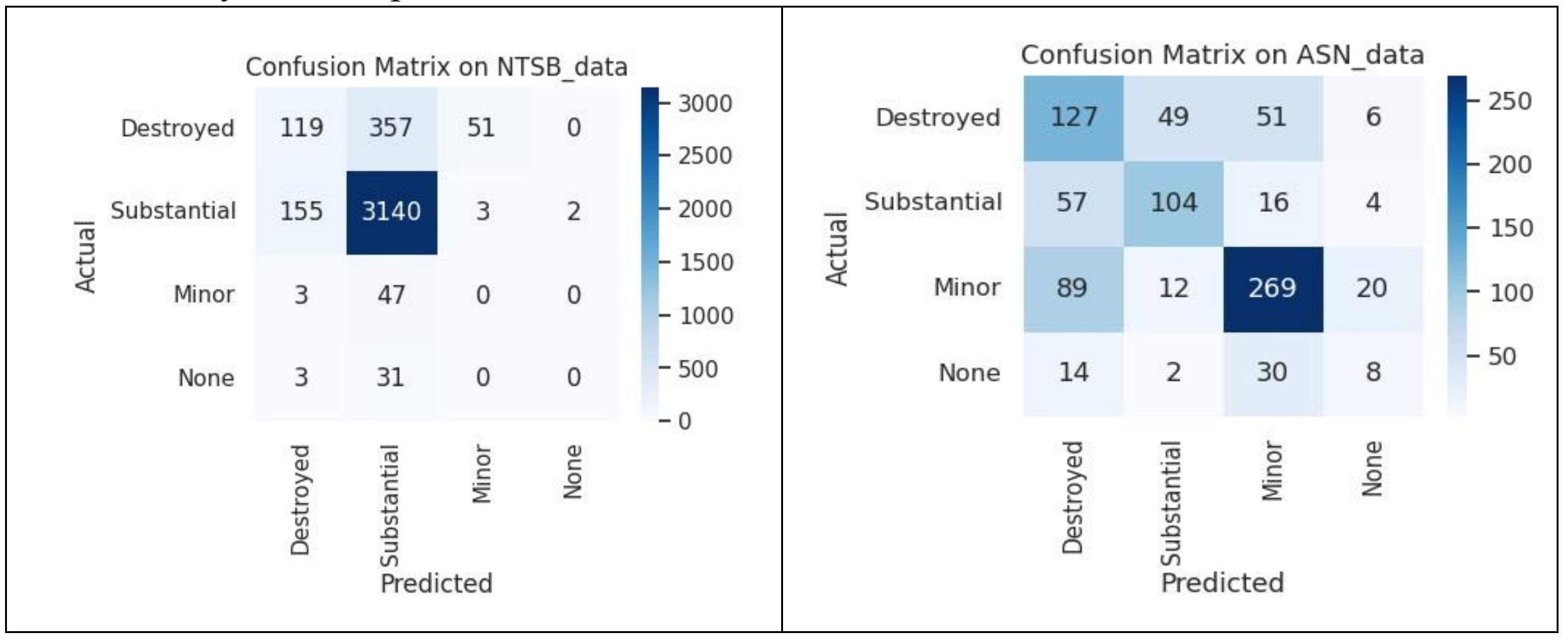

Figure 76. Confusion Matrices for CNN Model

*12.5.3 Performance Across Different Dataset Sizes*

Our investigation extends to analyzing the impact of dataset size on model performance. Table 26 summarizes the key performance metrics for each model architecture across various dataset sizes, providing a comprehensive view of their predictive capabilities.

Results show intriguing patterns as dataset sizes vary within the NTSB dataset. In the largest dataset (24,137 instances), all models exhibit consistent performance, with CNN slightly edging in accuracy and F1 score. However, as the dataset size decreases, particularly in the smallest NTSB dataset (4,526 instances), notable variations emerge. Specifically, the GRU model experiences a significant dip in performance, while CNN maintains a robust 0.84 accuracy. Conversely, LSTM and sRNN models maintain stable performance across varying dataset sizes, demonstrating their adaptability to different data volumes. Remarkably, when comparing the ASN dataset (also 4,526 instances) with the smallest NTSB dataset, we observe distinct model performance. LSTM, GRU, and CNN exhibit moderate accuracy in the ASN dataset, but their precision and recall are relatively lower. This discrepancy highlights the unique characteristics of the ASN dataset compared to the NTSB datasets.

Table 26. PERFORMANCE METRICS FOR DIFFERENT DATASET SIZES

| **Dataset Size** | **Model** | **Precision** | **Recall** | **F1_score** | **Accuracy** |
|---|---|---|---|---|---|
| | LSTM | 0.74 | 0.86 | 0.81 | 0.86 |
| 24,137 | GRU | 0.74 | 0.86 | 0.80 | 0.86 |
| | CNN | **0.85** | 0.86 | **0.85** | **0.87** |
| | sRNN | 0.74 | 0.87 | 0.81 | 0.86 |
| 18,104 | LSTM | 0.74 | 0.86 | 0.79 | 0.86 |
| | GRU | 0.75 | 0.86 | 0.80 | 0.86 |
| | CNN | **0.83** | **0.85** | **0.83** | **0.86** |
| | sRNN | 0.76 | 0.86 | 0.80 | 0.87 |
| 13,578 | LSTM | 0.74 | 0.86 | 0.80 | 0.86 |
| | GRU | 0.74 | 0.86 | 0.80 | 0.86 |
| | CNN | **0.84** | **0.87** | **0.85** | **0.86** |
| | sRNN | 0.76 | 0.87 | 0.81 | 0.87 |
| 9,052 | LSTM | 0.73 | 0.85 | 0.79 | 0.85 |
| | GRU | 0.76 | 0.87 | 0.81 | 0.869 |
| | CNN | **0.81** | **0.84** | **0.82** | 0.84 |
| | sRNN | 0.73 | 0.85 | 0.79 | 0.85 |
| 4,526 | LSTM | 0.74 | 0.86 | 0.80 | 0.86 |
| | GRU | 0.70 | 0.84 | 0.76 | 0.84 |
| | CNN | **0.80** | **0.82** | **0.82** | 0.83 |
| | sRNN | 0.73 | 0.86 | 0.79 | 0.81 |
| ASN_data (4,526) | LSTM | **0.63** | **0.62** | **0.62** | **0.62** |
| | GRU | 0.40 | 0.41 | 0.39 | 0.49 |
| | CNN | 0.60 | 0.59 | 0.59 | 0.60 |
| | sRNN | 0.51 | 0.58 | 0.52 | 0.58 |

These findings underscore the intricate relationship between dataset size and model performance in the aviation safety context. They hold significant implications for predictive system deployment in the aviation industry, emphasizing the need for adaptable models capable of effectively accommodating varying dataset sizes. As aviation safety standards continue to evolve, the insights from this research offer valuable guidance for stakeholders striving to enhance safety protocols and prevent incidents. Moreover, they contribute to the broader field of machine learning, providing empirical evidence on the influence of dataset size across different neural network architectures.

### 12.6 Conclusion

Safety stands as a cornerstone of the aviation industry. Incidents and accident reports in aviation, which include significant textual narratives, provide essential insights into events that compromise safety. Our study has investigated the intricacies of predictive model performance in the realm of aviation safety incident prediction, shedding light on critical factors. The analysis has uncovered the influence of dataset size, neural network architectures, and natural language processing (NLP) techniques on predictive models' accuracy and robustness.

Our findings confirm that dataset size wields a significant impact on model performance. Larger datasets yield higher accuracy and effectiveness. Notably, the smallest National Transport Safety Board (NTSB) dataset outperforms the Aviation Safety Network (ASN) dataset, underscoring the advantage of comprehensive data in aviation safety analysis. This research has significant implications for the aviation industry, advocating for investments in data collection and access to expansive datasets to bolster safety incident prediction.

Furthermore, our findings reinforce the importance of embracing advanced NLP techniques to harness the full potential of textual data. Future work will explore additional avenues for improvement, encompassing variations in training epochs, parameter fine-tuning, and the integration of state-of-the-art BERT models. By systematically evaluating these factors, we aspire to elevate model performance, further contributing to the safety of aviation and the effectiveness of safety protocols within the industry.

In summation, this study contributes to the field of aviation safety within the broader machine-learning community. It provides confirmation that consideration of dataset size and the deployment of advanced techniques need to be considered for innovation in aviation safety. Future work will look at the impact of combining the ASN and NTSB datasets. The modest performance of the ASN dataset highlights the need for improvements in data collection processes, suggesting a path towards a more robust and comprehensive dataset that can propel aviation safety to new heights.

# Chapter 13. Classification of Operational Records in Aviation Using Deep Learning Approaches.

[388], Nanyonga, A., & Wild, G. (2025, February). Classification of operational records in aviation using deep learning approaches. In *2025 International Conference on Pervasive Computational Technologies (ICPCT)* (pp. 997-1002). IEEE.

## *13.1* Abstract

Ensuring safety in the aviation industry is critical, even minor anomalies can lead to severe consequences. This study evaluates the performance of four different models for DP (deep learning), including Bidirectional Long Short-Term Memory (BLSTM), Convolutional Neural Networks (CNN), Long Short-Term Memory (LSTM), and Simple Recurrent Neural Networks (sRNN), on a multi-class classification task involving Commercial, Military, and Private categories using the Socrata aviation dataset of 4,864 records. The models were assessed using a classification report, confusion matrix analysis, accuracy metrics, validation loss and accuracy curves. Among the models, BLSTM achieved the highest overall accuracy of 72%, demonstrating superior performance in stability and balanced classification, while LSTM followed closely with 71%, excelling in recall for the Commercial class. CNN and sRNN exhibited 67% and 69% lower accuracies, with significant misclassifications in the Private class. While the results highlight the strengths of BLSTM and LSTM in handling sequential dependencies and complex classification tasks, all models faced challenges with class imbalance, particularly in predicting the Military and Private categories. Addressing these limitations through data augmentation, advanced feature engineering, and ensemble learning techniques could enhance classification accuracy and robustness. This study underscores the importance of selecting appropriate architectures for domainspecific tasks and contributes to advancing deep learning applications in multi-class classification problems

## 13.2 Introduction

The aviation industry generates vast amounts of operational data through daily flight operations, including maintenance logs, flight operation summaries, incident reports, and other safety-related documentation [166]. These records are essential for identifying trends, improving operational efficiency, and ensuring aviation safety across various sectors, including military, private, and commercial aviation [168]. However, with the increasing volume and diversity, manual classification and analysis methods have become impractical and time-consuming [63, 171]. Furthermore, the unstructured and diverse nature of operational narratives poses significant challenges for efficient classification and analysis. Automating these processes is essential for identifying safety risks, operational anomalies, and issues that may compromise aviation safety [116, 172]. Misclassifications in this context could delay responses to safety concerns or lead to inaccurate trend analysis, ultimately jeopardizing operational efficiency and safety [354].

Traditional approaches, such as rule-based systems and manual reviews, often fail to fully capture the nuanced, context-dependent information embedded in text-based records. With the exponential growth of data, their limitations have become more pronounced [328]. Modern Deep Learning (DL) and Natural Language Processing (NLP) techniques offer a viable alternative, having proven highly effective in automating complex classification tasks in various domains [101]. DL models such as Bidirectional LSTMs (BLSTMs), Convolutional Neural Networks (CNNs), Long Short-Term Memory (LSTM), and Recurrent Neural Networks (RNNs), networks, have shown remarkable performance in sequential data tasks like text classification, sentiment analysis, and speech recognition [331{Mikolov, 2010 #642{Mikolov, 2010 #642]}.

Aviation operational records require models capable of understanding context, sequence, and long-range dependencies. RNNs [325], designed for sequential data, are commonly used in text classification tasks [160, 355] but face challenges with long-term dependencies caused by "exploding or vanishing gradients". LSTM network [324] address these limitations through mechanisms that retain and manage information over longer sequences. BLSTM further enhances performance by capturing context from both forward and backward directions, making them particularly effective for tasks requiring sequence sensitivity. Additionally, CNNs, traditionally applied to image processing, have shown strong capabilities in text classification by capturing local features and patterns in textual data [122].

Despite these advances, the application of DL to aviation safety records has been relatively underexplored. Most research in this domain focuses on traditional rule-based methods, leaving the potential of DL largely untapped. This study aims to fill this gap by leveraging CNN, RNN, LSTM, and BLSTM models to classify aviation operational records from the Socrata dataset into three categories: military, private, and commercial sectors. The Socrata dataset, with its extensive collection of public aviation records, offers a valuable resource for evaluating these models. By applying these advanced models, this research explores how DL can enhance the automation and accuracy of classifying operational records, contributing to improved aviation safety. Accurate classification facilitates efficient risk assessment, regulatory compliance, and the timely identification of safety concerns, all of which are crucial for reducing incidents and enhancing safety standards in aviation.

This study provides a novel application of DL for classifying aviation operational records and contributes to the growing body of research on NLP and DL in aviation safety. The findings underscore the practical benefits of DL in this field and offer insights into improving automated systems for operational data classification across various aviation sectors.

The structure of this paper is as follows: the next section reviews related work, highlighting existing approaches and advancements in the domain, followed by a detailed methodology section that outlines the data preprocessing steps, model architecture, and evaluation metrics employed in this study. The results are then presented, followed by a discussion of the findings, providing an in-depth analysis of model performance and their implications. The paper concludes with a conclusion and directions for future research aimed at improving the classification of aviation operational records.

## 13.3 Related Work

The classification of operational records in aviation, particularly those related to safety, has garnered significant interest among researchers over the years. Traditionally, aviation safety data such as incident reports, maintenance logs, and flight operation summaries have been analyzed using rule-based systems, expert systems, and manual review methods. While these approaches have been effective to some extent, they are increasingly criticized for being time-consuming and incapable of addressing the complexities of the growing volume and diversity of operational records [169].

Historically, rule-based classification techniques have been employed to analyze aviation incident reports. These methods depend on predefined rules, such as keyword matching and heuristic approaches, to categorize records into specific classes. For instance, researchers have used decision trees (DT) and Naive Bayes (NB) [120{Bhanbhro, 2020 #490{Bhanbhro, 2020 #490]}, as well as random forest (RF), support vector machines (SVM), and NN [56], to be able to classify incident reports based on predefined features. However, the inherent limitation of these methods lies in their inability to understand the contextual nuances within text data, which is crucial for accurate classification.

Additionally, traditional machine learning algorithms such as Logistic Regression (LR) as well as Random Forests (RF) have been applied to predict the severity of aviation incidents. For example, Inan et al. [340] demonstrated the application of models such as LR, artificial neural networks (ANN), and DTs to classify aircraft damage based on various factors, including zones, weather, and incident severity. Although these models showed promise, they required extensive feature engineering and struggled to identify complex patterns within unstructured text data.

With advancements in DL, researchers have increasingly adopted more sophisticated methods to address the limitations of traditional approaches. CNN, RNN, LSTM and BLSTM [188] models have proven particularly effective for handling aviation safety data. These models excel in capturing complex patterns and dependencies, making them ideal for tasks like classifying aviation safety reports, which often involve textual or sequential information [170].

Several studies have highlighted the utility of DL in aviation safety analysis. For instance, Zhang et al., applied LSTM networks to classify aviation safety reports, leveraging the sequential nature of the data to improve prediction accuracy. Their results indicated that DL models outperformed traditional machine learning algorithms in classifying incident severity and conducting risk assessments [313].

Another significant study focused on classifying flight phases in safety reports from the Australian Transport Safety Bureau (ATSB). This study employed NLP and DL models, including CNN, LSTM, BLSTM, and simple RNN. The models were evaluated based on usual metrics; specifically, their accuracy, precision, recall, and F1 score. Among these, LSTM achieved the highest performance, with accuracy, precision, recall, and F1 scores of 87%, 88%, 87%, and 88%, respectively. These findings underscore the effectiveness of DL models in automating the analysis of safety occurrences, enabling more precise safety measures and efficient report handling [188].

Recent efforts have also explored combining DL models for improved performance [159, 189]. These studies used DL to try and classify the phase of flight in which an aviation safety occurrence was reported, by utilizing the information provided in the associated text narratives. These efforts demonstrated almost 90% accuracy.

Despite these advances, gaps remain in the literature regarding comparative analyses of DL models for aviation safety classification tasks. While numerous studies have explored CNN, LSTM, BLSTM, and RNN models on various datasets, there is limited research on direct comparisons of their effectiveness in classifying operational records across sectors such as military, private, and commercial aviation. Furthermore, the application of DL to the Socrata dataset for aviation operational record classification is largely unexplored.

This study addresses these gaps by evaluating and comparing the performance of CNN, RNN, LSTM, and BLSTM models in classifying aviation operational records. Leveraging the Socrata dataset, this research aims to analyze each model's strengths and limitations. The findings will offer insights into the potential of DL for improving aviation safety analysis, operational efficiency, and data classification

## 13.4 Methodology

The methodology employed in this study was to classify operator types (military, private, and commercial) using narratives from the Socrata aviation dataset. The methodology is divided into several key subsections: Data Annotation and Preprocessing, DL Model Architecture, and Model Performance Evaluation.

### *13.4.1 Data Annotation and Preprocessing*

The Socrata aviation dataset, which is publicly available online, serves as the primary data source for this study. The original dataset contained 4,995 records, with each record consisting of an operator type and a narrative describing aviation incidents. After cleaning the data, the dataset was reduced to 4,863 records. The focus of this study was on the *Operator* and narratives labelled as *Summary* fields. The Operator field represents the entity responsible for the operation of the aircraft, while the Narrative field provides a descriptive account of the incident.

The operators in the dataset were originally labelled with hundreds of categories. To streamline the classification process and ensure meaningful analysis, these categories were manually annotated and grouped into three primary classes: *Military*, *Private*, and *Commercial*. This annotation was performed by cross-checking each operator category with additional data sources and domain knowledge. This manual verification process helped to ensure that the categories were consistent and representative of the operators in the dataset. The categorization of operators into these three distinct classes allowed for a focused and meaningful classification task, ensuring that the models trained on this data could accurately predict operator types based on the narrative content.

### 13.4.2 Text Processing

Text preprocessing is needed to transform the raw text data from the reports into a format suitable for use in training the models. The text preprocessing pipeline in this study involved several stages, including text cleansing, tokenization, and word count distribution analysis.

The first step in text preprocessing was text cleansing, where irrelevant elements such as stop words (e.g., “the”, “and” “is”), punctuation and special characters were removed. This was done using popular NLP libraries, such as *NLTK* and *SpaCy*. The removal of stop words is particularly important because these words do not add significant meaning to the classification task but can introduce noise into the model, reducing its performance [356].

Following text cleansing, tokenization was performed using the Keras Tokenizer. Tokenization involves splitting each narrative into individual words or tokens and assigning each token a unique integer ID. This step is essential for converting text into a numerical format that DL models can process. The tokenized text was then padded to a fixed length of 200 words, ensuring that all input sequences had the same length. Narratives shorter than 200 words were padded with zeros, while longer ones were truncated to maintain consistency in sequence length across the dataset.

An analysis of the word count distribution within the Socrata dataset revealed that the narratives varied greatly in length. While some narratives were quite short, others contained hundreds of words. As seen in Fig. 1, this variance prompted the decision to standardize the input sequence length for the DL models. By ensuring that the models received consistent input, the approach aimed to optimize model performance and avoid issues associated with input length inconsistency.

Finally, the dataset was split into three parts for training, testing, and validation. The data splitting utilised the standard 80-10-10 ratio, where 80% of the data being utilised for model training, 10% for testing, and the remaining 10% for validation. This split is needed to be able to evaluate model performance utilizing unseen data helping to prevent overfitting the model based only on the training data.

### 13.4.3 DL Model Architecture

To ensure consistency and enable effective comparisons across all models, a unified architecture was used as the base, with slight modifications tailored to each model. The architecture consisted of three main components; the first is the embedding layer, then the hidden layers, and the output layer. Within hidden layers, activation utilized a Rectified Linear Unit (ReLU) function, which allowed the model to capture any complex, non-linear relationships. Within the output layer, the activation function utilized was SoftMax, enabling its use for multi-class classification. This approach allowed the model to output a probability distribution across the possible classes. The resultant class prediction utilized an argmax function to identify corresponding the index with the highest probability from the output of SoftMax function. A diagram illustrating the DL architecture used in this study is provided in Figure 77.

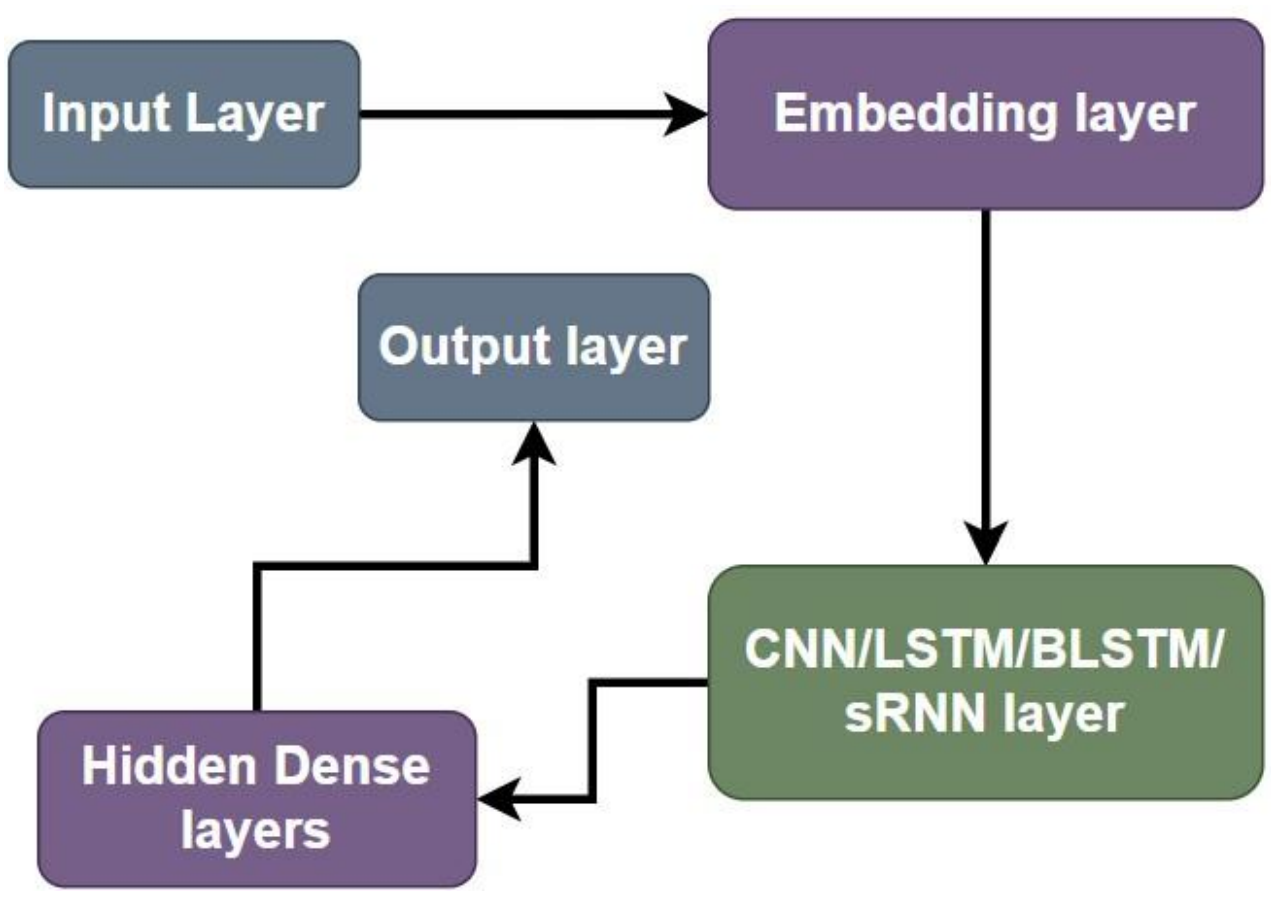

Figure 77. DL architecture

a) sRNNs.

The Simple RNN (sRNN) is a basic form of RNN designed to process sequential data by using the output from the previous time step as input for the current time step. Its architecture is straightforward, typically comprising a single hidden layer that enables the flow of information across successive time steps [325]. The mathematical formulation for an sRNN can be expressed as follows:

$$h_t = \sigma(W_h \cdot [h_{t-1}, x_t] + b_h),$$

where $h_t$ donates the hidden state at the time step $t$, while $x_t$ is the input at the corresponding time step, and $\sigma$ is the activation function used.

b) LSTM

The LSTM network is an advanced variant of the RNN that was designed to overcome the challenge of capturing long-term dependencies that may be present within sequential data. In contrast to basic RNNs, LSTMs are equipped with memory cells that allow for the storage and management of information over extended periods; this makes them particularly effective when tasks need an understanding of context and temporal patterns. The LSTM architecture is more intricate than that of a standard RNN, incorporating three gates, the forget, input, and output gates [324]. These gates collectively manage information flow, determining what data should be retained and what should be discarded.

c) CNN

The CNN is a DL model primarily used for image processing but is also effective in sequential data tasks like text classification. The CNN is made up of convolutional layers that utilize filters to analyze input data, detecting localized features and patterns. In text processing, CNNs help identify important wordlevel features, such as n-grams, which are crucial for understanding context [357].

d) BLSTM

The BLSTM model extends the traditional LSTM architecture by processing input sequences in both forward and backward directions. This bidirectional approach enables the model to capture information from both past and future contexts, thereby enhancing its ability to comprehend dependencies across the entire sequence. Such a feature makes BLSTM particularly effective in tasks requiring a holistic understanding of sequence context. Mathematically, the BLSTM model combines the operations of both forward and backward LSTMs, allowing for a more robust capture of long-term dependencies [22]. BLSTM is widely applied in domains such as speech recognition, language translation, and named entity recognition, where the integration of both past and future information is crucial for precise prediction and analysis

#### *13.4.4 Performance Evaluation*

This section elucidates the evaluation criteria utilized in this study to assess the models' performance. The primary focus of this research is multi-class classification, and as such, performance was gauged based on the accuracy of predictions across various classes. To comprehensively evaluate model performance, we employed a suite of standard prediction performance metrics; specifically, recall, F1-score, precision, and accuracy.

## 13.5 Results and Discussion

The evaluation of the four models CNN, sRNN, BLSTM, and LSTM on the given dataset, incorporating both training and validation metrics, is presented. The performance of each model is assessed through classification reports, confusion matrices, accuracy, macro averages, as well as validation loss and accuracy

#### *13.5.1 Model Performance: Classification Report*

The classification report (figure 78) evaluates the models' precision, recall, and F1 scores across three classes: Commercial, Military, and Private. Among the models, CNN excelled in classifying the Commercial class, achieving 81% precision and 79% recall. However, its performance dropped significantly for the Military and Private classes, with much lower recall and precision values. Similarly, the sRNN model achieved the highest recall (85%) for the Commercial class but failed to classify the Private class altogether, resulting in zero precision, recall, and F1-score for this category.

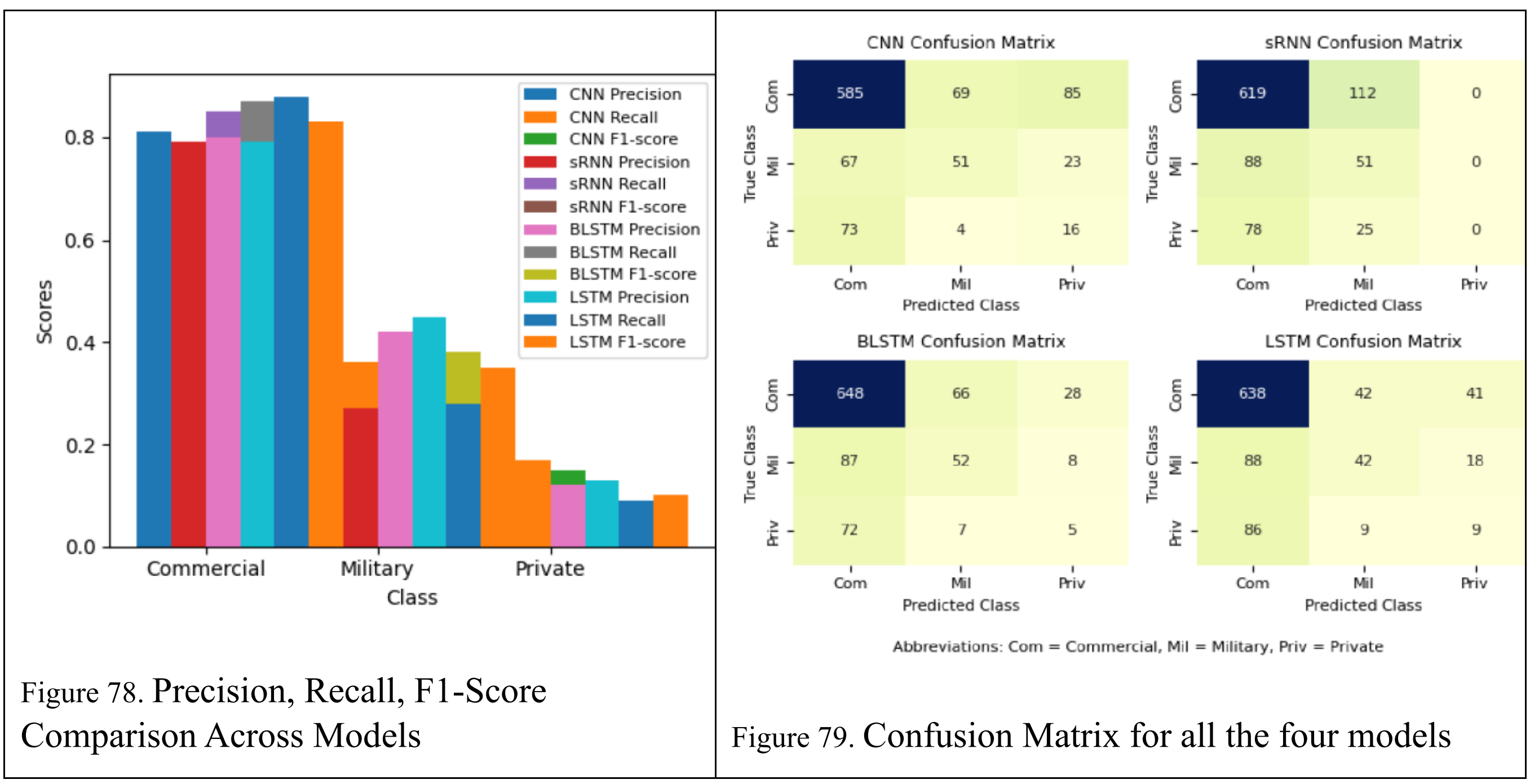

Figure 78. Precision, Recall, F1-Score Comparison Across Models

Figure 79. Confusion Matrix for all the four models

### *13.5.2 Confusion Matrix Analysis*

The confusion matrices (Figure 79) provide insights into the class-wise predictions and misclassifications. The CNN model demonstrated strong predictive accuracy for the Commercial class but showed higher misclassifications for the Military and Private classes, reflecting its challenges in distinguishing these categories. The sRNN model's confusion matrix highlights significant misclassification of the Private class, aligning with its low scores in the classification report.

On the other hand, BLSTM's confusion matrix showed relatively fewer misclassifications, particularly for the Commercial and Military classes, indicating its ability to capture complex dependencies within the data. LSTM also performed well for the Commercial class but exhibited a moderate level of misclassification for the Private class, performing better than sRNN but worse than BLSTM. These trends emphasize the models' varying capabilities to manage class imbalances and complexities in the dataset.

### *13.5.3 Model Accuracy Comparison*

The overall performance of each model, summarized in Table 27, highlights BLSTM's superior accuracy of 72%, closely followed by LSTM at 71%. These results demonstrate the ability of these models to correctly classify instances across classes. In contrast, sRNN and CNN achieved lower accuracies of 69% and 67%, respectively, reflecting their limitations in handling the Private and Military classes.

Notably, BLSTM and LSTM's ability to incorporate temporal dependencies likely contributed to their superior performance, as this approach enables better context understanding across sequences. CNN's comparatively lower accuracy suggests its limitations in capturing sequential relationships, while sRNN's architecture may have struggled with the dataset's complexity.

Table 27. DEEP LEARNING MODEL PERFORMANCE

| **Models** | Precision | Recall | F1 Score | Accuracy |
|---|---|---|---|---|
| **CNN** | 0.67 | 0.67 | 0.68 | 0.67 |
| **sRNN** | 0.69 | 0.69 | 0.66 | 0.69 |
| **LSTM** | **0.72** | **0.72** | **0.70** | **0.72** |
| **BLSTM** | 0.71 | 0.71 | 0.68 | 0.71 |

### *13.5.4 Macro Average Comparison*

Macro average precision, recall, and F1-scores, as shown in Figure 80, provide a balanced evaluation of the models across all classes, irrespective of class imbalances. BLSTM and LSTM outperformed CNN and sRNN in all three metrics, particularly excelling in recall for the Commercial class. However, all models showed reduced performance for the Private class, highlighting a consistent challenge in handling this category.

This underperformance could be attributed to class imbalance or insufficient distinguishing features within the dataset for the Private class. Addressing these challenges through techniques like data augmentation, oversampling, or feature engineering could improve overall performance.

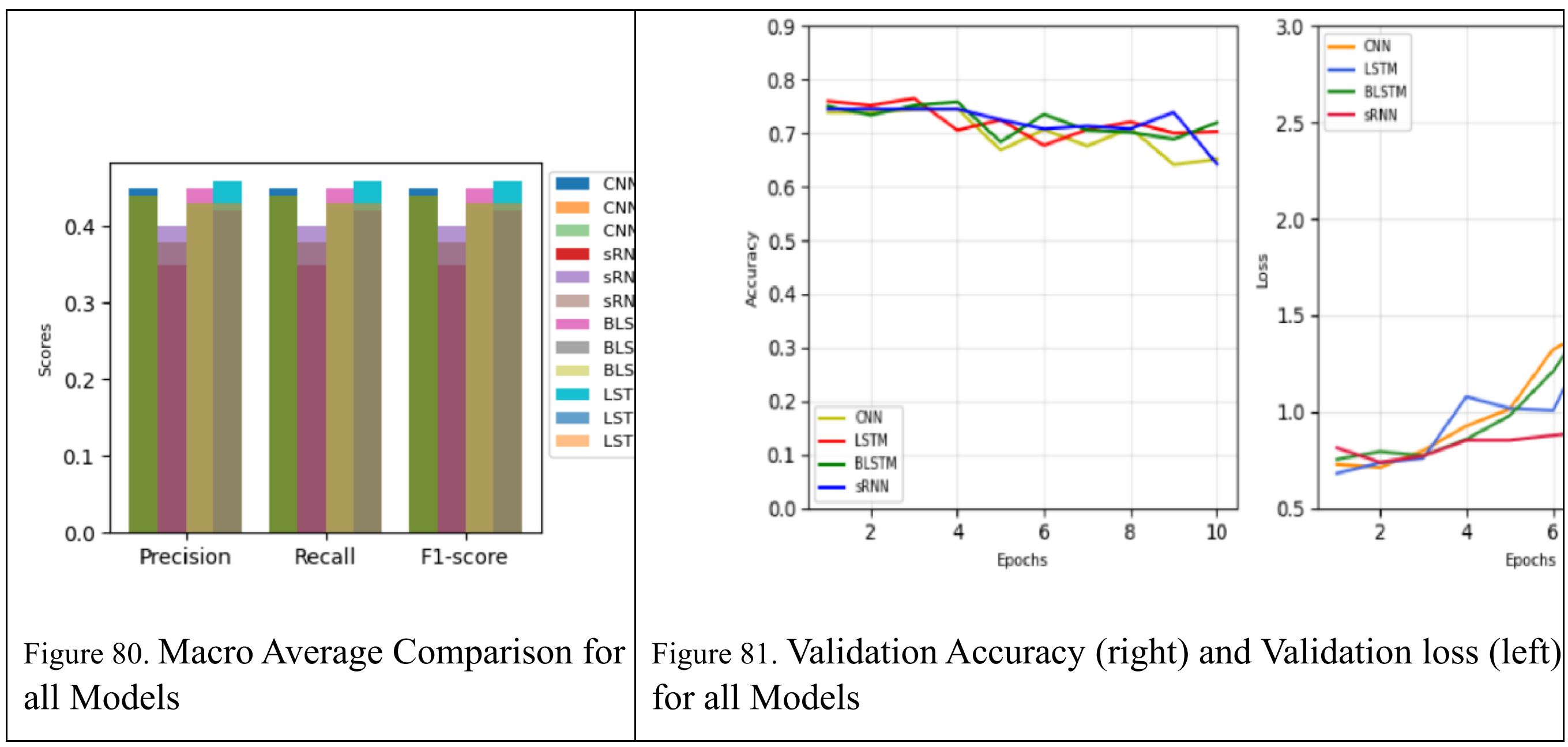

Figure 80. Macro Average Comparison for all Models

Figure 81. Validation Accuracy (right) and Validation loss (left) for all Models

#### *13.5.4 Validation metrics*

The validation loss and accuracy trends (Figure 81) reveal each model's generalization capabilities. BLSTM exhibited the lowest validation loss and the highest validation accuracy, reflecting its stable learning process and strong generalization to unseen data. LSTM followed closely, showing similar trends with slightly higher validation loss.

Conversely, CNN and sRNN showed higher validation losses, indicating weaker generalization and potential overfitting to the training data. These results underscore the importance of architectures that can effectively manage sequential dependencies, as demonstrated by BLSTM and LSTM.

## 13.6 CONCLUSION

This research employed four DL models CNN, sRNN, BLSTM, and LSTM to evaluate their performance in a multi-class classification task. The findings highlight the strengths and weaknesses of each of the models. CNN achieved an accuracy of 67%, displaying a balanced performance but facing challenges in classifying the Military and Private classes, as reflected in lower recall and F1 scores for these categories. sRNN performed slightly better with an accuracy of 69%, although it struggled particularly with the Private class, where recall and F1 scores were negligible.

BLSTM emerged as the most effective model, achieving the highest accuracy (72%) and demonstrating balanced classification across all classes, particularly excelling in recall for the Commercial class. LSTM followed closely with an accuracy of 71%, showing a strong recall for the Commercial class but encountering similar difficulties with the Military and Private classes. Validation loss and accuracy metrics further corroborated these results, with BLSTM and LSTM demonstrating lower validation loss and greater stability, indicating their superior generalization ability.

Despite the promising results, the study revealed key challenges, particularly in accurately classifying the Military and Private classes. These challenges likely stem from class imbalances and limited feature representation in the dataset. Future improvements could include the use of data augmentation and oversampling techniques to address class imbalance, the integration of attention mechanisms to enhance model focus on underrepresented categories, and the application of ensemble learning to leverage the strengths of multiple models. Additionally, domain-specific feature engineering could play a crucial role in improving classification performance for the minority classes.

In conclusion, while BLSTM proved to be the most effective model for this classification task, further refinements and advanced techniques are necessary to overcome the observed limitations, paving the way for more robust performance in multi-class classification problems.

# Section IV. Topic modelling

## Chapter 14. Topic Modeling Analysis of Aviation Accident Reports A Comparative Study between LDA and NMF Models

[390], Nanyonga, A., Wasswa, H., & Wild, G. (2023, December). Topic Modeling Analysis of Aviation Accident Reports: A Comparative Study between LDA and NMF Models. In *2023 3rd International Conference on Smart Generation Computing, Communication and Networking (SMART GENCON)* (pp. 1-2). IEEE.

### *14.1* Abstract

Aviation safety is paramount in the modern world, with a continuous commitment to reducing accidents and improving safety standards. Central to this endeavor is the analysis of aviation accident reports, rich textual resources that hold insights into the causes and contributing factors behind aviation mishaps. This paper compares two prominent topic modeling techniques, Latent Dirichlet Allocation (LDA) and Non-negative Matrix Factorization (NMF), in the context of aviation accident report analysis. The study leverages the National Transportation Safety Board (NTSB) Dataset with the primary objective of automating and streamlining the process of identifying latent themes and patterns within accident reports. The Coherence Value (C_v) metric was used to evaluate the quality of generated topics. LDA demonstrates higher topic coherence, indicating stronger semantic relevance among words within topics. At the same time, NMF excelled in producing distinct and granular topics, enabling a more focused analysis of specific aspects of aviation accidents.

## 14.2 Introduction

The aviation industry is a cornerstone of modern society, facilitating global travel, commerce, and cultural exchange. This sector continually advances its technology and safety protocols in pursuit of safer and more efficient operations [198, 341]. However, despite these efforts, aviation accidents continue to occur, prompting rigorous investigations to discern the root causes and develop preventive strategies. These investigations yield voluminous textual data in the form of accident reports, which harbour invaluable insights into enhancing aviation safety. Extracting meaningful information from these reports, given their volume and complexity, poses a considerable challenge. In this context, topic modelling emerges as a potential tool for analyzing aviation accident reports. Topic modelling, a natural language processing (NLP) technique, unveils latent thematic structures within a corpus of text documents [159, 170]. By systematically categorizing and exploring the content of accident reports, it contributes to a deeper understanding of the factors underlying aviation accidents.

The motivation behind studying aviation accident reports is grounded in the aviation industry's unwavering commitment to safety enhancement. This commitment has led to significant advancements in technology, training programs, and regulatory measures. However, accidents persist, often resulting from intricate interplays between human factors, technical failures, and environmental conditions. Timely and precise analysis of accident reports is imperative for preventing future incidents and elevating aviation safety measures.

This paper aims to demonstrate the effectiveness of topic modeling, particularly through the application of Latent Dirichlet Allocation (LDA) and Non-negative Matrix Factorization (NMF) models, in the analysis of aviation accident reports. Our objectives are threefold:

1. To identify and extract latent topics present in aviation accident reports, providing a structured and interpretable representation of their content thereby shedding light on potential areas for safety improvement.

2. To conduct a comparative analysis of LDA and NMF models, assessing their performance in uncovering meaningful topics from accident reports.

This study is of great significance as it has the potential to enhance aviation safety through data-driven insights. By systematically analyzing a substantial corpus of accident reports, we aim to contribute to identifying recurring themes and contributing factors, informing regulatory enhancements, guiding training programs, and driving innovation in aviation technology. Our research leverages the NTSB Aviation Dataset, encompassing comprehensive accident reports spanning years between 2000-2020, and employs rigorous methodologies for data preprocessing, model training, and evaluation based on topic coherence and interpretability metrics.

In the following sections of this paper, we will delve into the related work, and our methodology, present the experimental results, and engage in a comprehensive discussion of our findings. Furthermore, we will explore the implications of our research for the field of aviation safety and suggest potential avenues for future research.

## 14.3 RELATED WORK

First, In the realm of aviation safety analysis, the utilization of advanced techniques, such as topic modelling, is relatively novel but holds significant promise. Existing literature primarily focuses on traditional methods of accident investigation, expert analysis, and statistical approaches. However, a growing body of work acknowledges the potential of natural language processing (NLP) and machine learning techniques, including topic modelling, to glean insights from aviation accident reports [96, 150].

Historically, aviation accident investigations relied heavily on human expertise and manual examination of accident reports [329]. These methods, although valuable, are time-consuming and prone to human bias. Experts painstakingly sifted through textual narratives, findings, and recommendations to identify recurring patterns and contributing factors. While these approaches have yielded valuable insights, they are limited in their ability to handle the vast and complex aviation accident reports.

Recent developments in automated text analysis techniques have opened new avenues for aviation safety research. Researchers have recognized the potential of NLP and machine learning to extract actionable insights from textual data [114, 170]. Text mining and sentiment analysis have been applied to analyze reports such as online forums [358, 359]. These approaches offer the advantage of scalability and objectivity, reducing the human bias inherent in traditional analyses.

Topic modeling, a subset of NLP, has gained traction in aviation safety research. It offers a systematic way to uncover latent thematic structures within textual data, making it particularly suitable for aviation accident reports [131, 132]. [264] introduced Latent Dirichlet Allocation (LDA), a seminal topic modeling algorithm that has since become a cornerstone in this field. LDA has found applications in diverse textual datasets, ranging from news articles and social media content to scientific literature [360]. Its ability to unveil underlying topics and relationships has proven invaluable in numerous contexts. Also, [361] introduced Non-negative Matrix Factorization (NMF), a dimensionality reduction technique, that has found application in text mining and topic modeling. NMF has been employed in various studies to extract topics from text data, providing an alternative approach to LDA [360, 362].

A study by [131] employed LDA to extract topics from aviation safety reports and identified emerging safety concerns. Similarly, [96] explored topic modeling using LDA to categorize narratives in aviation accident reports, providing a structured representation of accident data. Also, [150] employed text-mining techniques to analyze aviation accident reports, focusing on identifying significant terms and phrases. Their study laid the foundation for applying computational methods to accident report analysis. The study in [336] introduced a framework that combined text mining and deep learning to automatically classify accident reports into categories based on accident prevention. Their approach showcased the potential for automating key aspects of accident analysis, offering efficiency and consistency.

In study [86] a compelling was presented on the application of structural topic modeling, specifically LDA, to aviation accident reports. The research demonstrated the feasibility of utilizing topic modeling to uncover latent themes within accident narratives. Moreover, it highlighted the potential for automating certain aspects of the analysis process, thereby increasing efficiency.

While individual studies have applied topic modeling to aviation safety, comparative studies between different topic modeling techniques, such as LDA and NMF, remain relatively scarce. Such comparative analyses offer valuable insights into the strengths and weaknesses of different approaches [361]. The adoption of advanced techniques, including topic modeling, in aviation safety research is steadily growing. Researchers recognize the potential of these methods to provide deeper insights into the causes and contributing factors of aviation accidents. Our study builds upon this foundation by specifically evaluating the performance of LDA and NMF models in extracting meaningful topics from aviation accident reports using the NTSB dataset, adding to the growing body of knowledge in this area.

## 14.4 METHODOLOGY

This section provides a detailed account of the methodology employed in this study, encompassing data collection, data preprocessing, the selection of topic modeling techniques (LDA and NMF), model training, and evaluation metrics.

### *14.4.1 Data Acquisition*

Aviation incident and accident investigation reports used in this study were exclusively sourced from the National Transport Safety Board (NTSB) dataset spanning years from 2000 to 2020. The dataset comprising a collection of more than 36,000 records in JSON format was obtained from the following source: https://www.ntsb.gov/Pages/AviationQuery.aspx. These reports encompass textual narratives, findings, and recommendations from NTSB investigations, making them an invaluable resource for the study.

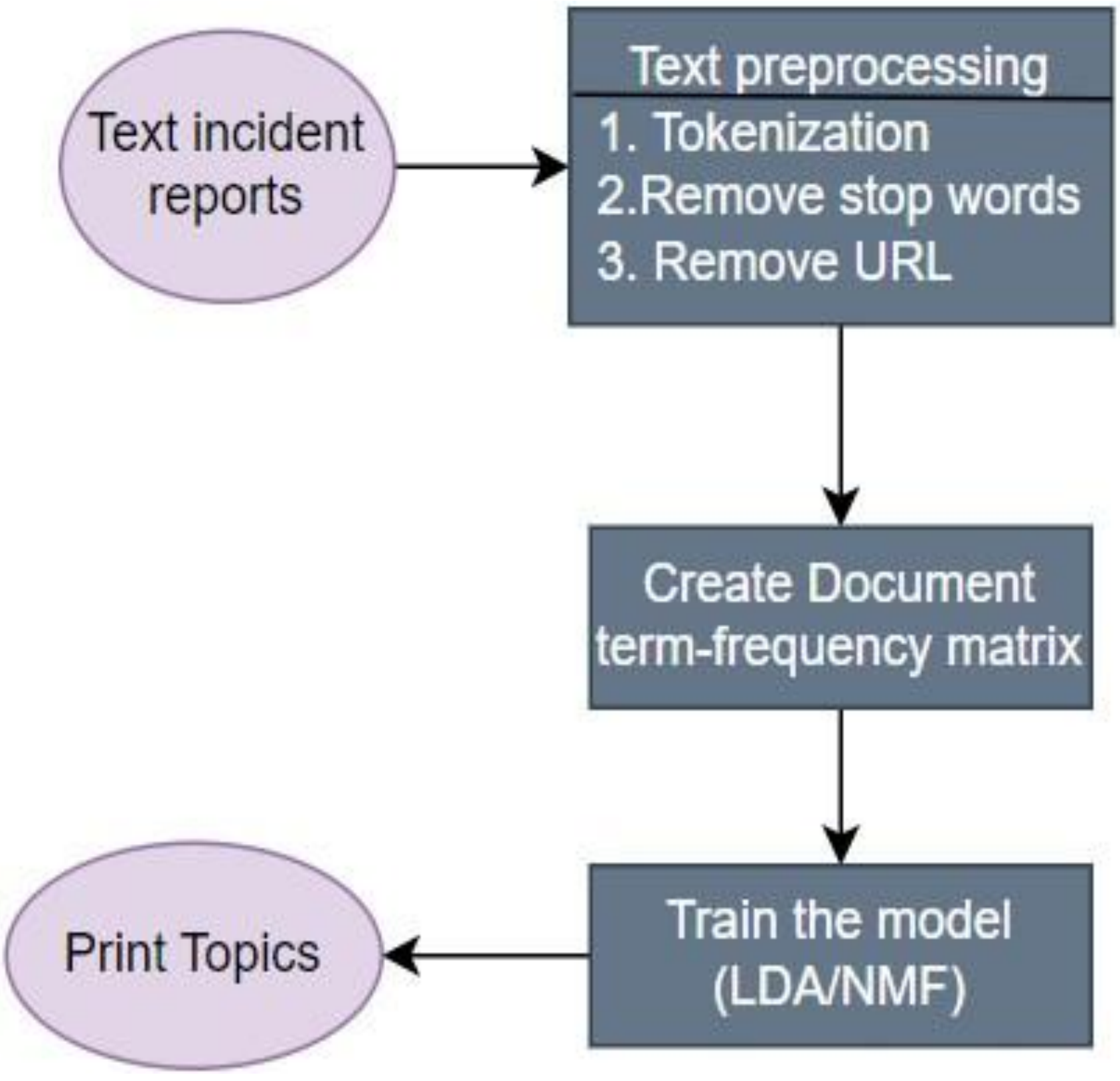

Figure 82. Methodological framework

*14.4.2 Text Processing*

Machine learning models inherently lack the capability to comprehend raw textual data. Our text preprocessing pipeline encompasses several essential stages to enhance data quality and improve model performance. These stages include tokenization, lowercasing, punctuation removal, stopword removal, and URL removal as shown in Figure 82. Lowercasing ensures consistency in the text, while punctuation removal streamlines the text for analysis. Tokenization breaks down narratives into individual words, enabling further analysis. Stopword removal eliminates common stopwords, reducing noise, and URL removal ensures web links do not interfere with the analysis.

Once narratives undergo these preprocessing steps, they become ready for feature extraction a vital transformation that converts textual data into numerical features suitable for machine learning models. For feature extraction, we utilized two distinct techniques: Term Frequency-Inverse Document Frequency (TFIDF) and Word Embeddings (Word2Vec). TF-IDF quantifies term importance within the narratives, capturing the semantic meaning of words. Word2Vec represents words as dense vectors, allowing models to understand semantic relationships. Additional preprocessing steps included the removal of HTML tags, non-alphanumeric characters, and other irrelevant elements. Finally, lemmatization reduced words to their base forms, improving topic modeling interpretability.

These comprehensive preprocessing and feature extraction steps ensure the textual data is transformed into a suitable format for subsequent topic modelling, leading to a more robust analysis of aviation accident reports.

*14.4.3 Topic Modeling Procedure*

The preprocessed textual data were skillfully transformed into a Document-Term Frequency Matrix. This matrix plays a pivotal role in representing the frequency of each word across all narratives contained in the accident reports. Essentially, it provides a structured numerical representation of the textual data, facilitating the subsequent topic-modeling process.

a) Latent Dirichlet Allocation (LDA)

For the task of topic modelling, we harnessed the power of LDA, a probabilistic generative model that operates on the premise that documents are mixtures of topics, and topics, in turn, are mixtures of words. LDA is renowned for its established track record in topic modelling tasks [264]. This model excels at uncovering the latent thematic structures present within textual data, making it a natural choice for our analysis. Figure 83 illustrates the operational mode of an LDA model. The model deploys a three-phase non-deterministic approach to assign topics to groups of words in each document. Phase one involves sampling a list of topics from a Dirichlet distribution of topics for each document [363]. This is followed by phase two where each document word is assigned a topic from the sampled topics in phase one. Lastly, in phase three, each word assigned to a topic in phase two is sampled from a multinomial prior over words related to that topic. In this model, φ denotes the matrix of topic distributions, with a multinomial distribution over N-word items for each of T topics being drawn independently from a symmetric Dirichlet(β) prior. θ is the matrix of document-specific mixture weights for these T topics, each being drawn independently from a symmetric Dirichlet(β) prior. For each word, z denotes the topic responsible for generating that word, drawn from the θ distribution for that document, and w is the word itself, drawn from the topic distribution φ corresponding to z. $N_d$ stands for the number of words in the document. D stands for the size of the document collection. Estimating φ and θ provides information about the topics in a collection and the weights of those topics in each document.

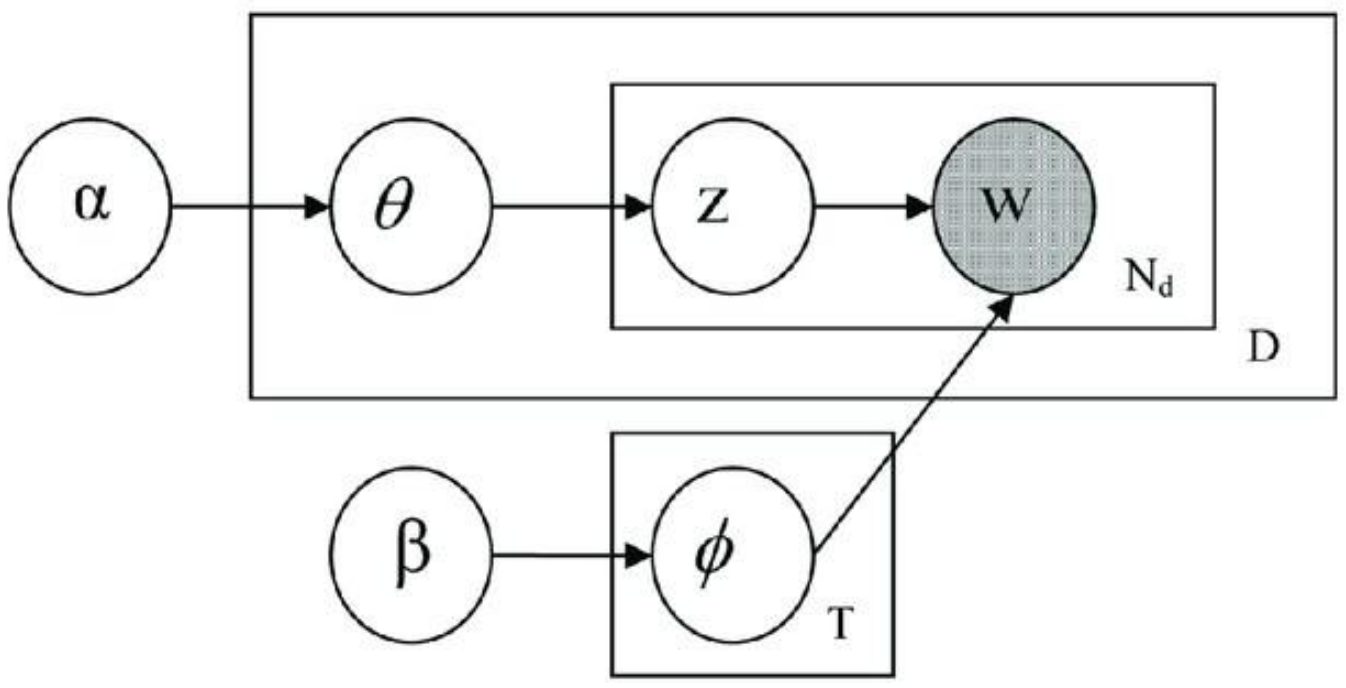

Figure 83. Graphical representation of the LDA model [21]

b) Non-negative Matrix Factorization (NMF)

As an alternative dimensionality reduction technique in topic modeling, Non-negative Matrix Factorization (NMF) was also harnessed in our analysis. NMF factorizes the Document-Term Frequency Matrix into two lower-dimensional matrices one representing topics and the other representing term distributions [361]. What sets NMF apart is its innate interpretability, which is valuable in extracting meaningful insights from aviation accident reports. The dataset is represented as a $w \times d$, matrix, $V$. where $w$ represents words in each document, $d$. Figure 84 illustrates a simple mechanism of how NMF breaks $V$ into its constituent components, $W$ and $H$ where $W$ is a $w \times t$ matrix, $H$ is a $t \times d$ matrix and t represents the distinct topics in $V$.

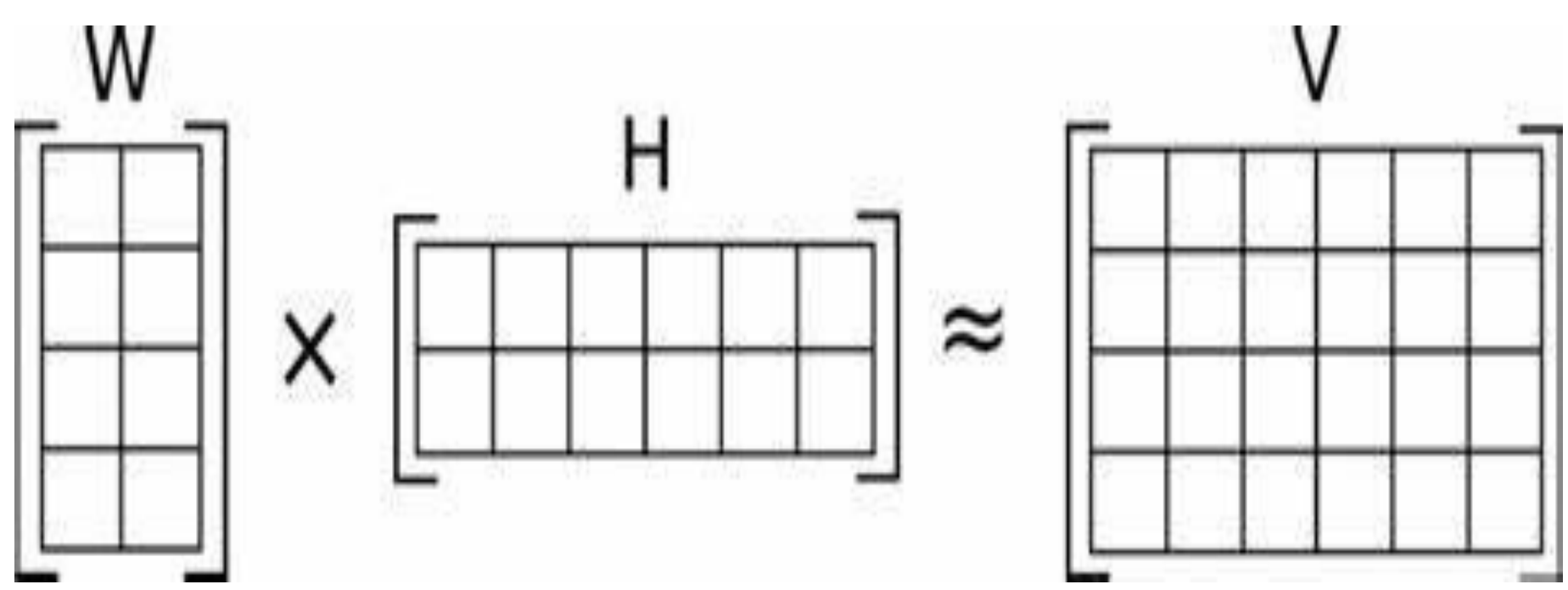

Figure 84. Non-negative matrix factorization diagram

### *14.5 RESULTS AND DISCUSSION*

In this section, we present the results of our comparative study between LDA and NMF models in extracting topics from aviation accident reports using the NTSB Aviation Dataset.

#### *14.5.1 Topic Extraction and Coherence Evaluation*

We employed the Coherence Value (C_v) as an evaluation metric to assess the quality of topics generated by the LDA and NMF models. The C_v measures the semantic coherence of topics, with higher values indicating more coherent topics. In our study, LDA yielded a C_v coherence score of 0.497, while NMF attained a C_v coherence score of 0.437.

#### *14.5.2 Topic Distribution*

We visualized the distribution of topics generated by both models. The topic distribution graph using the NMF model revealed distinct clusters of topics, suggesting meaningful separations. LDA also presented distinguishable topics, though with some overlapping themes as shown in Figures 86 and 87, respectively. It is clearly seen that each of the ten topics was derived from a varying number of documents (narratives). This indicates that the models were able to precisely cluster the various documents and ultimately assigned the appropriate topic to each document cluster. To facilitate a clear interpretation of the extracted information from a fitted LDA topic model, pyLDAvis was used to generate an intertropical distance map. A screenshot of the statistical proximity of the topics can be seen in Figure 85.

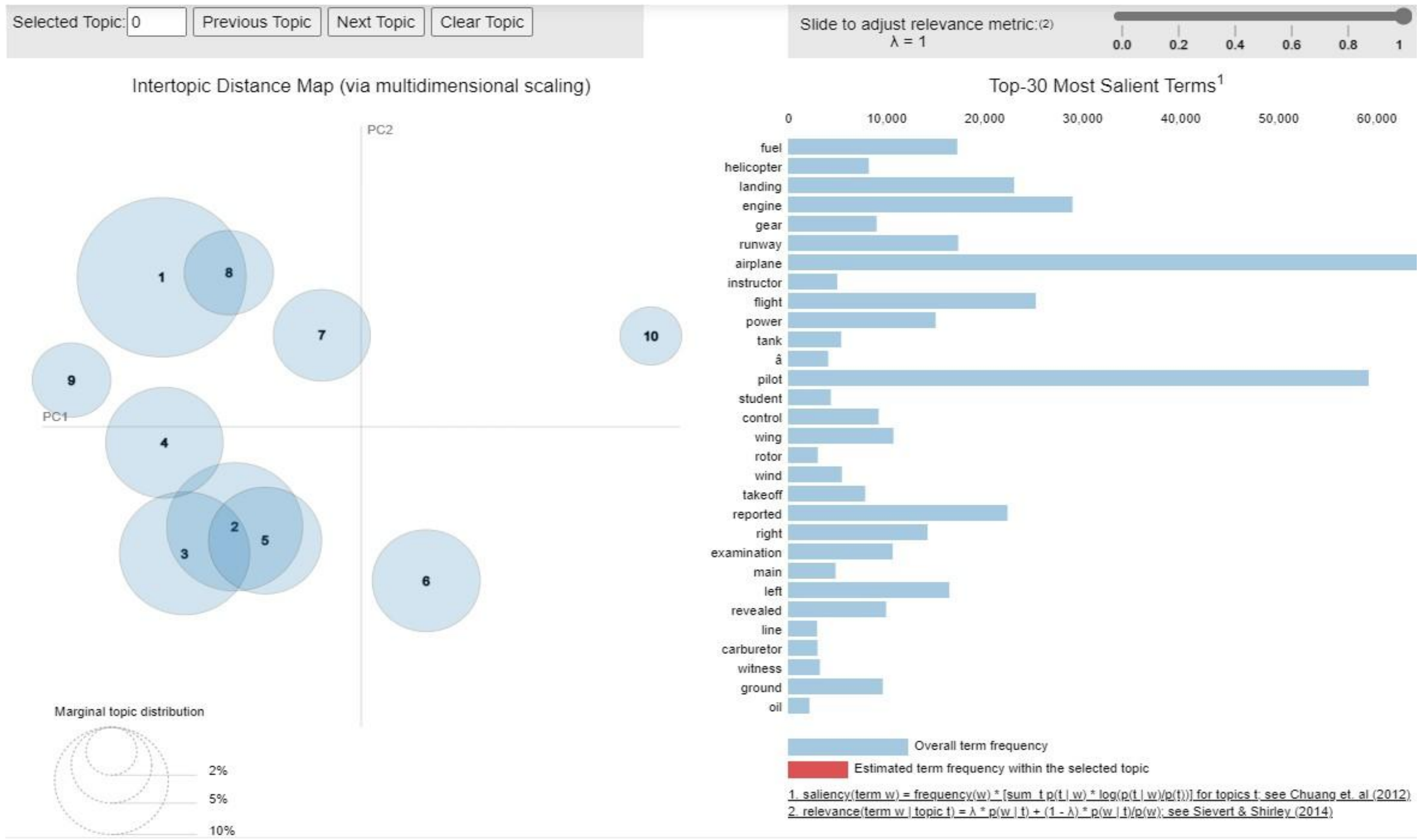

Figure 85. Topic modeling visualization for aviation reports

Table 28. TOP 5 WORDS FOR EACH OF THE 10 TOPICS THAT WERE CHOSEN BY BOTH MODELS

| **Models** | **Topic 1** | **Topic 2** | **Topic 3** | **Topic 4** | **Topic 5** | **Topic 6** | **Topic 7** | **Topic 8** | **Topic 9** | **Topic 10** |
|---|---|---|---|---|---|---|---|---|---|---|
| **LDA** | Pilot, | airplane | fuel | system | landing | engine | pilot | revealed | helicopter | airplane |
| | runway | pilot | tank | flight | gear | power | flight | examination | flight | pilot |
| | wind | reported | engine | control | pilot | airplane | airplane | inspection | instructor | accident |
| | knot | left | pilot | pilot | airplane | pilot | airport | bolt | pilot | flight |
| | accident | landing | airplane | seat | approach | loss | condition | fracture | rotor | witness |
| | airplane | fuel | engine | helicopter | gear | flight | student | wind | right | runway |
| **NMF** | reported | tank | power | rotor | landing | airplane | instructor | knot | left | airplane |
| | precluded | gallon | carburetor | tail | main | accident | flight | gust | airplane | Take off |
| | malfunction | engine | loss | blade | collapsed | condition | solo | gusting | rudder | approach |
| | operation | selector | icing | collective | nose | witness | cfi | accident | brake | foot |

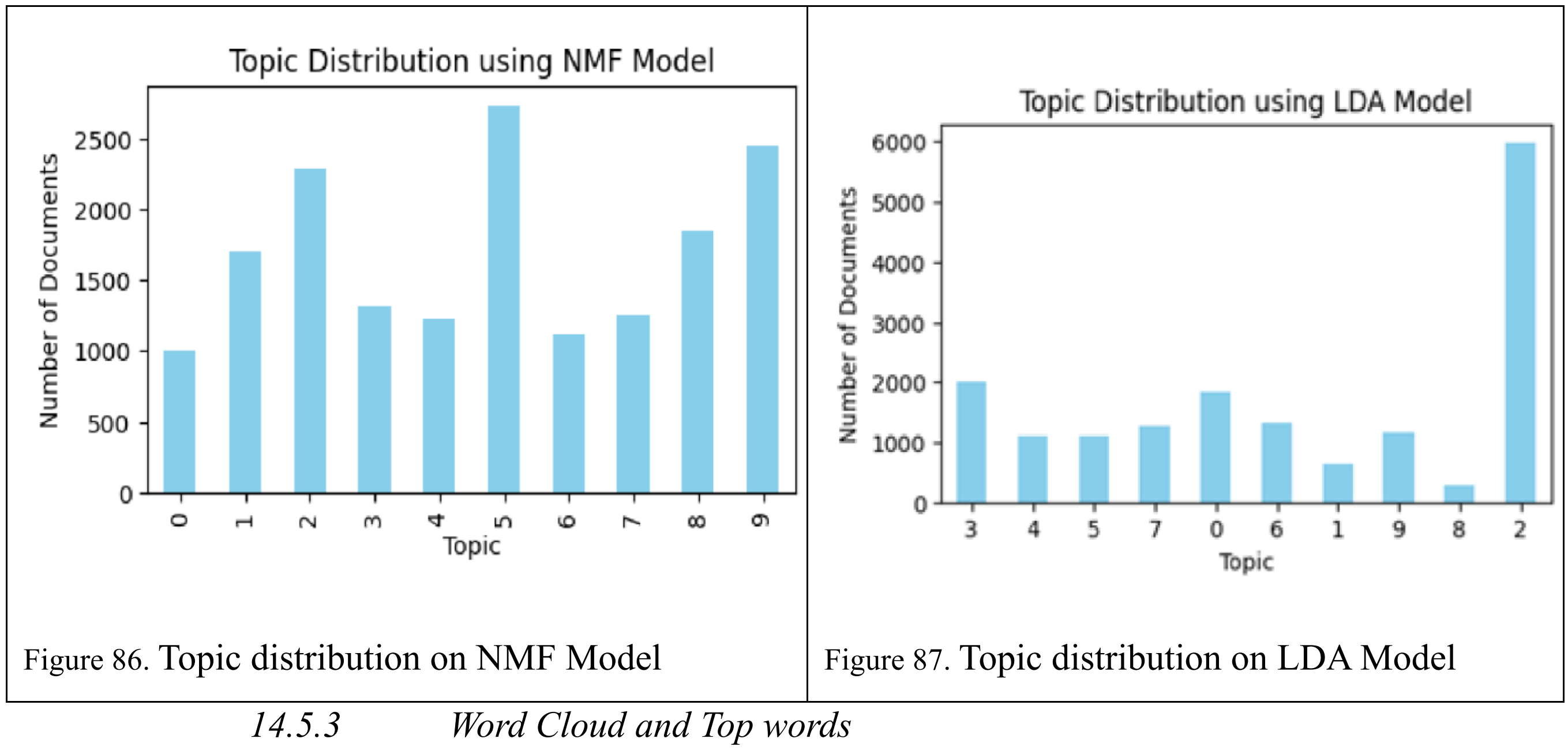

Figure 86. Topic distribution on NMF Model

Figure 87. Topic distribution on LDA Model

### *14.5.3 Word Cloud and Top words*

To enhance topic interpretability, we provided word clouds for both LDA and NMF-generated topics. Table 28 shows the top 5 words for each of the 10 topics for each model. To facilitate understanding of the central themes captured by each model, the results in Table 29 are further visualized in Figures 90 and 91 respectively. For each of the document clusters, the models learn the best defining topic words as shown in Figure 92 for LDA and 93 for NMF. It should be noted that a topic word can appear in more than one topic. For instance, "pilot" appears in 8 out of 10 topics for LDA while "airplane" appears in 5 out of the 10 topics in NMF. This feature distinguishes topic modeling from conventional unsupervised learningbased clustering.

Table 29. TOPIC EXTRACTION AND INTERPRETATION

| **LDA-Generated Topics** | **NMF-Generated Topics** |
|---|---|
| 1. Pilot and Aircraft Operations | 1. Aircraft Mechanical Failures |
| 2. Aircraft Mechanical Issues | 2. Fuel Systems and Engine Performance |
| 3. Fuel Systems and Engine Performance | 3. Engine Performance and Examination |
| 4. Flight Control Systems and Maintenance | 4. Helicopter Rotor and Tail Operations |
| 5. Landing and Gear Operations | 5. Landing Gear Issues |
| 6. Engine Performance and Examination | 6. Flight Conditions and Witness Reports |
| 7. Flight Conditions and Weather | 7. Flight Instruction and Solo Flights |
| 8. Maintenance and Component Examination | 8. Weather Conditions and Crosswind Effects |
| 9. Helicopter Operations and Instruction | 9. Aircraft Control and Runway Operations |
| 10. Witness Reports and Accident Examination | 10. Runway Operations and Takeoff |

For both models, the topics encompass a wide range of aviation-related themes, including aircraft operations, mechanical issues, fuel systems, engine performance, weather conditions, and flight instruction as shown in Table 29. The topics extracted by LDA and NMF provide a comprehensive overview of the factors contributing to aviation accidents.

*14.5.4 Topic Co-occurrence Matrix*

We presented a topic co-occurrence matrix for both models, shedding light on the relationships and overlaps between topics as seen in Figures 88 and 89.

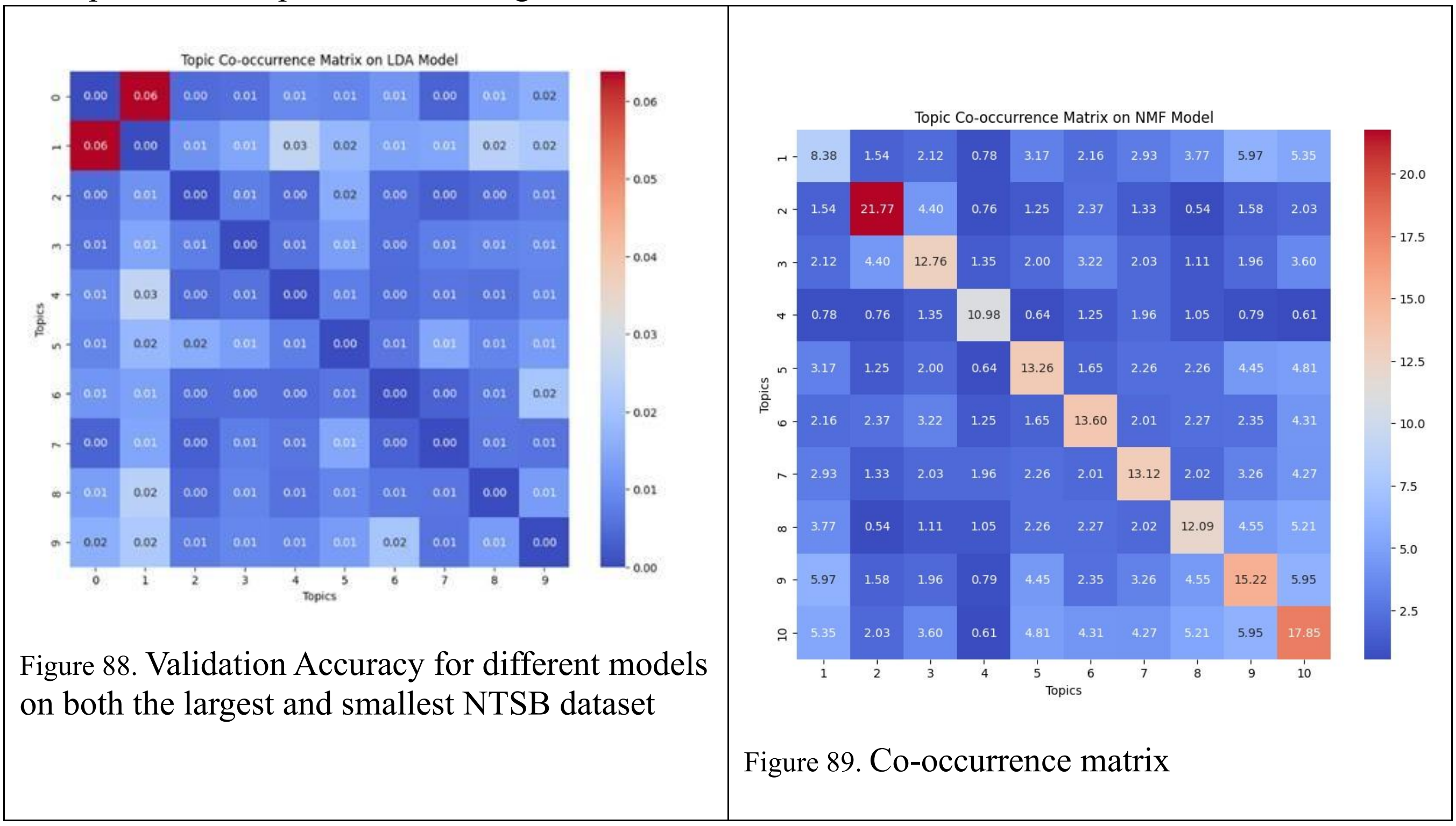

Figure 88. Validation Accuracy for different models on both the largest and smallest NTSB dataset

Figure 89. Co-occurrence matrix

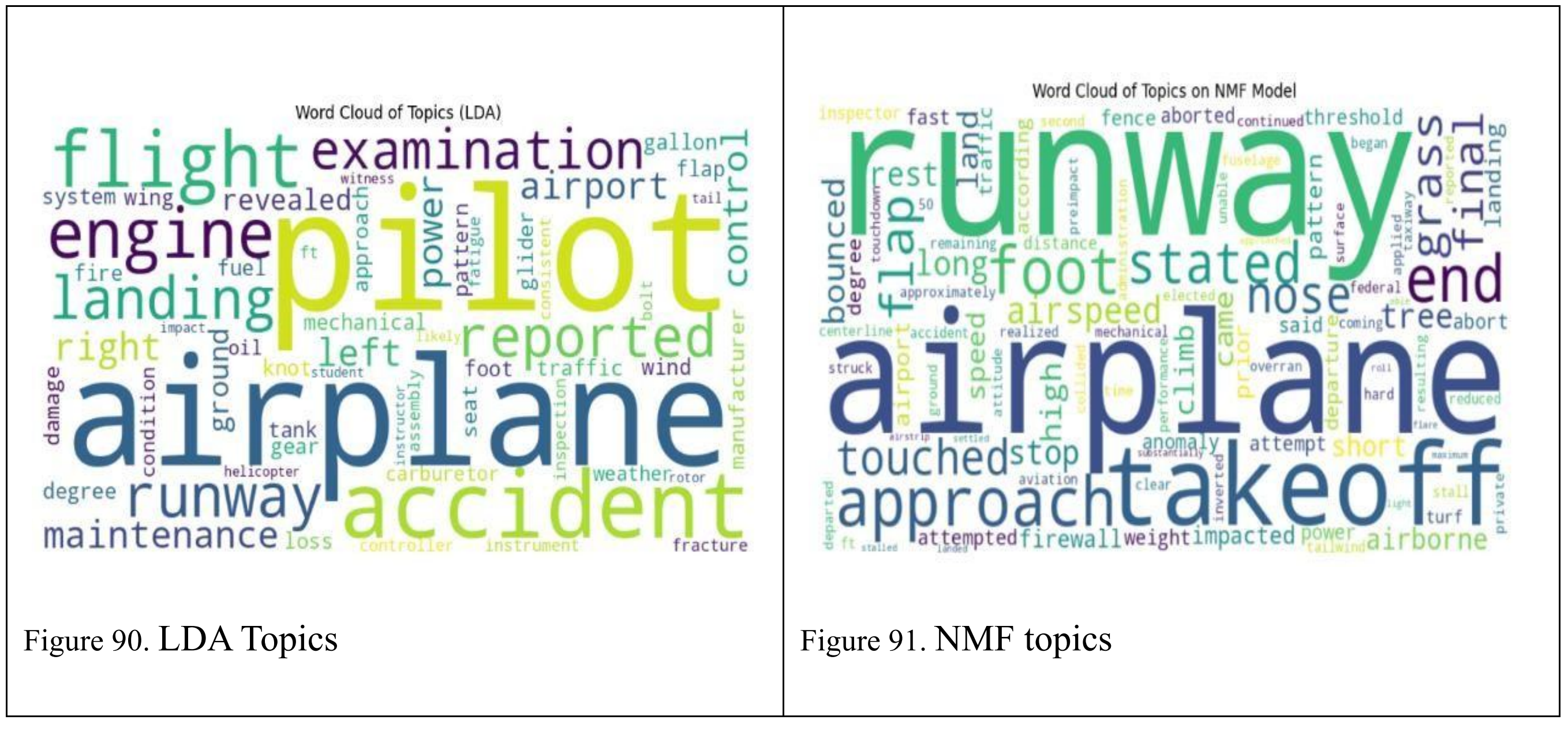

Figure 90. LDA Topics

Figure 91. NMF topics

Topic 1: airplane, reported, precluded, malfunction, operation, sustained, failure, normal, substantial, mechanical
Topic 2: fuel, tank, gallon, engine, selector, power, flight, pump, right, airplane
Topic 3: engine, power, carburetor, loss, icing, heat, oil, forced, examination, run
Topic 4: helicopter, rotor, tail, blade, collective, main, skid, ground, autorotation, rpm
Topic 5: gear, landing, main, collapsed, nose, extended, position, retracted, extend, revealed
Topic 6: flight, airplane, accident, condition, witness, instrument, airport, altitude, ft, turn
Topic 7: student, instructor, flight, solo, cfi, control, took, landing, instruction, instructional
Topic 8: wind, knot, gust, gusting, accident, weather, degree, crosswind, airport, runway
Topic 9: right, left, airplane, rudder, brake, wing, tailwheel, applied, veered, runway
Topic 10: runway, airplane, takeoff, approach, foot, end, stated, flap, nose, grass

Figure 92. Shows the topics that were selected by NMF

Topic 1: pilot, runway, wind, knot, accident, airplane, reported, airport, foot, degree
Topic 2: airplane, pilot, reported, left, landing, runway, right, wing, damage, mechanical
Topic 3: fuel, tank, engine, pilot, airplane, power, flight, right, left, gallon
Topic 4: system, flight, control, pilot, seat, maintenance, fire, engine, accident, manufacturer
Topic 5: landing, gear, pilot, airplane, approach, runway, glider, traffic, pattern, flap
Topic 6: engine, power, airplane, pilot, loss, carburetor, examination, oil, revealed, landing
Topic 7: pilot, flight, airplane, airport, condition, accident, weather, instrument, ft, controller
Topic 8: revealed, examination, inspection, bolt, fracture, maintenance, assembly, fatigue, accident, consistent
Topic 9: helicopter, flight, instructor, pilot, rotor, student, reported, ground, control, tail
Topic 10: airplane, pilot, accident, flight, witness, likely, engine, ground, examination, impact

Figure 93. Shows the topics that were selected by LDA

The comparative analysis reveals interesting findings regarding the performance of LDA and NMF in topic extraction from aviation accident reports. LDA outperforms NMF in terms of topic coherence, as indicated by the higher C_v score. This suggests that the topics generated by LDA exhibit stronger semantic relevance and coherence among the words within each topic. On the other hand, NMF exhibits its own strengths, particularly in selecting distinct and specific topics. NMF topics often focus on precise aspects of accidents, such as mechanical failures or weather conditions. This granularity can be advantageous for targeted analysis and decision-making. The word clouds figures, topic co-occurrence matrices figures, and top word table further illustrate the differences between the models' outputs. LDA tends to produce broader, more general topics, while NMF excels in capturing nuanced details.

## 14.6 CONCLUSION

In this study, we conducted a comparative analysis of LDA and NMF models for topic modeling in the domain of aviation accident reports. We found that LDA achieved higher topic coherence, indicating a stronger semantic relationship between words within topics. However, NMF exhibited the ability to generate more specific and granular topics, which may be valuable for in-depth accident analysis. Our findings underscore the importance of considering the specific goals of topic modeling when choosing between LDA and NMF. Researchers and aviation safety experts can benefit from selecting the model that aligns with their objectives, whether it be comprehensive topic coverage or a detailed focus on specific aspects of accidents.

As aviation safety continues to evolve, the insights derived from topic modeling analysis can contribute to informed decision-making, enhanced safety measures, and a deeper understanding of the complex factors influencing aviation accidents. Future research may explore hybrid approaches that leverage the strengths of both LDA and NMF to further advance accident report analysis in the aviation industry.

# Chapter 15. Exploring Aviation Incident Narratives Using Topic Modeling and Clustering Techniques

## 15.1 Abstract

Aviation safety is a global concern, requiring detailed investigations into incidents to understand contributing factors comprehensively. This study uses the National Transportation Safety Board (NTSB) dataset. It applies advanced natural language processing (NLP) techniques, including Latent Dirichlet Allocation (LDA), Non-Negative Matrix Factorization (NMF), Latent Semantic Analysis (LSA), Probabilistic Latent Semantic Analysis (pLSA), and K-means clustering. The main objectives are identifying latent themes, exploring semantic relationships, assessing probabilistic connections, and cluster incidents based on shared characteristics. This research contributes to aviation safety by providing insights into incident narratives and demonstrating the versatility of NLP and topic modelling techniques in extracting valuable information from complex datasets. The results, including topics identified from various techniques, provide an understanding of recurring themes. Comparative analysis reveals that LDA performed best with a coherence value of 0.597, pLSA of 0.583, LSA of 0.542, and NMF of 0.437. Kmeans clustering further reveals commonalities and unique insights into incident narratives. In conclusion, this study uncovers latent patterns and thematic structures within incident narratives, offering a comparative analysis of multiple-topic modelling techniques. Future research avenues include exploring temporal patterns, incorporating additional datasets, and developing predictive models for early identification of safety issues. This research lays the groundwork for enhancing the understanding and improvement of aviation safety by utilising the wealth of information embedded in incident narratives.

## 15.2 Introduction

Aviation safety is a paramount concern in contemporary society, necessitating investigations into aviation incidents to understand the multifaceted factors contributing to mishaps. The National Transportation Safety Board (NTSB) provides a comprehensive dataset of aviation incident narratives, a treasure trove of information on past incidents. Analyzing this dataset offers an invaluable opportunity to extract nuanced insights, contributing to the continuous enhancement of safety protocols [341]. The vast amount of textual data in aviation incident reports poses a significant challenge for manual analysis due to its volume, making the process time-consuming and impractical. Automated text analysis methods, specifically topic modelling, have emerged as effective solutions to this challenge. These techniques allow the identification of latent thematic structures within the textual data, facilitating the extraction of pertinent information, trends, and patterns [62, 97, 101].

The NTSB dataset, a repository of incident narratives spanning diverse scenarios and contributing factors, serves as a vital archive for the aviation industry. Each narrative provides a unique account of events, offering a nuanced perspective on the circumstances surrounding incidents. Leveraging advanced Natural Language Processing (NLP) techniques and machine learning methodologies can reveal latent patterns within these narratives, shedding light on recurring themes, hidden correlations, and potential areas for improvement.

This study aims to harness the power of advanced techniques, including Latent Dirichlet Allocation (LDA), Non-negative Matrix Factorization (NMF), Latent Semantic Analysis (LSA), Probabilistic Latent Semantic Analysis (pLSA), and K-means clustering, to delve into the intricate details of aviation incident narratives within the NTSB dataset. The specific objectives include:

1. Utilizing topic modelling techniques to identify latent themes within aviation incident narratives.
2. Applying matrix factorization techniques to decompose the dataset and identify key components.
3. Exploring latent semantic relationships within the narratives using LSA.
4. Assessing the probabilistic relationships between terms and topics through pLSA.
5. Employing K-means clustering to group incidents based on shared characteristics, providing a holistic view of narrative structures.

This research is significant on multiple fronts. Firstly, it contributes to the broader discourse on aviation safety by offering a nuanced understanding of incident narratives. Secondly, it showcases the versatility of advanced NLP and machine learning techniques in extracting valuable information from complex narratives, a methodological contribution to the NTSB dataset. Ultimately, the findings of this study have the potential to inform aviation authorities, empower safety professionals, and inspire further research, fostering a safer aviation environment.

In the subsequent sections, we will delve into the details of the related work (Section II), our methodology, data collection, and the application of advanced techniques, to reveal the untold stories within aviation incident narratives (Section III). The results (Section IV), conclusion, and avenues for future work (Section V) will follow in this exploration.

## 15.3 RELATED WORK

Topic modelling plays a pivotal role in text analysis and NLP, offering a potent method to unveil latent thematic structures within extensive textual data volumes. In recent years, the application of topic modelling techniques and NLP in analyzing aviation incident narratives has gained significant traction. Traditionally, accident investigation and safety analysis relied on expert analysis and statistical methods [62, 96, 364]. However, these methods, relying on manual examination of accident reports, are timeconsuming and susceptible to human bias [160].

Advancements in automated text analysis methods, notably in NLP and machine learning, have revolutionized aviation safety research [24, 189, 329]. Utilizing techniques such as text mining and sentiment analysis, researchers have gained scalability and objectivity while minimizing human bias when scrutinizing reports. Furthermore, within the realm of NLP, topic modelling has emerged as a significant tool in aviation safety research, providing a systematic method for revealing latent thematic structures within textual data. Key algorithms like LDA and NMF have been applied across diverse textual datasets, further enhancing the analytical capabilities in this field [132, 189, 264, 361].

The utilization of topic modelling in the analysis of aviation incident reports presents significant potential, as seen in studies utilizing pLSA and LDA [97]. Effective text preprocessing, such as NLP techniques [365], and methodologies like Latent Semantic Analysis (LSA) [171] and NMF [155], contribute to enhancing the quality of extracted topics.

Noteworthy, research [86, 132, 336, 364] has showcased the potential of computational methods, text mining, and machine learning in accident report analysis, emphasizing efficiency and consistency. Our research builds upon these foundations by conducting a comprehensive comparative analysis of pLSA, LSA, LDA, and NMF, along with k-means clustering, applied to the NTSB dataset. This work contributes to ongoing efforts to enhance aviation safety and risk assessment through advanced text analysis methodologies.

## 15.4 METHODOLOGY

In this section, we explain the methodology employed to explore aviation incident narratives within the NTSB dataset. Our approach integrates advanced NLP and ML techniques to unveil latent patterns, hidden correlations, and thematic structures embedded in the narrative. The next subsection gives a brief description of the data sets used in this study.

### *15.4.1 Data Acquisition*

Aviation incident and accident investigation reports used in this study were exclusively sourced from the NTSB dataset spanning years from 2000 to 2020. The dataset comprising a collection of more than 36,000 records in JSON format was obtained from the following source: https://www.ntsb.gov/Pages/AviationQuery.aspx. These reports encompass textual narratives, findings, and recommendations from NTSB investigations, making them an invaluable resource for the study.

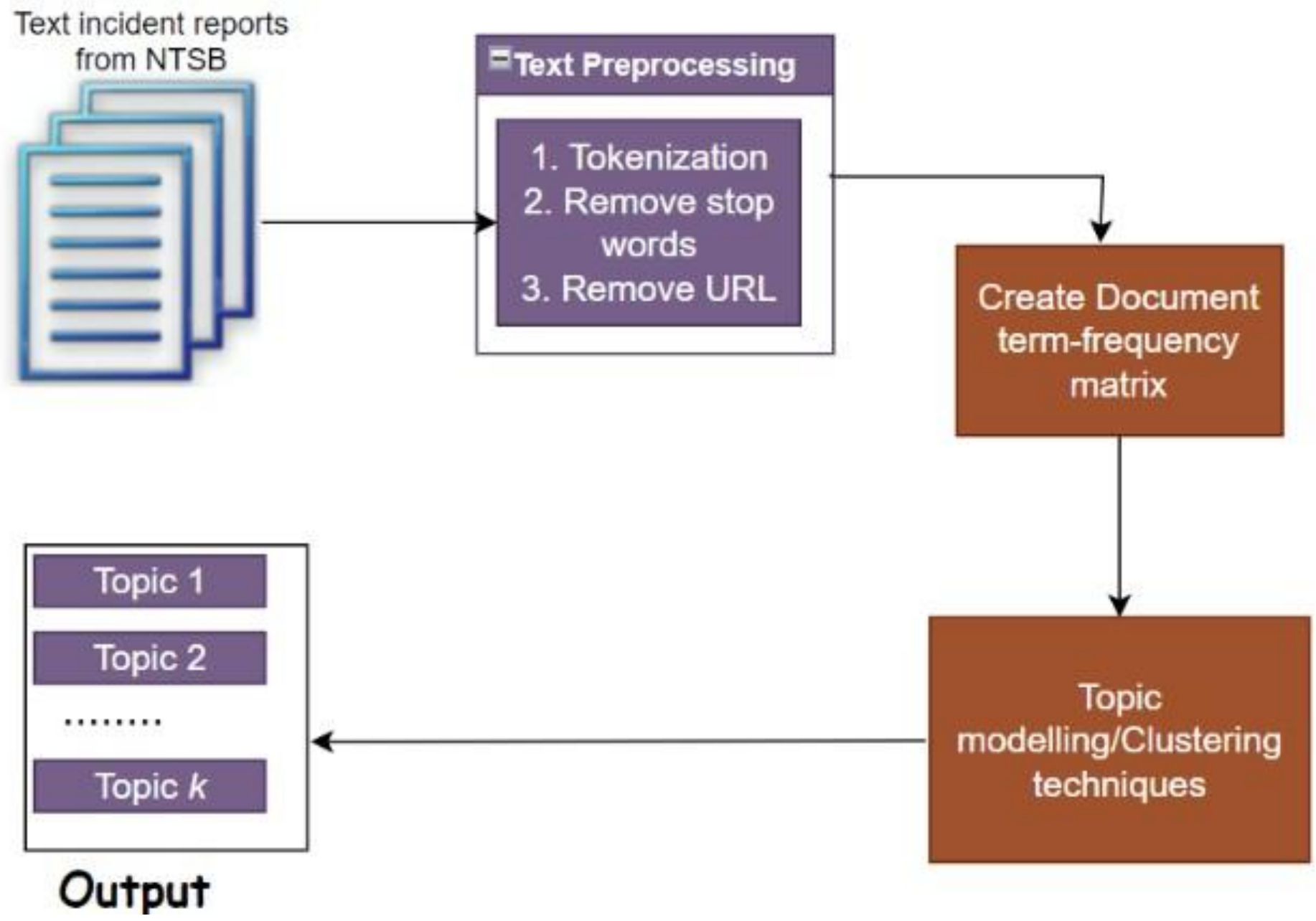

Figure 94. Methodological framework

*15.4.2 Text Processing*

Machine learning algorithms cannot inherently comprehend raw textual data. Our text preprocessing pipeline comprises multiple crucial stages aimed at enhancing data quality and optimizing model performance. These stages, illustrated in Figure 94, encompass tokenization, lowercasing, punctuation removal, stopword elimination, and URL removal. Lowercasing ensures uniformity in text representation, while punctuation removal simplifies text for analysis. Tokenization dissects narratives into individual words, facilitating further analysis. Eliminating stopwords reduces noise, and URL removal prevents web links from interfering with analysis [329].

After undergoing these preprocessing steps, narratives are primed for feature extraction, a pivotal process that transforms textual data into numerical features suitable for machine learning models. For feature extraction, we employed two distinct techniques: Term Frequency-Inverse Document Frequency (TF-IDF) and Word Embeddings (Word2Vec). TF-IDF assesses term importance within narratives, capturing semantic meaning. Word2Vec represents words as dense vectors, enabling models to comprehend semantic relationships. Supplementary preprocessing steps included HTML tag removal, non-alphanumeric character elimination, and exclusion of irrelevant elements. Finally, lemmatization standardizes words to their base forms, enhancing interpretability in topic modelling.

*15.4.3 Topic Modeling Techniques*

a) Latent Dirichlet Allocation (LDA)

LDA is employed to identify latent topics within the narratives. The algorithm models each document as a distribution of topics and each topic as a distribution of words. This facilitates the extraction of themes inherent in the narratives. Figure 95 illustrates the operational mode of an LDA model. The model utilizes a three-phase stochastic approach to allocate topics to clusters of words within each document. Initially, in phase one, topics are sampled from a Dirichlet distribution for each document *[86]*. Subsequently, during phase two, each word in the document is assigned a topic from the sampled topics obtained in phase one. Finally, in phase three, each word assigned to a topic in phase two is sampled from a multinomial distribution over words associated with that topic. In this model, the matrix φ represents the topic distributions, with a multinomial distribution over N-word items for each of T topics drawn independently from a symmetric Dirichlet (β) prior. θ represents the matrix containing document-specific mixture weights for a set of T topics, with each weight drawn independently from a symmetric Dirichlet (β) prior distribution. Within this framework, z identifies the topic attributed to generating a particular word, drawn from the θ distribution specific to the document, while w denotes the word itself, drawn from the topic distribution φ corresponding to z. Nd represents the total number of words in the document, and D signifies the size of the document collection. The estimation of φ and θ yields insights into the topics present in the collection and their respective weights within each document.

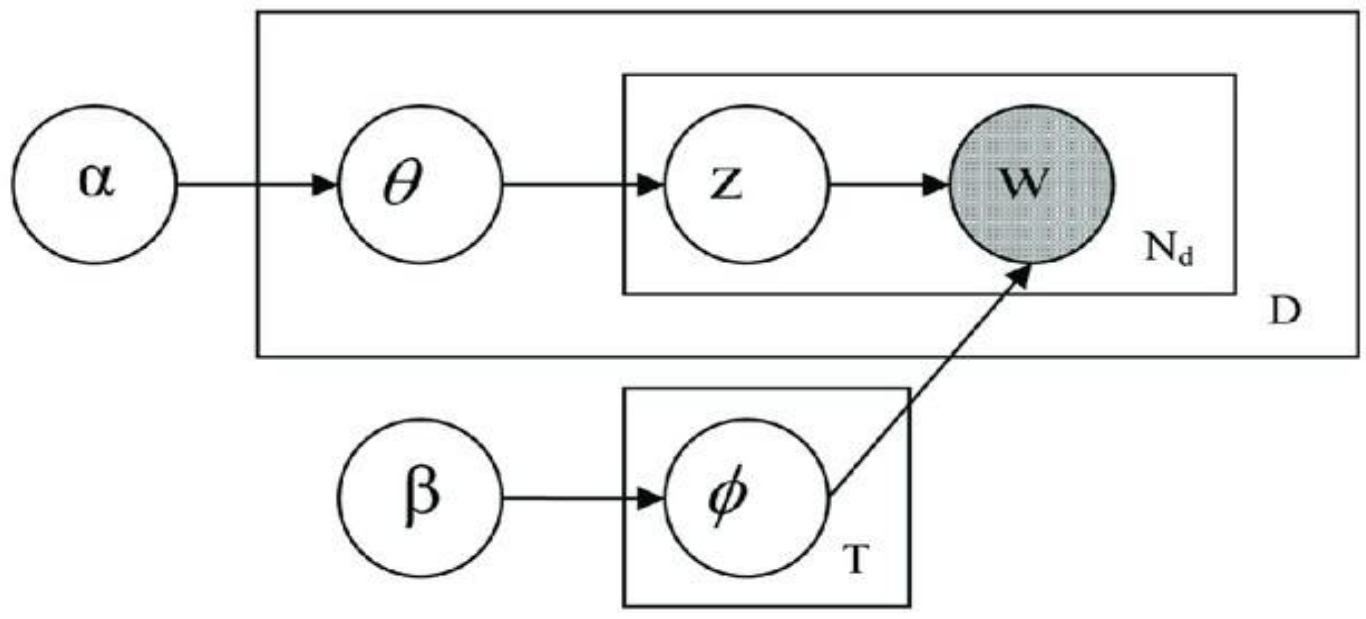

Figure 95. Graphical representation of the LDA model [21]

b) Non-Negative Matrix Factorization (NMF):

NMF is utilized for matrix factorization, decomposing the dataset into non-negative matrices representing key components. This technique aids in revealing underlying structures within the narratives. In our analysis, we also employed Non-negative Matrix Factorization (NMF) as an alternative approach for dimensionality reduction in topic modelling. NMF decomposes the Document-Term Frequency Matrix into two lower-dimensional matrices: one representing topics and the other representing term distributions [17]. What distinguishes NMF is its inherent interpretability, which proves valuable in extracting meaningful insights from aviation accident reports. The dataset is represented as a matrix V with dimensions w×d, where w denotes words in each document and d represents documents. Figure 96 illustrates a simplified depiction of how NMF decomposes V into its constituent parts, W and H, where W is a matrix of dimensions $w \times t$, H is a matrix of dimensions $t \times d$, and t represents the distinct topics in V.

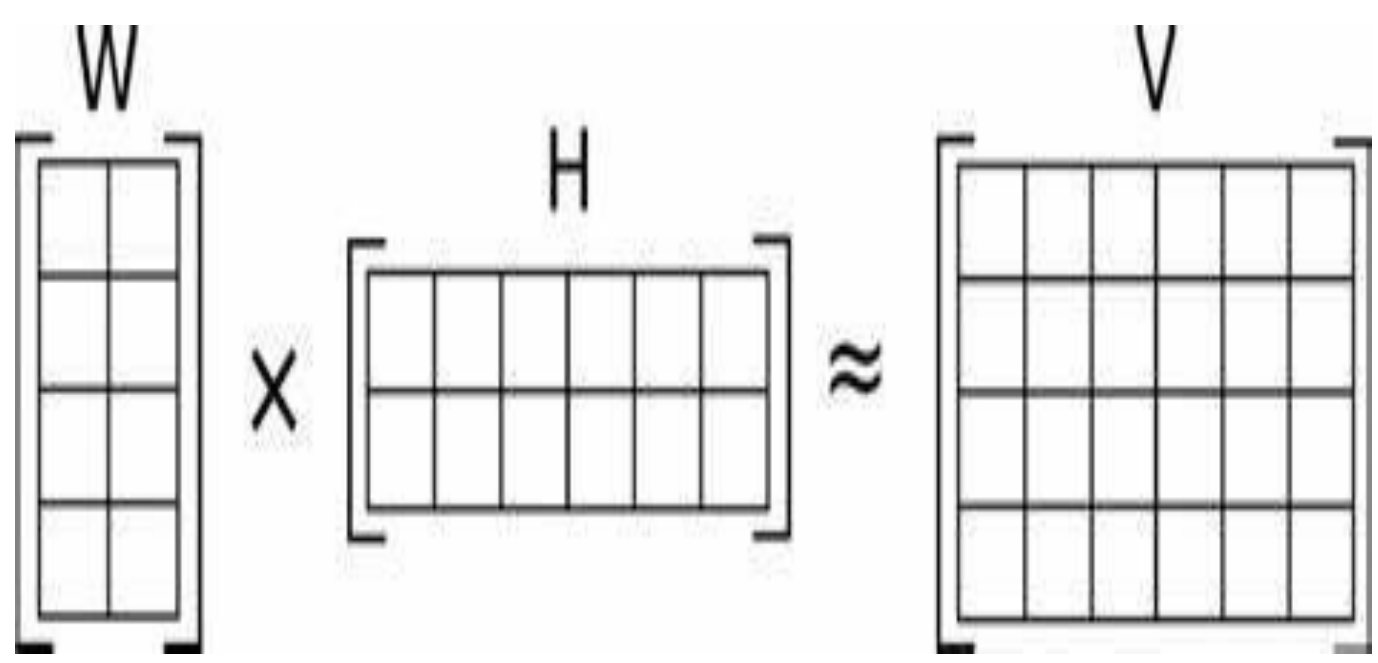

Figure 96. Non-negative matrix Factorization diagram

*15.4.4 Semantic Analysis*

a) Latent Semantic Analysis (LSA):

LSA is applied to capture latent semantic relationships within the narratives. By representing words and documents in a high-dimensional semantic space, it unveils nuanced connections between terms. LSA, as a method of dimensionality reduction, converts the document-term matrix into a space with fewer dimensions. It reveals the inherent structure of text data by discerning connections between terms and documents *[295]*. In our study, LSA was employed to extract topics from aviation safety reports, thus offering insights into the concealed themes present in the corpus.

b) Probabilistic Latent Semantic Analysis (pLSA):

pLSA is employed to model the probabilistic relationships between terms and topics. This probabilistic approach allows for a more nuanced understanding of the uncertainty inherent in incident narratives. pLSA takes a probabilistic approach, contrasting with Singular Value Decomposition (SVD) in tackling the topic modelling problem *[366]*. It constructs a probabilistic model featuring latent topics to generate the observed data in the document-term matrix. Specifically, it aims to establish a model $\boldsymbol{P(D, W)}$ that assigns probabilities to each entry in the document-term matrix for any given document D and word W. Adhering to the foundational assumptions of topic models, which propose that each document is a mixture of various topics and each topic comprises a set of words, pLSA introduces a probabilistic interpretation to these principles:

1. For document d, pLSA assigns topic z to that document with the likelihood denoted as $P(z|d)$.

2. When considering a topic z, pLSA models the probability of drawing a word w from that topic as $P(w|z)$.

The model is represented as

$$P(D, W) = P(D) \sum_z P(z|d)P(w|z)$$

where P(D), $P(Z|D)$, and $P(W|Z)$ are parameters. P(D) can be directly computed from the corpus data, while $P(Z|D)$ and $P(W|Z)$ are represented as multinomial distributions and can be trained using the expectation-maximization algorithm (EM) [63]. In simple terms, EM is a technique used to determine the most probable parameter estimates for a model relying on unobserved, latent variables (in this context, the topics).

#### *15.4.5 K-means Clustering*

K-means clustering is a widely used unsupervised machine learning technique that aims to partition a dataset into K distinct, non-overlapping subsets or clusters. Each data point is assigned to the cluster with the nearest mean, and the algorithm iteratively refines the cluster assignments to minimize the withincluster variance [367].

The algorithmic overview of the K-means is as follows:

- Randomly select K data points as the initial centroids of the clusters.
- Assign each data point to the cluster whose centroid is the nearest, typically using Euclidean distance.
- Recalculate the centroids based on the meaning of the data points within each cluster.
- Iterate the assignment and centroid update steps until convergence, which occurs when the assignments no longer change significantly and the result is K clusters, each characterized by its centroid.

#### *15.4.6 Evaluation Metrics*

To assess the chosen techniques' effectiveness, coherence scores and interpretability metrics are employed. Coherence scores measure semantic similarity between high-probability words in topics, while interpretability metrics evaluate the human interpretability of generated topics.

#### *15.4.7 Implementation*

The execution of our methodology is carried out using the Python programming language with Jupyter Notebook serving as the integrated development environment (IDE). We leverage the capabilities of prominent libraries, including NLTK, Gensim, and Scikit-Learn. Each technique is applied iteratively, with parameters fine-tuned to optimize performance. The selection of these libraries is grounded in their robust functionalities in natural language processing (NLP) and machine learning (ML), ensuring a comprehensive and efficient implementation of our techniques. This approach not only fosters transparency but also encourages reproducibility and extension of our analyses by future researchers.

## 15.5 RESULTS AND DISCUSSION

In this section, we present the outcomes of our analysis, including the identified topics from various techniques, NMF, PLSA, LSA, and LDA.

#### *15.5.1 Topic Extraction and Coherence Evaluation.*

We employed the Coherence Value (C_v) as an evaluation metric to assess the quality of topics generated by the models. The C_v measures the semantic coherence of topics, with higher values indicating more coherent topics. In our study, LDA yielded a C_v coherence score of 0.597, while outperforming other models as shown in Table 30.

Table 30. shows the coherence score of all models.

| **Technique** | **Coherence Value** |
|---|---|
| Non-Negative Matrix Factorization (NMF) | 0.437 |
| Latent Dirichlet Allocation (LDA) | 0.597 |
| Latent Semantic Analysis (LSA) | 0.542 |
| Probabilistic Latent Semantic Analysis (PLSA) | 0.583 |

*15.5.2 NMF and LDA model's performance*

The topic distribution from the document, as illustrated in Figure 97, showcases the contrasting approaches of Latent Dirichlet Allocation (LDA) and Non-Negative Matrix Factorization (NMF) models. In Figure 88, the LDA model offers a comprehensive view of topic distribution across all topics providing a holistic perspective on the document's content. Conversely, the NMF model selectively emphasizes topic distribution primarily focusing on topics 4, 6, and 8.

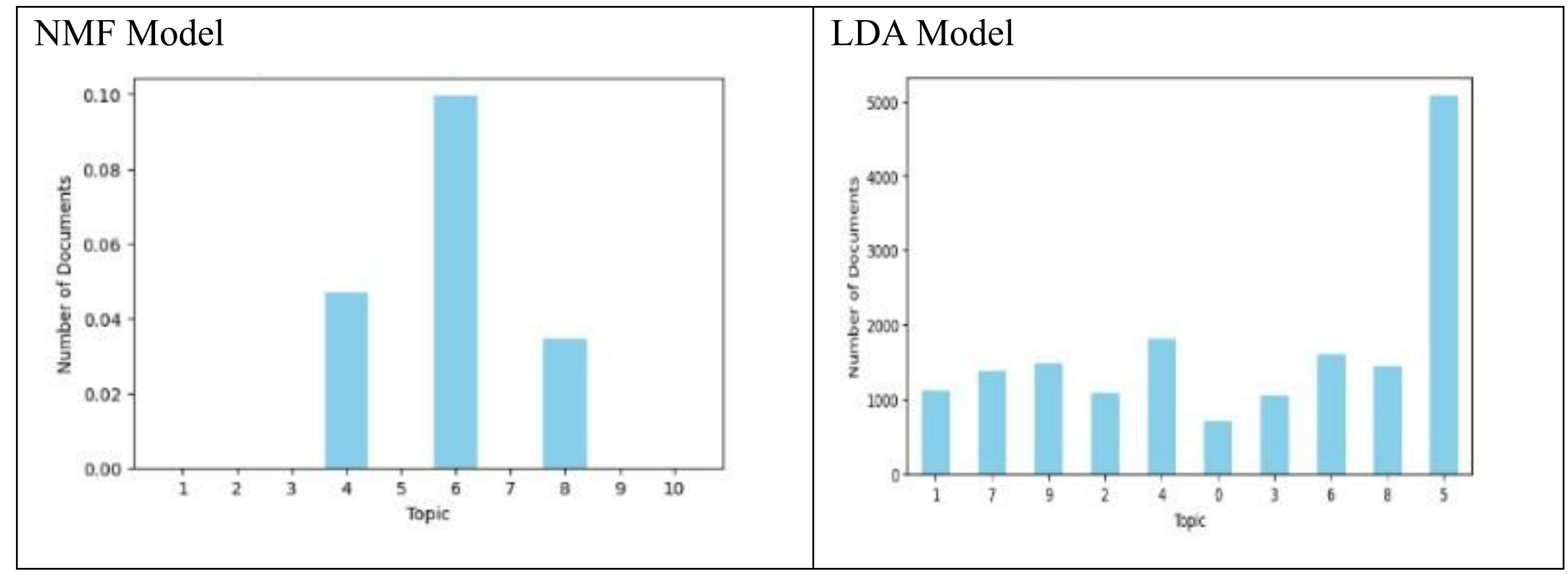

Figure 97. Shows the topic distribution of the words from the document on LDA and NMF models

*15.5.3 Top Words for Each Topic and Word Clouds*

We present an overview of the identified topics, their associated words, the distribution of topics across the dataset, and a thematic word cloud visualization. Examining Figure 98, which depicts the top words for each topic chosen by the pLSA model, it becomes evident that certain topics exhibit higher word scores than others. Notably, the words associated with topic three seen in Figure 98 stand out with elevated word scores, indicating their prominence within the document. This nuanced insight into word importance contributes to a more detailed understanding of the topics extracted by the LDA model.

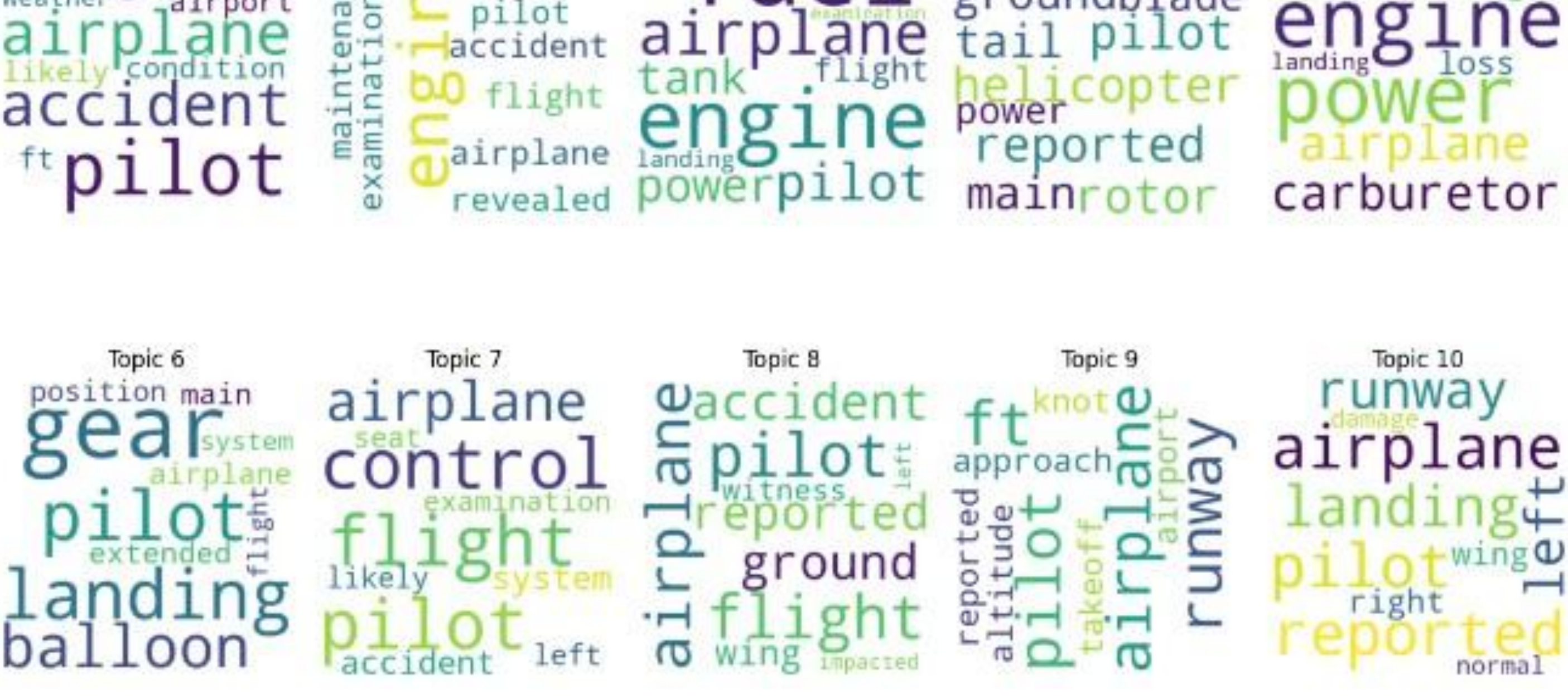

Figure 98. 10 words that were chosen by the pLSA model in each Topic

*15.5.4 pLSA and LSA models performance*

LSA leverages mathematical techniques to uncover underlying themes within aviation incident narratives. We present an overview of the identified topics, their associated words, and the proportion of variance explained, as well as a thematic word cloud visual representation of these topics. Also, Figure 99 shows the term weights of words selected by LSA on the first four topics. This visualization helps to determine the optimal number of topics to capture the underlying thematic structure of the aviation incident narratives.

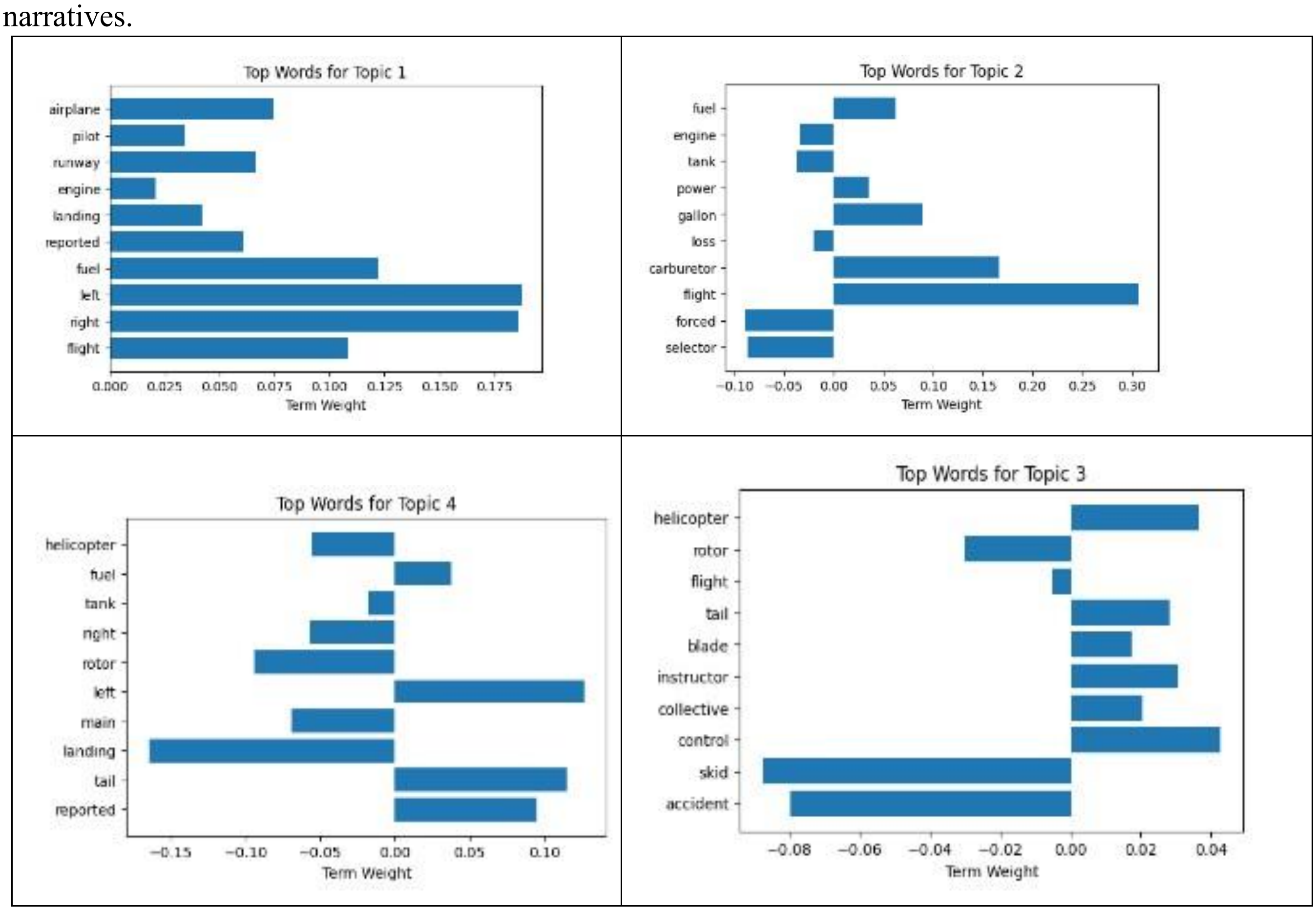

Figure 99. Shows the term weights of the words that were selected by LSA on the first four topics.

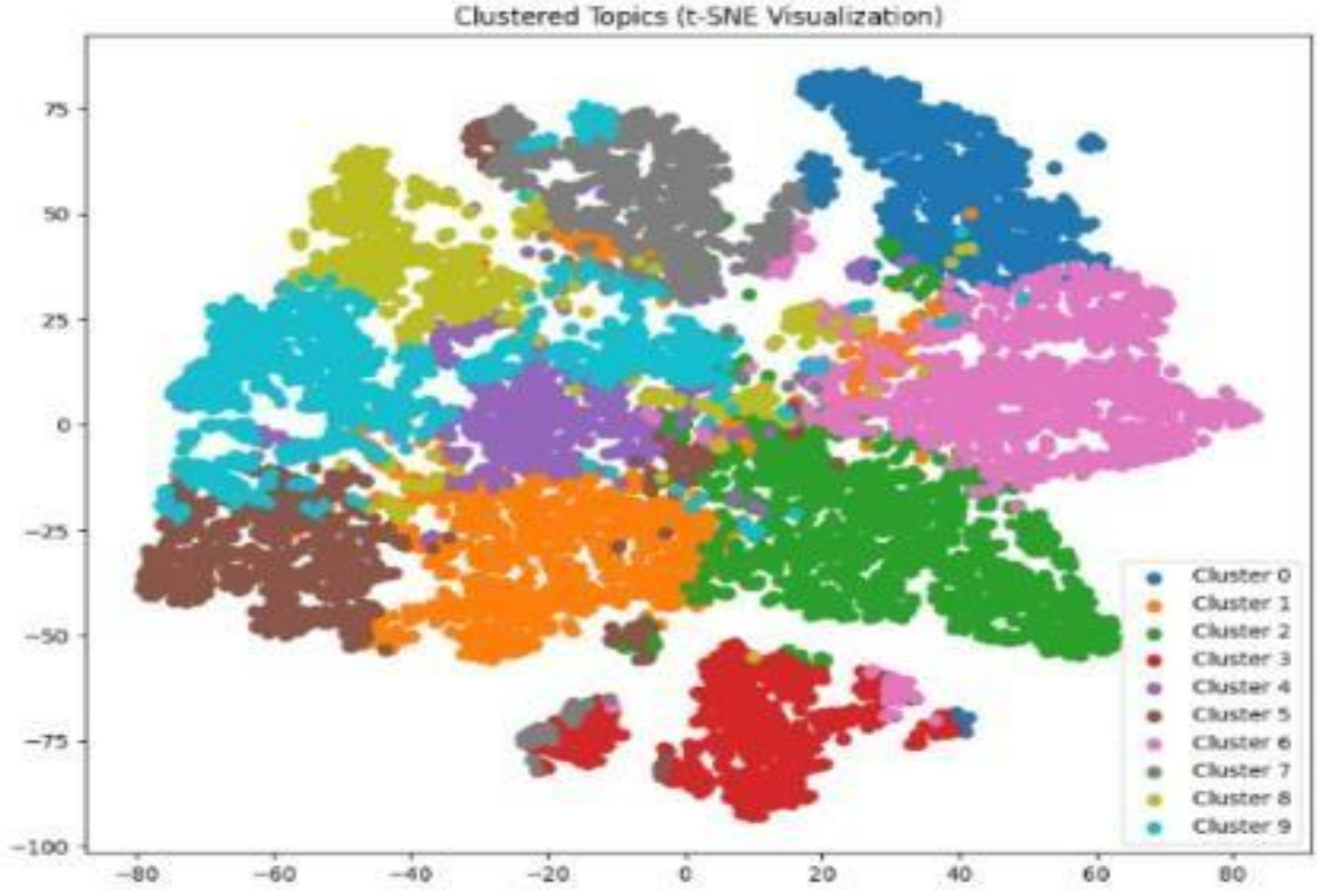

Figure 100. Clustered Topics (t-SNE Visualization).

Table 31. SHOWS THE TOPIC INTERPRETATION FOR ALL MODELS.

| No. Topics | NMF chosen Topics | Lda chosen topics | LSA chosen Topics | pLSA chosen topics |
|---|---|---|---|---|
| Topic 1 | Mechanical Failures | Maintenance and Inspection | Pilot-Centric Elements | Cabin Fumes and Engineering |
| Topic 2 | Fuel System Issues | Helicopter Operations | Fuel System Dynamics | ATC and Aircraft Separation |
| Topic 3 | Engine and Power Loss | Fuel System and Pilot Actions | Flight and Helicopter Dynamics | Engine Failure in Cruise |
| Topic 4 | Helicopter Operation | Landing Gear and Aircraft Control | Landing Gear and Helicopter Rotor | Bird Strikes |
| Topic 5 | Landing Gear Incidents | Engine and Power Loss | Engine and Fuel Interaction | Safety Officer Reports |
| Topic 6 | Flight Conditions | Accident Conditions | Runway and Flight Dynamics | Runway Incidents |
| Topic 7 | Student Pilot Incidents | Aviation Accident Dynamics | Instructor-Led Flight | Rejected Takeoff |
| Topic 8 | Wind and Weather Impact | Weather and Airport Conditions | Power Issues | Minor Damage Incidents |
| Topic 9 | Rudder and Brake Issues | Takeoff and Runway Conditions | Airport and Runway Dynamics | GPWS Alerts |
| Topic 10 | Runway Incidents | Passenger and Glider Dynamics | Runway and Aircraft Dynamics | Routine Inspection Findings |

#### 15.5.5 *Common Themes Across Models:*

Despite the differences in the algorithms, certain themes consistently emerge across multiple models as shown in Table 31. For instance, topics related to "Mechanical Failures," "Fuel System Issues," and "Engine and Power Loss" are recurrent, suggesting their prevalence in aviation incidents. This consistency reinforces the significance of these factors in contributing to safety concerns. In interpreting these results, it is essential to note the nuanced patterns and recurring themes across different models.

While NMF, LSA, pLSA, and LDA offer distinct perspectives on the dataset, the convergence of certain keywords across topics suggests common threads in aviation incident narratives. The identified topics provide a foundation for further exploration and analysis, contributing valuable insights to the broader discourse on aviation safety.

#### 15.5.6 *Unique Insights from Each Model:*

While common themes provide a baseline understanding, each modelling technique offers unique insights. NMF emphasizes specific aspects such as "Helicopter Operation" and "Flight Conditions," shedding light on scenarios that might be overlooked by other techniques. LSA captures nuanced relationships, revealing connections between "Wind and Weather Impact" and "Rudder and Brake Issues.". Table 32 depicts the top 10 words that were chosen by all models among the 10 topics.

#### *15.5.7 K-means Clustering of Incident Narratives*

In addition to topic modelling, K-means clustering is employed to categorize incidents based on shared characteristics, revealing underlying patterns and providing a complementary perspective to topic modelling. This approach enhances the interpretability of incident narratives by grouping them into distinct clusters, each defined by prevalent characteristics and keywords. These clusters, highlighted through K-means clustering, offer a structured view of narrative patterns, contributing to a more nuanced understanding of incident types and facilitating targeted interventions for safety improvement.

Clusters are formed by grouping incidents with similar attributes enabling a more focused analysis of shared characteristics. Each cluster encapsulates a set of incidents with commonalities, providing a clearer delineation of narrative patterns. For instance, a cluster may Centre around "Runway Incidents" or "Landing Gear and Helicopter Rotor," highlighting the thematic coherence within each group. Figure 100 visually represents the clustered topics using t-SNE (t-distributed Stochastic Neighbor Embedding) visualization. The figure illustrates ten distinctive clusters. The colour blue (Cluster 0), orange (Cluster 1), and green (Cluster 2) respectively provide a visual representation of the incident narratives grouped by Kmeans clustering. This visualization aids in identifying patterns and relationships between incidents within and across clusters, offering a valuable perspective for safety analysis.

### 15.6 Conclusion

This research employs advanced topic modelling and clustering techniques to delve into aviation incident narratives within the NTSB dataset. Through a comparative analysis of NMF, LSA, pLSA, and LDA, the study uncovers both common and unique insights, providing an understanding of recurring themes in aviation incidents. K-means clustering further enhances this comprehension by organizing incidents into meaningful groups, offering a nuanced perspective on narrative structures.

The study significantly contributes to aviation safety research by unveiling latent patterns and thematic structures within incident narratives. The exploration of incident clusters enhances the roughness of safety analysis, providing valuable insights into the complexities of aviation safety.

As avenues for future research, temporal patterns in incident narratives could be explored to identify evolving safety concerns. Furthermore, the incorporation of additional datasets could validate and expand the findings, offering a broader perspective on aviation safety. Additionally, the development of predictive models for the early identification of potential safety issues presents a promising direction for advancing the field. Therefore, this study acts as a foundational step for further advancements in understanding and improving aviation safety, leveraging the wealth of information embedded in incident narratives.

Table 32. TOP 5 WORDS FOR EACH OF THE 10 TOPICS THAT WERE CHOSEN BY ALL MODELS

| **Models** | **Topic 1** | **Topic 2** | **Topic 3** | **Topic 4** | **Topic 5** | **Topic 6** | **Topic 7** | **Topic 8** | **Topic 9** | **Topic 10** |
|---|---|---|---|---|---|---|---|---|---|---|
| **NMF** | Right | Fuel | Engine | Helicopter | Gear | Light | Student | Wind | Reported | Runway |
| | Left | Tank | Power | Rotor | Landing | Accident | Instructor | Knot | Mechanical | Approach |
| | Brake | Gallon | Carburetor | Tail | Nose | Condition | Flight | Gust | Malfunction | Takeoff |
| | Rudder | Engine | Loss | Blade | Collapsed | Witness | Control | Accident | Operation | Stated |
| | Wing | Selector | Oil | Collective | Approach | Instrument | Landing | Weather | Failure | Final |
| **LDA** | Maintenance | Helicopter | Fuel | Landing | Engine | Airplane | Airplane | Pilot | Airplane | Pilot |
| | Revealed | Flight | Tank | Gear | landing | Pilot | Pilot | Flight | Runway | Line |
| | Inspection | Instructor | Engine | Airplane | Power | Reported | Accident | Airplane | Pilot | Aircraft |
| | Manufacturer | Pilot | Pilot | Pilot | Pilot | Runway | Flight | Accident | Takeoff | Glider |
| | Examination | Rotor | Airplane | Brake | Airplane | Landing | Engine | Condition | Approach | passenger |
| **LSA** | Airplane | Fuel | Helicopter | Helicopter | Gear | Gear | Pilot | Left | Runway | Runway |
| | Pilot | Engine | Rotor | Fuel | Landing | Flight | Flight | Helicopter | Flight | Accident |
| | Runway | Tank | Flight | Tank | Engine | Landing | Helicopter | Right | Instructor | Flight |
| | Engine | Power | Tail | Rotor | Power | Helicopter | Left | Landing | Airplane | Helicopter |
| | Landing | Gallon | Instructor | Landing | Carburetor | Reported | Instructor | Gear | Reported | Pilot |
| **PLSA** | Pilot | Engine | Fuel | Helicopter | Engine | Gear | Control | Airplane | Airplane | Airplane |
| | Airplane | Airplane | Engine | Pilot | Power | Landing | Pilot | Pilot | Pilot | Pilot |
| | Flight | Pilot | Airplane | Rotor | Carburetor | Pilot | Flight | Flight | Runway | Reported |
| | Accident | Revealed | Pilot | Flight | Pilot | Balloon | Airplane | Accident | Approach | Landing |
| | Airport | Examine | Power | Reported | Airplane | Airplane | System | Reported | Knot | Runway |

# Chapter 16. Comparative Analysis of Topic Modeling Techniques on ATSB Text Narratives Using Natural Language Processing

[389], Nanyonga, A., Wasswa, H., Turhan, U., Joiner, K., & Wild, G. (2024, March). Comparative Analysis of Topic Modeling Techniques on ATSB Text Narratives Using Natural Language Processing. In *2024 3rd International Conference for Innovation in Technology (INOCON)* (pp. 1-7). IEEE.

## 16.1 Abstract

Improvements in aviation safety analysis call for innovative techniques to extract valuable insights from the abundance of textual data available in accident reports. This paper explores the application of four prominent topic modelling techniques, namely Probabilistic Latent Semantic Analysis (pLSA), Latent Semantic Analysis (LSA), Latent Dirichlet Allocation (LDA), and Non-negative Matrix Factorization (NMF), to dissect aviation incident narratives using the Australian Transport Safety Bureau (ATSB) dataset. The study examines each technique's ability to unveil latent thematic structures within the data, providing safety professionals with a systematic approach to gain actionable insights. Through comparative analysis, this research not only shows the potential of these methods in aviation safety but also elucidates their distinct advantages and limitations.

## 16.2 Introduction

The aviation industry plays a pivotal role in global transportation, ensuring the safe and efficient movement of people and goods across the world. As aviation activities continue to expand, safety remains a paramount concern. To ensure the highest levels of safety, aviation organizations worldwide collect and analyze a wealth of data, particularly incident reports. These reports, submitted to entities like the Australian Transport Safety Bureau (ATSB), contain crucial information about safety incidents, accidents, and near misses [341]. Analyzing and extracting insights from this vast textual data is an essential endeavour, as it can lead to improved safety measures, accident prevention, and enhanced risk assessment within the aviation sector.

The volume of textual data generated by aviation incident reports is immense, making manual analysis both time-consuming and impractical. To address this challenge, automated text analysis methods, known as topic modelling, have emerged as powerful tools. Topic modelling techniques are applied to discover latent thematic structures within textual data, enabling the extraction of relevant information, trends, and patterns [97, 150, 364].

In this paper, we present a comparative analysis of four leading topic modelling techniques; Probabilistic Latent Semantic Analysis (pLSA), Latent Semantic Analysis (LSA), Latent Dirichlet Allocation (LDA), and Non-negative Matrix Factorization (NMF) as applied to the text narratives found in the ATSB dataset.

We explore the potential of Natural Language Processing (NLP) techniques to preprocess the data and facilitate the application of these topic modelling methods.

The primary objective of this study is to evaluate and compare the performance of these topic modelling techniques in terms of their ability to uncover critical insights from aviation incident narratives. Our research aims to contribute to the ongoing efforts to enhance aviation safety and risk assessment through advanced text analysis.

In the following sections of this paper, we will delve into the related work in Section II, and our methodology in Section III, present the experimental result, and engage in a comprehensive discussion of our findings in Section IV. Furthermore, we will explore the implications of our research for the field of aviation safety and suggest potential avenues for future research in Section V.

## 16.3 Related work

Topic modelling is a critical issue of text analysis and natural language processing, providing a potent means to unveil latent thematic structures within extensive volumes of textual data. In recent years, the application of topic modelling techniques to aviation incident narratives and the utilization of NLP methods in this context have gained significant traction. Traditionally, accident investigation and safety analysis predominantly relied on expert analysis and statistical methods [96, 124, 150]. Nonetheless, these conventional methodologies are hampered by their dependence on the manual examination of accident reports, which is a time-consuming process and is susceptible to human bias [329]. Experts meticulously scrutinized textual narratives, findings, and recommendations to discern recurring patterns and contributing factors. While these methods have yielded valuable insights, they have limitations when dealing with the extensive and intricate datasets that aviation accident reports represent.

Recent advancements in automated text analysis techniques have ushered in a new era for aviation safety research. Researchers have acknowledged the potential of NLP and machine learning to extract actionable insights from textual data [114, 159, 170]. Text mining and sentiment analysis have been applied to scrutinize and analyze reports such as online forums [358, 359]. These approaches offer scalability and objectivity, mitigating the human bias inherent in traditional analyses.

Topic modelling, a subset of NLP, has garnered significant attention in the realm of aviation safety research. It offers a systematic approach to unveil latent thematic structures within textual data, making it particularly suited for aviation accident reports [132]. Blei et al. [264] introduced LDA, a seminal topic modelling algorithm that has become a cornerstone in this field. LDA has found applications in diverse textual datasets, spanning from news articles and social media content to scientific literature. Its proficiency in revealing underlying topics and relationships has proven invaluable across numerous contexts. Additionally, Lee and Seung introduced NMF, a dimensionality reduction technique applied in text mining and topic modelling. NMF has been utilized in various studies to extract topics from textual data, offering an alternative approach to LDA [361].

The application of topic modelling techniques to aviation incident reports holds substantial promise. pLSA and LDA have been utilized to unveil latent topics within textual narratives [4]. In a study by Luo and Shi, LDA was employed to identify topics related to accident causation and contributing factors within aviation accident reports, effectively showcasing the extraction of meaningful topics [64].

Effective text preprocessing is a vital step in text analysis. The application of NLP techniques, such as text tokenization, stop-word removal, and stemming, has become standard practice in the field. Research by Alghamdi & Alfalqi underscored the significance of NLP preprocessing in optimizing the performance of topic modelling techniques. Their survey conclusively demonstrated that proper preprocessing significantly enhances the quality of topics extracted from textual data [365].

Latent Semantic Analysis (LSA) has also found applications in aviation safety analysis. Researchers have employed LSA to unveil concealed structures in safety reports, thereby facilitating enhanced risk assessment and safety measures. LSA was deployed to analyze the latent semantic structure of safety reports within the aviation industry, with the intention of improving risk assessment and accident prevention strategies [368].

Non-negative Matrix Factorization (NMF) has gained prominence in text analysis for its capability to provide interpretable and non-negative factorization of document-term matrices. Within the context of aviation safety, NMF has been used to extract topics from incident reports. Li, Shen, and Xu delved into the application of NMF to identify safety topics from incident narratives, effectively demonstrating its capacity to yield meaningful and interpretable topics [155].

Robinson [132] harnessed LDA to extract topics from aviation safety reports and identify emerging safety concerns. Similarly, Ahadh et al. [96] delved into topic modelling to categorize narratives in aviation accident reports, furnishing a structured representation of accident data. [150] made use of text-mining techniques to analyze aviation accident reports, with a focus on identifying significant terms and phrases, thereby establishing the groundwork for the application of computational methods in accident report analysis. Zhong et al. [336] introduced a framework that seamlessly melded text mining and machine learning, enabling the automated classification of accident reports into categories based on contributing factors. Their approach served as a testament to the potential for automating key aspects of accident analysis, consequently enhancing efficiency and consistency.

The work by Rose et al. [86] delivered a compelling contribution by employing structural topic modelling, specifically LDA, to aviation accident reports. Their research not only affirmed the feasibility of employing topic modelling to reveal latent themes within accident narratives but also underscored its potential to automate certain facets of the analysis process, leading to enhanced efficiency. Their research emphasized the critical importance of selecting an appropriate topic modelling technique tailored to specific domains and datasets, accentuating the need for a nuanced approach to topic modelling in aviation safety analysis.

Collectively, these studies underscore the significance of text analysis and topic modelling in the enhancement of aviation safety. They emphasize the importance of adopting advanced techniques, including NLP preprocessing and conducting comparative evaluations of topic modelling methods to extract actionable insights from aviation incident reports. In our research, we aim to build upon these foundations by conducting a comprehensive comparative analysis of four prominent topic modelling techniques; pLSA, LSA, LDA, and NMF applied to the ATSB dataset. Our work contributes to the ongoing efforts to bolster aviation safety and risk assessment through advanced text analysis.

## 16.4 Methodology

This section outlines the processes and techniques employed in this research to conduct a comparative analysis of four prominent topic modelling techniques: pLSA, LSA, LDA, and NMF. This section provides an in-depth explanation of data collection, preprocessing, and the implementation of each topic modelling method.

### *16.4.1 Data Collection*

Aviation incident/accident investigation reports are collected and published by various organizations such as ATSB, the Aviation Safety Reporting System (ASRS), and the National Transportation Safety Board (NTSB). For this study, the researchers utilized the ATSB aviation incident/accident investigation reports. Depending on the nature of the problem, we considered text narratives that were recorded in Australia for the period of 10 years resulting in a dataset with 53,275 records where the data was sourced directly from the ATSB investigation authorities spanning from 1/01/2013 to 12/31/2022. This resulted in a dataset comprising 50,778 records following data preprocessing and cleaning.

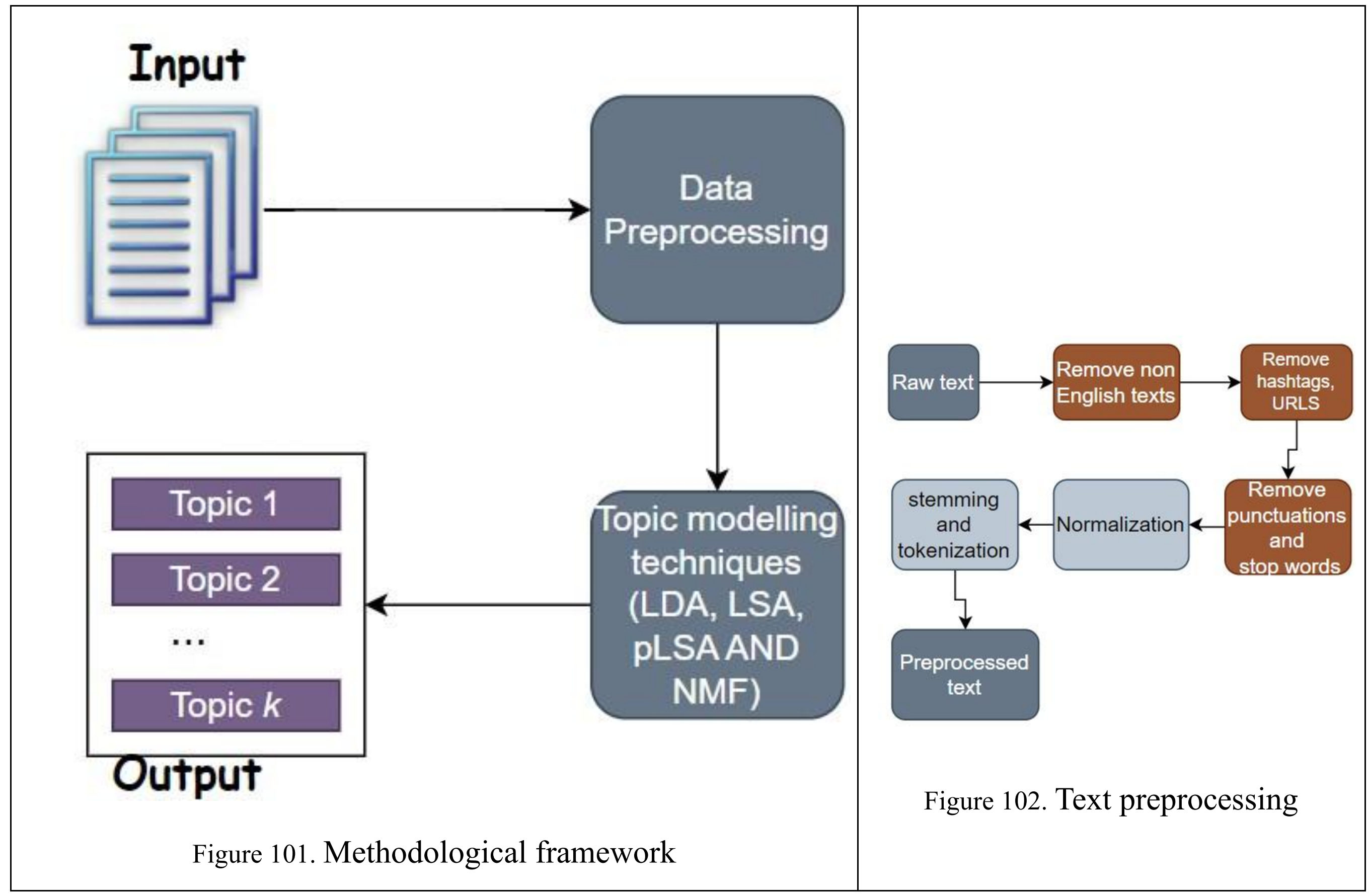

Figure 101. Methodological framework

Figure 102. Text preprocessing

### ***16.4.2*** *Text Preprocessing*

Machine learning models inherently lack the capability to comprehend raw textual data. Our text preprocessing pipeline encompasses several essential stages to enhance data quality and improve model performance. As shown in Figure 101, these stages include lowercasing, punctuation removal, tokenization, stopword removal, and URL removal. Lowercasing ensures consistency in the text, while punctuation removal streamlines the text for analysis. Tokenization breaks down narratives into individual words, enabling further analysis. Stopword removal eliminates common stopwords, reducing noise, and URL removal ensures web links do not interfere with the analysis.

Once narratives undergo these preprocessing steps, they become ready for feature extraction, a vital transformation that converts textual data into numerical features suitable for machine learning models. For feature extraction, we utilized two distinct techniques: Term Frequency-Inverse Document Frequency (TF-IDF) and Word Embeddings (Word2Vec). TF-IDF quantifies term importance within the narratives, capturing the semantic meaning of words. Word2Vec represents words as dense vectors, allowing models to understand semantic relationships. Additional preprocessing steps included the removal of HTML tags, non-alphanumeric characters, and other irrelevant elements. Consistent lowercase conversion and stopword removal reduced noise, and tokenization created a suitable corpus. Finally, lemmatization reduced words to their base forms, improving topic modelling interpretability. These comprehensive preprocessing and feature extraction steps ensure the textual data is transformed into a suitable format for subsequent topic modelling, leading to a more robust analysis of aviation accident reports.

#### *16.4.3 Topic Modelling Procedure*

After thorough text preprocessing, the next crucial steps involve transforming the preprocessed textual data into numerical features suitable for topic modelling techniques; pLSA, LSA, LDA, and NMF. Each technique was implemented separately, and topics were extracted from the dataset as shown in Figure 102.

The preprocessed textual data were skilfully transformed into a Document-Term Frequency Matrix. This matrix plays a pivotal role in representing the frequency of each word across all narratives contained in the accident reports. Essentially, it provides a structured numerical representation of the textual data, facilitating the subsequent topic modelling process.

a) Latent Dirichlet Allocation (LDA)

For the task of topic modelling, we harnessed the capabilities of LDA. LDA, a probabilistic generative model, is built upon the fundamental assumption that documents represent blends of various topics, and these topics, in turn, comprise assortments of words. LDA's (figure 103) reputation as a stalwart in the realm of topic modelling is well-documented [264]. This model stands out for its proficiency in unveiling the concealed thematic architectures inherent in textual data, rendering it an intuitive selection for our analytical endeavours.

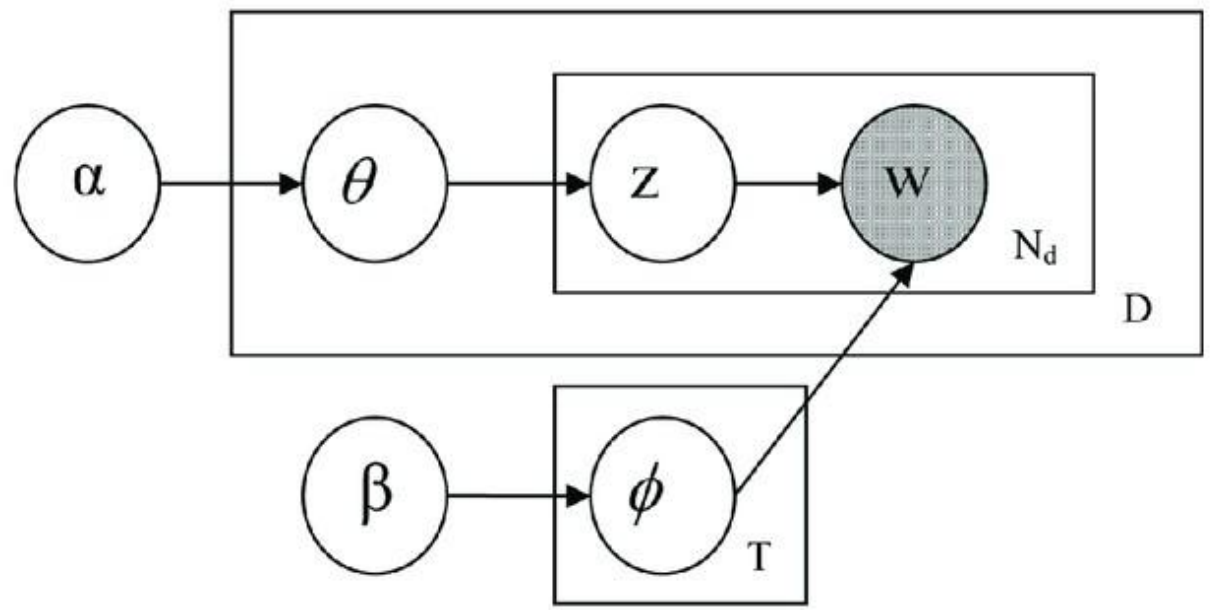

Figure 103. Shows a probabilistic graphical representation of the LDA model

b) Latent Semantic Analysis (LSA)

Latent Semantic Analysis (LSA) is a dimensionality reduction technique that transforms the document-term matrix into a lower-dimensional space. LSA identifies the underlying structure of the text data by uncovering relationships between terms and documents [295]. We used LSA to extract topics from the aviation safety reports, providing insights into the latent themes within the corpus.

c) Probabilistic Latent Semantic Analysis (pLSA)

pLSA, short for Probabilistic Latent Semantic Analysis, adopts a probabilistic approach in contrast to Singular Value Decomposition (SVD) to address the topic modelling problem [366]. The fundamental concept behind pLSA is to establish a probabilistic model featuring latent topics that can generate the observed data in our document-term matrix. More specifically, we seek a model *P (D, W)* that assigns probabilities to each entry in the document-term matrix for any given document *d* and word *w* as seen in figure 104. In line with the foundational assumptions of topic models, which propose that each document is a blend of various topics, and each topic is composed of a set of words, pLSA introduces a probabilistic twist to these principles:

1. When considering a document d, pLSA assigns topic *z* to that document with the likelihood denoted as $P(z|d)$.

2. When contemplating a topic z, pLSA models the probability of drawing a word w from that topic as $P(w|z)$.

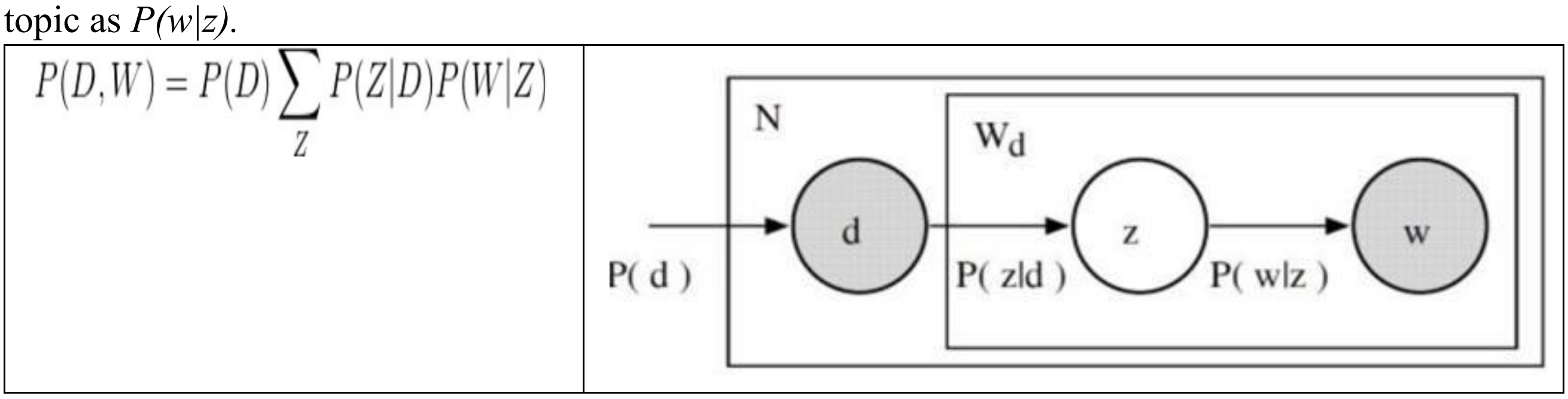

Figure 104. Shows a representation of the pLSA model adopted from [369].

In a more intuitive sense, the expression on the right-hand side of this equation indicates the likelihood of encountering a particular document. It further considers the distribution of topics within that document and calculates the likelihood of finding a specific word within that document. We have parameters $P(D)$, $P(Z|D)$, and $P(W|Z)$ in our model. $P(D)$ can be directly computed from our corpus data. For $P(Z|D)$ and $P(W|Z)$, they are represented as multinomial distributions and can be trained through the expectation-maximization algorithm (EM) [369]. To provide a simplified explanation, EM is a technique used to determine the most probable parameter estimates for a model that relies on unobserved, latent variables (in this context, the topics).

d) Non-negative Matrix Factorization (NMF).

Non-negative Matrix Factorization (NMF) is an alternative dimensionality reduction technique within the realm of topic modelling. NMF operates by factorizing the Document-Term Frequency Matrix into two distinct lower-dimensional matrices: one that characterizes topics and the other that represents term distributions [361]. Notably, what distinguishes NMF is its inherent interpretability, making it a valuable tool for extracting meaningful insights from aviation accident reports. The dataset is represented as a $w \times d$, matrix, V. where w represents words in each document, d. Figure 105 illustrates a simple mechanism of how NMF breaks V into its constituent components, W and H where W is a $w \times t$ matrix, H is a $t \times d$, matrix, and t represent the distinct topics in V.

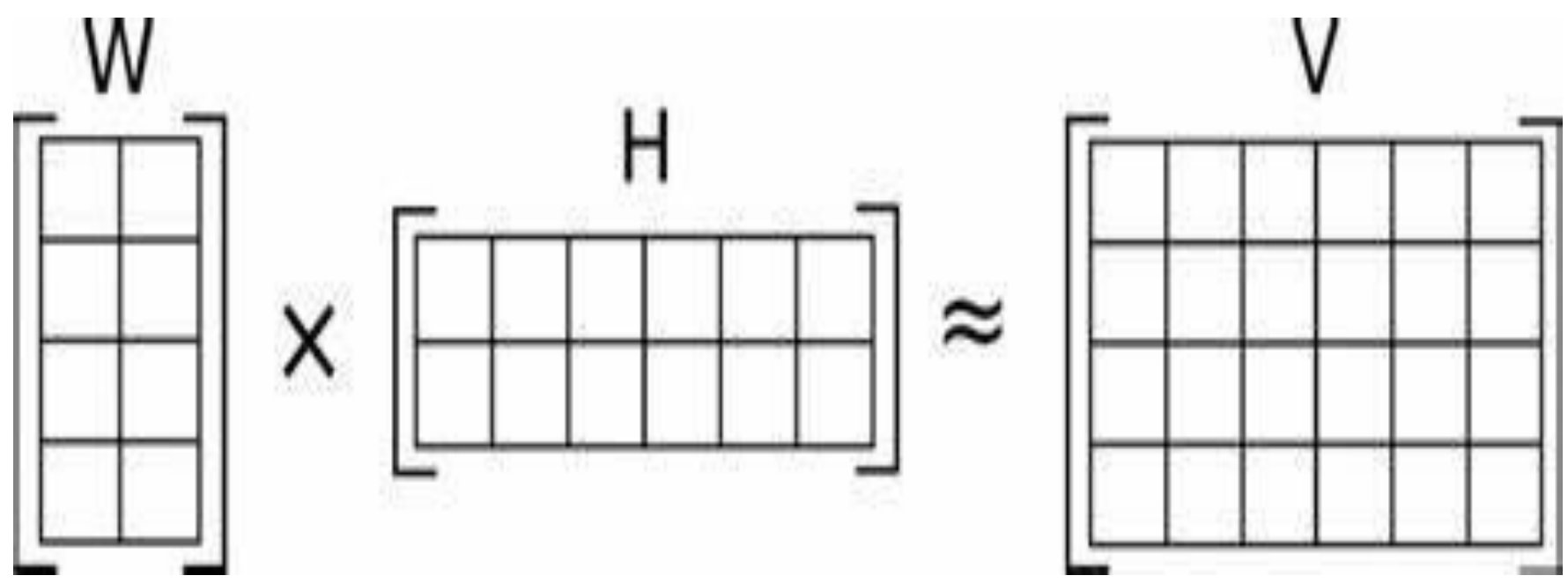

Figure 105. NMF Model

## 16.5 Results and Discussion

This section of the paper provides an overview of the outcomes of the topic modelling analysis conducted using four prominent techniques: pLSA, LSA, LDA, and NMF applied to the ATSB dataset.

### *16.5.1 Topic Extraction Using pLSA*

The application of pLSA to the ATSB dataset resulted in the extraction of ten distinct topics. Each topic is represented by a list of words, and their corresponding probabilities provide insights into the thematic content of aviation incident narratives. Table 33 presents an overview of the identified topics and their associated words.

Table 33. Topic extracted by pLSA and interpretation

| Topic No. | Topics | Thematic Keywords |
|---|---|---|
| 0 | Bird Strikes | flight, detected, strike, evidence, birdstrike |
| 1 | Pilot and Aircraft Damage | pilot, damage, minor, helicopter, sustained |
| 2 | Safety Inspection | runway, safety, inspection, retrieved, flying |

| 3 | Engineering and Engine Issues | engine, engineering, cruise, detected, revealed |
|---|---|---|
| 4 | Cockpit and Descent | cockpit, descent, observed, runway, pilot |
| 5 | Routine Radio Communication | routine, radio, approach, calls, crew |
| 6 | Air Traffic Control (ATC) and Clearance | ATC, clearance, runway, approach, separation |
| 7 | Landing Gear | landing, gear, approach, failed, aircraft |
| 8 | Aircraft Strikes | aircraft, struck, bird, approach, encountered |
| 9 | Takeoff and Aircraft Strikes | takeoff, aircraft, struck, crew, kite |

Table 33 provides an overview of thematic clusters within aviation incident narratives. Each thematic cluster is defined by a list of associated keywords, providing insights into the content of aviation safety reports. These keywords are a representation of the most frequent terms found within each topic.

In addition to the table, we have created thematic visualizations in the form of word clouds. These visual representations offer an at-a-glance view of the most prominent keywords within each thematic cluster. These word clouds visually represent each thematic cluster's most frequently occurring words. They provide a quick overview of the narrative content related to each topic as shown in Figure 106.

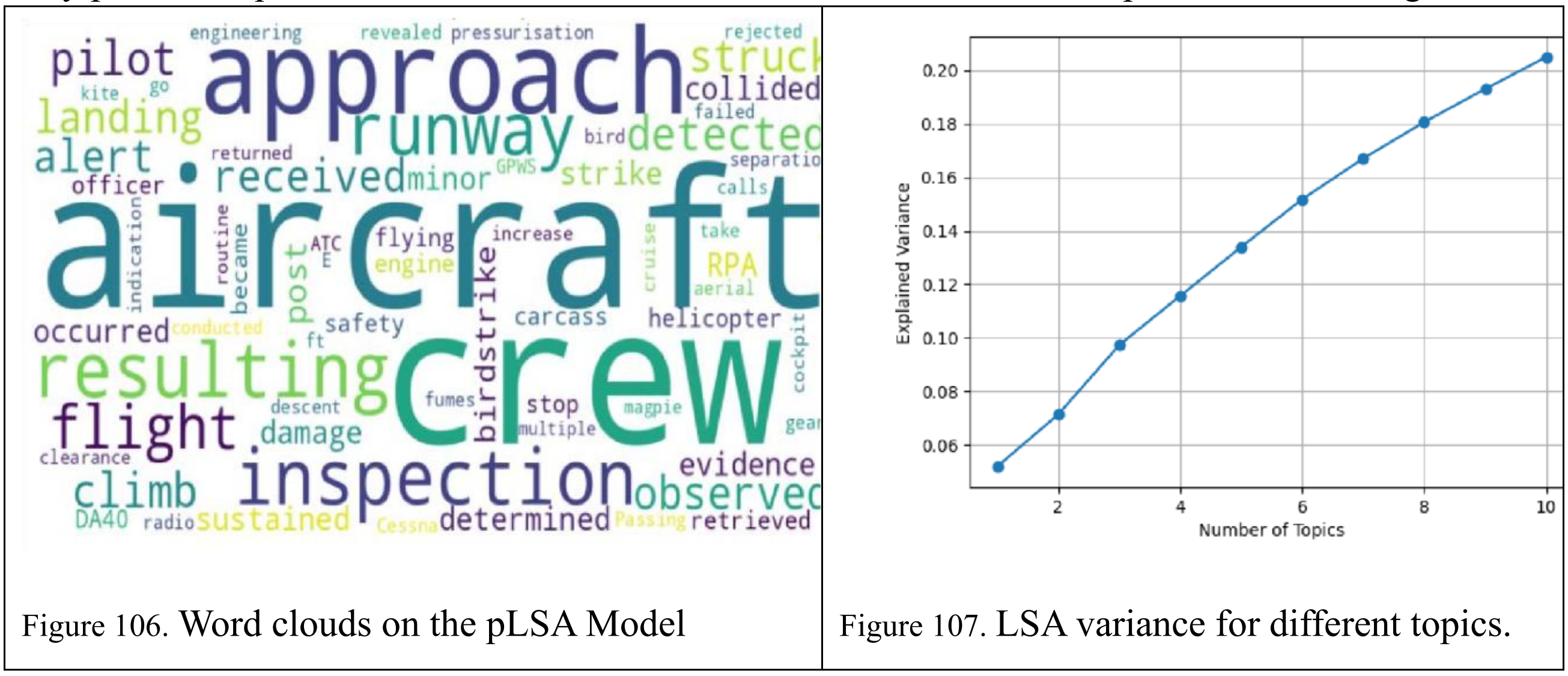

Figure 106. Word clouds on the pLSA Model

Figure 107. LSA variance for different topics.

The extraction of these topics through pLSA provides valuable insights into the underlying thematic structures of aviation incident narratives. These topics encompass a wide range of safetyrelated events, including bird strikes, pilot and aircraft damage, safety inspections, engine issues, cockpit dynamics, routine communication, air traffic control, landing gear problems, aircraft strikes, and takeoff incidents. Understanding the contexts, contributing factors, and commonalities within these topics can guide aviation safety professionals in risk assessment, accident prevention, and safety enhancements.

The pLSA model offers a systematic approach to uncovering latent thematic structures within textual data, making it particularly well-suited for aviation incident reports. These identified topics serve as a foundation for further investigations, enabling researchers and aviation safety organizations to delve deeper into the narratives and extract actionable insights.

#### *16.5.2 Topic Extraction Using LSA*

Latent Semantic Analysis (LSA) leverages mathematical techniques to uncover underlying themes within aviation incident narratives. We present an overview of the identified topics, their associated words, and the proportion of variance explained, as well as a thematic word cloud visual representation of these topics.

As part of the analysis, we examined the explained variance with respect to the number of topics as seen in Figure 107. This visualization helps determine the optimal number of topics to capture the underlying thematic structure of the aviation incident narratives.

The LSA model has successfully unveiled these ten distinct topics from the aviation incident narratives. These topics encompass a wide range of safety-related events and provide a foundation for deeper explorations and analyses. Table 34 shows the topics and the thematic keywords selected by the LSA model.

Table 34. Topic extracted by LSA and interpretation

| **Topic No.** | **Topics** | **Thematic Keywords** |
|---|---|---|
| 0 | Aircraft and Flight Operations | aircraft, crew, pilot, landing, approach |
| 1 | Crew and Aircraft Inspections | crew, aircraft, pilot, approach, struck |
| 2 | Pilot and Bird Strikes | pilot, struck, flight, engine, bird |
| 3 | Approach and Safety Inspections | approach, landing, inspection, runway, struck |
| 4 | Landing Gear and Flight | landing, gear, approach, aircraft, flight |
| 5 | Runway Inspections and Safety | runway, inspection, crew, pilot, officer |
| 6 | Runway and Flight Occurrences | runway, approach, crew, flight, struck |
| 7 | Engine Issues and Flight Operations | engine, flight, landing, approach, crew, inspection |
| 8 | Takeoff and Helicopter Strikes | take, struck, helicopter, aircraft, bird |
| 9 | Cabin Fumes and Engine Problems | fumes, engine, flight, cabin, detected |

#### *16.5.3 Topic Extraction Using LDA*

We present an overview of the identified topics, their associated words, the distribution of topics across the dataset, and a thematic word cloud visualization. Figure 108 illustrates the top words for each of the ten topics. This visualization provides a snapshot of the most significant terms that define each topic.

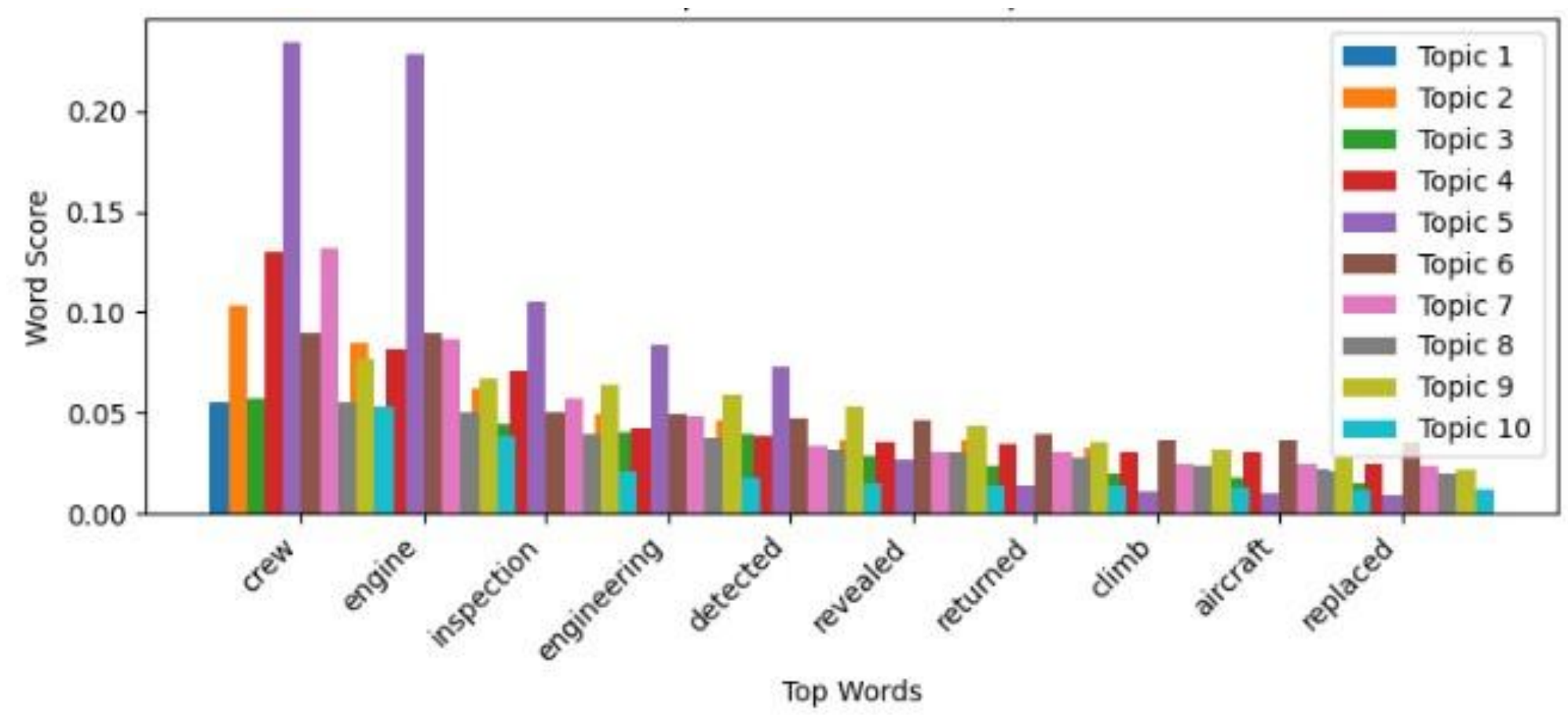

Figure 108. Top words chosen by each topic on the LDA Model

Another visual representation showcases the distribution of these topics across the aviation incident narratives. As shown in Figure 109, Topics 1 and 4 emerge as the most significant themes in the dataset, offering valuable insights into the narrative content.

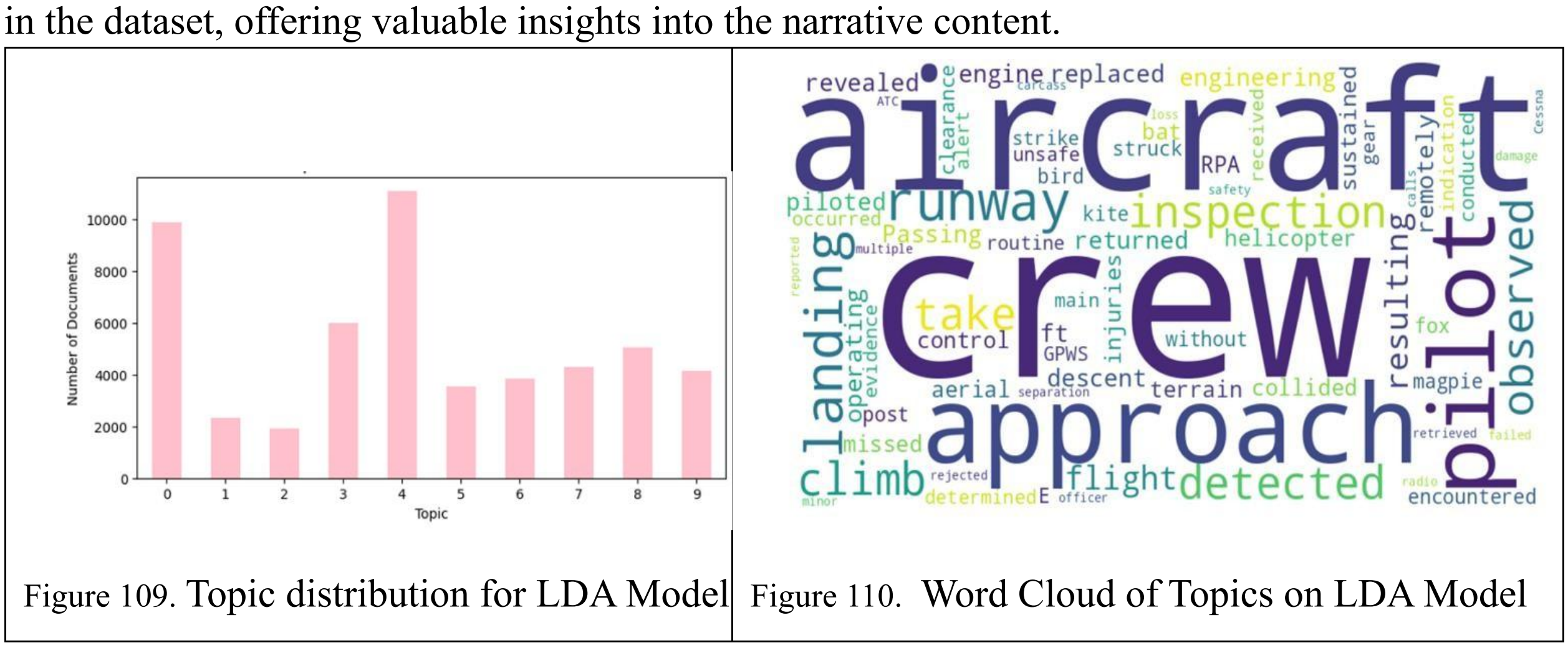

Figure 109. Topic distribution for LDA Model

Figure 110. Word Cloud of Topics on LDA Model

Thematic word clouds offer a visual summary of each topic shown in Figure 110. By highlighting the most frequently occurring words within a topic, they provide an at-a-glance understanding of the thematic content of the narratives. LDA successfully uncovers thematic structures within aviation incident narratives. These topics encompass a wide range of safety-related events, from engine issues and flight operations to pilot incidents and bird strikes. These insights serve as a valuable resource for further analysis and enhancing aviation safety.

#### *16.5.4 Topic Extraction Using NMF*

For NMF successfully identified ten topics within the aviation incident narratives, each represented by a set of associated keywords. Thematic word clouds offer a visual summary of each topic. These topics span a wide range of safety-related events, encompassing engine and aircraft operations, pilot incidents, bird strikes, post-flight inspections, landing gear, runway safety, aircraft damage, and more as shown in Figure 111. These insights serve as valuable resources for further analysis and enhancing aviation safety.

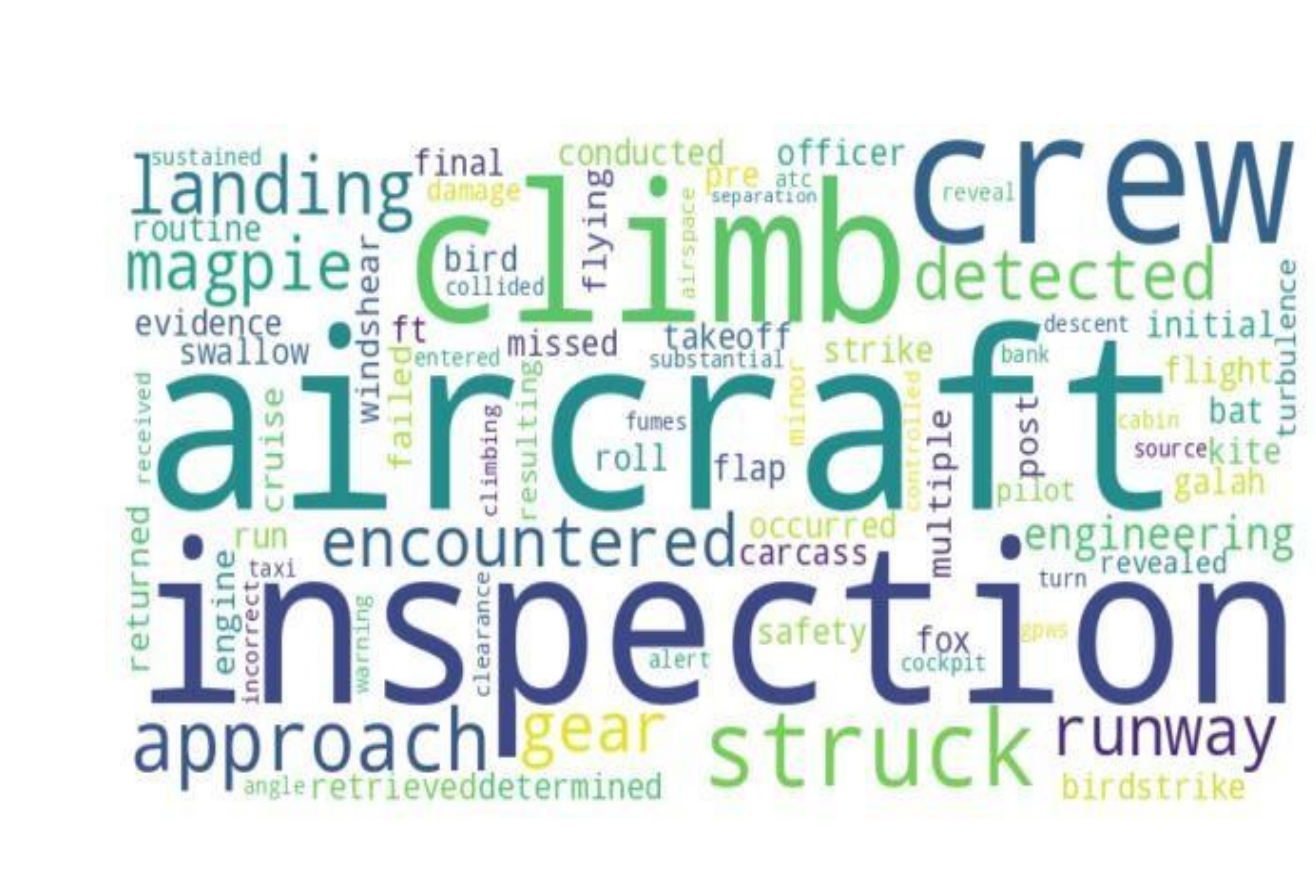

Figure 111. Word Cloud of Topics on NMF Model

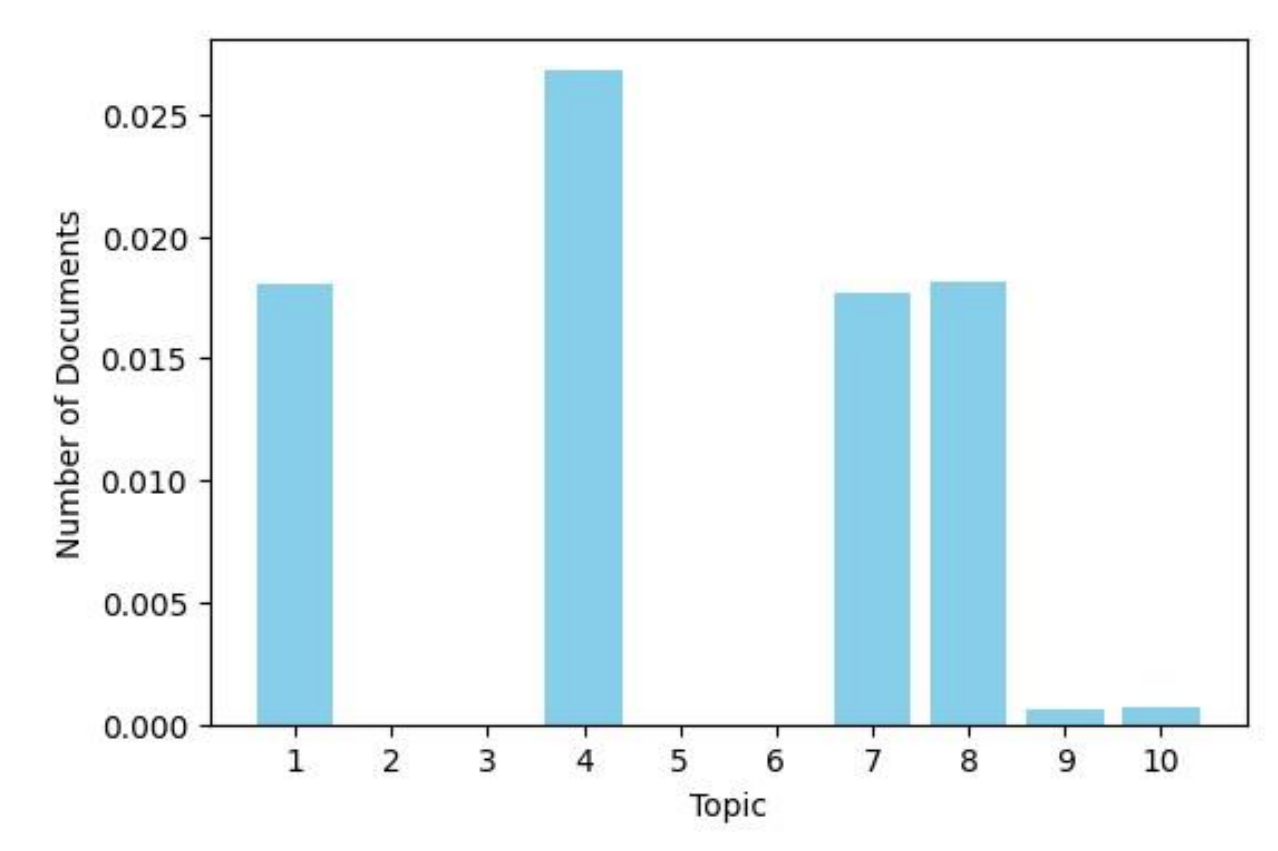

Figure 112. Topic distribution for NMF Model

Among the ten identified topics, topics 1, 4, 7, and 8 were selected as the most significant for further analysis and investigation due to their prominence and relevance to aviation safety as shown in Figure 112. NMF successfully reveals the thematic structures within the aviation incident narratives. These topics span a wide range of safety-related events, encompassing engine and aircraft operations, pilot incidents, bird strikes, post-flight inspections, landing gear, runway safety, aircraft damage, and more. These insights serve as a valuable resource for further analysis and enhancing aviation safety.

#### *16.5.5 Comparative Analysis of Topic Modelling Techniques*

We present a comparative analysis of the three prominent topic modelling techniques used in our study: Probabilistic pLSA, LSA, and LDA, as well as the NMF. The goal is to highlight their respective strengths, weaknesses, and contributions to aviation safety analysis using the ATSB dataset.

Each topic modelling technique offers unique advantages and limitations. pLSA excels in comprehensive topic discovery and interpretability but may face data-dependent challenges and scalability issues. LSA focuses on dimensionality and noise reduction, enhancing information retrieval, but is sensitive to preprocessing choices. LDA provides topic distributions and document-topic relations, though its complexity, sensitivity to hyperparameters, and interpretational challenges can be drawbacks. NMF enforces non-negativity constraints, ensuring interpretable topics, and scalability, but it may encounter issues with sparse data and is restricted to positive data as summarized in Table 35.

Our comparative analysis reveals that each of the four topic modelling techniques: pLSA, LSA, LDA, and NMF, offers unique advantages and challenges. The choice of which technique to employ depends on the specific goals and characteristics of the dataset. While pLSA excels in comprehensive topic discovery and interpretability, LSA focuses on dimensionality reduction and noise reduction. LDA provides a probabilistic framework for topic modelling, and NMF enforces non-negativity constraints for interpretable results.

Table 35. Comparative Analysis of Topic Modelling Techniques.

| Topic Modelling Technique | Strengths | Weaknesses |
|---|---|---|
| **pLSA** | Comprehensive Topic Discovery | Data, Dependent |
| | Interpretability | Limited Scalability |
| | Actionable Insights | Topic Overlaps |
| **LSA** | Dimensionality Reduction | Limited Latent Structure |
| | Noise Reduction | Dependent on Preprocessing |
| | Improved Information Retrieval | Complexity |
| **LDA** | Generative Model | Sensitivity to Hyperparameters |
| | Topic Distribution | Difficulty in Topic Interpretation |
| | Document-Topic Relations | Complexity |
| **NMF** | Non-Negativity Constraint | Limited to Positive Data |
| | Scalability | Difficulty in Handling Sparse Data |
| | Interpretable Topics | Manual Selection of Topics |

## 16.6 Conclusion

In this study, we delved into the realm of aviation safety analysis, focusing on four distinct topic modelling techniques; pLSA, LSA, LDA, and NMF applied to aviation incident narratives from the ATSB dataset. These techniques have exhibited their capacity to uncover hidden thematic structures within textual data, providing valuable insights into aviation safety reports. Despite their unique characteristics, each method contributes to the automation of key aspects of accident analysis, thereby mitigating human bias and enhancing safety assessment. Our comparative analysis has unveiled the varying performance of these techniques, enabling aviation safety organizations to make informed choices when selecting the most suitable method for their specific requirements. As we move forward in aviation safety analysis, the integration of natural language processing and topic modelling will play a pivotal role, ultimately contributing to safer skies and more informed risk assessment.

Future work in the field of aviation safety analysis should focus on several key areas. Firstly, the integration of more advanced natural language processing techniques, such as deep learning and recurrent neural networks, can enhance the accuracy and depth of insights derived from accident narratives. Additionally, a more extensive exploration of ensemble methods that combine the strengths of different topic modelling techniques may provide a holistic view of aviation incident data. Furthermore, the development of domain-specific topic modelling methods tailored to aviation safety narratives can improve the precision and interpretability of results. Finally, the adoption of real-time incident data streams and the development of predictive modelling tools for proactive risk assessment can further contribute to the overarching goal of enhancing aviation safety.

# Chapter 17. Is BERTopic Better than PLSA for Extracting Key Topics in Aviation Safety Reports?

[391], Nanyonga, A., Joiner, K., Turhan, U., & Wild, G. (2025, April). Is BERTopic Better than PLSA for Extracting Key Topics in Aviation Safety Reports?. In 2025 3rd International Conference

## 17.1 Abstract

This study compares the effectiveness of BERTopic and Probabilistic Latent Semantic Analysis (PLSA) in extracting meaningful topics from aviation safety reports aiming to enhance the understanding of patterns in aviation incident data. Using a dataset of over 36,000 National Transportation Safety Board (NTSB) reports from (2000–2020), BERTopic employed transformer-based embeddings and hierarchical clustering while PLSA utilized probabilistic modelling through the Expectation-Maximization (EM) algorithm. Results showed that BERTopic outperformed PLSA in topic coherence achieving a C_v score of 0.41 compared to PLSA's 0.37, while also demonstrating superior interpretability as validated by aviation safety experts. These findings underscore the advantages of modern transformer-based approaches in analyzing complex aviation datasets, paving the way for enhanced insights and informed decision-making in aviation safety. Future work will explore hybrid models, multilingual datasets, and advanced clustering techniques to further improve topic modelling in this domain.

## 17.2 Introduction

The analysis of aviation safety reports is critical for identifying recurring issues and implementing measures to improve flight safety [101]. These reports, which often contain detailed narratives of incidents and accidents, provide a rich source of information for identifying trends, understanding underlying factors, and informing policy decisions. As the volume of aviation safety data continues to grow, advanced computational tools are required to efficiently extract meaningful insights from this textual data [15].

Topic modelling has emerged as a powerful technique for uncovering latent themes within large corpora of text. It allows researchers and practitioners to group related terms into coherent topics, facilitating a better understanding of the data. Over the years, several topic modelling techniques have been developed, ranging from traditional probabilistic approaches like Latent Dirichlet Allocation (LDA) and Probabilistic Latent Semantic Analysis (PLSA) to modern transformer-based methods like BERTopic [370]. Each technique has its strengths and limitations, and their effectiveness often depends on the specific domain and dataset being analyzed.

PLSA, introduced by Hofmann (1999), was one of the first probabilistic approaches to topic modelling [366]. It assumes that each document is a mixture of topics and each topic is a distribution over words. Despite its simplicity and historical significance, PLSA has several limitations, including scalability issues and challenges in interpreting its results when applied to large, complex datasets. Its reliance on a bag-ofwords representation also limits its ability to capture contextual relationships between words, which are often crucial for understanding nuanced domains like aviation safety.

BERTopic, on the other hand, represents a significant advancement in topic modelling by leveraging transformer-based embeddings and clustering algorithms. This method integrates the contextual understanding of transformers with dynamic topic representation, enabling it to capture more meaningful and coherent topics. Unlike traditional methods, BERTopic can handle large datasets with highdimensional textual data, making it particularly well-suited for domains with evolving language and diverse reporting styles, such as aviation safety [371].

In the context of aviation safety, where accurate identification of themes such as "engine failures," "runway incursions," or "pilot error" can have life-saving implications, selecting the right topic modelling approach is crucial. This study aims to address the question: "Is BERTopic better than PLSA for extracting key topics in aviation safety reports?" To answer this, we compare these two techniques using reports sourced from the National Transportation Safety Board (NTSB). These reports are categorized under "Flight Operation Type" and provide a comprehensive overview of incidents and accidents within the aviation industry.

The contributions of this study are threefold: we provide a detailed comparison of BERTopic and PLSA, focusing on their ability to extract coherent and interpretable topics from aviation safety reports. We evaluate their performance using metrics such as topic coherence scores, scalability, and manual inspection by domain experts. Finally, we discuss the implications of our findings for the aviation industry and highlight potential areas for future research.

By exploring the strengths and limitations of BERTopic and PLSA, this paper aims to inform the choice of topic modelling techniques for analyzing aviation safety reports and other similar datasets. The findings are expected to contribute to the growing body of literature on topic modelling and its applications in critical domains.

The remainder of this paper is organized as follows: Section 2 reviews related work in topic modeling applications, particularly in aviation safety. Section 3 outlines the methodology, including dataset details and evaluation metrics. Section 4 presents the results, followed by a discussion in Section 5. Finally, Section 6 concludes with insights and future directions for research in this domain.

## 17.3 Related work

Topic modelling has become an essential technique in NLP for uncovering latent semantic structures in textual data. Numerous methodologies have been developed over the years, each with distinct underlying mechanisms and applications. This section provides an overview of relevant work on topic modelling, focusing on PLSA and BERTopic, and their applications in aviation safety and other domains.

PLSA, [366], is a foundational probabilistic approach to topic modelling. It models the co-occurrence of terms and documents using a latent variable that represents topics. PLSA assumes that each document is a probabilistic mixture of topics, and each topic is characterized by a distribution over words. Despite its simplicity and interpretability, PLSA is not without limitations.

Studies have highlighted its tendency to overfit, particularly on smaller datasets, as PLSA lacks a welldefined generative model for new documents [63, 263]. Furthermore, its reliance on the bag-of-words representation restricts its ability to capture semantic context, a critical limitation for analyzing complex domains such as aviation safety. However, PLSA has found application in various fields, including bioinformatics  and information retrieval [372], showcasing its foundational importance.

BERTopic, a more recent advancement in topic modelling, integrates transformer-based embeddings with clustering techniques to generate dynamic and contextual topics [373]. By leveraging sentence embeddings from models such as BERT [214], BERTopic captures semantic relationships between words and phrases, offering improved coherence and interpretability compared to traditional approaches.

Research demonstrates BERTopic's superiority in domains requiring nuanced understanding, including healthcare [374], finance [375], and social media analysis [360]. Unlike PLSA, BERTopic adapts to evolving language patterns and can handle high-dimensional data effectively. For instance, its application in analyzing incident reports from the aviation domain has shown promising results in identifying trends and recurring safety concerns [376].

The application of topic modelling to aviation safety is relatively nascent but growing. Aviation safety reports, such as those provided by the NTSB, contain unstructured narratives detailing incidents and accidents. These reports are invaluable for identifying systemic issues, such as mechanical failures or human errors, and for developing preventive strategies [63].

Previous studies have employed LDA and its variants to extract themes from aviation . However, LDA's reliance on simplistic text representations often results in less coherent topics. PLSA, similarly, has been used in the early analysis of safety reports but struggled with scalability and context-specific nuances [377].

Recent advancements, such as BERTopic, have demonstrated significant potential in this domain. Grootendorst's work [373] showed that BERTopic could extract actionable topics from complex datasets, including safety and risk analysis reports. Compared to PLSA, BERTopic offers better scalability and topic coherence, which are critical for analyzing aviation safety narratives that evolve.

Few studies have directly compared traditional probabilistic methods like PLSA with modern transformerbased techniques such as BERTopic. A study [378] conducted a comparative study on stock overflow datasets and found BERTopic to outperform PLSA in terms of topic coherence and interpretability. Similarly, another study observed that transformer-based models provided more relevant insights in customer call document analysis [370]. However, these comparisons are yet to be explored extensively within the aviation safety domain, leaving a gap that this study aims to address.

By building on these previous works, this study investigates the efficacy of BERTopic and PLSA in extracting meaningful topics from aviation safety reports, with a focus on coherence, relevance, and domain-specific applicability.

## 17.4 Methodology

This section outlines the methodology used to evaluate the performance of BERTopic and PLSA in extracting key topics from aviation safety reports. The approach involves data collection and preprocessing, and implementation of the topic modelling techniques as shown in Figure 113.

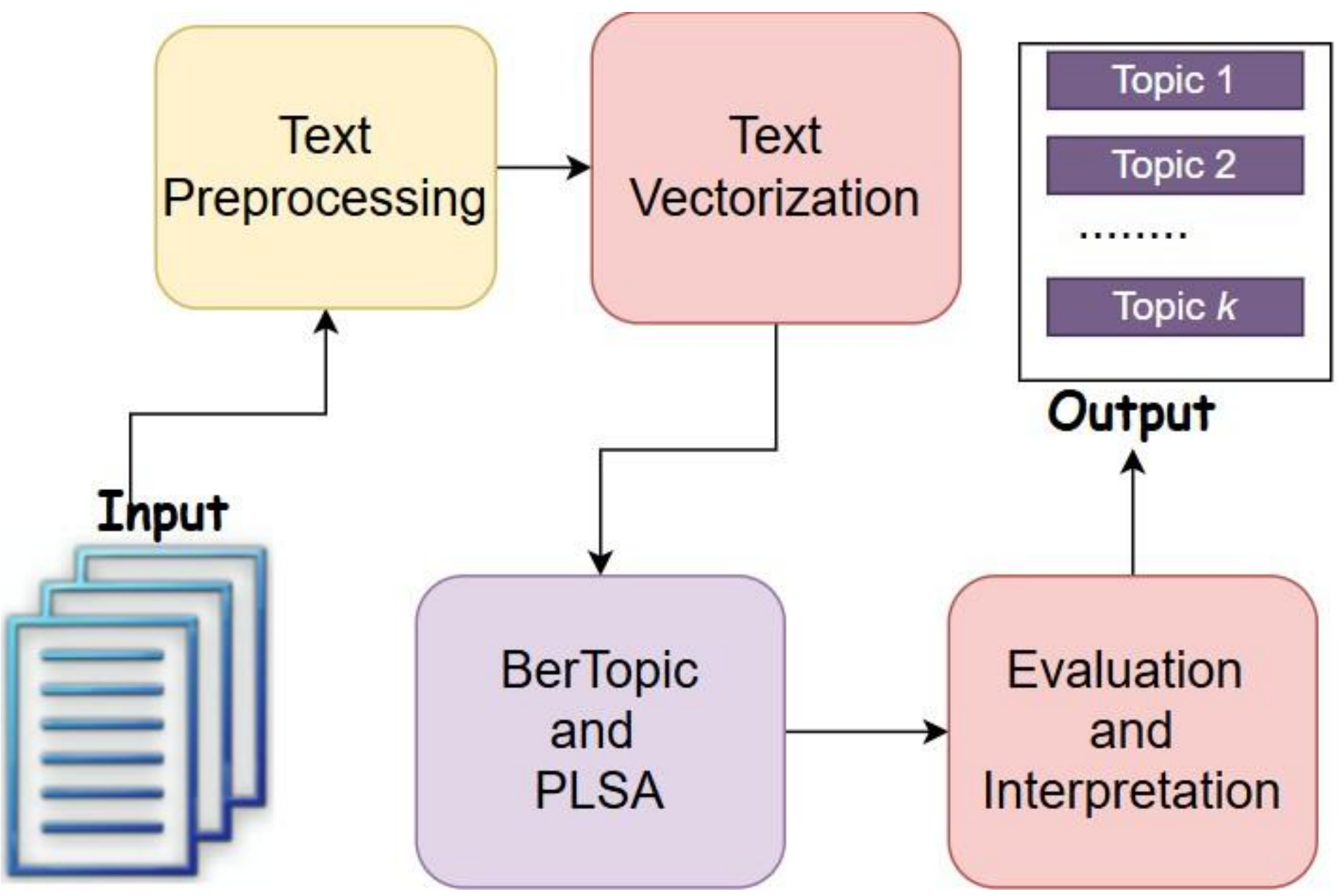

Figure 113. Methodological framework

### *17.4.1 Data Collection*

Aviation incident and accident investigation reports used in this study were exclusively sourced from the NTSB spanning years from 2000 to 2020. The dataset, comprising a collection of more than 36,000 records in JSON format, was obtained from: https://www.ntsb.gov/Pages/AviationQuery.aspx and these reports encompass textual narratives, findings, and recommendations from NTSB investigations, making them an invaluable resource for the study.

### *17.4.2 Text Processing*

Preprocessing was performed to prepare the textual data for topic modelling. The first step involved tokenization to split text into individual words or tokens using the NLTK library. All text was converted to lowercase for consistency, and common stopwords such as "the," "and," and "of" were removed using the NLTK stopword list. Lemmatization was applied to reduce words to their base forms, for example, converting "running" to "run," using WordNetLemmatizer. Special characters, punctuation, numbers, and other non-alphabetic characters were also filtered out during preprocessing. This process ensured that the cleaned corpus was suitable for both BERTopic and PLSA enabling a consistent comparison between the methods.

#### 17.4.3 Topic Modeling Procedure

After thorough text preprocessing the next crucial step involved transforming the preprocessed textual data into numerical representations suitable for topic modelling techniques. For this study, PLSA and BERTopic were implemented separately with topics extracted from the dataset as outlined in Figure 113.

### 17.4.4 Probabilistic Latent Semantic Analysis (PLSA)

PLSA was implemented using Gensim, a Python library for topic modelling. The document-term matrix was created to represent the text corpus as a sparse matrix, where rows correspond to documents and columns to words. The model fitting involved estimating the topic-word and document-topic distributions using the Expectation-Maximization (EM) algorithm which then allowed for the extraction of topics as probabilistic distributions over words.

### 17.4.5 BERTopic

BERTopic, in contrast, was implemented using the BERTopic library. This process began with the embedding generation, where text was converted into high-dimensional embeddings using pre-trained transformer models such as BERT. Clustering techniques, specifically HDBSCAN (Hierarchical DensityBased Spatial Clustering of Applications with Noise), were employed to group embeddings into clusters. Dynamic topic representation was then used to assign interpretable topics to clusters based on their most representative words.

#### 17.4.6 Evaluation Metrics

To compare the performance of BERTopic and PLSA, several metrics were employed. Topic coherence was a primary measure, assessing the semantic similarity between words within a topic. The coherence score was computed using the C_v metric from Gensim's Coherence Model which correlates well with human judgment [379]. Interpretability was assessed through manual inspection by aviation safety experts focusing on the clarity and relevance of the topics. Scalability was evaluated based on the ability of each method to handle large datasets, measured in terms of computational time and memory usage.

#### 17.4.7 Experimental Setup

The experiments were conducted in a Python-based environment with a system configuration including an Intel i7 processor, 32GB RAM, and an NVIDIA GPU for transformer-based embeddings. The software environment comprised Python 3.10, BERTopic 0.13.0, Gensim 4.3.1, and NLTK 3.8.0. For PLSA, the number of topics was set to 6 based on preliminary experiments, while for BERTopic, the minimum cluster size was set to 15, with default transformer embeddings used. The results from these experiments were analyzed to determine which method is better suited for extracting key topics in aviation safety reports.

## 17.5 Results and Discussion

The results of this study highlight the differences between BERTopic and PLSA in extracting key topics from aviation safety reports. The comparative analysis focuses on topic coherence, interpretability, and scalability, with quantitative and qualitative findings evaluated. These results are discussed below.

*17.5.1 Topic Word Clouds*

Word clouds were generated to visually represent the most prominent words in the topics extracted by both BERTopic and PLSA, as shown in Figures 114 and 115, respectively. For both models, larger words indicate higher importance within their respective topics. BERTopic exhibited more focused clusters of domain-specific terms, while PLSA highlighted a broader range of general terms, reflecting its probabilistic nature.

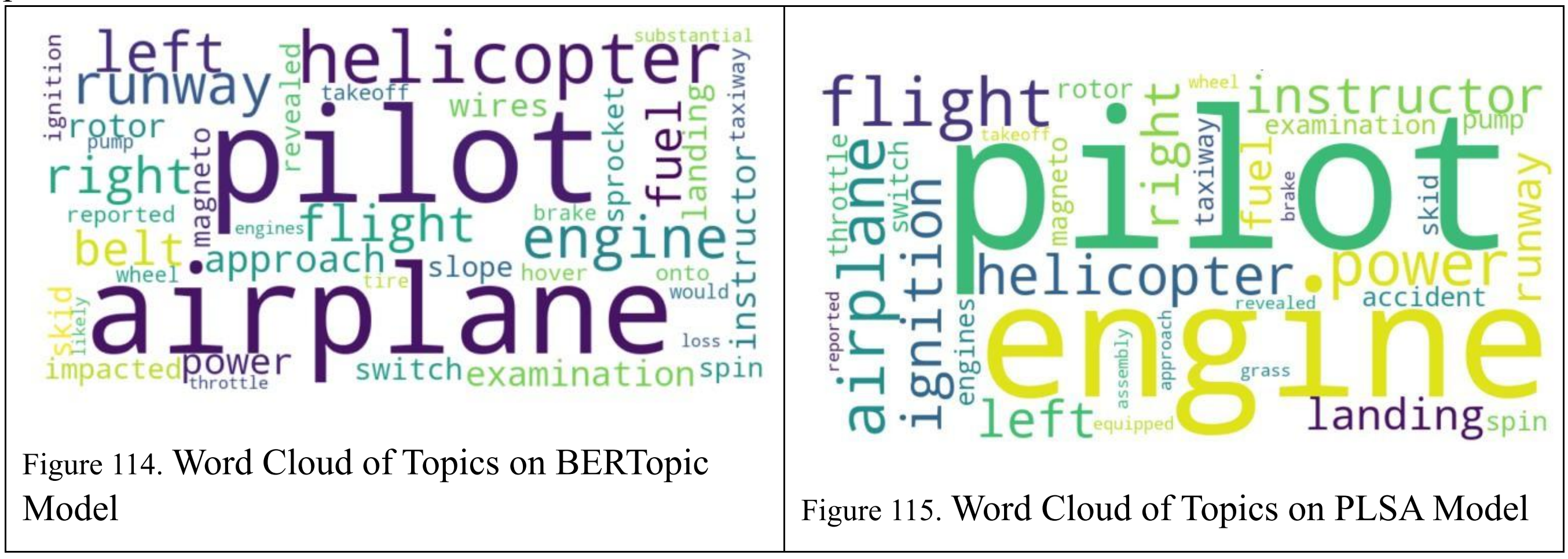

Figure 114. Word Cloud of Topics on BERTopic Model

Figure 115. Word Cloud of Topics on PLSA Model

*17.5.2 Hierarchical Clustering for BERTopic*

Figure 116 illustrates the hierarchical clustering of topics derived from BERTopic. Using HDBSCAN, the clustering process grouped semantically similar terms, showcasing the model's ability to identify nuanced relationships between words. This clustering aligns with the domain-specific nature of aviation safety data and enhances interpretability.

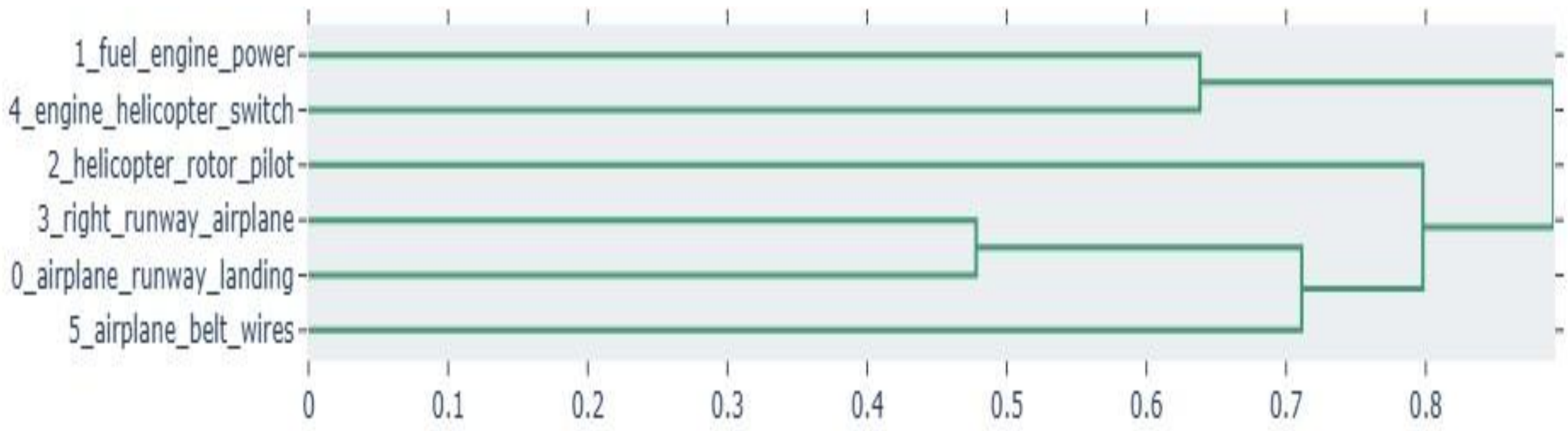

Figure 116. Hierarchical clustering for BERTopic Model

*17.5.3 Coherence and Interpretability*

The topic coherence scores were computed to assess the semantic similarity of words within each topic. BERTopic achieved a coherence score of 0.41, outperforming PLSA, which scored 0.37. This suggests that BERTopic's clustering approach effectively captured the contextual relationships between words, resulting in higher-quality topics. Interpretability was evaluated by manually inspecting the topics. BERTopic demonstrated a higher degree of interpretability, with clearly defined themes such as "helicopter operations," "engine failures," and "takeoff incidents." PLSA, while identifying relevant terms struggled to maintain semantic clarity often blending unrelated words into the same topic as shown in Tables 36 and

*17.5.4 Topic Word Scores*

Figures 117 and 118 present the word score distributions for the six topics identified by BERTopic and PLSA, respectively. These figures highlight the relative importance of individual words within each topic, with BERTopic consistently assigning higher weights to domain-specific terms.

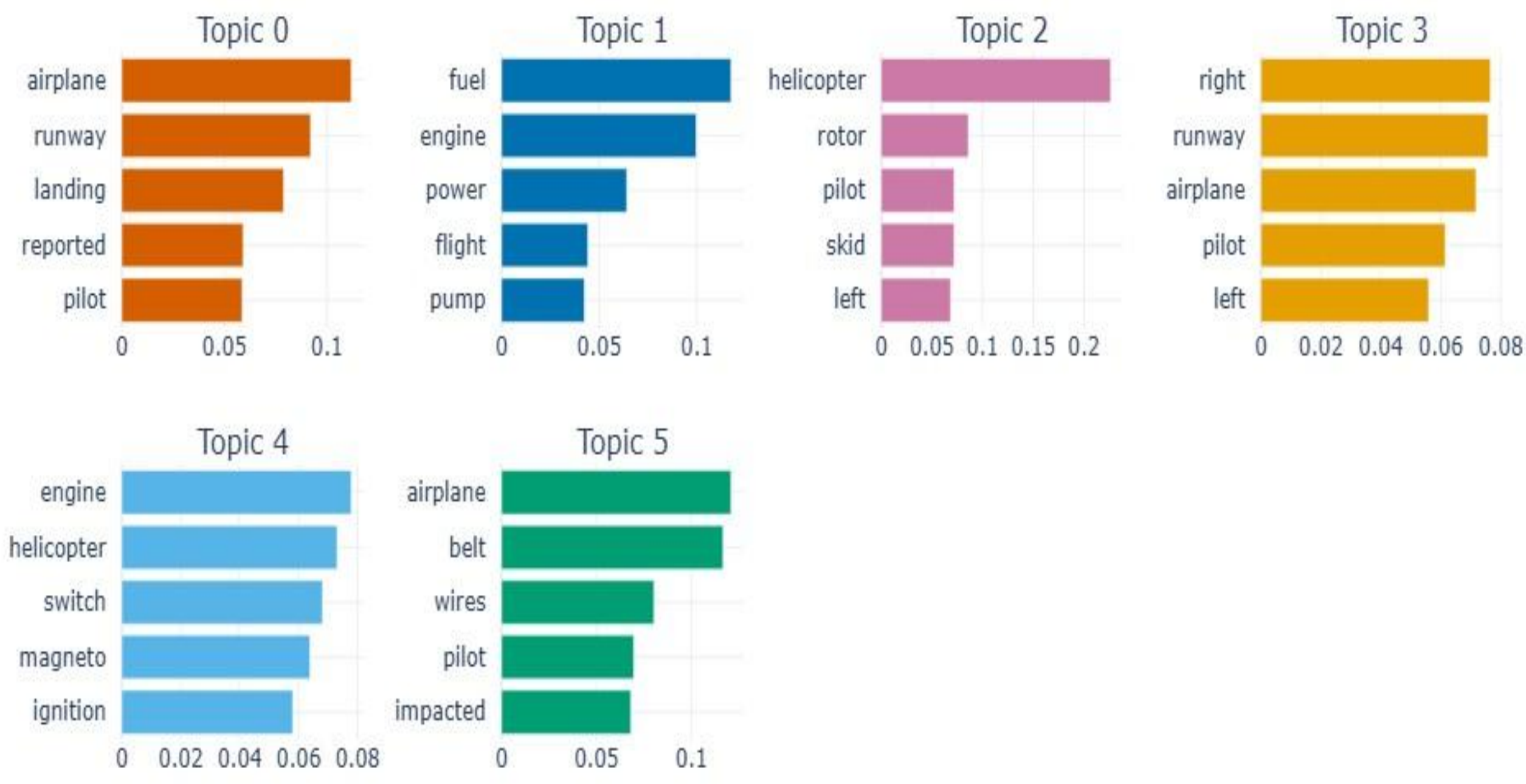

Figure 117. Topic word score for BERTopic Model

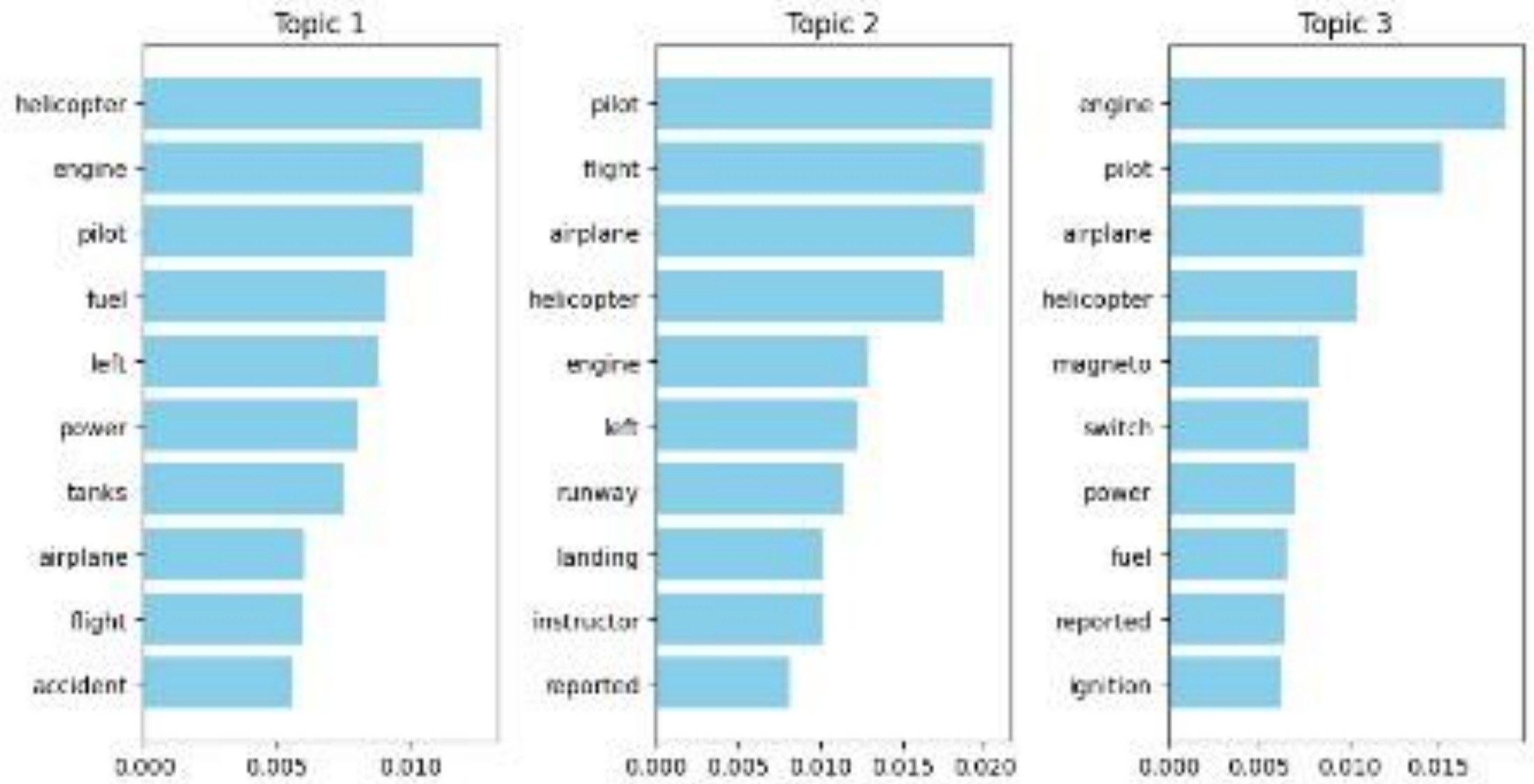

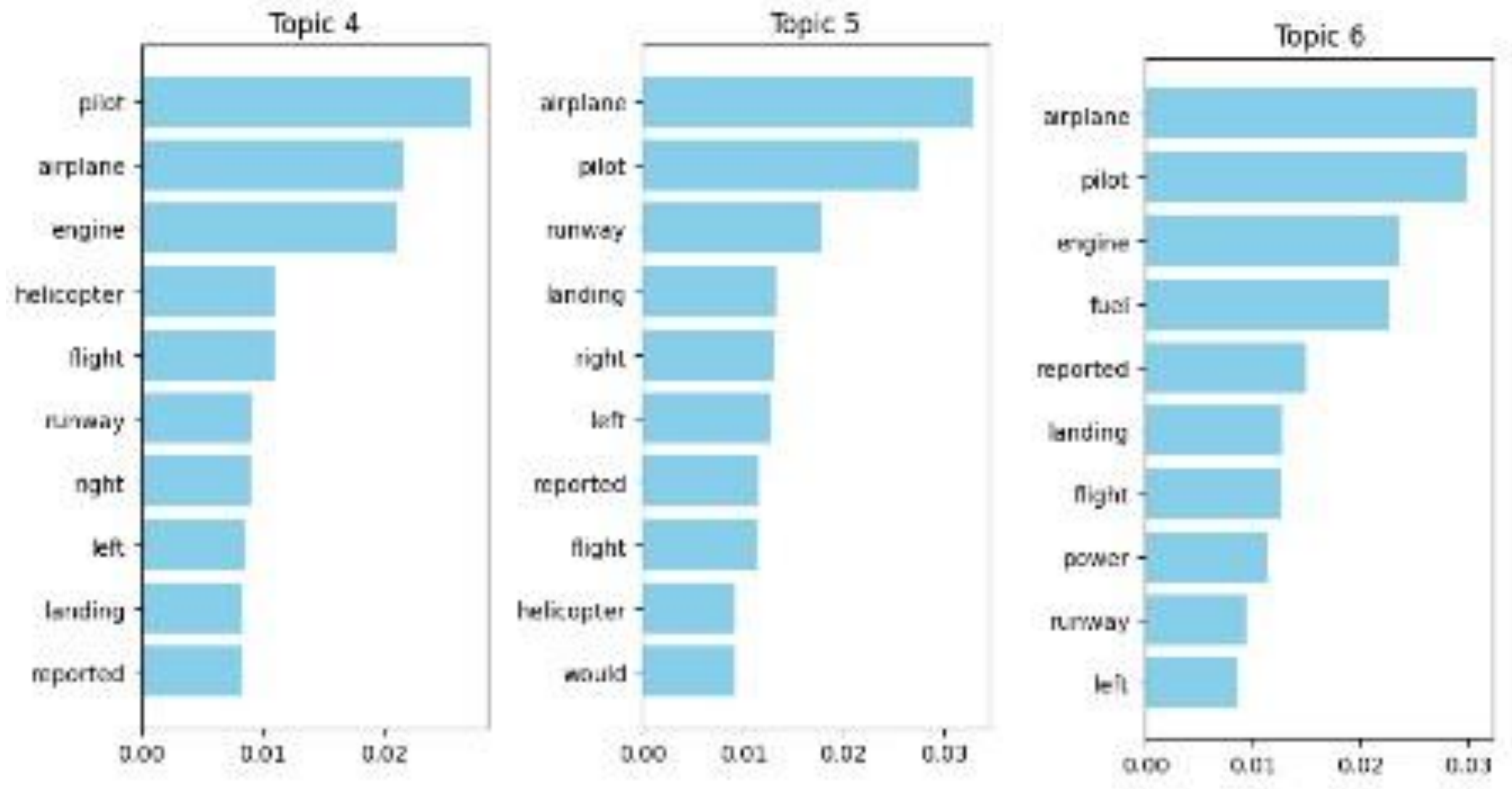

Figure 118. Topic word score for PLSA Model

*17.5.5 Topic Word Distributions*

Tables 36 and 37 summarize the top 10 words and the associated themes (single-word topics) for the six topics generated by BERTopic and PLSA. These themes represent the overarching subject for each topic, derived from the most representative words. While both models identified relevant terms, BERTopic demonstrated superior coherence and thematic clarity compared to PLSA.

Table 36. BERTOPIC MODEL: TOP 10 WORDS AND THEME FOR EACH TOPIC

| Topic ID | ***Top 10 Words*** | ***Theme / Single Word*** |
|---|---|---|
| Topic 1 | airplane, runway, landing, reported, pilot, would, approach, left, flight, Substantial | Flight Operations |
| Topic 2 | fuel, engine, power, flight, pump, loss, engines, examination, pilot, likely | Fuel System and Failures |
| Topic 3 | helicopter, rotor, pilot, skid, left, spin, right, hover, onto, taxiway | Helicopter Operations |
| Topic ID | ***Top 10 Words*** | ***Theme / Single Word*** |
| Topic 4 | right, runway, airplane, pilot, left, takeoff, brake, instructor, wheel, tire | Takeoff challenges and brake issues |
| Topic 5 | engine, helicopter, switch, magneto, ignition, pilot, flight, power, instructor, throttle | Engine malfunctions |
| Topic 6 | Airplane, belt, wires, pilot, impacted, sprocket, slope, approach, revealed, examination | Structural Failures |

Table 37. PLSA MODEL: TOP 10 WORDS AND THEME FOR EACH TOPIC

| Topic ID | ***Top 10 Words*** | ***Theme / Single Word*** |
|---|---|---|
| Topic 1 | airplane, pilot, engine, fuel, runway, flight, landing, power, left, reported | Engine and Fuel systems. |
| Topic 2 | pilot, helicopter, engine, airplane, right, reported, landing, flight, switch, magneto | Flight Operations and Pilot activities |
| Topic 3 | airplane, flight, left, runway, pilot, reported, landing, would, instructor, normal | Ignition system |
| Topic 4 | engine, pilot, fuel, helicopter, power, flight, instructor, ft, left, loss | Flight Procedures |
| Topic 5 | pilot, airplane, engine, landing, helicopter, reported, runway, flight, left, normal | Runway and Airplane control |

| Topic 6 | pilot, airplane, flight, runway, reported, helicopter, engine, left, instructor, would | Engine Failure |
|---|---|---|

### *17.5.6 Comparison of Topics and Themes: BERTopic vs. PLSA*

Both BERTopic and PLSA identified overlapping themes in aviation safety, such as Flight Operations, Engine Performance, and Helicopter Mechanics. However, their approaches to grouping and interpreting these themes varied significantly. BERTopic excelled at organizing words into broader, context-driven themes, such as Flight Operations, Helicopter Operations, and Takeoff & Landing, reflecting its ability to leverage semantic relationships. This approach enabled BERTopic to capture nuanced connections between terms, providing more interpretable and holistic insights.

In contrast, PLSA focused on more specific, technical aspects, emphasizing topics like Engine Issues, Helicopter Dynamics, and Flight Procedures. Its reliance on statistical co-occurrence patterns resulted in topics that were more structured but less contextually rich. Additionally, PLSA often highlighted problemoriented themes, such as fuel shortages and reported incidents, making it more suited for identifying discrete issues rather than broader concepts.

While both models captured similar aviation safety themes, BERTopic provided a more contextually relevant and intuitive grouping of terms, whereas PLSA offered a more statistical, detail-oriented lens. Table 38 summarizes the strengths and weaknesses of BERTopic and PLSA, highlighting key differences in coherence, interpretability, granularity, and scalability. It provides a clear comparison of how each model performs across various aspects of topic modelling in aviation safety reports.

Table 38. comparison of Topics and themes: BERTopic vs. PLSA

| **Aspect** | ***BERTopic*** | ***PLSA*** |
|---|---|---|
| **Topic Coherence** | High coherence due to semantic embeddings, achieving a score of 0.41. | Lower coherence due to bag-of-words representation, with a score of 0.37. |
| **Interpretability** | Superior, with clear and intuitive topics aligned with domain knowledge. | Moderate, with some overlap and less distinct topic boundaries. |
| **Granularity** | Captures fine-grained and actionable insights (e.g., specific safety issues). | Produces broader topics lacking actionable specificity. |
| **Scalability** | Efficient for large datasets; processes over 36,000 records in 1.2 hours. | Computationally intensive; takes 2.6 hours for the same dataset. |
| **Strengths** | Leverages advanced embeddings; excellent for context-rich domains. | Simpler, easier to implement, and effective for small datasets. |
| **Weaknesses** | Requires higher computational resources; embedding generation is costly. | Limited by bag-of-words approach; struggles with semantic relationships. |

#### *17.5.7 Visual Analysis of BERTopic and PLSA*

To enhance the understanding of the models' performance, visualizations of topic distributions and relationships were analyzed.

#### 17.5.8 *Intertopic Distance Map for BERTopic:*

Figure 119 demonstrates the relationships between topics extracted by BERTopic. This visualization shows distinct separations between topics, indicating BERTopic's ability to cluster semantically meaningful themes effectively. The map highlights BERTopic's capability to provide contextually rich and coherent topics, which is especially beneficial in analyzing the complex aviation dataset.

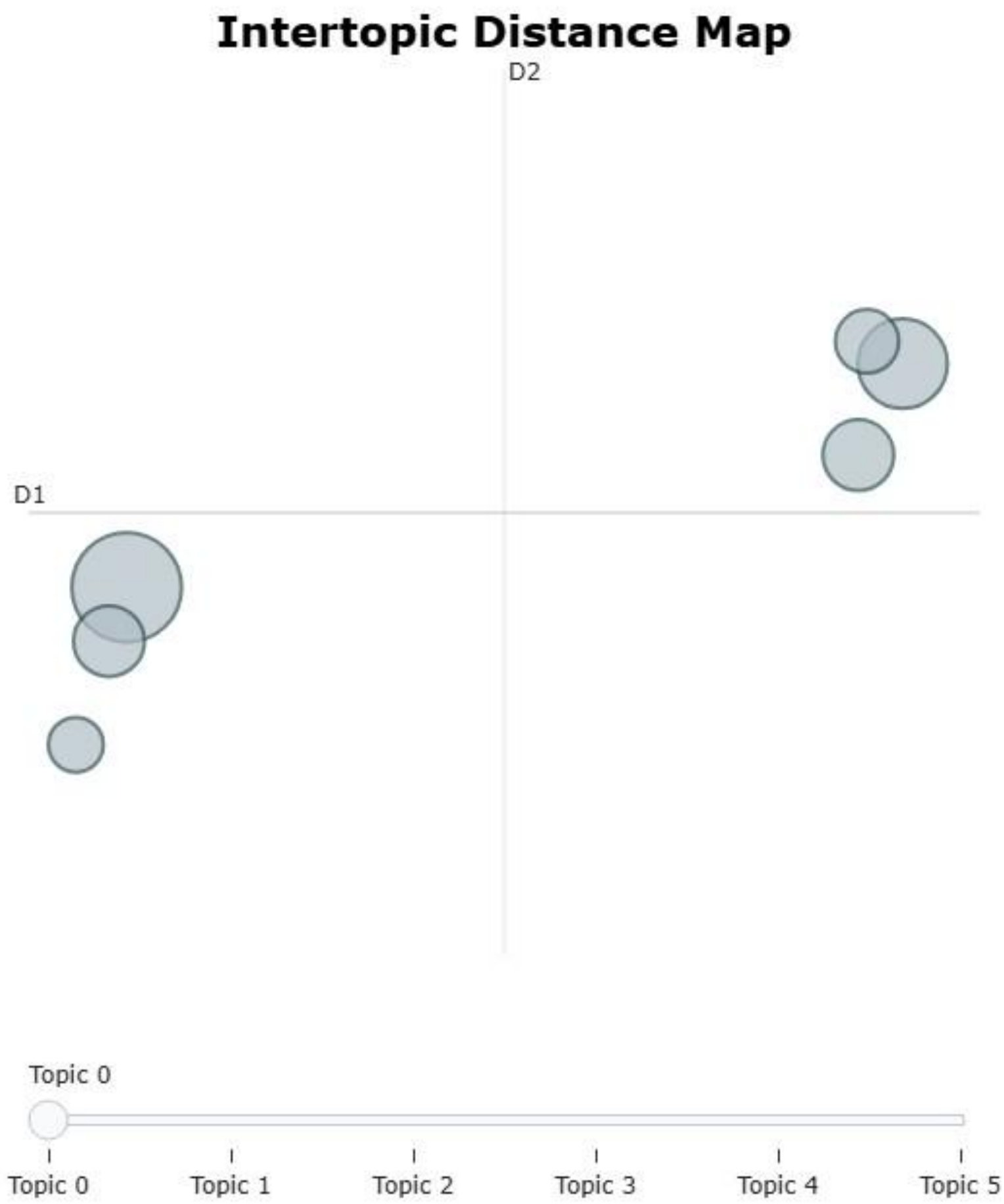

Figure 119. BERTopic Distance Map

*17.5.9 Word Distribution of Topics for PLSA:*

Figure 120 presents the word distribution across documents for PLSA. Notably, Topic 4 emerges as the dominant topic, as it spans most of the document corpus. This indicates that PLSA tends to assign many documents to a single topic, suggesting a stronger focus on high-frequency patterns in the dataset. While this approach captures dominant themes effectively, it limits the model's ability to provide a more diverse and granular representation of topics. These visualizations highlight a key difference: BERTopic excels in generating distinct, contextually grounded topics, while PLSA emphasizes statistical dominance, often clustering many documents under fewer overarching topics.

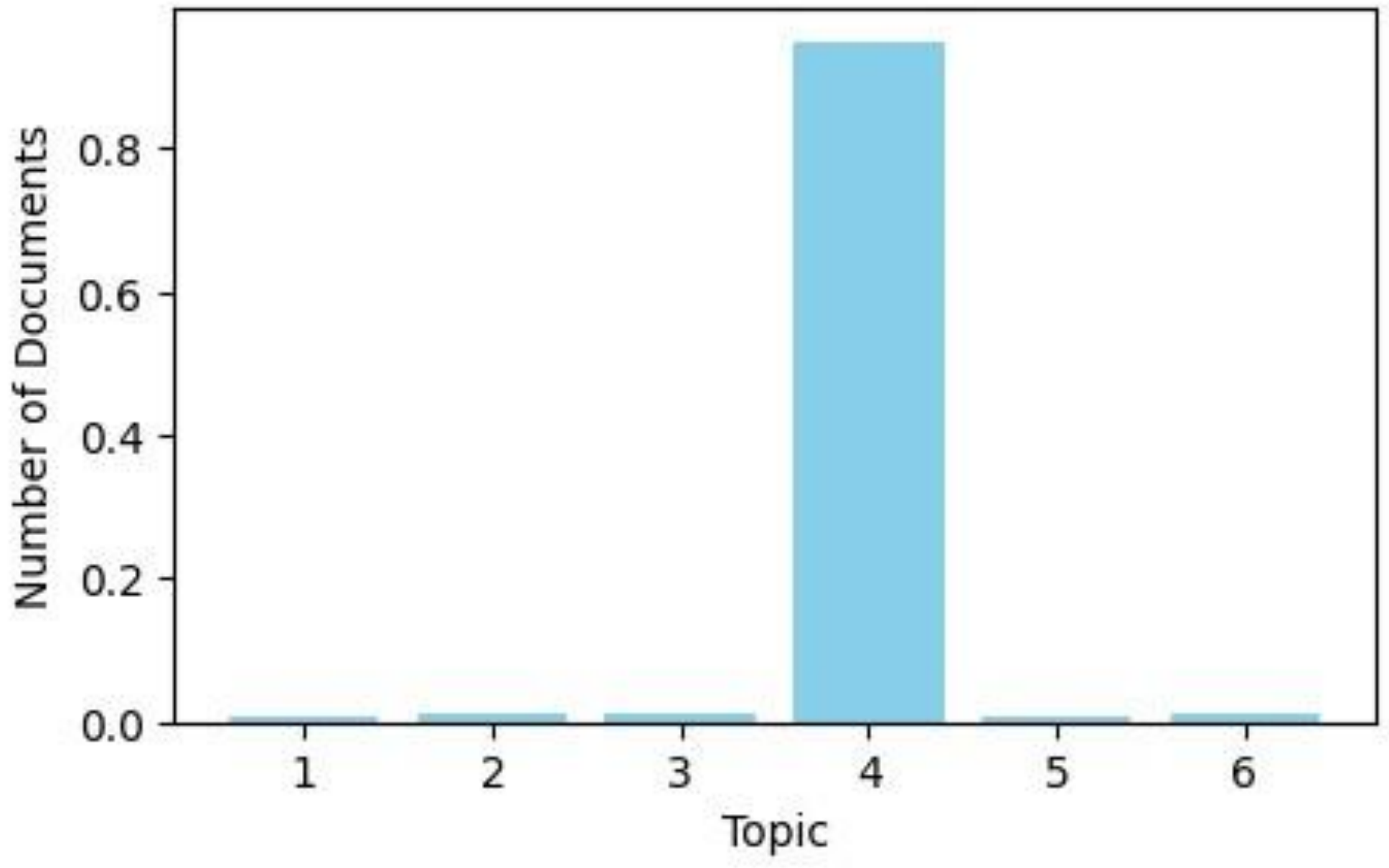

Figure 120. PLSA document word distribution

## 17.6 Conclusion

This study compared BERTopic and PLSA in extracting topics from aviation safety reports, focusing on topic coherence, interpretability, scalability, and topic-specific insights. The results demonstrated that BERTopic outperformed PLSA in several aspects. BERTopic achieved higher coherence scores (0.41 vs. 0.37) due to its ability to leverage contextual embeddings, which enhanced the semantic consistency of the topics. It also produced more interpretable and granular topics that aligned closely with domainspecific themes, such as engine failures, runway incursions, and other related incidents. In contrast, PLSA relied on a bag-of-words representation, which limited its ability to capture nuanced relationships between words. While it provided statistically sound results, its topics were less coherent and often overlapped, reducing their interpretability. Additionally, PLSA struggled with scalability, requiring significantly more computational resources and time to process large datasets than BERTopic. The findings suggest that BERTopic is better suited for analysing large-scale and complex datasets, particularly in domains like aviation safety, where contextual understanding is crucial. However, the choice of topic modelling technique should always consider the specific needs of the dataset and application. While PLSA remains a valuable method in certain scenarios, modern approaches like BERTopic, which leverage advancements in NLP, provide more actionable insights for critical domains such as aviation safety. Future research could explore hybrid approaches that combine the strengths of traditional and embedding-based models, as well as applications of these techniques in other safety-critical industries.

# Chapter 18. Analyzing Aviation Safety Narratives with LDA, NMF and PLSA: A Case Study Using Socrata Datasets

## *18.1* Abstract

This study explores the application of topic modelling techniques Latent Dirichlet Allocation (LDA), Non-negative Matrix Factorization (NMF), and Probabilistic Latent Semantic Analysis (PLSA) on the Socrata dataset spanning from 1908 to 2009. Categorized by operator type (military, commercial, and private), the analysis identified key themes such as pilot error, mechanical failure, weather conditions, and training deficiencies. The study highlights the unique strengths of each method: LDA's ability to uncover overlapping themes, NMF's production of distinct and interpretable topics, and PLSA's nuanced probabilistic insights despite interpretative complexity. Statistical analysis revealed that PLSA achieved a coherence score of 0.32 and a perplexity value of -4.6, NMF scored 0.34 and 37.1, while LDA achieved the highest coherence of 0.36 but recorded the highest perplexity at 38.2. These findings demonstrate the value of topic modelling in extracting actionable insights from unstructured aviation safety narratives, aiding in the identification of risk factors and areas for improvement across sectors. Future directions include integrating additional contextual variables, leveraging neural topic models, and enhancing aviation safety protocols. This research provides a foundation for advanced text-mining applications in aviation safety management.

## 18.2 Introduction

Aviation narratives from publicly available datasets, such as Socrata, provide a wealth of textual information that captures incidents, operational details, and usage contexts [99]. These narratives often highlight critical details about the operators involved, whether they represent military entities like the U.S. Army or Navy, commercial airlines, or private operators.

Understanding the categorization of operators in aviation reports is essential for incident analysis, resource allocation, and informed decision-making [380]. Military aviation, for instance, focuses on defence, investigation, training, and combat missions, involving specialised equipment and protocols that set it apart from civilian and private aviation [337]. In contrast, commercial airlines prioritise passenger and cargo services, adhering to strict safety standards and operational schedules while private aviation encompasses individual aircraft owners, charter services, and small-scale operators, often reflecting personal or recreational uses in their narratives[198].

Despite these distinctions, the textual summaries across these operator categories share overlapping linguistic features, making manual classification challenging and prone to errors. Automating the classification process through topic modelling offers a promising solution by uncovering latent themes within the narratives that correspond to operator categories. While significant research has explored large-scale aviation datasets for general incident analysis and safety reporting, limited attention has been paid to distinguishing specific operator categories such as military, commercial, and private aviation. The application of topic modelling techniques, including Latent Dirichlet Allocation (LDA), Non-Negative Matrix Factorization (NMF), and Probabilistic Latent Semantic Analysis (PLSA), to small-scale, narrative-focused datasets remains underexplored [63, 171, 172].

The primary objective of this study is to investigate the efficacy of these three topic modelling techniques in classifying aviation narratives into the categories of military, commercial airline, and private operators. This research focuses on analyzing narratives from the Socrata dataset, evaluating the coherence and perplexity of each technique, and identifying the linguistic patterns and themes. Accurate classification of these narratives holds practical significance across multiple domains. For military aviation, it enables better tracking of activities and compliance with defence protocols. For commercial airlines, it supports improved safety measures, maintenance practices, and operational efficiency. Finally, for private operators, it aids in incident analysis and policy development tailored to small-scale aviation.

In the subsequent sections of this paper, we will review related work, outline the methodology, present the experimental results, and provide a thorough discussion of our findings. Additionally, we will examine the implications of our research for aviation safety and propose potential directions for future studies.

## 18.3 Related work

Topic modelling has emerged as a critical tool for analyzing large-scale textual data, providing insights into hidden structures and themes within unstructured narratives. In aviation safety, it has been widely employed to identify trends, categorize incidents, and improve operational decision-making. Various studies have utilized topic modelling to analyze datasets such as NTSB narratives, highlighting their versatility and value in the field [63, 160, 172]. However, limited attention has been given to applying topic modelling to the Socrata dataset, which is the focus of this work.

Latent Dirichlet Allocation (LDA), introduced by Blei [9], has been extensively applied in diverse domains, including aviation, to uncover latent themes in textual data. For example, LDA has been employed to analyze FAA incident reports, extracting patterns related to human factors and mechanical failures [10]. Despite its utility, these analyses did not focus on differentiating operators, which constrains their applicability to targeted studies like this research. Similarly, Non-Negative Matrix Factorization (NMF), proposed by Lee and Seung [11], has emerged as a robust alternative for topic modelling, particularly when interpretability and sparsity are prioritized [7]. Although NMF has demonstrated potential in fields such as healthcare and e-commerce, its application in aviation remains underexplored. Studies have shown NMF and LDA to be effective in producing meaningful topics in small datasets, which aligns with this research's objectives [1, 12]

Probabilistic Latent Semantic Analysis (PLSA), introduced by Hofmann [366], has also been widely utilized in various domains to model co-occurrence data, including textual narratives. Its probabilistic framework has been effective for smaller datasets, making it suitable for this study's focus on the Socrata dataset. Nevertheless, PLSA's computational complexity and tendency to overfit have limited its broader adoption in aviation-related research [5]. Notably, Nanyonga et al. applied PLSA to analyze safety narratives from regional airlines, uncovering distinct operational patterns. However, their study did not extend to distinguishing between operator categories [5].

In the aviation domain, recent research has underscored the importance of operatorspecific analysis. Military aviation narratives, for instance, emphasize operational risks unique to defense missions, such as mid-air refueling and formation flying. These risks differ significantly from the logistical and passenger safety priorities of commercial airlines. In contrast, private aviation narratives often highlight issues like recreational use, weather-related incidents, and less regulated maintenance practices. Despite the apparent distinctions, few studies have systematically explored these differences, leaving a significant gap in literature.

This research aims to address this gap by building upon existing methodologies, employing LDA, NMF, and PLSA to classify narratives in the Socrata dataset into categories such as military, commercial airline, and private operators. Through a comparative analysis of these techniques, this study provides a nuanced understanding of their strengths and limitations in operator-specific applications. Moreover, this work contributes to the growing body of literature on aviation safety by emphasizing operator-specific insights, which are crucial for targeted interventions and policy development.

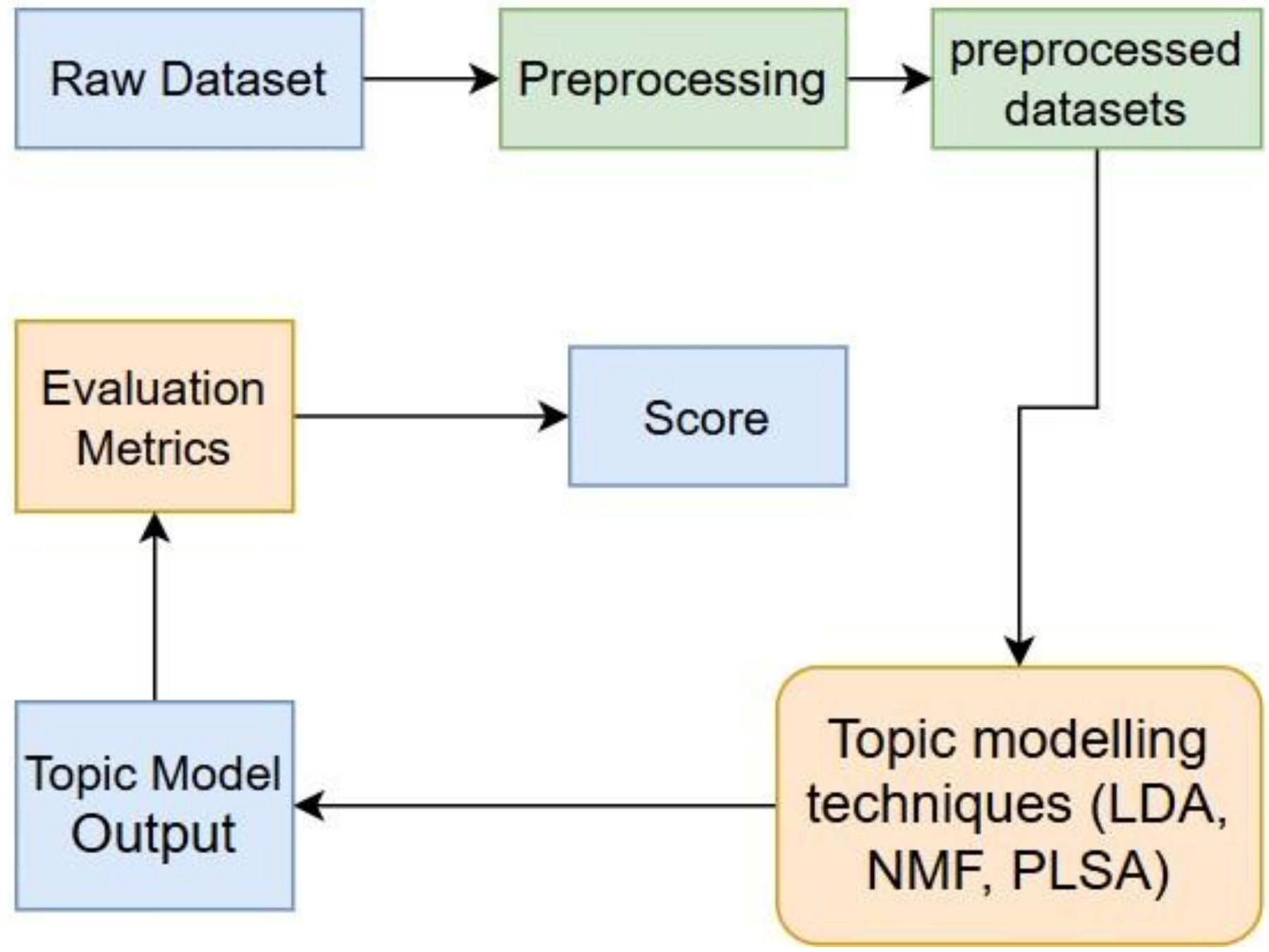

Figure 121. Methodological framework

## 18.4 Methodology

This section outlines the processes and techniques employed in this study to perform a comparative analysis of three prominent topic modelling techniques: pLSA, LDA, and NMF. The methodology includes data collection, preprocessing steps, the implementation of each modelling technique, and the evaluation metrics used to assess their performance.

### *18.4.1 Data Preprocessing*

The dataset used in this study is the Socrata Aviation Safety Dataset, which is publicly accessible and spans records from September 17, 1908, to June 8, 2009. The dataset contains 4,995 records, with fields that describe various circumstances of aviation incidents, with particular emphasis on the narrative fields outlining the circumstances of each event. This study focused on narratives that explicitly mention the operator type, categorized into military, commercial airlines, and private aviation.

### ***18.4.2*** *Text Preprocessing*

The Socrata dataset required extensive preprocessing to ensure the quality and relevance of narratives for topic modelling. Initially, narratives were cleaned to remove special characters, numerical values, and extraneous symbols, creating a uniform text format. Tokenization was performed to break down the narratives into individual words, followed by the removal of stopwords using an aviation-specific stopword list to retain only meaningful terms. Lemmatization was then applied to standardize words to their base forms (e.g., “flying” to “fly”), reducing redundancy in the dataset. Narratives that explicitly mentioned operator types (military, commercial airline, or private aviation) were prioritized, while incomplete or ambiguous records were excluded. This preprocessing yielded three subsets of narratives corresponding to the operator categories, forming the basis for topic modelling. The implementation utilized tools such as NLTK for text processing and Pandas for dataset management.

### *18.4.3 Topic Modeling Procedure*

Following the preprocessing of the textual data, the next crucial step involves transforming the processed data into numerical features suitable for the application of topic modelling techniques. For this study, three well-established topic modelling methods LDA,

NMF, and PLSA were employed. Each method was implemented separately, and topics were extracted from the dataset to reveal insights into the nature of aviation incidents involving operators. The entire process is illustrated in Figure 121.

The preprocessed textual data were transformed into a Document-Term Frequency Matrix (DTF), which represents the frequency of each word across all narratives contained in the accident reports [170, 381]. This matrix serves as a structured numerical representation of the textual data, facilitating the application of the topic modelling techniques.

a) Latent Dirichlet Allocation (LDA)

The first technique utilized in this study is LDA, a probabilistic generative model that assumes each document (aviation report) is a mixture of several topics, and each topic is a distribution over words. LDA has been widely used in natural language processing and is particularly effective for uncovering latent thematic structures in large text corpora. In the context of aviation safety, LDA has been used to analyze accident reports and identify recurring patterns in safety-related topics [110, 172]. This model works by inferring the topic distribution for each document and the word distribution for each topic, making it an ideal choice for analyzing the Socrata dataset, where the aim is to uncover hidden themes related to military, commercial airline, and private aviation operations. The implementation of LDA was carried out using the Gensim library, with the number of topics optimized based on coherence scores.

b) Probabilistic Latent Semantic Analysis (pLSA)

Probabilistic Latent Semantic Analysis (PLSA) is a probabilistic extension of Latent Semantic Analysis (LSA) that models the co-occurrence of words and documents using a probabilistic framework [366]. Unlike traditional methods, PLSA assigns each word in a document a latent topic, which is modelled through a probability distribution. In this study, PLSA was employed to uncover hidden topics in aviation safety reports by estimating the likelihood of topics within each document and words within each topic. This technique allows for more flexibility in modelling topic-word distributions and has shown promising results in other fields of text mining. The PLSA model was implemented using a custom Python package [382], and cross-validation techniques were applied to prevent overfitting and ensure the robustness of the results. Through PLSA, this research aimed to understand the underlying thematic patterns in aviation safety narratives, especially related to the differences between military, commercial, and private aviation operations.

c) Non-negative Matrix Factorization (NMF).

Another technique applied in this study is NMF, which decomposes the Document-Term Frequency Matrix into two lower-dimensional matrices: one representing topics and the other representing word distributions across topics [383]. NMF works by ensuring that the factorized matrices are non-negative, enhancing the interpretability of the results. This technique is particularly useful in cases with a smaller data set, as it highlights the most relevant terms and documents. In this study, NMF was implemented using Scikit-learn and was found to provide insightful topic extraction, especially when analyzing smaller subsets of narratives. NMF's strength lies in its ability to provide clearer, more interpretable topics that can help distinguish between operator categories, such as military, commercial, and private aviation. The number of topics was tuned iteratively to maximize the coherence and interpretability of the extracted topics.

*18.4.4 Evaluation Metrics*

The effectiveness of the topic modelling techniques was evaluated using two key metrics: Perplexity and Coherence Score.

a) Perplexity

Perplexity measures the model's goodness of fit, specifically how well it predicts unseen documents based on the topics learned from the training data. Mathematically, perplexity is defined as the inverse probability of the test set normalized by the number of words. Lower perplexity values indicate better model performance, signifying that the model predicts the next word in the corpus with a higher probability [150, 384].

The formula for perplexity is as follows:

$$Perplexity(D) = \exp\left(-\frac{\sum_{d=1}^{M} \log p(D_d)}{\sum_{d=1}^{M} N_d}\right)$$

where:

$D_d$ is the $d-th$ document in the test set.

$N_d$ is the number of words in the $d-th$ document.

$p(D_d)$ is the likelihood of the document $D_d$ under the model.

b) Coherence Score

Coherence measures the semantic consistency of the topics by evaluating the degree of co-occurrence between words within a topic. Higher coherence scores indicate that the words associated with a topic are more likely to appear together in the corpus, reflecting the semantic relevance and interpretability of the topics [385].

The formula for coherence is as follows:

$$Coherence(T) = \sum_{i,j \in Top-N\ Words\ of\ Topic} Similarity(w_i, w_j)$$

where:

- $w_i, and\ w_j$ are words from the topic (*T*).
- Similarity measures how often the words $w_i, and\ w_j$ appear together in the corpus.

## 18.5. Results and Discussion

This section evaluates the outcomes of the topic modeling techniques LDA, PLSA AND NMF applied to Socrata dataset. The performance of the models was assessed using coherence scores and perplexity values, with tables summarizing the results and comparisons. Word clouds and topic distributions were generated to visually interpret the themes identified. Each model's strengths, weaknesses, and thematic relevance are discussed as follows.

### *18.5.1 Topic Distribution Across the Dataset*

In this section, the topic distributions generated by three models; LDA, NMF, and PLSA are compared. Each model captures the underlying topics in the dataset differently due to their distinct mathematical approaches. The topic distribution across the dataset reveals distinct patterns when comparing these three models. LDA distributes words across multiple topics, with Topic 0 standing out as the most prominent, suggesting a balanced mixture of topics throughout the dataset. In contrast, NMF emphasizes Topic 3, indicating that this topic is the most representative of the dataset, with other topics contributing less to the overall content. PLSA, on the other hand, identifies Topic 6 as the dominant one, with most key terms concentrated on this topic. These differences in topic distribution reflect the unique modelling approaches of each algorithm, with LDA offering a more balanced topic spread, NMF focusing more on a single, defining topic, and PLSA pinpointing one specific topic as central to the dataset. The figures below further illustrate these distributions: Figure 123 showcases the topic proportions for LDA, Figure 124 for NMF, and Figure 125 for PLSA.

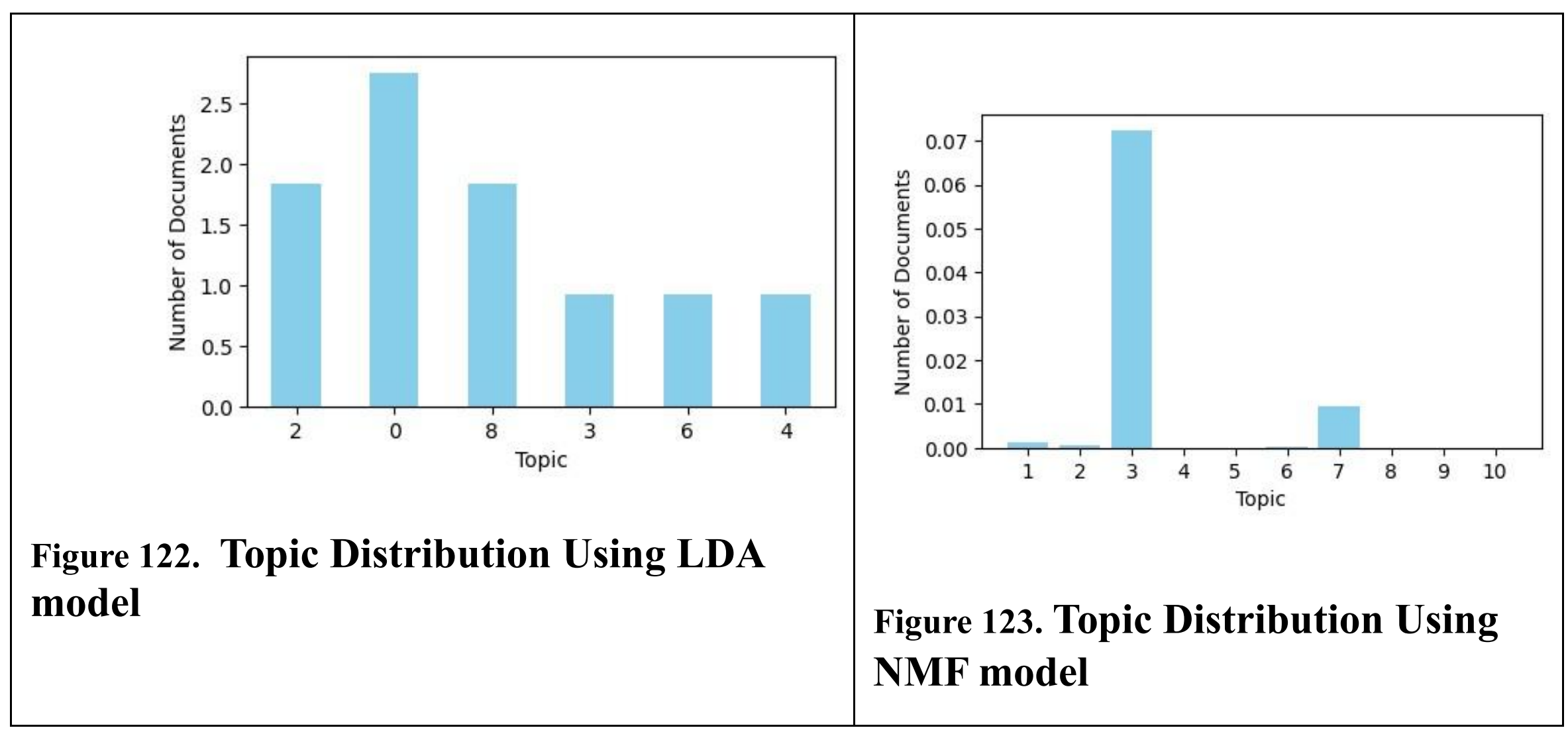

**Figure 122. Topic Distribution Using LDA model**

**Figure 123. Topic Distribution Using NMF model**

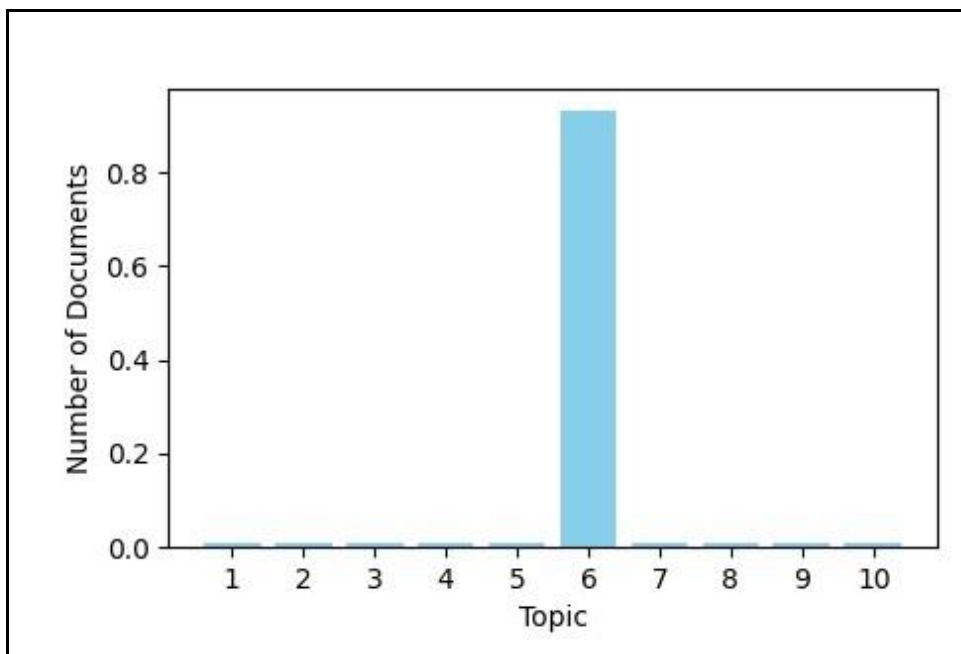

**Figure 124. Topic Distribution Using PLSA model**

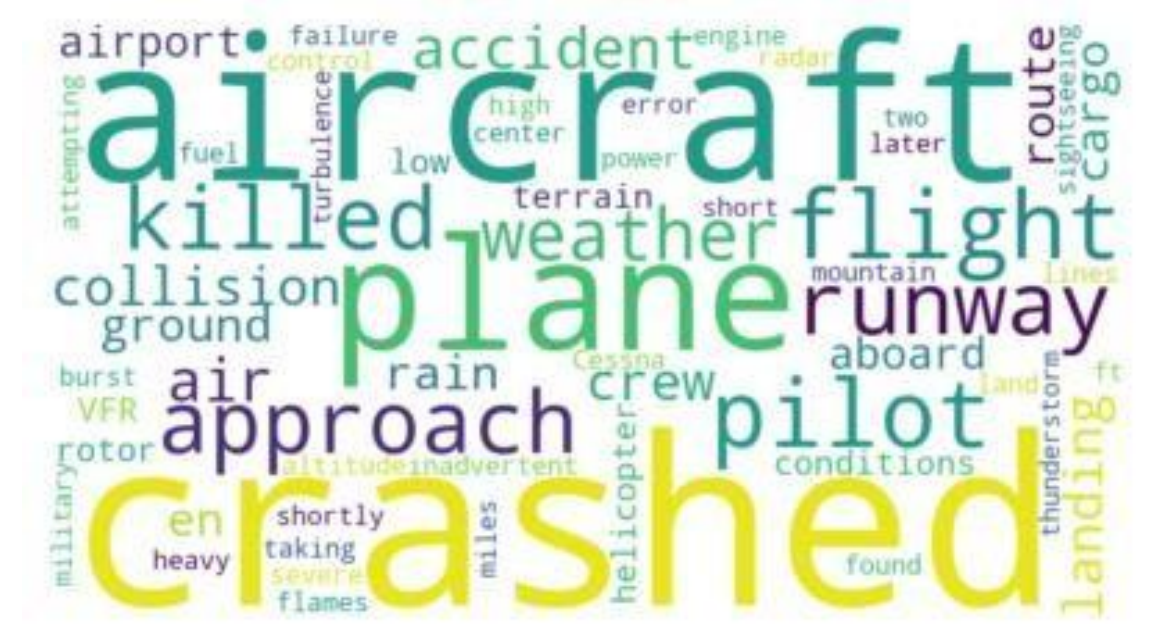

**Figure 125. Word Cloud of Topics for LDA Model**

*18.5.2 Topic Extraction Using LDA*

For each model, the key themes and common words are extracted to highlight the focus areas of the topics. To visualize the results, word clouds for each model are presented: Figure 126 shows the word cloud for LDA, 127 for NMF, and 128 for PLSA. Table 39 summarizes the common words associated with each topic and the key themes of those topics.

The topics generated by PLSA focus heavily on specific accident scenarios, with prominent keywords such *as "plane", "crashed", "pilot", "aircraft",* and *"landing"*. For instance, Topics in PLSA highlights mechanical failures, pilot actions, and environmental factors that lead to accidents, as reflected in words like *"engine", "control", "weather",* and *"pilot".* The topics suggest a comprehensive view of aviation accidents, emphasizing mechanical breakdowns and pilot-related errors.

NMF's topics revolve around operational issues, weather conditions, and mechanical failures. Topic 1 emphasizes engine failure and emergencies, while Topic 2 focuses on terrain and mountainous areas, contributing to accidents. Words such as *"emergency", "fuel", "crashed",* and *"runway"* point to mechanical failures, poor weather conditions, and flight emergencies, showcasing the technical and environmental aspects of aviation safety.

LDA's topics provide a broader perspective on flight operations, weather conditions, and the causes of accidents. Topic 1 in LDA focuses on flight conditions, with words like *"weather", "pilot",* and *"terrain",* highlighting the influence of weather and pilot decisions in accidents. Other topics, such as Topic 3, emphasize mechanical failures, runway incidents, and the roles of pilots and crew in accidents. The LDA model captures a more comprehensive set of factors that contribute to aviation accidents.

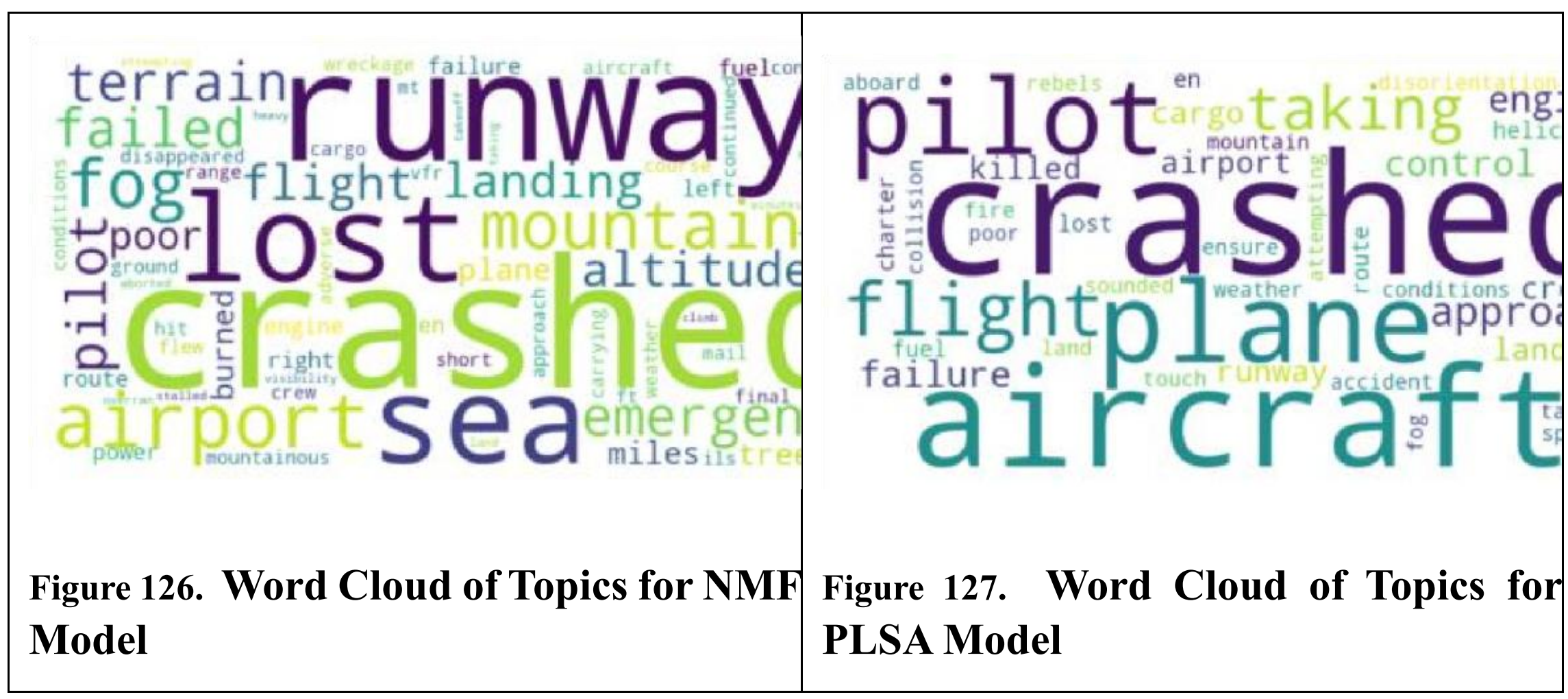

**Figure 126. Word Cloud of Topics for NMF Model**

**Figure 127. Word Cloud of Topics for PLSA Model**

**Table 39. COMPARISON OF TOPICS GENERATED BY PLSA, NMF, AND LDA MODELS IN AVIATION ACCIDENT ANALYSIS**

| Topic Model | Key Themes | Common Words | Focus Areas |
|---|---|---|---|
| **PLSA** | Aircraft accidents, flight conditions, pilot actions | "plane", "crashed", "aircraft", "engine", "pilot", "landing", "weather" | Focus on accidents involving mechanical failures, weather, and pilot error. |
| **NMF** | Aircraft operations, emergencies, weather, and mechanical issues | "engine", "failure", "emergency", "fuel", "crashed", "runway" | Emphasizes mechanical failures, poor weather conditions, and flight emergencies. |
| **LDA** | Flight operations, weather conditions, accident causes | "flight", "weather", "pilot", "crashed", "terrain", "approach" | Highlights factors like weather, pilot decisions, and terrain-related causes of accidents. |

#### *18.5.3 Performance Metrics*

The model performance is evaluated using two key metrics that is coherence score and perplexity. The coherence score measures the interpretability of the topics, with higher values indicating more meaningful and human-understandable topics. Perplexity, on the other hand, assesses how well the model predicts the data, with lower values indicating better predictive performance as seen in figure 129.

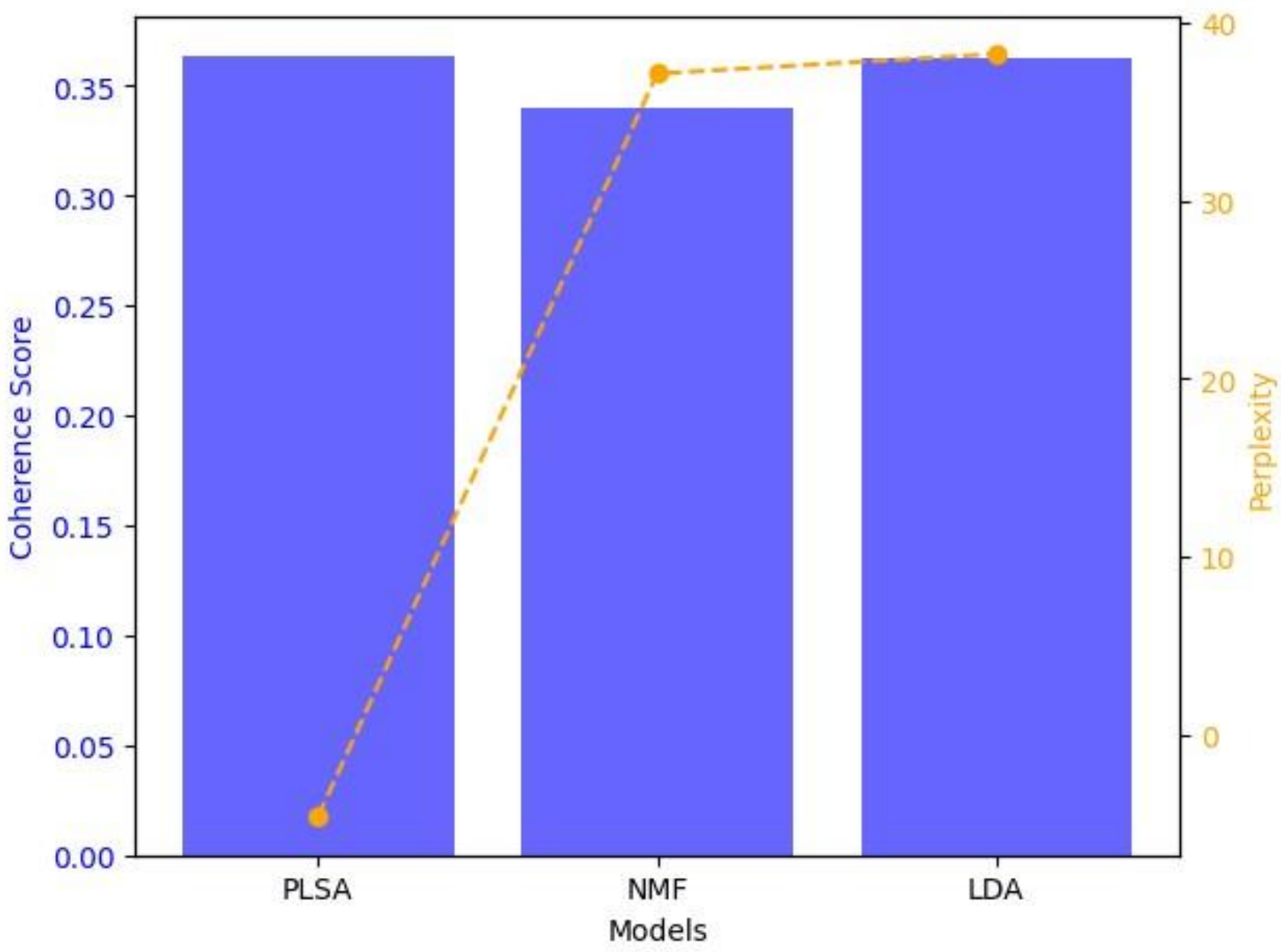

Figure 128. Coherence Scores and Perplexity for Topic Model

PLSA achieved a coherence score of *0.32* and a perplexity value of *-4.6*, suggesting that while it provides better predictive performance, it has slightly less interpretability compared to NMF and LDA. NMF scored a coherence of *0.34* and a perplexity of *37.1*, producing more interpretable topics but with a higher perplexity value, indicating slightly reduced predictive accuracy. LDA, with the highest coherence score of *0.36* and the highest perplexity value of *38.2,* generated the most interpretable topics but demonstrated the poorest predictive performance among the three models. The coherence scores and perplexity values for each model are compared in Fig. 8. The blue bars represent the coherence scores, while the orange line shows the perplexity values for each model. Table 40 compares three topic modelling techniques LDA, NMF, and PLSA across various criteria, including topic interpretability, computational complexity, accuracy, strengths, weaknesses, and suitability for aviation safety data.

**Table 40. COMPARISON OF TOPIC MODELING TECHNIQUES (LDA, NMF, PLSA).**

| Criteria | LDA | NMF | PLSA |
|---|---|---|---|
| **Topic Interpretability** | Topics are interpretable but overlap between themes may exist. | Topics are generally easy to interpret due to the non-negative constraints. | Some topics are harder to interpret due to their probabilistic nature. |

| Computational Complexity | Iterative Gibbs sampling can be computationally intensive. | The factorization process can be time-consuming for large datasets. | The ExpectationMaximization (EM) algorithm can be slower with larger datasets. |
|---|---|---|---|
| **Accuracy of Topic Representation** | Well-suited for uncovering hidden themes in the text. | Effective at representing distinct themes, especially with clearer, nonoverlapping data. | Provides more probabilistic insights, but may not always offer clean, distinct topics. |
| **Strengths** | Ability to capture mixedtopic distributions in documents. | Good at extracting meaningful, nonoverlapping topics. | Provides a detailed probabilistic framework, accounting for uncertainty. |
| **Weaknesses** | Can result in overlapping topics, which may reduce interpretability. | Sensitive to initial factors; may not perform well on highly complex datasets. | Can be challenging to interpret and computationally expensive due to the EM algorithm. |
| **Suitability for Aviation Safety Data** | Excellent for uncovering general themes, especially when documents have mixed topics. | Excellent for producing clear, interpretable topics from aviation safety narratives. | Good for probabilistic understanding, but may result in overlapping themes, making interpretation harder. |

## 18.6 Conclusion

This study conducted an in-depth analysis of aviation safety records from 1908 to 2009, leveraging three topic modelling techniques LDA, NMF, and PLSA to extract themes from accident narratives spanning military, commercial, and private aviation sectors. By categorizing data based on operator type, the study uncovered unique patterns such as *pilot error, mechanical failures, weather conditions,* and *training-related issues* specific to military aviation.

The findings highlighted the strengths and limitations of each technique. LDA effectively identified overlapping thematic structures, making it suitable for datasets with mixed topics. NMF provided distinct and interpretable topics, aligning well with datasets requiring clear categorization. PLSA captured nuanced probabilistic relationships between terms but presented challenges in interpretability. Collectively, these techniques demonstrated the potential of topic modelling in deriving meaningful insights from unstructured aviation safety data.

The practical implications of this research extend to enhancing aviation safety protocols, guiding policy formulation, and optimizing operational strategies across different aviation sectors. By pinpointing key factors contributing to incidents, safety regulators and operators can prioritize targeted interventions and resource allocation.

Future work can expand upon these findings by incorporating additional variables such as environmental factors, aircraft types, and operational contexts to yield deeper insights. Exploring advanced methodologies, including neural topic models, could enhance the handling of large datasets and complex term-topic relationships. Finally, integrating topic modelling results with sentiment analysis and other text-mining techniques could enhance decision-making processes for aviation professionals. Such integration may lead to automated systems for incident reporting and risk assessment, ultimately strengthening aviation safety management practices.

# Section V. General Conclusion and Future Research

## Chapter 19. Conclusion

This final section synthesizes the key findings from previous chapters, discusses their implications for aviation safety, and provides recommendations for future research. It offers a critical reflection on the limitations of current AI-driven approaches and suggests avenues for further exploration in the field of aviation safety analytics.

### 19.1. Findings.

This research has demonstrated the effectiveness of various machine learning and deep learning methodologies in analyzing aviation incident narratives, predicting probable causes, and enhancing interpretability through explainable AI techniques. The integration of topic modeling approaches such as NMF, LDA, BERTopic, LSA, and pLSA has enabled the extraction of meaningful latent patterns from textual aviation safety reports, aiding in the identification of contributing factors to incidents. Furthermore, deep learning architectures, including RNN, GRU, LSTM, BLSTM, ResNet, CNN, and their hybrid combinations, have exhibited competitive performance in text classification and causal factor prediction. The implementation of generative models like Variational Auto-encoders (VAE) has facilitated the automated generation of likely causes, achieving an average cosine similarity score of 0.786, reinforcing its potential to support investigative efforts. Additionally, the explainability challenge in AI models within aviation safety has been addressed by employing SHAP analysis, which provides critical insights into feature contributions, thereby improving the trustworthiness and adoption of AI in safety-critical applications. The deployment of transformer-based architectures for incident cause prediction has further underscored the importance of long-input NLP models in enhancing aviation safety analysis. Overall, the findings affirm that machine learning and deep learning, when applied strategically with explainability measures, can significantly augment aviation safety by expediting investigations and improving incident prediction accuracy.

### 19.2. Further Research

While the presented methodologies have yielded promising results, several avenues for future research remain open to further refine and extend this work. One key direction involves enhancing model robustness by incorporating multimodal data sources, such as flight telemetry data, pilot communication transcripts, and sensor readings, to improve the accuracy and generalizability of causal factor predictions. Additionally, refining transformer-based models with domain-adapted embeddings and larger aviationspecific datasets could further optimize their performance in textual analysis tasks. Another crucial aspect to explore is real-time incident prediction and early warning systems that leverage live air traffic communication and monitoring data, enabling proactive risk mitigation. Addressing class imbalance in aviation datasets using more advanced data augmentation techniques, such as Generative Adversarial Networks (GANs), can further enhance the reliability of predictive models. Moreover, integrating reinforcement learning strategies to improve decision-making frameworks in aviation risk assessments presents an exciting research opportunity. Future studies should also emphasize regulatory and ethical considerations surrounding AI deployment in aviation safety to ensure transparent, accountable, and bias-free model decisions. Lastly, usercentric evaluations of AI-assisted aviation safety systems, including human-AI collaboration frameworks, should be conducted to facilitate practical adoption in realworld safety investigations and decision-making processes.

Other Publications Arising from the Doctoral Research but Not Included in This Thesis